\pdfoutput=1
\documentclass{article}

\usepackage[utf8]{inputenc} 
\usepackage[T1]{fontenc}    
\usepackage{iclr2027_conference,times}
\usepackage{anyfontsize}

\usepackage{url}            
\usepackage{booktabs}       
\usepackage{longtable}      
\usepackage{tabularx}       
\usepackage{adjustbox}      
\usepackage{makecell}       
\usepackage{array}          
\usepackage{colortbl}       
\usepackage{fontawesome5}   
\usepackage{amsfonts}       
\usepackage{nicefrac}       
\usepackage{microtype}      
\usepackage{needspace}      
\usepackage{xcolor}         

\DeclareUnicodeCharacter{2705}{\textcolor{green!50!black}{\faIcon{check-circle}}}
\DeclareUnicodeCharacter{25D0}{\textcolor{orange!80!black}{\faIcon{adjust}}}
\DeclareUnicodeCharacter{26A0}{\textcolor{orange!90!black}{\faIcon{exclamation-triangle}}}
\DeclareUnicodeCharacter{FE0F}{}
\definecolor{darkyellow}{RGB}{180,140,0}

\usepackage{mathrsfs}
\usepackage{mathtools}
\usepackage{graphicx}

\usepackage{subcaption}

\usepackage{cancel}
\usepackage{multirow}
\usepackage{tikz}
\usetikzlibrary{arrows.meta,positioning,fit,backgrounds,calc}
\definecolor{paperViolet}{HTML}{6B5ECF}
\definecolor{paperOrange}{HTML}{D95F02}
\definecolor{paperTeal}{HTML}{527D79}
\definecolor{paperTealMist}{HTML}{DFF3EC}
\definecolor{paperIvory}{HTML}{FFF2D9}
\colorlet{exammain}{paperViolet}
\colorlet{examhard}{paperOrange}
\definecolor{paperSeparator}{HTML}{8C9994}
\DeclareRobustCommand{\markclosed}{\textcolor{paperOrange}{$\bullet$}}
\DeclareRobustCommand{\markopen}{\textcolor{paperViolet}{\raisebox{0.12ex}{\scriptsize$\blacktriangle$}}}
\newcommand{\examMain}[1][Realistic]{\textcolor{exammain}{#1}}
\newcommand{\examHard}[1][Hard]{\textcolor{examhard}{#1}}
\colorlet{hbFullColor}{green!55!black}
\colorlet{hbHalfColor}{darkyellow}
\colorlet{hbQuarterColor}{orange!90!black}
\colorlet{hbNoneColor}{red!80!black}
\newlength{\hbr}
\DeclareRobustCommand{\hbFull}{\tikz[baseline=-0.55ex]{\fill[hbFullColor] (0,0) circle[radius=\hbr];}}
\DeclareRobustCommand{\hbHalf}{\tikz[baseline=-0.55ex]{\draw[hbHalfColor,line width=0.5pt] (0,0) circle[radius=\hbr];\fill[hbHalfColor] (0,0) -- (90:\hbr) arc[start angle=90,end angle=-90,radius=\hbr] -- cycle;}}
\DeclareRobustCommand{\hbQuarter}{\tikz[baseline=-0.55ex]{\draw[hbQuarterColor,line width=0.5pt] (0,0) circle[radius=\hbr];\fill[hbQuarterColor] (0,0) -- (90:\hbr) arc[start angle=90,end angle=0,radius=\hbr] -- cycle;}}
\DeclareRobustCommand{\hbNone}{\tikz[baseline=-0.55ex]{\draw[hbNoneColor,line width=0.5pt] (0,0) circle[radius=\hbr];}}
\usepackage{listings}
\usepackage{algorithm}
\usepackage{algorithmic}
\renewcommand{\algorithmicrequire}{\textbf{Input:}}
\renewcommand{\algorithmicensure}{\textbf{Output:}}
\usepackage{caption}
\usepackage{hyperref}       
\makeatletter
\renewcommand*{\pdfannot@link@off@@}{\pdfrunninglinkoff}
\renewcommand*{\pdfannot@link@on@@}{\pdfrunninglinkon}
\let\iclr@footnoterule\footnoterule
\def\footnoterule{\pdfrunninglinkoff\iclr@footnoterule}
\makeatother

\usepackage{xr-hyper}

 \usepackage[disable,textsize=small]{todonotes}

\usepackage{macros}

\lstdefinestyle{causaldsprompt}{
    basicstyle=\ttfamily\footnotesize,
    breakatwhitespace=true,
    breaklines=true,
    breakindent=1.5em,
    breakautoindent=true,
    postbreak=\mbox{\textcolor{gray}{$\hookrightarrow$}\space},
    columns=fullflexible,
    keepspaces=true,
    showstringspaces=false,
    gobble=8,
    xleftmargin=0pt,
    aboveskip=0.35\baselineskip,
    belowskip=0pt
}
\lstnewenvironment{promptblock}{\lstset{style=causaldsprompt}}{}

\title{CausalDS: Benchmarking Causal Reasoning in Data-Science Agents}

\author{%
    Andrej Leban\thanks{Correspondence: \texttt{leban@umich.edu}. Code: \href{https://github.com/andleb/CausalDS}{\nolinkurl{github.com/andleb/CausalDS}}.} \\
    Department of Statistics \\
    University of Michigan
    \And
    Theo Saarinen \\
    Department of Statistics \\
    University of California, Berkeley \\
    and AIA Labs, Bridgewater
    \And
    Yuekai Sun \\
    Department of Statistics \\
    University of Michigan
}

\iclrfinalcopy

\begin{document}

    \maketitle\vspace{-0.15in}
    \lhead{Preprint}

    \begin{abstract}
        Large language models (LLMs) increasingly act as \emph{data-science agents}, combining abstract
        reasoning with advanced tool use. Yet the relevant benchmark landscape largely divides into symbolic causal
        reasoning benchmarks without realistic data analysis \textit{or} data analysis benchmarks without
        causal data-generating structure or a range of causal data science tasks.
        Existing causal evaluation datasets are often restricted to curated examples from
        existing sources, with diversity coming from limited templatized variations rather than from systematic
        generation of novel synthetic causal structures.
        We introduce \textbf{CausalDS}, a benchmark for evaluating \emph{causal reasoning in agentic data-science
        workflows}. Each benchmark instance is a \emph{scene} consisting of a sampled structural causal model (SCM) with
        generated observational data and an accompanying synthetic natural-language story grounded in a real
        domain.
        We ground the \textit{composition} of the benchmark components in empirical distributions
        obtained from real-world datasets, thus retaining empirical structure while reducing the memorization risk through
        completely synthetic generation.
        From each scene, we then derive tasks spanning all three of Pearl's rungs. Most tasks include a data science coding component, where the model typically needs
        to use several tools to arrive at the final answer due to imperfect observations. Abstention is tested directly: for targets the graph does not identify, the graded answer is to abstain. The benchmark thus jointly evaluates symbolic causal reasoning,
        data science, uncertainty quantification, abstention, and tool use/coding. Because the generator is fully synthetic, the exam is not a fixed dataset: composition, difficulty, and size are parameters, and fresh exams can be drawn at will.
    \end{abstract}

    \section{Introduction}
    \label{sec:intro}

    Modern LLMs are routinely used in data-science workflows
    \citep{jing_dsbench_2025,chan_mle-bench_2025,gu_blade_2024,majumder_discoverybench_2025}, yet their actual \emph{causal} reasoning capabilities remain contentious \citep{zecevic_causal_2023,jin_cladder_2023, yu_caliper_2026}.
    In realistic causal data science, the relevant task is not merely to answer a causal question in text.
    An analyst must interpret a domain description, reason about the implied causal structure, inspect observational data,
    decide what is identifiable, and then either estimate the target quantity or decline to answer when the available
    information is insufficient.

    \textbf{CausalDS} evaluates this setting along five axes that are usually tested separately. The first is symbolic causal reasoning: interpreting a causal scenario, reasoning over graph structure, and distinguishing associational, interventional, and counterfactual targets. The second is data-science execution: using tabular data and  analysis tools to produce estimates and predictions. The third is uncertainty quantification: attaching calibrated uncertainty to those estimates. The fourth is epistemic abstention: recognizing when the requested causal claim is not warranted by the data and assumptions. The fifth is tool use and coding: carrying the analysis out through code. These axes are separable in principle, but realistic causal analysis requires their interaction.

    Existing evaluations tend to isolate parts of this problem. Data-science agent benchmarks stress coding and open-ended analysis but usually lack a hidden causal data-generating structure \citep{lai_ds-1000_2022,jing_dsbench_2025,chan_mle-bench_2025}. Causal reasoning benchmarks test graph reasoning, intervention, or counterfactual logic, but often do so on a purely symbolic level \citep{jin_cladder_2023,sheth_causalgraph2llm_2024,chen_clear_2024}.
    Benchmarks built from published datasets or familiar causal examples add realism,
    but this can make it difficult to guarantee novelty and avoid contamination: models may rely on \emph{amortized causal inference} rather than genuine structure-sensitive reasoning ---
    the ``causal parrot'' failure mode \citep{jin_cladder_2023,zecevic_causal_2023}.

    CausalDS, on the other hand, addresses this gap by generating synthetic causal data-science scenes with private SCM-derived ground truth, natural-language problem descriptions, tabular data, and deterministic scoring.
    \paragraph{Main Contributions}
    \begin{itemize}
        \item \textbf{SCM-grounded synthetic scenes with empirically anchored composition.}
        The problem settings are fully synthetic, and their composition can be empirically anchored along axes such as variable type, graph structure, and identifiability.
        \item \textbf{Graph-faithful free-form verbalization.}
        CausalDS generates fully synthetic natural-language stories audited against the generating causal graph.
        \item \textbf{A separate observation layer for data-analysis difficulty.}
        The data-science difficulty can be varied while the causal estimand and its identifiability remain fixed.
        \item \textbf{A broad causal data-science task suite.}
        Tasks span Pearl's hierarchy and include multiple output variants per task.
        \item \textbf{Abstention-aware deterministic evaluation.}
        Deliberately non-identifiable causal queries where the correct behavior is to abstain are treated as a first-class scored outcome.
    \end{itemize}

    We evaluate contemporary agents on both an exam compositionally grounded in real-world corpora and a targeted hard exam, and show that model performance dissociates along the five axes above.

    \section{Background}
    \label{sec:background}

    \paragraph{Causal graphs, SCMs, and Pearl's hierarchy.}
    A structural causal model (SCM) over a directed acyclic graph (DAG) $G$ on variables $V = (V_1, \dots, V_n)$ assigns each
    node a \textit{structural equation } $V_i = f_i(\mathrm{Pa}_i, U_i)$ from its graph parents $\mathrm{Pa}_i$ and an exogenous
    disturbance $U_i$; the joint distribution of the $\{U_i\}$ together with the mechanisms $\{f_i\}$ induces the
    observational distribution $P(V)$ \citep{pearl_causal_2021}. The same SCM
    defines interventional distributions $P(V \mid \mathrm{do}(X=x))$ obtained by replacing $f_X$ with the constant $X =
    x$, as well as counterfactual quantities that compare multiple ``parallel worlds'' sharing the same exogenous draw
    of $U$. We use Pearl's three-rung hierarchy as the organizing axis of the task suite: \emph{Rung~1} (associational
    -- $P(V)$, which includes usual data science tasks such as prediction), \emph{Rung~2} (interventional, $P(Y\mid
    \mathrm{do}(X))$), and \emph{Rung~3} (counterfactual).

    \paragraph{Effects and identifiability.}
    A causal \textit{effect} is \emph{identifiable} from $P(V)$ given $G$ when the corresponding interventional functional can be
    expressed in observational terms alone. Pearl's do-calculus together with the ID algorithm of
    \citet{shpitser_complete_2008} gives a complete decision procedure; classical sufficient conditions include the
    backdoor criterion and the front-door criterion \citep{pearl_causal_2021}.
    At Rung~2 the target is the \emph{population} average treatment effect (ATE), $\mathbb{E}[Y\mid\mathrm{do}(X{=}x_1)] -
    \mathbb{E}[Y\mid\mathrm{do}(X{=}x_0)]$, and effect-estimation tasks are \emph{identification-gated}: the agent must first
    decide whether this population estimand is identifiable from the conceptual observational law under the story-implied
    graph, and abstain when it is not---even though a finite observational sample is always provided.

    Counterfactual reasoning (Rung 3) yields estimands that Rung 2 effects cannot express, including the \emph{effect of
    treatment on the treated} (ETT) $\mathbb{E}[Y_{x_1} - Y_{x_0} \mid X=x_1]$ and Pearl's \emph{natural direct} and
    \emph{natural indirect} effects (NDE, NIE) for mediation analysis \citep{pearl_causal_2021,pearl_causality_2022}.
    Identifiability for these estimands is governed by the \textit{ID*/IDC*} algorithms
    \citep{shpitser_complete_2008}. In particular: a graph for which a Rung 2 effect, e.g. the ATE, is identifiable need not yield an identifiable ETT, NDE, or NIE. CausalDS therefore tracks Rung-3 identifiability separately from Rung-2 identifiability.
    CausalDS assesses identifiability on conceptual variables, as opposed to their noisy observations, which are discussed in Sec.~\ref{sec:observation_layer}.

    \section{The CausalDS Benchmark}
    \label{sec:benchmark}

\begin{figure}[ht!]
        \centering
        \includegraphics[width=.65\textwidth]{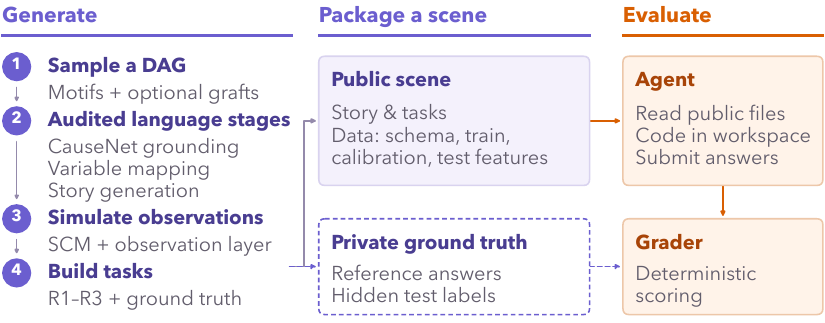}
        \caption{\textbf{CausalDS pipeline.}}
        \label{fig:pipeline}
    \end{figure}

    A CausalDS dataset instance is a \emph{scene} (Fig.~\ref{fig:running_example}): a narrative story, a tabular dataset, a lightweight data schema, and a list of tasks.
    Fig.~\ref{fig:pipeline} sketches the generation pipeline; Sec.~\ref{sec:graph_data} to Sec.~\ref{sec:scene_format} follow its stages.
    A composition configuration sets the distribution over motif, graft complexity, identifiability regime,
    treatment and
    outcome type, SCM profile, and observation variant; for the
    realistic exam these follow empirical corpora (App.~\ref{app:composition_grounding}). For benchmarking, the agent receives only the public scene directory
    and writes answer files; the grader separately loads the private artifacts and scores deterministically.

    \begin{figure}[ht!]
        \centering
        \input{figs/scene_example.tex}
        \caption{\textbf{A CausalDS scene.}}
        \label{fig:running_example}
    \end{figure}

    \subsection{Causal graphs}
    \label{sec:graph_data}

    We sample DAGs starting from canonical \textit{motifs} (chain, fork, confounding, and others; the full
    catalog is reproduced in App.~\ref{app:motif_catalog}, Fig.~\ref{fig:appendix_motif_catalog}). To assemble
    larger, dynamic graphs, we additionally support
    \emph{anchor-based grafting}: after drawing and verbalizing a coherent variable mapping for the  \textit{main graph}, we attach one or more small \textit{auxiliary motifs} through exactly one shared node --- the anchor --- each. Grafting builds larger graphs while keeping the mapping problem local: each fragment is named around an already-verbalized anchor.
    Some motifs include latent variables (e.g., unobserved confounders), enabling stress tests for identifiability and adjustment reasoning.

    \subsection{Mapping variables and story generation}
    \label{sec:verbalization}
    Abstract node identifiers are mapped to domain-relevant variable names (e.g., \texttt{v1} $\to$ \texttt{Exercise})
    using an LLM. 
    Optionally, a knowledge-grounding step seeds names for a subset of nodes using CauseNet
    \citep{heindorf_causenet_2020}, inducing the remaining nodes to conform to them, which carries real-world diversity into the whole verbalization.
    Because this step injects recognizable real-world causal pairs into the prose, we verify that answers do not
    hinge on the particular pairs seeded: a matched verbalization-swap ablation that varies only the seeded story and variable names shows stability (App.~\ref{app:causenet_ablation}).

    A mapper--auditor loop checks variable names against the graph, and a second audit checks the generated story.
    App.~\ref{app:scene_subalgorithms} gives the control flow and subroutines (Alg.~\ref{alg:scene-synthesis});
    production settings and audit details are in App.~\ref{app:audit_details}, and generation prompts in App.~\ref{app:pipeline_prompts}.

    \subsection{SCM-generated tabular data}
    \label{sec:scm_data}

    Given the resulting DAG, we instantiate a structural causal model (SCM) supporting a mix of continuous and binary
    variables, with mechanisms organized into typed \emph{SCM profiles}: each scene is assigned a
    continuous and a binary profile drawn from independent registries.
    Each registry mixes \emph{empirical profiles} whose mean-function families derive from real biochemical-equation
    and Boolean-rule corpora with \emph{synthetic profiles} for additional diversity.
    The registry, its mechanism families and noise options, and the deployed mixture
    weights are listed in App.~\ref{app:scm_profiles}.

    \subsection{Observation layer}
    \label{sec:observation_layer}
    A central design choice in CausalDS is to distinguish \emph{causal structure} from \emph{measurement}.
    After the conceptual DAG is fixed, the public data table may withhold a small subset of conceptual variables (typically
    causally important nodes such as treatments, outcomes, mediators, or confounders) and replace each with a bundle of
    noisy measurements. Measured variables stay in the narrated graph and are observed directly only in a small \textit{calibration} subset of the training rows; latent SCM nodes are never observed.

    Formally, for a selected $Z_j$ the public table contains $W_j = (W_{j1}, \dots, W_{jd_j})$, with continuous measurements
    $W_{jr} = h_{jr}(Z_j) + \epsilon_{jr}$ and binary measurements $W_{jr} \mid Z_j \sim \mathrm{Bernoulli}(p_{jr}(Z_j))$.
    The links $h_{jr}$ and $p_{jr}$ and the noise families are drawn from the same mechanism families as the SCM
    registry (App.~\ref{app:scm_profiles}, App.~\ref{app:obsmodel_details}).
    Causal identifiability is always evaluated on the conceptual graph: the observation layer is a calibrated \emph{measurement} layer that changes how hard it is to \emph{numerically estimate} an identified functional. Each observation model bundle has a single conceptual parent, so the observation model never creates or removes a confounder, instrument, or mediator, and never flips an identifiability label (this is not proximal causal inference; App.~\ref{app:obsmodel_details}).

    Four observation variants are released: \texttt{clean} (no measurement layer), \texttt{proxy}, \texttt{proxy\_hard} (more measured non-outcome nodes, stronger corruption, smaller calibration subset), and, for the hard exam, \texttt{proxy\_severe}. Candidate measurement bundles are screened for informativeness via a Fisher information criterion. App.~\ref{app:obsmodel_details} gives the screening criteria, private recoverability diagnostics, and the comparison with proximal causal inference.

    \subsection{Task suite}
    \label{sec:tasks}
    Tasks span all three rungs of Pearl's hierarchy, with each task specified by a question family and output variant
    (App.~\ref{app:scoring_rules}, Tab.~\ref{tab:task_families}).
    Rung-2 effect and identification tasks are gated on \emph{population-ATE} identifiability, decided with DoWhy
    \citep{sharma_dowhy_2020}. Rung-3 identifiability is decided separately with the \texttt{y0} \textit{ID*/IDC*}
    algorithms\footnote{\href{https://github.com/y0-causal-inference/y0}{\nolinkurl{github.com/y0-causal-inference/y0}}}.

    \subsection{Scene format}
    \label{sec:scene_format}

    Public scene artifacts comprise the \textit{story}, relevant dataset \textit{parquet} files, and the dataset schema.
    For scenes with a measurement observation model, \texttt{calibration.parquet} contains a \textit{calibration subset}:
    rows selected from \texttt{data.parquet} and augmented with gold values for the conceptual variables represented by
    measurement bundles.
    Private scoring artifacts include the ground truth, the observation model diagnostics, and, when
    needed, hidden test set labels (for \textit{Kaggle}-style prediction tasks).
    For the results presented, we constructed a dataset with 953 scenes.

    \subsection{Evaluation harness}
    \label{sec:harness}

    We evaluate models using \textit{mini-swe-agent} \citep{yang2024sweagent}, a lightweight agent harness in which the
    model interacts with the environment by emitting \texttt{bash} shell commands.
    All tool execution happens inside a sandboxed Docker (or similar)
    container with network access disabled, and each task is run as a fresh agent conversation.
    The container ships the standard data-science stack.
    We log efficiency signals (tool-call counts and API-reported token usage) to support joint evaluation of reasoning quality and tool use. Technical details
    (containerization, shell processes, and available packages) are given in App.~\ref{app:harness_ops}.

    \subsection{Scoring}
    \label{sec:scoring}

    Grading is fully deterministic: for each task, the grader checks that the required answer file is present and
    parseable, then computes a task-specific metric from the answer and the private ground truth.
    For non-identifiable task instances, task schemas include
    explicit abstain targets (e.g., \texttt{null} or \texttt{unknown}), allowing the benchmark to quantify a model's
    propensity to hallucinate when no answer is available.
    Each task emits one atomic score with a metric name; the per-metric formulas (RMSE, ROC-AUC, Brier, \ldots) and the task-specific scoring rules by rung are listed in App.~\ref{app:scoring_rules}
    (Tabs.~\ref{tab:scoring_r1}--\ref{tab:scoring_r3}).
    For every variant where the answer might not be identifiable,
    the grader uses \textit{mutually exclusive routing}: a task receives a binary score
    whenever \textit{either side} abstains --- i.e., the ground truth is non-identifiable, \textit{or} the model wrongly abstains; otherwise it is graded by its content metric.

    To summarize models for a leaderboard, we introduce three pooled quality categories and two single-number composites.
    Let $D$ be the discrete-task slice; let
    $T_{\mathrm{NR}}$ be the set of continuous-output tasks; let $T_{F_1}$ be the set of $F_1$-graded tasks.
    Define the per-task normalized error (Eq.~\ref{eq:nrel_error}):
        {\small
    \begin{equation}
        \mathrm{NRelErr}_i = \frac{\sqrt{n_i^{-1}\sum_j(\theta_{ij}-\hat\theta_{ij})^2}}{1+s_i},
        \label{eq:nrel_error}
    \end{equation}
    }
    where $n_i$ counts target values and $s_i$ is a scale defined by task type
    (App.~\ref{app:scoring_rules}, Tab.~\ref{tab:scoring_base_metrics}).
    The headline columns are then
    defined in Eq.~\ref{eq:headline_columns}:
        {\small
    \begin{equation}
        \adjustbox{max width=\dimexpr\linewidth-4em\relax}{$\displaystyle
            \mathrm{PassRate} = |D|^{-1}\sum_{i\in D}\mathbf{1}[\hat{y}_i=y_i],\quad
            \mathrm{Med.\ NRel.\ Err} = \mathop{\mathrm{med}}\limits_{i\in T_{\mathrm{NR}}}\mathrm{NRelErr}_i,\quad
            \mathrm{Med.\ F_1\text{-}Loss} = \mathop{\mathrm{med}}\limits_{i\in T_{F_1}}(1-F_{1,i}).
        $}
        \label{eq:headline_columns}
    \end{equation}
    }
    Pass Rate pools both actual binary content and the abstention binaries; the medians keep the typical-task headline robust to the long-tailed outliers. We partition $T_{\mathrm{NR}} = T_{\mathrm{NR}}^{\mathrm{pt}} \cup T_{\mathrm{NR}}^{\mathrm{int}}$ into point-graded tasks and interval-graded tasks
    with the aggregator $S_{\mathrm{NR}}$ in Eq.~\ref{eq:nrel_aggregator}:
        {\small
    \begin{equation}
        \adjustbox{max width=\dimexpr\linewidth-4em\relax}{$\displaystyle
            S_{\mathrm{NR}} =
            \frac{|T_{\mathrm{NR}}^{\mathrm{pt}}|m_{\mathrm{pt}}
                + |T_{\mathrm{NR}}^{\mathrm{int}}|m_{\mathrm{int}}}{|T_{\mathrm{NR}}|},\quad
            m_{\mathrm{pt}}=\mathop{\mathrm{med}}\limits_{i \in T_{\mathrm{NR}}^{\mathrm{pt}}}\mathrm{NRelErr}_i,\quad
            m_{\mathrm{int}}=\mathop{\mathrm{mean}}\limits_{i \in T_{\mathrm{NR}}^{\mathrm{int}}}\min(\mathrm{NRelIS}_i,c),
        $}
        \label{eq:nrel_aggregator}
    \end{equation}
    }
    with $c = 10$, using a median for robustness to extreme point errors and a \textbf{capped mean} to retain interval-miss penalties while limiting extreme contributions (App.~\ref{app:scoring_rules}).
    For an absolute single-number summary we define the \textbf{CausalDS} score (Eq.~\ref{eq:causalds_score}):
        {\small
    \begin{equation}
        \mathrm{CausalDSScore}
        = w_p(1-\mathrm{PassRate}) + w_rS_{\mathrm{NR}} + w_f\,\mathrm{Med.\ F_1\text{-}Loss},
        \label{eq:causalds_score}
    \end{equation}
    }
    with $w_k=N_k/(N_p+N_r+N_f)$ for $k\in\{p,r,f\}$, where $N_p=|D|, N_r=\left|T_{\mathrm{NR}}\right|$, and $N_f=\left|T_{F_1}\right|$ are the per-pool task counts; \textit{lower is better}.
    CausalDSScore is comparable across exams but is dominated, by construction, by whichever pool is largest unless re-weighted.
    As a complementary, exam-relative summary we compute the \textbf{Composite Rank}: per-column average ranks normalized to $[0,1]$, then averaged, computed on the same three metrics as CausalDSScore (Pass Rate, $S_{\mathrm{NR}}$, and Med.\ $F_1$-Loss). It is comparable only within one exam; however, it is not sensitive to large numeric outliers.

    Missing or malformed outputs are handled by metric family:
    for both \textit{bounded} metrics (binary and $F_1$ scores), missing answers \textit{count as failures}. For \textit{unbounded} outputs, we avoid assigning an arbitrary numeric
    loss to missing values; instead, NRel.\ metrics are computed on valid numeric submissions and we report the fraction of continuous-style tasks with a valid answer.
    While a single-number metric can be attractive as a rough orientation, it is not the final word. For this reason, in Sec.~\ref{sec:results} we report both the intermediate (the three components of the CausalDS score) and atomic metrics, whenever appropriate. We also report \emph{Tok./task} as the mean API-reported prompt-plus-completion tokens summed over all calls per task,
    including re-sent context (App.~\ref{app:cost_efficiency}).

    \section{Main Results}
    \label{sec:results}
    \begin{table}[ht!]
        \centering
        \caption{\textbf{Realistic-exam leaderboard (100 tasks, three runs).} Three-run mean with SD beneath; CausalDSScore (Eq.~\ref{eq:causalds_score}) shows mean with best--worst run. Comp.\ rank: Composite Rank (Sec.~\ref{sec:scoring}). Pass Rate pools the content and abstention binaries; brackets: per-run pool size. Any-of-3 / all-of-3: share of the fixed ground-truth binary pool ($n{=}34$) passed in at least one / all three runs. $S_{\mathrm{NR}}$: Eq.~\ref{eq:nrel_aggregator}. Valid cont.: valid continuous answers, mean / total, tinted by within-exam rank third (green / orange / red). Rows sorted by mean score; best per column in bold. \textcolor{red!65!black}{$^\dagger$} provider-side safety refusals. \markclosed\ closed, \markopen\ open-weight.}
        \label{tab:realistic_leaderboard}
        \begingroup
\DeclareFontFamily{T1}{LeaderboardSans}{}
\DeclareFontShape{T1}{LeaderboardSans}{m}{n}{<-> s*[.96] phvr8t}{}
\DeclareFontShape{T1}{LeaderboardSans}{m}{it}{<-> s*[.96] phvro8t}{}
\DeclareFontShape{T1}{LeaderboardSans}{b}{n}{<-> s*[.96] phvb8t}{}
\DeclareFontShape{T1}{LeaderboardSans}{b}{it}{<-> s*[.96] phvbo8t}{}
\DeclareFontShape{T1}{LeaderboardSans}{bx}{n}{<-> ssub * LeaderboardSans/b/n}{}
\DeclareFontShape{T1}{LeaderboardSans}{bx}{it}{<-> ssub * LeaderboardSans/b/it}{}
\renewcommand{\sfdefault}{LeaderboardSans}
\definecolor{LBInk}{HTML}{393340}
\definecolor{LBMuted}{HTML}{625D6C}
\definecolor{LBRule}{HTML}{DCE3DF}
\definecolor{LBValidTop}{HTML}{28734F}
\colorlet{LBValidMiddle}{paperOrange!80!black}
\definecolor{LBValidBottom}{HTML}{AE3E43}
\newcommand{\LBPrimary}{\sffamily\fontsize{7.8}{9.2}\selectfont}
\newcommand{\LBDetail}{\sffamily\fontsize{6.3}{7.2}\selectfont}
\newcommand{\LBBest}[1]{{\bfseries #1}}
\newcommand{\LBDagger}{\kern.4pt\raisebox{.35ex}{\textcolor{LBValidBottom}{\fontfamily{phv}\bfseries\fontsize{7.5}{7.5}\selectfont\textdagger}}}
\newcommand{\LBStack}[2]{%
  \begin{tabular}[c]{@{}c@{}}\LBPrimary #1\\[-1.5pt]
    \LBDetail\color{LBMuted}#2\end{tabular}%
}
\newcommand{\LBReliability}[2]{%
  \begin{tabular}[c]{@{}c@{}}\LBPrimary #1\\[-1.5pt]
    \LBPrimary #2\end{tabular}%
}
\newcommand{\LBModel}[2]{%
  \makebox[11pt][l]{\LBDetail\color{LBMuted}#1}%
  {\sffamily\fontsize{7.4}{9}\selectfont #2}%
}
\newcommand{\LBHeader}[2]{%
  \begin{tabular}[c]{@{}c@{}}\sffamily\bfseries\fontsize{7}{8.5}\selectfont #1\\[-1pt]
    \sffamily\fontsize{6.1}{7.5}\selectfont\color{LBMuted}#2\end{tabular}%
}
\newcommand{\LBValidity}[3]{%
  \cellcolor{#1!6}%
  \begin{tabular}[c]{@{}c@{}}\LBPrimary\color{LBInk}#2\\[-1.5pt]
    \LBDetail\color{LBMuted}#3\end{tabular}%
}
\makeatletter
\let\LBSavedRuleColor\CT@arc@
\newcommand{\LBFinish}{\global\let\CT@arc@\LBSavedRuleColor}
\makeatother
\newcommand{\LBTableSetup}{%
  \LBPrimary\color{LBInk}%
  \setlength{\tabcolsep}{0pt}\renewcommand{\arraystretch}{1}%
  \arrayrulecolor{LBRule}%
  \setlength{\aboverulesep}{2pt}\setlength{\belowrulesep}{2pt}%
}

        \colorlet{LBAccent}{paperViolet}
\LBTableSetup
\begin{tabular}{@{}>{\raggedright\arraybackslash}m{.22\linewidth} >{\columncolor{LBAccent!6}}>{\centering\arraybackslash}m{.148\linewidth}
  >{\centering\arraybackslash}m{.064\linewidth} >{\centering\arraybackslash}m{.108\linewidth} >{\centering\arraybackslash}m{.094\linewidth}
  >{\centering\arraybackslash}m{.088\linewidth} >{\centering\arraybackslash}m{.086\linewidth} >{\centering\arraybackslash}m{.096\linewidth} >{\centering\arraybackslash}m{.096\linewidth}@{}}
  & \multicolumn{2}{c}{\color{LBAccent!80!black}\sffamily\bfseries\fontsize{7}{8}\selectfont Overall}
  & \multicolumn{2}{c}{\color{LBAccent!80!black}\sffamily\bfseries\fontsize{7}{8}\selectfont Binary tasks}
  & \multicolumn{3}{c}{\color{LBAccent!80!black}\sffamily\bfseries\fontsize{7}{8}\selectfont Numeric answers}
  & {\color{LBAccent!80!black}\sffamily\bfseries\fontsize{7}{8}\selectfont Effort}\\[2pt]
  \cmidrule(lr{3pt}){2-3}\cmidrule(lr{3pt}){4-5}\cmidrule(lr{3pt}){6-8}\cmidrule(l{3pt}){9-9}
  \LBHeader{Model}{score order}
  & \LBHeader{CausalDSScore $\downarrow$}{mean; best--worst}
  & \LBHeader{Comp.}{rank $\downarrow$}
  & \LBHeader{Pass Rate $\uparrow$}{\%\quad$\pm$ SD [$N$]}
  & \LBHeader{any-of-3 $\uparrow$}{all-of-3 $\uparrow$}
  & \LBHeader{Med.\ NRel.}{Err $\downarrow$}
  & \LBHeader{$S_{\mathrm{NR}}$ $\downarrow$}{mean; $\pm$ SD}
  & \LBHeader{Valid cont.}{answers $\uparrow$}
  & \LBHeader{Tokens $\downarrow$}{k / task}\\[4pt]
  \midrule
                  \LBModel{1}{\markclosed\,GPT-6 Astra} & \LBStack{\LBBest{0.2112}}{\LBBest{0.1645}\,--\,0.3018} & \LBBest{0.288} & \LBStack{\LBBest{96.1}}{$\pm$\LBBest{1.7}\,[34]} & \LBReliability{\LBBest{100.0}}{\LBBest{94.1}} & \LBStack{0.084}{$\pm$\,0.041} & \LBStack{0.507}{$\pm$\,0.186} & \LBValidity{LBValidTop}{\LBBest{39.0\,/\,39}}{} & \LBStack{\LBBest{5.1}}{$\pm$\,\LBBest{0.2}} \\[1.7pt]
                \LBModel{2}{\markopen\,GLM 5.3} & \LBStack{0.2682}{0.2573\,--\,\LBBest{0.2786}} & 0.348 & \LBStack{72.8}{$\pm$2.0\,[34--35]} & \LBReliability{82.4}{64.7} & \LBStack{\LBBest{0.080}}{$\pm$\,\LBBest{0.011}} & \LBStack{\LBBest{0.454}}{$\pm$\,\LBBest{0.047}} & \LBValidity{LBValidMiddle}{38.3\,/\,39}{$\pm$\,0.6} & \LBStack{235.9}{$\pm$\,33.7} \\[1.7pt]
                \LBModel{3}{\markclosed\,Gemini 3.1 Pro} & \LBStack{0.2754}{0.2255\,--\,0.3684} & 0.409 & \LBStack{78.7}{$\pm$3.2\,[34--35]} & \LBReliability{88.2}{67.6} & \LBStack{0.140}{$\pm$\,0.081} & \LBStack{0.525}{$\pm$\,0.197} & \LBValidity{LBValidMiddle}{38.3\,/\,39}{$\pm$\,0.6} & \LBStack{160.3}{$\pm$\,13.6} \\[1.7pt]
                \LBModel{4}{\markopen\,DeepSeek V4.1 Flash} & \LBStack{0.2805}{0.2750\,--\,0.2867} & 0.379 & \LBStack{73.5}{$\pm$2.9\,[34]} & \LBReliability{85.3}{64.7} & \LBStack{0.149}{$\pm$\,0.042} & \LBStack{0.489}{$\pm$\,0.035} & \LBValidity{LBValidTop}{\LBBest{39.0\,/\,39}}{} & \LBStack{247.4}{$\pm$\,30.6} \\[1.7pt]
                \LBModel{5}{\markclosed\,Claude Fable 5.1\LBDagger} & \LBStack{0.2940}{0.2784\,--\,0.3135} & 0.500 & \LBStack{68.6\LBDagger}{$\pm$2.9\,[35]} & \LBReliability{79.4\LBDagger}{61.8\LBDagger} & \LBStack{0.103}{$\pm$\,0.012} & \LBStack{0.501}{$\pm$\,0.048} & \LBValidity{LBValidBottom}{35.0\,/\,39\LBDagger}{$\pm$\,0.0} & \LBStack{25.0}{$\pm$\,2.4} \\[1.7pt]
                \LBModel{6}{\markopen\,Kimi K3} & \LBStack{0.2943}{0.2477\,--\,0.3211} & 0.470 & \LBStack{66.7}{$\pm$6.1\,[34]} & \LBReliability{82.4}{52.9} & \LBStack{0.099}{$\pm$\,0.013} & \LBStack{0.464}{$\pm$\,0.052} & \LBValidity{LBValidTop}{\LBBest{39.0\,/\,39}}{} & \LBStack{74.9}{$\pm$\,21.6} \\[1.7pt]
                \LBModel{7}{\markopen\,K2-MoVA 36B-A4B} & \LBStack{0.3003}{0.2481\,--\,0.3784} & 0.561 & \LBStack{72.1}{$\pm$4.8\,[34--35]} & \LBReliability{85.3}{58.8} & \LBStack{0.166}{$\pm$\,0.059} & \LBStack{0.527}{$\pm$\,0.162} & \LBValidity{LBValidMiddle}{38.3\,/\,39}{$\pm$\,0.6} & \LBStack{455.8}{$\pm$\,79.4} \\[1.7pt]
                \LBModel{8}{\markclosed\,Claude Opus 4.8} & \LBStack{0.3537}{0.2762\,--\,0.4151} & 0.409 & \LBStack{81.4}{$\pm$1.7\,[34]} & \LBReliability{88.2}{76.5} & \LBStack{0.186}{$\pm$\,0.033} & \LBStack{0.747}{$\pm$\,0.167} & \LBValidity{LBValidMiddle}{38.7\,/\,39}{$\pm$\,0.6} & \LBStack{21.5}{$\pm$\,3.3} \\[1.7pt]
                \LBModel{9}{\markopen\,Kimi K2.6} & \LBStack{0.3848}{0.2413\,--\,0.4760} & 0.561 & \LBStack{76.8}{$\pm$10.0\,[34--35]} & \LBReliability{91.2}{61.8} & \LBStack{0.186}{$\pm$\,0.038} & \LBStack{0.790}{$\pm$\,0.259} & \LBValidity{LBValidMiddle}{38.7\,/\,39}{$\pm$\,0.6} & \LBStack{439.1}{$\pm$\,149.6} \\[1.7pt]
                \LBModel{10}{\markopen\,Qwen 3.6 35B} & \LBStack{0.3925}{0.3431\,--\,0.4269} & 0.712 & \LBStack{66.5}{$\pm$3.8\,[34--38]} & \LBReliability{82.4}{55.9} & \LBStack{0.203}{$\pm$\,0.041} & \LBStack{0.764}{$\pm$\,0.167} & \LBValidity{LBValidBottom}{35.0\,/\,39}{$\pm$\,6.1} & \LBStack{200.7}{$\pm$\,56.3} \\[1.7pt]
                \LBModel{11}{\markclosed\,GPT-5.5} & \LBStack{0.5603}{0.4738\,--\,0.6498} & 0.530 & \LBStack{81.0}{$\pm$1.2\,[34--36]} & \LBReliability{94.1}{70.6} & \LBStack{0.223}{$\pm$\,0.042} & \LBStack{1.308}{$\pm$\,0.252} & \LBValidity{LBValidBottom}{37.7\,/\,39}{$\pm$\,0.6} & \LBStack{31.2}{$\pm$\,17.2} \\[1.7pt]
                \LBModel{12}{\markopen\,Gemma 4 26B} & \LBStack{0.6621}{0.6213\,--\,0.7274} & 0.833 & \LBStack{59.2}{$\pm$3.0\,[34--35]} & \LBReliability{79.4}{41.2} & \LBStack{0.267}{$\pm$\,0.041} & \LBStack{1.352}{$\pm$\,0.168} & \LBValidity{LBValidMiddle}{38.7\,/\,39}{$\pm$\,0.6} & \LBStack{32.7}{$\pm$\,1.0} \\[1.7pt]
  \bottomrule[.55pt]
\end{tabular}\par
        \LBFinish
        \endgroup
    \end{table}
    \begin{table}[ht!]
        \centering
        \caption{\textbf{Hard-exam leaderboard (200 tasks, three runs).} Columns, sorting, and tints as in Tab.~\ref{tab:realistic_leaderboard}; fixed ground-truth binary pool $n{=}97$.}
        \label{tab:hard_leaderboard}
        \begingroup
        
        \colorlet{LBAccent}{paperOrange}
\LBTableSetup
\begin{tabular}{@{}>{\raggedright\arraybackslash}m{.22\linewidth} >{\columncolor{LBAccent!6}}>{\centering\arraybackslash}m{.148\linewidth}
  >{\centering\arraybackslash}m{.064\linewidth} >{\centering\arraybackslash}m{.108\linewidth} >{\centering\arraybackslash}m{.094\linewidth}
  >{\centering\arraybackslash}m{.088\linewidth} >{\centering\arraybackslash}m{.086\linewidth} >{\centering\arraybackslash}m{.096\linewidth} >{\centering\arraybackslash}m{.096\linewidth}@{}}
  & \multicolumn{2}{c}{\color{LBAccent!80!black}\sffamily\bfseries\fontsize{7}{8}\selectfont Overall}
  & \multicolumn{2}{c}{\color{LBAccent!80!black}\sffamily\bfseries\fontsize{7}{8}\selectfont Binary tasks}
  & \multicolumn{3}{c}{\color{LBAccent!80!black}\sffamily\bfseries\fontsize{7}{8}\selectfont Numeric answers}
  & {\color{LBAccent!80!black}\sffamily\bfseries\fontsize{7}{8}\selectfont Effort}\\[2pt]
  \cmidrule(lr{3pt}){2-3}\cmidrule(lr{3pt}){4-5}\cmidrule(lr{3pt}){6-8}\cmidrule(l{3pt}){9-9}
  \LBHeader{Model}{score order}
  & \LBHeader{CausalDSScore $\downarrow$}{mean; best--worst}
  & \LBHeader{Comp.}{rank $\downarrow$}
  & \LBHeader{Pass Rate $\uparrow$}{\%\quad$\pm$ SD [$N$]}
  & \LBHeader{any-of-3 $\uparrow$}{all-of-3 $\uparrow$}
  & \LBHeader{Med.\ NRel.}{Err $\downarrow$}
  & \LBHeader{$S_{\mathrm{NR}}$ $\downarrow$}{mean; $\pm$ SD}
  & \LBHeader{Valid cont.}{answers $\uparrow$}
  & \LBHeader{Tokens $\downarrow$}{k / task}\\[4pt]
  \midrule
                  \LBModel{1}{\markopen\,GLM 5.3} & \LBStack{\LBBest{0.5591}}{\LBBest{0.4711}\,--\,\LBBest{0.6512}} & 0.288 & \LBStack{73.0}{$\pm$2.8\,[103--107]} & \LBReliability{89.7}{68.0} & \LBStack{\LBBest{0.121}}{$\pm$\,\LBBest{0.021}} & \LBStack{\LBBest{0.988}}{$\pm$\,\LBBest{0.238}} & \LBValidity{LBValidBottom}{82.7\,/\,93}{$\pm$\,1.2} & \LBStack{1006.9}{$\pm$\,121.7} \\[1.7pt]
                \LBModel{2}{\markclosed\,Claude Fable 5.1} & \LBStack{0.6270}{0.5750\,--\,0.6944} & \LBBest{0.227} & \LBStack{85.4}{$\pm$2.2\,[97--99]} & \LBReliability{90.7}{81.4} & \LBStack{0.125}{$\pm$\,0.009} & \LBStack{1.221}{$\pm$\,0.127} & \LBValidity{LBValidTop}{90.0\,/\,93}{$\pm$\,2.0} & \LBStack{79.0}{$\pm$\,2.6} \\[1.7pt]
                \LBModel{3}{\markopen\,DeepSeek V4.1 Flash} & \LBStack{0.6390}{0.5521\,--\,0.6920} & 0.288 & \LBStack{83.9}{$\pm$1.4\,[100--103]} & \LBReliability{92.8}{81.4} & \LBStack{0.140}{$\pm$\,0.012} & \LBStack{1.258}{$\pm$\,0.189} & \LBValidity{LBValidMiddle}{88.7\,/\,93}{$\pm$\,1.5} & \LBStack{610.2}{$\pm$\,74.6} \\[1.7pt]
                \LBModel{4}{\markclosed\,GPT-6 Astra} & \LBStack{0.7325}{0.6894\,--\,0.7595} & 0.258 & \LBStack{\LBBest{90.4}}{$\pm$\LBBest{0.6}\,[97]} & \LBReliability{\LBBest{94.8}}{\LBBest{86.6}} & \LBStack{0.155}{$\pm$\,0.014} & \LBStack{1.475}{$\pm$\,0.087} & \LBValidity{LBValidTop}{\LBBest{93.0\,/\,93}}{} & \LBStack{\LBBest{16.3}}{$\pm$\,\LBBest{4.4}} \\[1.7pt]
                \LBModel{5}{\markopen\,Kimi K2.6} & \LBStack{0.9496}{0.8021\,--\,1.0387} & 0.439 & \LBStack{78.5}{$\pm$2.9\,[100--101]} & \LBReliability{88.7}{73.2} & \LBStack{0.213}{$\pm$\,0.023} & \LBStack{1.886}{$\pm$\,0.269} & \LBValidity{LBValidMiddle}{89.0\,/\,93}{$\pm$\,1.0} & \LBStack{1119.5}{$\pm$\,31.7} \\[1.7pt]
                \LBModel{6}{\markclosed\,Gemini 3.1 Pro} & \LBStack{0.9631}{0.8989\,--\,1.0538} & 0.530 & \LBStack{64.5}{$\pm$2.3\,[98--100]} & \LBReliability{73.2}{57.7} & \LBStack{0.232}{$\pm$\,0.047} & \LBStack{1.727}{$\pm$\,0.173} & \LBValidity{LBValidTop}{91.3\,/\,93}{$\pm$\,1.2} & \LBStack{440.2}{$\pm$\,29.1} \\[1.7pt]
                \LBModel{7}{\markopen\,Kimi K3} & \LBStack{0.9663}{0.9382\,--\,1.0075} & 0.591 & \LBStack{64.0}{$\pm$2.0\,[98--100]} & \LBReliability{75.3}{54.6} & \LBStack{0.221}{$\pm$\,0.061} & \LBStack{1.728}{$\pm$\,0.069} & \LBValidity{LBValidTop}{91.0\,/\,93}{$\pm$\,1.0} & \LBStack{391.7}{$\pm$\,66.7} \\[1.7pt]
                \LBModel{8}{\markopen\,K2-MoVA 36B-A4B} & \LBStack{0.9669}{0.9202\,--\,1.0184} & 0.530 & \LBStack{71.7}{$\pm$3.7\,[106--108]} & \LBReliability{88.7}{69.1} & \LBStack{0.172}{$\pm$\,0.022} & \LBStack{1.972}{$\pm$\,0.106} & \LBValidity{LBValidBottom}{82.0\,/\,93}{$\pm$\,1.0} & \LBStack{1615.7}{$\pm$\,47.5} \\[1.7pt]
                \LBModel{9}{\markclosed\,Claude Opus 4.8} & \LBStack{1.1364}{1.0117\,--\,1.2484} & 0.621 & \LBStack{65.2}{$\pm$1.1\,[102--108]} & \LBReliability{78.4}{61.9} & \LBStack{0.275}{$\pm$\,0.034} & \LBStack{2.237}{$\pm$\,0.334} & \LBValidity{LBValidMiddle}{85.7\,/\,93}{$\pm$\,3.2} & \LBStack{55.1}{$\pm$\,2.6} \\[1.7pt]
                \LBModel{10}{\markopen\,Qwen 3.6 35B} & \LBStack{1.3367}{1.3077\,--\,1.3831} & 0.742 & \LBStack{57.6}{$\pm$3.2\,[106--109]} & \LBReliability{77.3}{47.4} & \LBStack{0.375}{$\pm$\,0.055} & \LBStack{2.640}{$\pm$\,0.101} & \LBValidity{LBValidMiddle}{84.0\,/\,93}{$\pm$\,1.7} & \LBStack{452.3}{$\pm$\,15.6} \\[1.7pt]
                \LBModel{11}{\markclosed\,GPT-5.5} & \LBStack{1.4157}{1.3160\,--\,1.5117} & 0.652 & \LBStack{69.0}{$\pm$2.9\,[108--113]} & \LBReliability{90.7}{63.9} & \LBStack{0.437}{$\pm$\,0.057} & \LBStack{3.171}{$\pm$\,0.138} & \LBValidity{LBValidBottom}{77.7\,/\,93}{$\pm$\,2.1} & \LBStack{64.2}{$\pm$\,16.8} \\[1.7pt]
                \LBModel{12}{\markopen\,Gemma 4 26B} & \LBStack{1.5941}{1.4995\,--\,1.7550} & 0.833 & \LBStack{47.4}{$\pm$0.6\,[106--112]} & \LBReliability{75.3}{32.0} & \LBStack{1.022}{$\pm$\,0.122} & \LBStack{3.249}{$\pm$\,0.388} & \LBValidity{LBValidBottom}{80.3\,/\,93}{$\pm$\,2.3} & \LBStack{103.2}{$\pm$\,5.5} \\[1.7pt]
  \bottomrule[.55pt]
\end{tabular}\par
        \LBFinish
        \endgroup
    \end{table}
    \paragraph{Exams and models.}
    We report two exams (compositions in App.~\ref{app:exam_results}). The \emph{realistic exam} (100 tasks) is drawn from a $953$-scene pool with composition weights from the empirical distributions of App.~\ref{app:composition_grounding}. The \emph{hard exam} (200 tasks) fixes identifiability and graph-structure strata, concentrates on identification, bias diagnostics, and R2/R3 effect estimation, and places nearly all data-backed tasks in the two hardest observation tiers, including a \texttt{proxy\_severe} tier built for it.%
     Before sampling the hard exam (and unlike the realistic exam), we exclude candidate scenes that trigger a safety refusal in a single Claude Fable 5.1 screen pass.
    We evaluate twelve agents, five closed and seven open-weight, at high reasoning effort where available, with three complete replications of every exam (App.~\ref{app:harness_ops}).
    The leaderboards use two binary pools. Pass Rate follows the grader's routing, in which abstaining on an identifiable target is a failure, so its pool varies with the model's answers; any-of-3 / all-of-3 use the fixed ground-truth pool ($n{=}34$ realistic, $97$ hard) and are comparable across models (Sec.~\ref{sec:scoring}).

    \paragraph{Headline results.}
    Tables~\ref{tab:realistic_leaderboard} and~\ref{tab:hard_leaderboard} rank by CausalDSScore (Eq.~\ref{eq:causalds_score}). On the realistic exam GPT-6 Astra leads, is best on every pass-rate and reliability column, and spends the fewest tokens; GLM 5.3 is second with the best numeric columns.
    The hard exam has a different leader: GLM 5.3, Claude Fable 5.1, and DeepSeek V4.1 Flash lead through estimation under severe measurement. Astra is fourth despite the best Pass Rate, any-of-3 / all-of-3, and validity, because its intervals are weaker than its point estimates (Fig.~\ref{fig:ate_coverage}) and $S_{\mathrm{NR}}$ enters the score by magnitude.
    The same mechanism separates GPT-5.5's two composites: third-highest Pass Rate on the realistic exam, eleventh of twelve on CausalDSScore on both, because its estimates and intervals have a heavy tail on the hardest observation variants (Tab.~\ref{tab:realistic_obs_variant}).
    No model leads on every axis on either exam. We take the axes of Sec.~\ref{sec:intro} in turn, then stability across repeated attempts.

        \begin{figure}[ht!]
        \centering
        {\phantomsubcaption\label{fig:token_efficiency_main}}%
        {\phantomsubcaption\label{fig:token_efficiency_hard}}%
        \includegraphics[width=.75\textwidth]{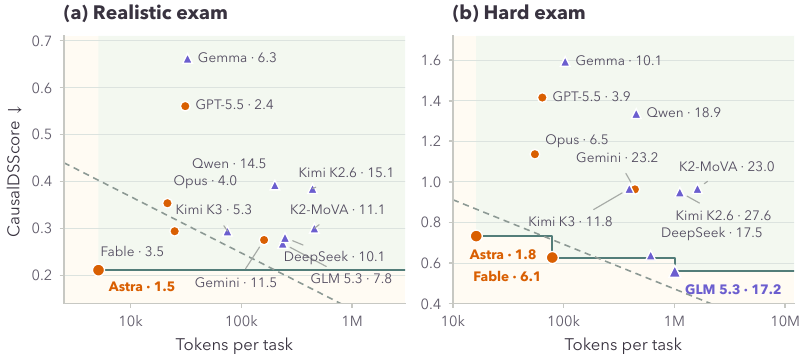}
        \caption{\textbf{Score versus token cost.} Mean tokens per task (log scale) against mean CausalDSScore (lower is better) on (a) the realistic and (b) the hard exam. Orange circles: closed models; violet triangles: open-weight. Labels: model and mean bash calls per task. Solid teal line: best mean score attainable within each token budget. Dashed line: illustrative separator for the current-generation closed models (GPT-6 Astra, Claude Fable 5.1).}
        \label{fig:token_efficiency}
    \end{figure}
    \begin{itemize}
        \item \textbf{Structure recovery is saturated; identification is not.}
        Every model recovers the story-implied graph: $F_1$-loss on the structure tasks is essentially zero on both exams (App.~\ref{app:exam_results}). The symbolic axis therefore reduces to the identification call, where pass rates spread widely (Tabs.~\ref{tab:realistic_perrung}, \ref{tab:hard_perrung}). Symbolic and data-backed pass rates are only loosely related across models (Tab.~\ref{tab:realistic_input_mode}); several models that identify well estimate poorly, and the reverse. The most common reasoning error on both exams is a claim of identification the graph does not support, taken up in the third point.
        \item \textbf{Imperfect observation degrades estimates more than decisions.}
        Numeric error varies much more than Pass Rate across the hard exam's observation tiers (App.~\ref{app:exam_results}, Tab.~\ref{tab:realistic_obs_variant}); the matched-scene ablation confirms the measurement-difficulty ordering (App.~\ref{app:observation_ablation}). Nominal $95\%$ ATE intervals under-cover on both exams; App.~\ref{app:error_intervals} reports their coverage and examines the sources of interval error (Fig.~\ref{fig:ate_coverage}).
        \item \textbf{Content correctness and abstention separate.}
        Abstention drives most of the realistic-exam Pass Rate spread, while the hard exam also separates models on content correctness and false abstention (App.~\ref{app:exam_results}, Tab.~\ref{tab:realistic_pr_split}).
    \end{itemize}

    \paragraph{Tokens versus score.}
    Tool and runtime exits are the smallest failure mechanism on both exams and rarely recur across restarts (App.~\ref{app:error_taxonomy}, Fig.~\ref{fig:error_persistence}); tool use shows up in cost, not correctness. In Fig.~\ref{fig:token_efficiency} the field splits into near-one-shot models, which write one script and read its output, and iterative models that probe the data over many calls; the split follows model family rather than capability, as other agent benchmarks also report \citep{wang_mcp-bench_2025, xu_roadmapbench_2026, zhang_deepplanning_2026, xu_autolab_2026}. Token use spans nearly two orders of magnitude on each exam and has no monotone relation to score: on the realistic exam the best and the worst model are both among the cheapest, and models at similar scores differ by a factor of ten or more in tokens. Within each of the three families with two generations (GPT, Claude, Kimi), the newer model typically scores as well as or better than its predecessor with fewer calls and no more tokens. The two current-generation closed models sit apart in the low-token, low-score corner on both exams (dashed lines). Comparing within approximate model generations, a closed--open separation persists for older generations, too (e.g., Opus 4.8 and Gemini 3.1 vs Kimi K2.6 and Qwen 3.6).

    \paragraph{Repeated-attempt stability.}
    The three replications behind the any-of-3 / all-of-3 columns are analyzed per axis in App.~\ref{app:pass@k}; to go beyond three runs we built a 45-task mini-exam of realistic- and hard-exam tasks, grouped by ground truth into set-valued, point, interval, and binary/abstention pools, and ran GPT-6 Astra, GLM 5.3, and Qwen 3.6 35B for $N{=}10$ attempts each (App.~\ref{app:pass@k}).
    Binary tasks use the hypergeometric pass@$k$ and pass\^{}$k$ estimators over the ten attempts \citep{chen_codex_2021, yao_tau_bench_2024}; loss-scored tasks use the continuous analogue, the expected best-of-$k$ and worst-of-$k$ loss over a random $k$-subset of the attempts (App.~\ref{app:pass@k}, Eq.~\ref{eq:subset_extrema}).

\begin{figure}[ht!]
        \centering
        \includegraphics[width=.95\textwidth]{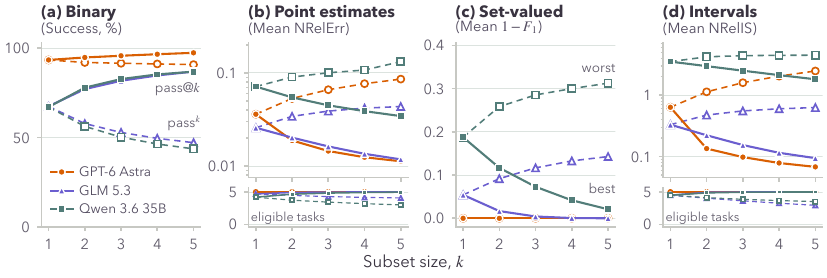}
        \caption{\textbf{Reliability over $k$ of $N{=}10$ attempts on the 45-task mini-exam.} (a) Binary pool: pass@$k$ (solid, filled markers) and pass\^{}$k$ (dashed, hollow). (b)--(d) Point, set-valued, and interval losses: best-of-$k$ (solid) and worst-of-$k$ (dashed), Eq.~\ref{eq:subset_extrema}, as equal-task means; (b) and (d) on log axes; strips: expected number of eligible tasks. Orange circles: GPT-6 Astra; violet triangles: GLM 5.3; teal squares: Qwen 3.6 35B.}
        \label{fig:targeted_reliability}
        \label{fig:targeted_passk}
        \label{fig:targeted_continuous}
    \end{figure}

    In Fig.~\ref{fig:targeted_reliability}, Astra's pass@$k$ and pass\^{}$k$ curves are nearly flat and close together, so its success barely depends on repetition. For GLM and Qwen the curves diverge widely by $k{=}5$, pass@$k$ approaching Astra's while pass\^{}$k$ falls; single-run pass rates overstate their reliability. On point estimates, best-of-$k$ converges to very small error for all three while worst-of-$k$ stays an order of magnitude higher, so the error tail comes from a minority of attempts. Intervals are Astra's weakest pool: its worst-of-$k$ interval score rises steeply with $k$, consistent with its $S_{\mathrm{NR}}$ on the hard exam.
    Run-to-run variation is one source of uncertainty in these leaderboards; the exam draw is the other, since CausalDS generates its exams. App.~\ref{app:leaderboard_ci} bootstraps tasks within grading categories and construction strata and reports which orderings survive both.

    \section{Related work}%
    \label{sec:related_work}
    Tab.~\ref{tab:main-competitors} compares the closest benchmarks across the main design axes; App.~\ref{app:related_work} provides the extended discussion.

    \begin{table}[ht!]
        \centering
        \fontsize{7.25pt}{8.25pt}\selectfont
        \setlength{\tabcolsep}{2.6pt}
        \renewcommand{\arraystretch}{1.08}
        \newcommand{\Full}{\hbFull}
        \newcommand{\Part}{\hbHalf}
        \newcommand{\Narrow}{\hbQuarter}
        \newcommand{\No}{\hbNone}
        \newcommand{\rot}[1]{\makebox[\dimexpr2\hbr+0.5pt\relax][l]{\rotatebox[origin=lb]{45}{#1}}}
        \caption{\textbf{Related benchmark overview.}}
        \label{tab:main-competitors}
        \begin{tabularx}{\textwidth}{@{}l ccccc >{\hspace{6pt}}ccccc >{\leftskip=6pt\relax}X@{}}
            \toprule
            & \rot{Synth.\ inst.}
            & \rot{Gen.\ graph}
            & \rot{Free-form story}
            & \rot{Causal-model GT}
            & \rot{Obs.\ layer}
            & \rot{Agentic execution}
            & \rot{R1--R3 tasks}
            & \rot{Non-ID cases}
            & \rot{Scored abstention}
            & \rot{Stat.\ uncertainty}
            & \hfill\begin{minipage}[b]{\dimexpr\linewidth-30pt\relax}
                \raggedright
                \Full{} central feature;\\
                \Part{} partially present;\\
                \Narrow{} narrow, probe-level, or adjacent;\\
                \No{} absent or N/A.
              \end{minipage}
            \\
            \cmidrule(lr){2-6}\cmidrule(l{9pt}r){7-11}
            Work & \multicolumn{5}{c}{\scshape\color{paperViolet} generation} & \multicolumn{5}{c}{\scshape\color{paperOrange} evaluation} & Comment \\
            \midrule
            \rowcolor{exammain!7}
            \textbf{CausalDS} (ours)
            & \Full & \Full & \Full & \Full & \Full & \Full & \Full & \Full & \Full & \Full
            & Integrated exams for causal data-science agents: graph-audited stories, observational files, and calibration data; identification, estimates, intervals, and abstention scored against hidden SCM ground truth.
            \\
            CLadder
            & \Full & \Narrow & \Narrow & \Full & \No & \No & \Full & \No & \No & \No
            & Templated causal QA over fixed graph motifs; oracle answers from Bernoulli CBNs, with identifying probabilities supplied in the prompt.
            \\
            \rowcolor{paperIvory!20}
            UniCo
            & \Full & \Full & \Full & \Full & \No & \No & \Full & \Part & \No & \No
            & Causal QA from sampled models and supplied probabilities, in symbolic, code, or narrated form. Includes non-ID judgments; numerical queries solvable by construction.
            \\
            CausalT5k
            & \Part & \No & \Part & \Narrow & \No & \No & \Full & \Part & \Full & \No
            & Narrative claim judgments and scored refusal under insufficient evidence; human/LLM-authored cases, with qualitative graph checks on labels for rungs 2--3.
            \\
            \rowcolor{paperIvory!20}
            CausalProfiler
            & \Full & \Full & \No & \Full & \No & \No & \Part & \Narrow & \No & \No
            & Synthetic data generator for causal-ML estimators: ATE, CATE, and counterfactual effects; hidden-confounding and misspecification regimes scored by estimation error, without per-query non-ID labels.
            \\
            CauSciBench
            & \Part & \Narrow & \Part & \Part & \No & \Full & \Narrow & \No & \No & \No
            & Causal-effect analysis on study/textbook tasks and synthetic data with fixed outcome models and known effects; scores method selection and executed estimates.
            \\
            \rowcolor{paperIvory!20}
            \makecell[l]{CausalReasoning\\Benchmark}
            & \No & \No & \Part & \No & \No & \Full & \Narrow & \No & \No & \Full
            & Causal identification and executed effect estimation on real datasets; estimates, standard errors, and interval overlap scored against reference analyses.
            \\
            CausalGame
            & \Full & \Narrow & \Narrow & \Full & \Full & \Full & \Part & \No & \No & \No
            & Interactive causal discovery across 14 fixed game narratives, with noisy measurements and sandboxed data analysis; reports scored on supported causal claims.
            \\
            \rowcolor{paperIvory!20}
            CausaLab
            & \Full & \Full & \Narrow & \Full & \No & \Full & \Part & \No & \No & \No
            & Interactive causal discovery in a fixed laboratory narrative over sampled SCMs; budgeted interventions, held-out prediction, and mechanism recovery.
            \\
            \bottomrule
        \end{tabularx}

        \vspace{0.35em}
        \begin{minipage}{0.98\textwidth}
            \fontsize{7.25pt}{8.25pt}\selectfont
            \textbf{Generation (left to right):} fresh synthetic instances; sampled graph structures beyond fixed motifs;
            graph-tied prose beyond templates; SCM/CBN-derived targets; noisy measurements.
            \textbf{Evaluation:} tool/environment interaction; Pearl-ladder coverage; non-identifiability given supplied
            information; rewarded abstention; scored intervals or standard errors.
            Comments qualify broader non-ID regimes and specify uncertainty references.
        \end{minipage}
    \end{table}

    \section{Discussion}%
    \label{sec:discussion}
    CausalDS pairs hidden SCMs and graph-faithful stories with SCM-derived ground truth, including identifiability, and an observation model that corrupts the released data, so one scene scores symbolic causal reasoning, estimation, uncertainty quantification, abstention, and tool use together.
    On both exams these axes dissociate.
    Every model recovers the story-implied structure, yet the identification call spreads widely, and the most common discrete error is claiming identification the graph does not support, usually from an instrument.
    The realistic exam is led by the model that makes the identification and abstention calls best (GPT-6 Astra); the hard exam by the models that estimate best under severe measurement (GLM 5.3, Claude Fable 5.1, DeepSeek V4.1 Flash), with Astra fourth due to poor interval estimation performance there.
    Intervals are the weakest axis for every model: nominal $95\%$ coverage is missed on both exams, and three quarters of the misses are intervals centered on the truth but too narrow: correct point estimation, but overconfident.
    Models that answer nearly every identifiable target correctly still abstain on only about half of the non-identifiable ones, and reliability follows the same line: over ten attempts, best-of-$k$ converges across models while worst-of-$k$ does not, with the variance in intervals and abstention flips rather than point estimates.
    In tool use/efficiency, the current-generation closed models (GPT-6 Astra and Claude Fable 5.1) are separated from the remaining evaluated models in the log-token--score plane on both exams (Fig.~\ref{fig:token_efficiency}).
    Tool and runtime errors are the smallest and least persistent failure class; reasoning errors sit at the identification step; data-science errors recur most across restarts. A possible next step would be targeted exams for abstention, estimation, and counterfactual reasoning; App.~\ref{app:limitations} lists what the present evidence does not yet establish.
    The models master the parts, reading the structure and producing the estimates, and part ways on the whole: knowing how good the estimate is, and whether any answer is licensed at all. GPT-6 Astra comes closest to a well-rounded \textit{causal data-science agent}, and even it fails the interval axis under severe measurement noise.

    \clearpage


    \subsection*{AI use statement}
    Large language models were used to assist with typesetting and polishing author-written text, literature discovery, routine software development, and the generation of components of the synthetic benchmark.
    As described in Sec.~\ref{sec:benchmark}, LLMs are part of the CausalDS synthetic generation pipeline and their outputs are checked by the benchmark's auditing and verification procedures. For literature discovery, the authors verified citations and claims against the original sources. For software assistance, all resulting code used in the experiments was reviewed and tested by the authors. The authors take responsibility for the final manuscript, implementation, experiments, and conclusions.

%

    \subsection*{Reproducibility statement}
    The CausalDS repository at \href{https://github.com/andleb/CausalDS}{\nolinkurl{github.com/andleb/CausalDS}} provides the complete source code and configurations, together with the frozen task lists, public inputs, and private grading data for the realistic exam and the observation-layer and CauseNet-seeding ablations.
    Its \texttt{README.md} documents installation, container setup, scene generation, exam execution, and regrading.
    Besides the code, the execution settings are additionally described in App.~\ref{app:harness_ops}, and the prompt templates in App.~\ref{app:prompt_templates}.

    \bibliography{bibliography}

    \clearpage
    \appendix
    \raggedbottom

    \section{Appendix}

    \subsection{Motif catalog}
    \label{app:motif_catalog}

    Fig.~\ref{fig:appendix_motif_catalog} shows the base motif templates used in this work. These are the graph skeletons used by the sampler, both as main graphs and as grafted auxiliary fragments. Gray dashed nodes denote latent variables.
    \begin{figure}[ht!]
        \centering
        \tikzset{
    motif-observed/.style={
        circle,
        draw=black,
        fill=white,
        minimum size=7mm,
        inner sep=0pt,
        font=\small
    },
    motif-latent/.style={
        circle,
        draw=gray!80,
        dashed,
        fill=gray!15,
        text=gray!30!black,
        minimum size=7mm,
        inner sep=0pt,
        font=\small
    },
    motif-edge/.style={
        -Latex,
        semithick,
        shorten >=1pt,
        shorten <=1pt
    }
}

\captionsetup[subfigure]{justification=centering}
\begin{subfigure}[t]{0.23\textwidth}
    \centering
    \begin{tikzpicture}[x=1cm,y=1cm]
        \node[motif-observed] (X) at (0,0) {X};
        \node[motif-observed] (M) at (1.1,0) {M};
        \node[motif-observed] (Y) at (2.2,0) {Y};
        \draw[motif-edge] (X) -- (M);
        \draw[motif-edge] (M) -- (Y);
    \end{tikzpicture}
    \caption{\texttt{chain}}
\end{subfigure}\hfill
\begin{subfigure}[t]{0.23\textwidth}
    \centering
    \begin{tikzpicture}[x=1cm,y=1cm]
        \node[motif-observed] (X) at (0,0) {X};
        \node[motif-observed] (M) at (1.1,0.9) {M};
        \node[motif-observed] (Y) at (2.2,0) {Y};
        \draw[motif-edge] (X) -- (M);
        \draw[motif-edge] (M) -- (Y);
        \draw[motif-edge] (X) -- (Y);
    \end{tikzpicture}
    \caption{\texttt{mediation}}
\end{subfigure}\hfill
\begin{subfigure}[t]{0.23\textwidth}
    \centering
    \begin{tikzpicture}[x=1cm,y=1cm]
        \node[motif-observed] (Z) at (1.1,1) {Z};
        \node[motif-observed] (X) at (0,0) {X};
        \node[motif-observed] (Y) at (2.2,0) {Y};
        \draw[motif-edge] (Z) -- (X);
        \draw[motif-edge] (Z) -- (Y);
        \draw[motif-edge] (X) -- (Y);
    \end{tikzpicture}
    \caption{\texttt{confounding}}
\end{subfigure}\hfill
\begin{subfigure}[t]{0.23\textwidth}
    \centering
    \begin{tikzpicture}[x=1cm,y=1cm]
        \node[motif-observed] (Z) at (1.1,1) {Z};
        \node[motif-observed] (X) at (0,0) {X};
        \node[motif-observed] (Y) at (2.2,0) {Y};
        \draw[motif-edge] (Z) -- (X);
        \draw[motif-edge] (Z) -- (Y);
    \end{tikzpicture}
    \caption{\texttt{fork}}
\end{subfigure}

\par\medskip

\begin{subfigure}[t]{0.23\textwidth}
    \centering
    \begin{tikzpicture}[x=1cm,y=1cm]
        \node[motif-observed] (X) at (0,0) {X};
        \node[motif-observed] (Y) at (2.2,0) {Y};
        \node[motif-observed] (Z) at (1.1,0.9) {Z};
        \draw[motif-edge] (X) -- (Z);
        \draw[motif-edge] (Y) -- (Z);
    \end{tikzpicture}
    \caption{\texttt{collider}}
\end{subfigure}\hfill
\begin{subfigure}[t]{0.23\textwidth}
    \centering
    \begin{tikzpicture}[x=1cm,y=1cm]
        \node[motif-observed] (V1) at (0,1) {V1};
        \node[motif-observed] (X) at (0,0) {X};
        \node[motif-observed] (V2) at (1.1,0.5) {V2};
        \node[motif-observed] (Y) at (2.3,0.5) {Y};
        \draw[motif-edge] (V1) -- (V2);
        \draw[motif-edge] (X) -- (V2);
        \draw[motif-edge] (V1) to[bend left=25] (Y);
        \draw[motif-edge] (X) to[bend right=25] (Y);
        \draw[motif-edge] (V2) -- (Y);
    \end{tikzpicture}
    \caption{\texttt{arrowhead}}
\end{subfigure}\hfill
\begin{subfigure}[t]{0.23\textwidth}
    \centering
    \begin{tikzpicture}[x=1cm,y=1cm]
        \node[motif-observed] (X) at (0,0.5) {X};
        \node[motif-observed] (M1) at (1.1,1) {M1};
        \node[motif-observed] (M2) at (1.1,0) {M2};
        \node[motif-observed] (Y) at (2.2,0.5) {Y};
        \draw[motif-edge] (X) -- (M1);
        \draw[motif-edge] (M1) -- (Y);
        \draw[motif-edge] (X) -- (M2);
        \draw[motif-edge] (M2) -- (Y);
    \end{tikzpicture}
    \caption{\texttt{diamond}}
\end{subfigure}\hfill
\begin{subfigure}[t]{0.23\textwidth}
    \centering
    \begin{tikzpicture}[x=1cm,y=1cm]
        \node[motif-observed] (V1) at (0,1) {V1};
        \node[motif-observed] (X) at (0,0) {X};
        \node[motif-observed] (V2) at (1.3,1) {V2};
        \node[motif-observed] (Y) at (1.3,0) {Y};
        \draw[motif-edge] (V1) -- (V2);
        \draw[motif-edge] (V1) -- (X);
        \draw[motif-edge] (X) -- (Y);
        \draw[motif-edge] (V2) -- (Y);
    \end{tikzpicture}
    \caption{\texttt{diamondcut}}
\end{subfigure}

\par\medskip

\begin{subfigure}[t]{0.23\textwidth}
    \centering
    \begin{tikzpicture}[x=1cm,y=1cm]
        \node[motif-latent] (U) at (1.1,1) {U};
        \node[motif-observed] (X) at (0,0) {X};
        \node[motif-observed] (Z) at (1.1,0) {Z};
        \node[motif-observed] (Y) at (2.2,0) {Y};
        \draw[motif-edge] (U) -- (X);
        \draw[motif-edge] (U) -- (Y);
        \draw[motif-edge] (X) -- (Z);
        \draw[motif-edge] (Z) -- (Y);
    \end{tikzpicture}
    \caption{\texttt{frontdoor}}
\end{subfigure}\hfill
\begin{subfigure}[t]{0.23\textwidth}
    \centering
    \begin{tikzpicture}[x=1cm,y=1cm]
        \node[motif-observed] (Z) at (0,1) {Z};
        \node[motif-latent] (U) at (1.1,1) {U};
        \node[motif-observed] (X) at (1.1,0) {X};
        \node[motif-observed] (Y) at (2.2,0) {Y};
        \draw[motif-edge] (Z) -- (X);
        \draw[motif-edge] (U) -- (X);
        \draw[motif-edge] (U) -- (Y);
        \draw[motif-edge] (X) -- (Y);
    \end{tikzpicture}
    \caption{\texttt{iv}}
\end{subfigure}\hfill
\begin{subfigure}[t]{0.23\textwidth}
    \centering
    \begin{tikzpicture}[x=0.72cm,y=0.95cm]
        \node[motif-latent] (U) at (1.8,1.9) {U};
        \node[motif-observed] (A) at (1.0,0.9) {A};
        \node[motif-observed] (Y) at (2.6,0.9) {Y};
        \node[motif-observed] (Z) at (0,0) {Z};
        \node[motif-observed] (W) at (3.6,0) {W};
        \draw[motif-edge] (U) -- (A);
        \draw[motif-edge] (U) -- (Y);
        \draw[motif-edge] (U) to[bend right=30] (Z);
        \draw[motif-edge] (U) to[bend left=30] (W);
        \draw[motif-edge] (A) -- (Y);
    \end{tikzpicture}
    \caption{\texttt{double\_nc}}
\end{subfigure}\hfill
\begin{subfigure}[t]{0.23\textwidth}
    \centering
    \begin{tikzpicture}[x=1cm,y=1cm]
        \node[motif-observed] (X) at (0,0) {X};
        \node[motif-observed] (Y) at (2.0,0) {Y};
        \node[motif-observed] (V3) at (1.0,0.9) {V3};
        \draw[motif-edge] (X) -- (Y);
        \draw[motif-edge] (X) -- (V3);
        \draw[motif-edge] (Y) -- (V3);
    \end{tikzpicture}
    \caption{\texttt{triangle}}
\end{subfigure}
        \caption{\textbf{Base motif templates.} Latent variables are shown as gray dashed nodes.}
        \label{fig:appendix_motif_catalog}
    \end{figure}

    \subsection{Extended related work}
    \label{app:related_work}

    Tab.~\ref{tab:main-competitors} summarizes the closest comparisons; the following discussion places
    them within four broader lines of work.

    \paragraph{Real-study causal analysis benchmarks.}
    The closest real-data benchmarks evaluate causal analysis over scientific or policy studies.
    \textit{CauSciBench} \citep{acharya_causcibench_2026} combines research-paper and textbook tasks
    with synthetic scenarios whose true effects are sampled under fixed outcome models. Agents select
    a causal method and implement it in Python; the reported metrics score method-selection accuracy
    and relative effect error, with requested standard errors excluded from scoring. The real-paper
    subset uses reference estimates, while the synthetic subset supplies model-derived targets.
    \textit{CausalReasoningBenchmark} \citep{sawarni_causalreasoningbenchmark_2026} evaluates
    structured identification specifications and executed effect estimation on curated real datasets.
    Its scoring includes point estimates, standard errors, and confidence-interval overlap against
    reference analyses. Thus, statistical-uncertainty evaluation has a direct precedent, with a
    different reference: CausalDS scores estimates and intervals against hidden SCM-derived targets.

    \textit{InterveneBench} emphasizes intervention-centered study-design reasoning in real
    social systems without predefined graphs or structural equations \citep{shi_intervenebench_2026}.
    The study-derived instances provide strong external realism, but their dependence on existing
    papers, datasets, or study designs makes them potentially vulnerable to data contamination
    (the ``causal parrot'' issue).

    \paragraph{Synthetic causal benchmarks and causal reasoning probes.}
    A second line evaluates formal causal reasoning in controlled settings.
    \textit{CLadder} \citep{jin_cladder_2023} constructs templated natural-language questions from
    fixed graph motifs and Bernoulli causal Bayesian networks, with oracle-computed answers spanning
    all three rungs of Pearl's ladder. Its queries are identifiable by construction, and the required
    probabilities are supplied in the prompt.

    \textit{UniCo} \citep{dai_towards_2026} samples graphs and conditional probability tables,
    obtains exact-inference answers for 18 query types across the ladder, and renders questions in
    symbolic, code, and LLM-narrated forms. CausalDS addresses a different statistical problem:
    UniCo supplies probabilities or mechanisms needed to answer a query, while CausalDS withholds
    generating mechanisms and requires estimation from finite observational samples. UniCo's generated
    benchmark does not evaluate tool-executed data analysis or statistical intervals, and has no
    noisy-measurement layer; its code form is an input representation of the causal model. Its identification
    questions do include cases with no identifying variable set, but its numerical queries are
    answerable by construction and do not score abstention on non-identifiable estimates. CausalDS
    combines tool-executed estimation, interval scoring, and such abstention, including scenes with
    noisy measurements and calibration data. The benchmarks therefore measure different capabilities:
    reasoning from supplied causal information in UniCo, and causal data-science analysis of narrated
    observational files in CausalDS.

    \textit{CauSim} \citep{astorga_causim_2026} builds executable SCMs of increasing complexity
    and renders them in natural language, checking each translation by mapping it back to an
    executable SCM. This supplies supervision for reasoning with a given causal simulator or its
    verbalization. CausalDS instead keeps the generating mechanisms private and asks an agent to
    estimate identified quantities from observational data. The distinction is therefore between
    answering queries about a supplied causal model and performing statistical analysis of data
    generated by a hidden one.

    Related graph-centric or format-dependent tests of causal understanding include
    \textit{CausalGraph2LLM}, \textit{CLEAR}, and \textit{CausalBench}
    \citep{sheth_causalgraph2llm_2024,chen_clear_2024,zhou_causalbench_2024}.
    \textit{Executable Counterfactuals} focuses on code-mediated counterfactual reasoning
    \citep{vashishtha_executable_2026}.

    \textit{CausalProfiler} \citep{panayiotou_causalprofiler_2026} is a synthetic data generator for
    causal-ML estimators. It samples causal models, observational data, queries, and ground truth for
    finite-sample estimation of ATE, CATE, and counterfactual total effects. Its estimation-error
    comparisons include identification-consistent, misspecified-graph, and hidden-confounding regimes,
    without assigning query-level non-identifiability labels or scoring abstention. It therefore
    provides a precedent for finite-sample effect evaluation against known causal-model targets.
    CausalDS combines that statistical setting with narrated scenes and tool-using agents that perform
    the analysis.

    \textit{Language Models as Causal
    Effect Generators} introduces sequence-driven SCMs for treatment-effect benchmarking
    \citep{bynum_language_2025}; \textit{CausalSynth} realizes
    SCM-sampled records as unstructured text under iterative consistency checks \citep{cheng_causalsynth_2026}.
    \textit{NoisyCausal} evaluates natural-language causal QA under structured noise (distractors,
    perturbed values, latent confounders, partial masking) \citep{xu_noisycausal_2026}.

    CausalDS differs by making the benchmark unit a
    narrated full-fledged scene: a hidden SCM is realistically grounded into domain variables,
    rendered as a graph-audited free-form story, packaged with synthetic SCM-driven data files, and the
    agent's tool-backed answers are then scored deterministically.

    \paragraph{Answerability and abstention.}
    A third line asks whether models recognize when a causal answer is unwarranted.
    \textit{CausalT5k} \citep{geng_causalt5k_2026} evaluates narrative causal-claim judgments across
    all three rungs of Pearl's ladder and scores informed refusal under insufficient evidence.
    Human-authored case types are expanded by an LLM and reviewed by humans; solver checks validate
    labels for rungs 2 and 3 against qualitative causal structures. Its utility/safety split penalizes
    both unsupported answers and over-refusal.
    \textit{CausalPitfalls} stresses classical statistical traps such as confounding, Simpson's paradox,
    and selection bias \citep{du2026ice}, and \textit{CauGym} generates probability-style questions
    from random DAGs and, in its insufficient-information split, requires the answer
    \texttt{LACK\_CONDITION} under exact-match scoring \citep{chen_can_2026}.
    CausalDS's abstention behaves differently: a formally specified estimand is judged non-identifiable
    from the observational law under the story-implied graph, and the same scene couples the outcome
    of that judgment to numerical estimation.
    This abstention criterion is complementary to verifier-style scoring approaches proposed for causal expressions
    \citep{he_uncovering_2025}: a verifier checks whether an emitted causal expression is formally correct, whereas
    abstention scoring asks the orthogonal question of whether the agent recognizes that no formally correct answer
    exists.

    \paragraph{Agentic data-science benchmarks and causal-agent systems.}
    Outside causality, \textit{DS-1000}, \textit{DSBench}, \textit{MLE-bench}, \textit{MLE-Dojo},
    \textit{BLADE}, and \textit{DiscoveryBench} evaluate code generation, ML engineering, or
    open-ended data-analysis workflows
    \citep{lai_ds-1000_2022,jing_dsbench_2025,chan_mle-bench_2025,qiang_mle-dojo_2025,
        gu_blade_2024,majumder_discoverybench_2025}, often using interactive execution frameworks
    such as \textit{InterCode} and \textit{SWE-agent}/\textit{mini-swe-agent}
    \citep{yang_intercode_2023,yang2024sweagent}. These benchmarks motivate our file-backed,
    tool-using evaluation protocol, but they do not provide hidden causal structure, interventional or
    counterfactual task families, or abstention targets for non-identifiable estimands. Within causal
    inference, \textit{CAIS} automates causal method selection and execution from a dataset-plus-query
    input \citep{verma_causal_2025}.

    \textit{CausalGame} \citep{chen_causalgame_2026} evaluates active experimental discovery across
    14 fixed game narratives driven by SCMs, with hidden confounders, injected measurement noise, and
    selection bias. Agents collect data and can run sandboxed Python analyses; their explanation
    reports are scored on supported causal claims.
    \textit{CausaLab} \citep{yang_causalab_2026} uses a fixed laboratory narrative with a freshly
    sampled hidden SCM in each episode. Agents choose budgeted interventions through an experimental
    interface, receive measurements, and are scored on held-out-target prediction and DSL-based
    mechanism recovery. Its hidden disturbances enter the structural dynamics.
    Both evaluate an experimental causal discoverer, including its choice of experiments;
    CausalDS asks whether an agent can act as a causal data
    scientist on a common exam: every model receives the same evidence and is evaluated against the
    same ground truth, allowing controlled comparison on the several axes CausalDS examines.

    Prior work establishes precedents for individual components of CausalDS. To the best of our knowledge,
    no other benchmark integrates sampled causal graphs, graph-audited free-form stories, SCM-generated
    tabular data with clean and noisy-measurement variants, and tool-using agent tasks spanning all three
    rungs of Pearl's ladder with deterministic scoring of identification, numerical estimates, statistical
    intervals, and abstention on non-identifiable estimands against hidden causal-model ground truth.

    \subsection{Scene synthesis algorithms}
    \label{app:scene_subalgorithms}

    Naming the abstract nodes of a sampled graph can introduce semantic drift and ``accidental edges'': a choice of variable names that implies a direct cause the graph does not contain, or hides one it does. Scene synthesis therefore audits both the variable mapping and the story against the graph and repairs them iteratively. Alg.~\ref{alg:scene-synthesis} gives the top-level control flow; the audit rules and yields are in App.~\ref{app:audit_details}, and the prompt templates in App.~\ref{app:pipeline_prompts}.

    \begin{figure}[ht!]
        \centering
        \setlength{\fboxsep}{3pt}%
        \fbox{%
            \begin{minipage}{\dimexpr\textwidth-2\fboxsep-2\fboxrule\relax}
                \captionsetup{skip=3pt}%
                \captionof{algorithm}{Scene synthesis: variable mapping and story verbalization}
                \label{alg:scene-synthesis}
                {\linespread{0.95}\footnotesize
                    \begin{algorithmic}[1]
                    \REQUIRE DAG $G$; optional stage plan $P$; budgets $R,T,S$ for seeding, mapping audit/repair, and story verification.
                    \ENSURE domain $D$, mapping $M$, verified story $\sigma$, or failure.
                    \STATE If $P$ is absent, set $P \leftarrow [G]$; let $H_0$ be the first stage and initialize its fixed-name assignment $F_0 \leftarrow \emptyset$.
                    \IF{CauseNet seeding is enabled}
                    \STATE Try up to $R$ CauseNet matches on $H_0$; take the first passing \textsc{PreAudit} (Alg.~\ref{alg:preaudit}), or fail if seeding is mandatory.
                    \ENDIF
                    \STATE $(D, M) \leftarrow \textsc{MapStage}(H_0, F_0, T)$ (Alg.~\ref{alg:mapstage}); \textbf{return} failure if it fails.
                    \FOR{each auxiliary stage $H_i$ with shared anchor $a_i$}
                    \STATE Extend $M$ with $\textsc{MapStage}(H_i, F_i, T)$, where $F_i$ fixes $a_i$ to its current
                    name and forbids names from other stages; fail on mapping failure, anchor drift, or name duplication across stages.
                    \ENDFOR
                    \IF{there are auxiliary stages and the merged-graph audit is enabled}
                    \STATE Run the audit/repair loop on $M$ over the full $G$ for up to $T$ rounds, preserving anchor meanings; \textbf{return} failure if it fails.
                    \ENDIF
                    \STATE $\sigma \leftarrow \textsc{Verbalize}(D, M, G, S)$ (Alg.~\ref{alg:story}); \textbf{return} failure if it fails.
                    \STATE \textbf{return} $(D, M, \sigma)$.
                    \end{algorithmic}
                }
            \end{minipage}%
        }
    \end{figure}

    Alg.~\ref{alg:scene-synthesis} calls three subroutines.
    \textsc{PreAudit} (Alg.~\ref{alg:preaudit}) is the LLM feasibility check on a CauseNet seeding: whether the fixed real-world concepts can be completed into a plausible graph before any mapping is attempted.
    \textsc{MapStage} (Alg.~\ref{alg:mapstage}) names the variables of one (sub)graph with a \textit{mapper}--\textit{auditor} loop: the mapper proposes a domain and names, the auditor checks every edge and non-edge for plausibility and flags names that imply links the graph lacks, and the mapper regenerates from the auditor's feedback. Grafted graphs are mapped stage-wise through the same loop: the host graph first, then each auxiliary fragment with the shared anchor fixed and names from other stages forbidden, and finally a global audit of the merged graph. The mapper may use web search to pin down realistic variable names, units, and story details.
    \textsc{Verbalize} (Alg.~\ref{alg:story}) drafts the narrative from the mapped graph and revises it against a story-to-DAG verifier until no hard violation (missing variable, missing edge, reversed direction) remains.

    \begin{algorithm}[ht!]
        \caption{LLM pre-audit for CauseNet seeding}
        \label{alg:preaudit}
        \begin{algorithmic}[1]
            \REQUIRE stage graph $H$ with CauseNet-fixed nodes $F$.
            \ENSURE feasibility verdict $(\texttt{feasible}, \texttt{confidence}, \texttt{reason})$.
            \STATE Extract directed edges $E$, all unordered non-edges $\mathcal{N}$, and the fixed--fixed subset $\mathcal{N}_F \subseteq \mathcal{N}$.
            \STATE Prompt with fixed nodes, remaining placeholders, $E$, $\mathcal{N}_F$, and $\mathcal{N}$.
            \STATE Ask whether a skilled mapper could interpret the fixed concepts and name placeholders so the graph constraints remain plausible.
            \STATE Reject iff a fixed--fixed non-edge has an unavoidable mainstream direct-cause reading, or a fixed--fixed edge is impossible.
            \STATE Otherwise accept, using \texttt{low} confidence when only narrow mainstream interpretations make the constraints work; uncertain cases default to feasible (as the following stages can still reject them).
            \STATE \textbf{return} the feasibility verdict.
        \end{algorithmic}
    \end{algorithm}
    \begin{algorithm}[ht!]
        \caption{\textsc{MapStage}: name the (sub)graph variables with audit/repair}
        \label{alg:mapstage}
        \begin{algorithmic}[1]
            \REQUIRE (sub)graph $H$; fixed-name assignment $F$ (possibly empty); audit retry budget $T$.
            \ENSURE mapping $M$ for $H$, or failure.
            \STATE The mapper proposes $M$ under fixed-name, completeness, and uniqueness constraints.
            \FOR{$t=1$ to $T$}
            \STATE The auditor checks edge plausibility, absence of plausible direct links for non-edges, and conditional independence/type-consistency signals.
            \IF{no violations remain}
            \STATE \textbf{return} $M$
            \ENDIF
            \IF{a violation cannot be fixed without changing fixed names}
            \STATE \textbf{return} failure
            \ENDIF
            \STATE The mapper regenerates $M$ from a feedback prompt provided by the auditor, listing the violations.
            \ENDFOR
            \STATE \textbf{return} failure.
        \end{algorithmic}
    \end{algorithm}
    \begin{algorithm}[ht!]
        \caption{\textsc{Verbalize}: story generation with story-to-DAG verifier loop}
        \label{alg:story}
        \begin{algorithmic}[1]
            \REQUIRE domain $D$, mapping $M$, DAG $G$, story retry budget $S$.
            \ENSURE verified story $\sigma$, or failure.
            \STATE Build yellow-flag context $\Phi$ by scanning $M$ for derived-name markers and restrictive qualifiers (cf. App.~\ref{app:audit_details}).
            \STATE The story writer drafts $\sigma$ from $(D, M, \Phi)$.
            \FOR{$s=1$ to $S$}
            \STATE The verifier checks $\sigma$ against $(G, M)$: missing variables, missing edges, and direction
            contradictions are hard violations; extra direct effects, domain drift, graph jargon, and plausibility/coherence issues are warnings.
            \IF{there are no hard violations}
            \STATE \textbf{return} $\sigma$
            \ENDIF
            \STATE The story writer revises $\sigma$ from the verifier's feedback listing the hard violations, warnings, and $\Phi$.
            \ENDFOR
            \STATE \textbf{return} failure.
        \end{algorithmic}
    \end{algorithm}

    \subsection{Mapping and narration audit details}
    \label{app:audit_details}

    \paragraph{Production configuration and yield.}
    The initial production batch used Kimi K2.5 for variable mapping, narration, and their audits. A regeneration run was done with Kimi K2.6 to retry all 151 failed instances, complete five previously unattempted instances, and replace one scene removed after a quality check. Tab.~\ref{tab:generation_yield} reports the generation yield, showing that mapping was the main failure stage in both batches.
    In the results presented in this work, we adopted a policy of seeding the outcome node and one of its neighbors (on the realistic exam, $83$ scenes seeded with two nodes, $17$ with one). Seeding is not fully deterministic: the mapper may still rename a seeded node within its topic.
    \begin{table}[ht!]
        \centering
        \caption{\textbf{Generation yield.} Failures are classified by stopping stage.}
        \small
        \setlength{\tabcolsep}{5pt}
        \begin{tabular}{@{}l r r r r r@{}}
            \toprule
            Batch (model) & Attempted & \makecell{Mapping\\failures} & \makecell{Story\\failures} & Completed & Yield \\
            \midrule
            Initial (Kimi K2.5) & 895 & 146 & 5 & 744 & 83.1\% \\
            Regeneration (Kimi K2.6) & 157 & 43 & 0 & 114 & 72.6\% \\
            \bottomrule
        \end{tabular}
        \label{tab:generation_yield}
    \end{table}

    Both the mapping and story auditors are augmented with a deterministic preprocessor that scans variable meanings
    for derived-name markers (e.g. ``corrected'', ``residual'', ``count of'', ``index based on'') and qualifiers (e.g.\ ``based only on'', ``not influenced by other factors''), and appends a yellow-flag for the auditor. This targets a recurring failure mode where a construct is narrated as directly causing a record/ID meant to merely keep track of it.

    \paragraph{Story-to-graph fidelity check.} 
    As both our generation and auditing are LLM-based, we use the exam results as an independent check that the released stories reliably encode their hidden graphs.
    The \texttt{causal\_sketch} family asks an agent to determine the graph implied by
    the story alone. It is scored by edge-set $F_1$ against the hidden graph, so a narrated but unrecoverable edge lowers recall and an edge the prose over-implies lowers precision. 
    Over twelve models and three replications of the realistic exam,
    the $17$ sketch tasks ($12$ directed, $5$ ``skeleton'') are recovered with $F_1{=}1$ in $606$ of $612$ task-runs
    (Tab.~\ref{tab:realistic_per_task_type}). 
    Three of the misses are provider-side safety refusals by Claude Fable~5.1 on one
    scene; the remaining three are single-run slips ($F_1$ of $0.5$ twice on one skeleton task, $0.83$ once on a grafted directed graph) by models that recovered the same scenes perfectly in their other runs.
    The four sketch tasks of the verbalization-swap ablation (App.~\ref{app:causenet_ablation}) are also recovered perfectly by both models.
    Twelve models from independent providers thus agree with the hidden graph on every scene they are asked
    about, which indicates more-than-sufficient story quality for the purposes of this benchmark; the stories are also plausible enough to trigger Anthropic's  refusal mechanism, as mentioned previously.

    \subsection{SCM profile registry}
    \label{app:scm_profiles}

    CausalDS supports a rich family of functional relations in the generated SCMs: each scene draws one continuous and one binary profile from the two profile registries in Tab.~\ref{tab:scm_profile_registry}, so that the node types can be either continuous or binary.
    A profile determines the functional mechanism family and the parameter ranges for the node. Root nodes are noise-only (continuous) or Bernoulli with a profile-specific base-rate range (binary). 
    Continuous noise is additive and drawn from Gaussian, Laplace, Student-$t$, or Gaussian-mixture
    families. Binary mechanisms produce a Bernoulli output through a logistic, threshold, or noisy-gate link with no separate noise term, the Bernoulli draw being the source of stochasticity. 
    These noise specifications support Monte Carlo evaluation of treatment effects, mediation, and counterfactuals using the same SCM.
    The empirical profiles take their families from the equation and rule corpora described in
    App.~\ref{app:composition_grounding}. 
    The profile mixture weights in the last column of Tab.~\ref{tab:scm_profile_registry} were likewise derived from the same corpora, and the synthetic profiles serve as a uniform coverage prior in the mixture.
    Synthetic regimes include neural-network mechanisms, mixture noise, and sharp thresholds.
    The SCM profiles are not a targeted difficulty axis for the realistic exam,
    as their weights and compositions are determined by the empirical grounding (the ``realistic'' part); for the hard exam, we tilt them toward more complex mechanisms in the observation model (cf. App.~\ref{app:obsmodel_details}, App.~\ref{app:exam_results}).
    \begin{table}[ht!]
        \centering
        \caption{\textbf{SCM profile registry.}}
        \label{tab:scm_profile_registry}
        \small
        \setlength{\tabcolsep}{4pt}
        \begin{tabularx}{\textwidth}{@{}>{\raggedright\arraybackslash}p{0.28\textwidth} l >{\raggedright\arraybackslash}X r@{}}
            \toprule
            Profile & Source & Mechanism & Weight \\
            \midrule
            \multicolumn{4}{@{}l}{\emph{Continuous profiles}} \\
            \texttt{interaction\_\allowbreak response} & empirical & product, saturating, or cooperative (Hill) response of the parents & $0.234$ \\
            \texttt{symbolic\_\allowbreak transform} & empirical & rational, power, exp/log, or piecewise transform of a linear parent index & $0.053$ \\
            \texttt{linear\_\allowbreak additive} & empirical & affine mean function $b + \sum_i w_i x_i$ & $0.013$ \\
            \texttt{generic\_\allowbreak additive\_\allowbreak baseline} & synthetic & affine mean function with random-sign Gaussian weights and Gaussian bias & $0.140$ \\
            \texttt{generic\_\allowbreak smooth\_\allowbreak handcrafted} & synthetic & linear term, shifted $\tanh$ nonlinearity, and pairwise parent interactions & $0.140$ \\
            \texttt{generic\_\allowbreak neural\_\allowbreak blackbox} & synthetic & random $\tanh$ network (2--4 layers, 8--64 units, spectral-normalized weights), output calibrated to $[-1, 1]$ & $0.140$ \\
            \texttt{generic\_\allowbreak mixture\_\allowbreak noise} & synthetic & handcrafted family; each node draws Gaussian-mixture noise (2--6 components) with probability $0.65$ & $0.140$ \\
            \texttt{generic\_\allowbreak heteroscedastic} & synthetic & handcrafted family; each node draws a parent-dependent noise scale with probability $0.7$ & $0.140$ \\
            \midrule
            \multicolumn{4}{@{}l}{\emph{Binary profiles}} \\
            \texttt{multi\_\allowbreak input\_\allowbreak logic\_\allowbreak gate} & empirical & each node draws a random Boolean gate (prob.\ $0.64$; each input negated with prob.\ $0.52$), noisy-AND ($0.19$), or noisy-OR ($0.16$) & $0.204$ \\
            \texttt{single\_\allowbreak regulator\_\allowbreak logistic} & empirical & logistic link on a linear-plus-softsign index & $0.096$ \\
            \texttt{generic\_\allowbreak shifted\_\allowbreak logistic} & synthetic & logistic link with a shifted softsign nonlinearity & $0.140$ \\
            \texttt{generic\_\allowbreak sharp\_\allowbreak threshold} & synthetic & sharp logistic threshold on the weighted parent sum & $0.140$ \\
            \texttt{generic\_\allowbreak noisy\_\allowbreak or} & synthetic & noisy-OR gate over binarized parents (flip probability $2$--$10\%$) & $0.140$ \\
            \texttt{generic\_\allowbreak noisy\_\allowbreak and} & synthetic & noisy-AND gate over binarized parents (flip probability $2$--$10\%$) & $0.140$ \\
            \texttt{generic\_\allowbreak signed\_\allowbreak logic\_\allowbreak high\_\allowbreak negation} & synthetic & random Boolean gate; each input negated with probability $0.55$ & $0.140$ \\
            \bottomrule
        \end{tabularx}
    \end{table}

    \subsection{Observation-model implementation details}
    \label{app:obsmodel_details}

    \paragraph{Mechanism families and noise.}
    As noted, every observation bundle has a single conceptual parent $Z$ (``measured node''). Continuous \textit{observed} columns use either the handcrafted form $h_r(z) = w z + b + w_{\mathrm{nl}} \tanh(z - c)$ or, with probability $p_{\mathrm{nn}}$, a small
    spectral-normalized $\tanh$ network calibrated to the realized range of $Z$.
    Binary columns use a logistic link on $w z + b + w_{\mathrm{nl}}\,\mathrm{softsign}(z - c)$.
    The shift $c$ is drawn per measurement so that the flat region of the nonlinearity, where $h'_r(z) \approx 0$ and the Fisher information vanishes, does not systematically coincide with the realized treatment-contrast region (cf. ``\textit{Admissibility checks}'' below).
    Continuous noise is additive with a base scale $\sigma_0$ drawn per measurement; its family is a Gaussian mixture
    with probability $p_{\mathrm{mix}}$ and otherwise Gaussian, Laplace, or Student-$t$ with equal probability. With
    probability $p_{\mathrm{het}}$ the scale is heteroscedastic, $\sigma(z) = \mathrm{softplus}(w_\sigma z + b_\sigma)$  clamped to $[\sigma_{\min}, \sigma_{\max}]$ and multiplying $\sigma_0$, so the local Gaussian approximation to the measurement's Fisher information is location-dependent: $I_r(z) \approx (h'_r(z))^2 / (\sigma_0\,\sigma(z))^2$.
    A released variant fixes these probabilities together with the number of measured nodes, the number of
    measurements per node, the range of $\sigma_0$, and the fraction of training rows used for calibration. \texttt{proxy} measures $1$--$3$
    non-outcome nodes with $2$--$10$ columns each, $(p_{\mathrm{nn}}, p_{\mathrm{mix}}, p_{\mathrm{het}}) = (0.3,
    0.3, 0.3)$, $\sigma_0 \in [1, 3]$, and a calibration fraction of $15\%$; \texttt{proxy\_hard} measures $2$--$5$ nodes     with $(0.45, 0.45, 0.6)$, $\sigma_0 \in [1.5, 4]$, and $10\%$; the hard exam's \texttt{proxy\_severe} measures $3$--$6$ nodes, allows the outcome node to be among the candidates, with $2$--$4$ columns each (fewer columns make the problem harder), $(0.5, 0.5, 0.7)$, $\sigma_0 \in [2.5, 5]$, and $4\%$.

    \paragraph{Admissibility checks on candidate bundles.}
    Candidate bundles for a continuous conceptual variable $Z$ are screened by estimated \emph{Fisher information}
    to reject nearly uninformative measurements. Let $I_r(z)$ denote the estimated local Fisher information about
    $Z$ carried by measurement $W_r$ at $Z=z$, and let $I(z) = \sum_r I_r(z)$ denote the bundle's total information.
    These estimates use local fits to the generated $(Z,W_r)$ pairs: a linear-Gaussian approximation for continuous
    measurements and a Bernoulli logistic model for binary measurements.
    We evaluate $I(z)$ on a grid over the realized support of $Z$ and accept the bundle only if its Fisher-information
    estimates pass all three checks: (i) $\min_z I(z) \geq \tau_{\min}$ (no dead zones),
    (ii) $\bar{I} \geq\tau_{\mathrm{avg}}$ (sufficient average information), and
    (iii) the lower-tail quantiles $I_{0.10}$ and $I_{0.25}$ clear their respective thresholds.
    Here, $\bar{I}$ is the mean over grid points and $I_q$ is the $q$-quantile of the grid values.
    For binary $Z$, we sum class-separation scores across measurements: squared standardized mean differences for
    continuous measurements, and mutual information for binary measurements.
    Scenes that also place a measurement bundle on the outcome apply the same checks with stricter thresholds.
    The sampler retries the mechanism draw up to a configured budget and fails the bundle afterward. The setting used for the results presented is $\tau_{\min} = 0.03$.

    \paragraph{Calibration rows.}
    The calibration dataset is a coverage-oriented selected subset of the public training rows.
    We first select rows from each quantile bin of \textit{every} measured conceptual variable.
    We then fill the remaining slots within each bin by selecting rows whose variable ranks are farthest (in the joint space of all selected conceptual variables to be ``proxied'') from those already selected.
    
    This preserves each measured variable's marginal support, which fitting the measurement model $W_{jr}\mid Z_j$ across the range of $Z_j$ requires, and since every measurement depends on a single conceptual parent with independent noise, the conditional law of $W_{jr}$ given $Z_j$ is the same in the calibration rows as in the population. The \textit{joint dependence} among measured variables is attenuated by the same construction: spreading points in rank space favors regions of low joint density --- in the exam scenes, pairwise rank correlations between measured variables drop to roughly a quarter to a third of their population values. 
    Population relationships among conceptual variables are thus hard to estimate from the calibration set; in particular, when the outcome is measured, an outcome regression directly fitted on the calibration rows is biased toward no effect.

\begin{figure}[ht!]
        \centering
        \includegraphics[width=.5\linewidth]{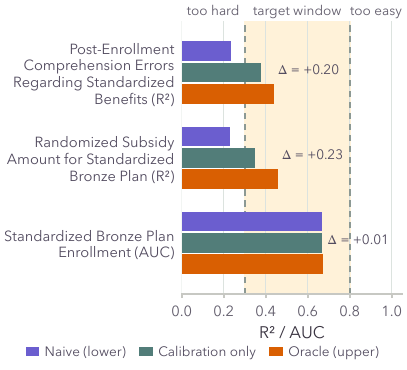}
        \caption{\textbf{Per-scene recoverability diagnostic.} The \texttt{proxy\_severe} variant of one hard-exam-pool scene with three measured conceptual variables (two continuous, one binary). Violet bars show the naive lower bound, teal bars show the calibration-only fit, orange bars show the ``oracle'' upper bound.            The shaded band marks the target window ($0.3$--$0.8$) used during scene generation: below it is ``too hard,'' above it ``too easy.'' We display the gap (Eq.~\ref{eq:delta_gap}) adjacent to the bars.
        }
        \label{fig:recoverability_diagnostic_example}
    \end{figure}

    \paragraph{Recoverability diagnostics in the private bundle.}
    Beyond the Fisher-information screen at generation time, each measured conceptual variable $Z_j$ is privately scored by how well it can be recovered from its $d_j$ measurements.
    Every score is evaluated on the scene's held-out test
    split and expressed as $R^2$ for continuous $Z_j$ and AUC for binary $Z_j$, so scores are comparable across scenes.
    For diagnostics, we use three reference fits: the \emph{lower bound} $S_{\mathrm{low}}$ is the best single-measurement \textit{linear} model (OLS or logistic regression) fitted on the public calibration rows, the score a naive agent would obtain. $S_{1}$ is the best single-measurement \textit{nonlinear} model.
    The \emph{upper bound} $S_{\mathrm{upp}}$ is a gradient-boosting model on all $d_j$ measurements
    jointly, fitted on the full training rows with their gold conceptual values (i.e., labels no agent ever sees), defining an oracle-like ceiling that caps even an oracle's downstream performance.
    The gap between the bounds:
    \begin{equation}
        \Delta \;=\; S_{\mathrm{upp}} - S_{\mathrm{low}}
        \;=\; \underbrace{(S_{1} - S_{\mathrm{low}})}_{\text{misspecification gap}} \;+\;
        \underbrace{(S_{\mathrm{upp}} - S_{1})}_{\text{multi-measurement/data information gain}},
        \label{eq:delta_gap}
    \end{equation}
    measures how much a skilled data-science agent can gain over the naive fit: the first part is what a flexible nonlinear model recovers from the single best column, the second what having all $d_j$ columns \textit{and} all rows would add on top. 
    A large $\Delta$ means the scene rewards a careful measurement model, and a small $\Delta$ means one column already carries most of the recoverable signal. 
    A further fit, $S_{\mathrm{cal}}$, applies the same gradient-boosting model to all $d_j$ measurements using the
    public calibration rows alone. It is recorded next to the two bounds as the score an agent can reach from the
    released files with a reasonably flexible model.\footnote{In practice, the agents are expected to exceed this since we do not optimize the model iteratively.} Fig.~\ref{fig:recoverability_diagnostic_example} shows $S_{\mathrm{low}}$,
    $S_{\mathrm{cal}}$, and $S_{\mathrm{upp}}$ on a representative scene.

    \paragraph{Realized measurement mechanisms and difficulty.}
    Within one observation variant the bundles still vary from scene to scene: the number of measured nodes, the share of neural mechanisms, and the noise family and scale are all random draws. Unlike the SCM profile (App.~\ref{app:scm_profiles}), whose mechanism shape reaches an agent only through the joint distribution it estimates from,
    a measurement mechanism has
    to be \textit{inverted}: the agent recovers the measured variable from its bundle through the calibration rows.
    The functional form of the link thus bears more directly on the task, and, e.g.,  a non-monotone neural network mechanism is what widens the misspecification gap of Eq.~\ref{eq:delta_gap} for an agent that fits linear calibration models. 
    This is the axis along which the variants are constructed to be hard, and it shows within variants as well. Regressing a per-task unit loss (clipping and scaling continuous-task losses to $[0,1]$) on model identity and task composition (output variant, structural label, treatment and outcome type) within one variant slice, with the realized share of neural-network mechanisms added as a covariate, a larger share of neural-network mechanisms goes with a higher loss in the realistic exam's \texttt{proxy\_hard} slice and the hard exam's
    \texttt{proxy\_severe} slice ($68$ scenes), by $0.22$ and $0.08$ loss units per standard deviation of the share
    ($p = 0.02$ and $0.015$ from reshuffling the share across scenes).

    \paragraph{Relation to proximal causal inference.}
    The calibrated observation layer is distinct from proximal causal inference \citep{tchetgen_introduction_2024, miao_identifying_2018}. Proximal causal inference addresses causal-effect identification when relevant confounding mechanisms are not directly observed, using auxiliary measurements or negative controls under additional assumptions about their relation to the unobserved confounding structure. In \textit{CausalDS}, the observation model has no such role: the conceptual DAG and SCM are fixed before the observation layer is applied, and each measurement bundle is generated as a noisy observation of \textit{one} conceptual variable. These released columns are not additional causal variables in the conceptual graph and cannot provide a new route to identification as they cannot create or remove confounding, instruments, or mediators. Thus, the observation layer changes only the estimation difficulty from the released files; the conceptual estimand and its identifiability are unchanged.

    \subsection{Optional empirical grounding of composition axes}
    \label{app:composition_grounding}

    CausalDS is fully synthetic at the scene level, but a number of \emph{composition axes} can be grounded from
    external corpora when we want a specific benchmark mixture to better resemble real causal-analysis workloads, as is the case for our \textit{realistic exam} (Sec.~\ref{sec:results}).
    
    The key design choice is to \emph{ground composition axes}: we derive axis-specific
    empirical distributions from several benchmark corpora and companion data, weight them by how directly they expose the quantity of interest (as our concepts such as question type might not map one-to-one), 
    and then allow benchmark-design tilts when needed for difficulty, coverage, or stress-testing (e.g.\ the exam-side difficulty knob of App.~\ref{app:difficulty_knob}). 
    The composition axes are split into a \emph{generation} side and an \emph{exam} side. Generation
    axes (variable types; main-motif weights, auxiliary-motif weights, and graft probability; identifiability
    proportion; typed SCM mechanism profiles; observation-model profile composition) determine the type of scenes that get sampled and built, while exam axes shape which task, required output type, and measurement view are  selected from a scene in a given exam. This split lets the same scene support multiple released exams without re-running expensive verbalization.

    The exam side  would require estimating the full joint $P(\text{motif}, \text{question type}, \text{output
    variant}, \text{observation variant})$ from available data, which is not possible given the available empirical sources. Thus, we factor the \textit{exam side} through the question type:
    \[
        P(S, Q, \text{output}, \text{obs})
        =
        P(S)\,P(Q \mid S)\,P(\text{output} \mid Q)\,P(\text{obs} \mid Q),
        \qquad
        \text{output} \ind \text{obs} \mid Q,
    \]
    where  $Q$ denotes the question type, and $S$ a structural label: motif-based or grafted DAG. 
    This factorization is illustrated in Fig.~\ref{fig:exam_composition_dag}.
    \begin{figure}[ht!]
        \centering
        \resizebox{0.65\linewidth}{!}{
\begingroup
\definecolor{compositionInk}{HTML}{393340}
\definecolor{compositionMuted}{HTML}{625D6C}
\colorlet{compositionStructure}{paperViolet!78!black}
\colorlet{compositionExam}{paperOrange!78!black}
\begin{tikzpicture}[
  x=1pt, y=1pt,
  font=\sffamily\fontsize{11.2}{13}\selectfont,
  card/.style={
    rounded corners=4pt,
    line width=.65pt,
    align=center,
    text=compositionInk,
    inner xsep=6pt,
    inner ysep=6pt,
    outer sep=0pt,
    minimum height=50pt
  },
  structure/.style={
    card, text width=92pt,
    draw=paperViolet!35, fill=paperViolet!6
  },
  question/.style={
    card, text width=92pt,
    draw=paperOrange!40, fill=paperOrange!7
  },
  variant/.style={
    card, text width=110pt,
    draw=paperOrange!30, fill=paperOrange!3
  },
  edge/.style={
    -{Stealth[length=5pt,width=3.6pt]},
    line width=.9pt, draw=compositionMuted!80,
    shorten <=2pt, shorten >=2pt
  }
]
  \node[structure] (S) at (52,0) {%
    {\bfseries\color{compositionStructure}Structural label $S$}\\[3pt]
    {\fontsize{8.8}{10}\selectfont\color{compositionMuted}(motif / grafted bucket)}
  };
  \node[question] (Q) at (174,0) {%
    {\fontsize{9}{10}\selectfont\bfseries\color{compositionExam}Axis 1}\\[2pt]
    {\bfseries Question type $Q$}\\[3pt]
    {\fontsize{10.5}{12}\selectfont $P(Q\mid S)$}
  };
  \node[variant] (O) at (311,32) {%
    {\fontsize{9}{10}\selectfont\bfseries\color{compositionExam}Axis 2}\\[2pt]
    {\bfseries Output variant}\\[3pt]
    {\fontsize{10.5}{12}\selectfont $P(\text{output}\mid Q)$}
  };
  \node[variant] (V) at (311,-32) {%
    {\fontsize{9}{10}\selectfont\bfseries\color{compositionExam}Axis 3}\\[2pt]
    {\bfseries Observation variant}\\[3pt]
    {\fontsize{10.5}{12}\selectfont $P(\text{obs}\mid Q)$}
  };

  \draw[edge] (S.east) -- (Q.west);
  \draw[edge] (Q.east) -- (O.west);
  \draw[edge] (Q.east) -- (V.west);
\end{tikzpicture}
\endgroup}
        \caption{\textbf{Exam-composition factorization.} Conditional on the structural label $S$ (motif or the synthetic
        \texttt{grafted} bucket), the exam builder draws the question type $Q$, then independently draws the output
        variant and the observation variant given $Q$. Each conditional corresponds to one exam-side axis: $P(Q \mid S)$
            is Exam Axis 1, $P(\text{output} \mid Q)$ is Exam Axis 2, and $P(\text{obs} \mid Q)$ is Exam Axis 3.}
        \label{fig:exam_composition_dag}
    \end{figure}

    Tab.~\ref{tab:grounding_source_key} lists the external sources used, Tab.~\ref{tab:grounding_sources} shows which
    source served which axis and in what role, and Tab.~\ref{tab:grounding_axes} gives, per axis, the estimated
    quantity, how the sources are combined, and the benchmark-design tilt applied on top of the empirical estimate for the realistic exam. Further details follow in \textit{Notes on individual axes.}
    When numbers are quoted, we refer to the specific composition chosen for the presented \textit{realistic} exam.
    \begin{table}[ht!]
        \centering
        \caption{\textbf{External grounding sources.} \textit{Abbr.}: abbreviations used in Tab.~\ref{tab:grounding_sources}.}
        \label{tab:grounding_source_key}
        \small
        \setlength{\tabcolsep}{5pt}
        \begin{tabular}{@{}l l l@{}}
            \toprule
            Abbr. & Source & Kind \\
            \midrule
            EC & \textit{EconCausal} \citep{lee_econcausal_2026} & real-study causal benchmark \\
            CSB & \textit{CauSciBench}, real-study subset \citep{acharya_causcibench_2026} & real-study causal benchmark \\
            CSB-QR & \textit{CauSciBench}, QR subset \citep{liu_qrdata_2024} & textbook benchmark \\
            CRB & \textit{CausalReasoningBenchmark} \citep{sawarni_causalreasoningbenchmark_2026} & real-study causal benchmark \\
            IB & \textit{InterveneBench} \citep{shi_intervenebench_2026} & real-study causal benchmark \\
            ReCITE & \textit{ReCITE} \citep{saklad_can_2025} & real-paper causal graphs \\
            CLadder & \textit{CLadder} \citep{jin_cladder_2023} & synthetic motif benchmark \\
            EI & \textit{EvidenceInference} \citep{deyoung_evidence_2020} & biomedical RCT reports \\
            TMW & \textit{TellMeWhy} \citep{lal_tellmewhy_2021} & narrative why-questions \\
            CG & \textit{CausalGame} \citep{chen_causalgame_2026} & challenge suite \\
            CP & \textit{CausalPitfalls} \citep{du2026ice} & challenge suite \\
            BM & \textit{BioModels} \citep{malik_sheriff_biomodels_2020} & biochemical equations \\
            SABIO & \textit{SABIO-RK} \citep{wittig_sabiork_2018} & biochemical equations \\
            BBM & \textit{BiodivineBooleanModels} \citep{pastva_biodivine_2023} & Boolean rules \\
            KDP & \textit{Kadelka DesignPrinciples} \citep{designprinciplesgenenetworks_2024} & Boolean rules \\
            \bottomrule
        \end{tabular}
    \end{table}
    \begin{table}[ht!]
        \centering
        \caption{\textbf{Source roles by composition axis.} $\bullet$: the source contributes frequency mass to the
        axis's empirical prior; $\circ$: the source is a companion that calibrates structure or
        coverage. Source abbreviations as in Tab.~\ref{tab:grounding_source_key}.}
        \label{tab:grounding_sources}
        \small
        \setlength{\tabcolsep}{6pt}
        \begin{tabular}{@{}l *{15}{c}@{}}
            \toprule
            Axis & \rotatebox{90}{EC} & \rotatebox{90}{CSB} & \rotatebox{90}{CSB-QR} & \rotatebox{90}{CRB} & \rotatebox{90}{IB} & \rotatebox{90}{ReCITE} & \rotatebox{90}{CLadder} & \rotatebox{90}{EI} & \rotatebox{90}{TMW} & \rotatebox{90}{CG} & \rotatebox{90}{CP} & \rotatebox{90}{BM} & \rotatebox{90}{SABIO} & \rotatebox{90}{BBM} & \rotatebox{90}{KDP} \\
            \midrule
            \multicolumn{16}{@{}l}{\emph{\color{paperViolet}Generation axes}} \\
            Variable types & $\bullet$ & $\bullet$ & $\bullet$ & $\bullet$ & & & & & & & & & & & \\
            Main motif & $\bullet$ & $\bullet$ & & $\bullet$ & $\bullet$ & $\circ$ & $\circ$ & & & & & & & & \\
            Aux.\ motifs / grafts & & & & & & $\bullet$ & $\circ$ & & & & & & & & \\
            Identifiability & $\bullet$ & $\bullet$ & & & & & & & & $\circ$ & $\circ$ & & & & \\
            SCM profiles & & & & & & & & & & & & $\bullet$ & $\bullet$ & $\bullet$ & $\bullet$ \\
            \midrule
            \multicolumn{16}{@{}l}{\emph{\color{paperOrange}Exam axes}} \\
            Question type & $\bullet$ & $\bullet$ & & $\bullet$ & & & $\circ$ & $\bullet$ & & & & & & & \\
            Output variant & & $\bullet$ & $\bullet$ & $\bullet$ & & & & & & & & & & & \\
            Observation variant & & & & $\circ$ & & $\bullet$ & & $\bullet$ & $\circ$ & & & & & & \\
            \bottomrule
        \end{tabular}
    \end{table}
    \begin{table}[ht!]
        \centering
        \caption{\textbf{Per-axis estimation and design tilts.} The quantities estimated,
        how the sources of Tab.~\ref{tab:grounding_sources} are combined, and the manual tilt applied on top of the empirical estimate.}
        \label{tab:grounding_axes}
        \footnotesize
        \setlength{\tabcolsep}{4pt}
        \renewcommand{\arraystretch}{1.1}
        \begin{tabularx}{\textwidth}{@{}>{\raggedright\arraybackslash}p{0.12\textwidth} >{\raggedright\arraybackslash}p{0.19\textwidth} >{\raggedright\arraybackslash}X >{\raggedright\arraybackslash}p{0.25\textwidth}@{}}
            \toprule
            Axis & Estimated quantity & Combination of sources & Design tilt \\
            \midrule
            \multicolumn{4}{@{}l}{\emph{\color{paperViolet}Generation axes}} \\
            Variable types & Continuous vs.\ non-continuous share, separately for treatment and outcome  & Types coded from source columns, metadata, and variable descriptions; sources credibility-weighted by how well their constructs map to ours & Slight tilt toward continuous variables \\
            \addlinespace[3pt]
            Main motif & Weights for the motif inventory & Design-family labels (adjusted observational, DiD/event study, RD, IV, matching) mapped to motifs explicitly; per-source distributions credibility-weighted; \textit{ReCITE} supplies the relative weights when a label is mapped to multiple motifs & \textit{CLadder}'s distribution used as coverage for motifs absent from design-family labels \\
            \addlinespace[3pt]
            Auxiliary motifs and grafts & Auxiliary-pool weights; graft-count distribution & \textit{ReCITE} structural catalog over explicit real-paper graphs; \textit{CLadder} for coverage & Graft-count prior hand-authored with a small adjustment toward \textit{ReCITE}'s distribution \\
            \addlinespace[3pt]
            Identifiability & Share of scenes with a non-identifiable population ATE (as a stand-in) & Empirical sources are $99.6\%$ identifiable; challenge suites give only the \emph{kinds} of obstacles & Manually requested non-identifiable quota\\
            \addlinespace[3pt]
            SCM profiles & Profile weights per node type & We exclude rare equation and rule types and group the remainder into three continuous and two binary profiles. Profile weights are estimated
            separately for continuous and binary nodes, independently of graph motif. & Convex mix with a uniform synthetic coverage prior (weight $0.3$ on the empirical part; App.~\ref{app:scm_profiles}) \\
            \midrule
            \multicolumn{4}{@{}l}{\emph{\color{paperOrange}Exam axes}} \\
            Question type & $P(Q \mid S)$ & Credibility-weighted empirical marginal over task-family labels, mixed with a uniform coverage prior, then reshaped by a \textit{CLadder} structural lift (Eq.~\ref{eq:structural_lift}) & Floor keeps every rung-1 and rung-3 family off zero under the R2-heavy empirical marginal; small lift exponent\\
            \addlinespace[3pt]
            Output variant  & $P(\text{output} \mid Q)$ & We manually map source answer formats to our output variants. For \texttt{identification} and \texttt{effect\_estimate}, we mix source
            frequencies with uniform weights over compatible variants; other families  start with uniform weights  & Tilt toward harder answer contracts in prediction, association, and causal sketch, which are among the easier tasks \\
            \addlinespace[3pt]
            Observation variant (Exam Axis 3) & $P(\text{obs} \mid Q)$ over \texttt{clean}, \texttt{proxy}, \texttt{proxy\_hard} & Source items are grouped into the four accessibility levels.
            We estimate distributions directly for graph-recovery and effect-estimation questions, reuse them for related question families, and  use the overall source distribution for the remaining families. & The two hardest source levels both map onto \texttt{proxy\_hard} \\
            \bottomrule
        \end{tabularx}
    \end{table}

    \paragraph{Notes on individual axes.}
    
    \emph{Variable types.} Some high-cardinality discrete variables are better modeled as effectively continuous in
    downstream estimation, and a modest tilt toward continuous variables raises data-analysis difficulty without
    changing the causal semantics.\\
    \emph{Auxiliary motifs and grafts.} The graft-count prior is hand-authored rather than taken from corpus frequencies because the available real-paper graph complexity source (\textit{ReCITE}) violates $P(2~\text{grafts}) < P(1~\text{graft})$, a constraint we hold to be structurally justified.\\
    \emph{Identifiability.} Real-world data on non-identifiable cases is scarce as ``solvability'' is typically desirable for published problems. This generation axis uses the population ATE as a one-item proxy for identifiability, while on the exam, identifiability is checked for each requested estimand type.
    Under the benchmark's assumptions, the ATE is non-identifiable in \texttt{double\_nc} and \texttt{iv}.
    The generation configuration requests a quota of non-identifiable cases ($10\%$ for the presented realistic exam). If the two motifs above cannot satisfy it, we inject latent confounding $U \to \{X, Y\}$ into other motifs and accept only graphs
    whose population-ATE identifiability check returns \texttt{False} (i.e., we use rejection sampling).\\
    \emph{SCM profiles.} The empirical evidence comes from a biology-heavy source pool and is concentrated on
    multi-parent interaction shapes and multi-input gates. Because of this narrow domain and limited mechanism
    diversity, the empirical part is demoted to $\alpha = 0.3$ in favor of the uniform coverage prior, and the compact
    profile names keep the registry domain-neutral.\\
    \emph{Question type.} Let $\hat p_{\mathrm{emp}}(Q)$ be the credibility-weighted family marginal of the sources,
    which has mass on identification and effect estimation only, and $p_0(Q) = 0.15\,\hat p_{\mathrm{emp}}(Q) +
    0.85/|\mathcal{Q}|$ its mixture with a uniform floor over the $|\mathcal{Q}| = 10$ deployable families. The
    structural lift is
    \begin{equation}
        P(Q \mid S) \;\propto\; p_0(Q)\,
        \left(\frac{P_{\textit{CLadder}}(Q \mid S)}{P_{\textit{CLadder}}(Q)}\right)^{\lambda},
        \qquad \lambda = 0.15,
        \label{eq:structural_lift}
    \end{equation}
    \textit{CLadder} being the only source with joint graph $\times$ question-type rows. It is generated with equal
    counts across supported combinations per graph-family and question-type cell, so its question marginal carries no frequency information; only the ratio does, albeit weakly, which is why we pick a small $\lambda$.\\
    \emph{Observation variant.} We use information \textit{accessibility} as a proxy for observation difficulty and order it as: \texttt{clean} (the needed information is stated in the given passage) $<$ \texttt{partly\_explicit} (only partly stated there and the rest must be inferred) $<$ \texttt{document\_context} (it is stated elsewhere in the document, outside the given
    passage) $<$ \texttt{external\_implicit} (it is not in the document at all).

    \subsection{Exam-side difficulty knob}
    \label{app:difficulty_knob}

    After the distributions of App.~\ref{app:composition_grounding} are built, a separate \textbf{difficulty knob}
    $d \in [0, 1]$ tilts the three exam-side distributions: the question-type conditional $P(Q \mid S)$, the within-family
    output-variant distribution $P(\text{output} \mid Q)$, and the released
    observation-variant distribution $P(\text{obs} \mid Q)$.
    We partition question families by Pearl rung $r(Q) \in \{1, 2, 3\}$,
    index output variants by a within-family difficulty rank $r_q(v)$ with midpoint $c_q$, and index observation
    variants by $o(\text{obs}) \in \{0, 1, 2\}$ for $\{\texttt{clean}, \texttt{proxy}, \texttt{proxy\_hard}\}$.%
    \footnote{\texttt{proxy\_severe} was constructed for the hard exam specifically.}
    We multiply the initial weights by the following difficulty factors, then normalize them to sum to one:
    \begin{equation}
        \begin{aligned}
            \hat p_{d}(Q \mid S)
            &\propto \hat p(Q \mid S)
            \exp\!\bigl(\beta\,(2d{-}1)\,(r(Q){-}2)\bigr), \\
            \hat p_{d}(v \mid Q)
            &\propto \hat p(v \mid q)
            \exp\!\bigl(\delta\,(2d{-}1)\,(r_q(v) - c_q)\bigr), \\
            \hat p_{d}(\text{obs} \mid Q)
            &\propto \hat p(\text{obs} \mid Q)
            \exp\!\bigl(\gamma\,(2d{-}1)\,(o(\text{obs}){-}1)\bigr),
        \end{aligned}
        \label{eq:difficulty_tilt}
    \end{equation}
    In Eq.~\ref{eq:difficulty_tilt}, $d = 0.5$ leaves the initial weights unchanged; $d = 0$ favors R1, easier output
    variants, and \texttt{clean}, while $d = 1$ favors R3, harder output variants, and \texttt{proxy\_hard}.
    The constants control the strength of each adjustment: $\beta$ for the Pearl-rung mix, $\delta$ for output-variant
    difficulty within each question family, and $\gamma$ for observation-variant difficulty. All default to $1$ and
    can be set independently.
    These difficulty rankings reflect benchmark-design choices: Pearl rung is a coarse indicator, and output variants
    are ranked within each question family, for example with CIs ranked as more difficult than point estimates.
    The observation ordering \texttt{clean} $<$ \texttt{proxy} $<$ \texttt{proxy\_hard} follows increasing measurement corruption.
    The realistic exam uses $d = 0.5$, leaving the empirical weights from
    App.~\ref{app:composition_grounding} unchanged.

    \paragraph{Post hoc check of the three factors.}
    We examine how model performance varies with observation variant, Pearl rung, and output variant on both exams, using the twelve models and three replications of the leaderboards (Tab.~\ref{tab:difficulty_axis_evidence}).
    \emph{Score} is the CausalDSScore of a level's tasks pooled over all model runs; \emph{Band} is the $5$--$95\%$ range over $500$ resamples of the level's tasks; \emph{Model-centered $\Delta$} subtracts each model's full-exam score from its score on the level and averages over models; \emph{Adjusted $\Delta$} compares the mean per-task loss ($0/1$ for discrete tasks, normalized error for point tasks, capped normalized interval score for intervals, $F_1$-loss for graph and set tasks) with a reference level after an additive regression adjustment for model and the other factors.
    Positive differences indicate worse performance.

    The observation ranking holds. On the realistic exam \texttt{proxy} is no harder than \texttt{clean} (adjusted $\Delta$ of $+0.03$, overlapping bands) while \texttt{proxy\_hard} is harder on every summary, with an adjusted $\Delta$ of $+0.19$ that survives adjustment for model, rung, and output variant; on the hard exam the three proxy tiers order as intended, with adjusted $\Delta$ of $+0.48$ (\texttt{proxy\_hard}) and $+0.73$ (\texttt{proxy\_severe}) relative to \texttt{proxy}.
    The rung ranking does not hold: performance does not worsen from R1 to R3 on either exam, and on the realistic exam R3 is the easiest rung once observation variant is adjusted for ($-0.27$ against R1), largely because most of its tasks are counterfactual-identification binaries with high pass rates and its numeric errors are small, whereas R1 carries the prediction tasks with the largest normalized errors (Tab.~\ref{tab:realistic_perrung}).
    The output-variant ranking is supported in the aggregate but not family by family: a regression with fixed effects for model, task family, and observation variant estimates a loss increase of $0.25 \pm 0.03$ per rank step on the realistic exam and $0.96 \pm 0.03$ on the hard exam, and the interval outputs are the hardest variants of their families on both, yet the direction of the within-family relationship agrees with the chosen ranking in only 5 of 9 realistic-exam families and 3 of 4 hard-exam families.
    Given this mixed evidence, the hard exam specifies its sampling weights directly and sets $\beta = \delta = \gamma = 0$ (App.~\ref{app:exam_results}).

    \begin{table}[ht!]
        \centering
        \caption{\textbf{Model performance by observation variant and Pearl rung.} Twelve models, three replications, corrected grades. \emph{Score}: CausalDSScore of the level's tasks pooled over all model runs; \emph{Band}: $5$--$95\%$ range over $500$ task resamples within the level; \emph{Model-centered $\Delta$}: mean over models of the level score minus that model's full-exam score; \emph{Adjusted $\Delta$}: mean per-task loss relative to the reference level (ref.) after additive adjustment for model and, for observation variants, rung and output variant; rung is adjusted for model and observation variant only, since output variants are nested in rungs. Positive is harder. The hard exam has no R1 tasks, and its \texttt{clean} slice holds only symbolic output variants, so its adjusted contrast is not separable (n/s) and \texttt{proxy} is the reference there.}
        \label{tab:difficulty_axis_evidence}
        \footnotesize
        \begin{adjustbox}{max width=\textwidth}
            \begin{tabular}{@{}l l l r c c c c@{}}
                \toprule
                Exam & Axis & Level & Tasks & Score & Band & Model-centered $\Delta$ & Adjusted $\Delta$ \\
                \midrule
                \multirow{3}{*}{\examMain} & \multirow{3}{*}{Observation} & \texttt{clean} & 48 & 0.280 & $[0.159, 0.439]$ & $-0.076$ & ref. \\
                 &  & \texttt{proxy} & 32 & 0.155 & $[0.087, 0.229]$ & $-0.200$ & $+0.031$ \\
                 &  & \texttt{proxy\_hard} & 20 & 0.901 & $[0.463, 1.340]$ & $+0.540$ & $+0.193$ \\
                \addlinespace[3pt]
                \multirow{3}{*}{\examMain} & \multirow{3}{*}{Rung} & R1 & 28 & 0.486 & $[0.260, 0.798]$ & $+0.137$ & ref. \\
                 &  & R2 & 51 & 0.376 & $[0.208, 0.562]$ & $+0.020$ & $-0.092$ \\
                 &  & R3 & 21 & 0.136 & $[0.075, 0.206]$ & $-0.218$ & $-0.267$ \\
                \midrule
                \multirow{4}{*}{\examHard} & \multirow{4}{*}{Observation} & \texttt{clean} & 33 & 0.164 & $[0.105, 0.231]$ & $-0.827$ & n/s \\
                 &  & \texttt{proxy} & 24 & 0.552 & $[0.296, 0.904]$ & $-0.431$ & ref. \\
                 &  & \texttt{proxy\_hard} & 55 & 1.063 & $[0.709, 1.493]$ & $+0.073$ & $+0.483$ \\
                 &  & \texttt{proxy\_severe} & 88 & 1.376 & $[1.030, 1.778]$ & $+0.384$ & $+0.734$ \\
                \addlinespace[3pt]
                \multirow{2}{*}{\examHard} & \multirow{2}{*}{Rung} & R2 & 104 & 0.978 & $[0.726, 1.261]$ & $-0.013$ & ref. \\
                 &  & R3 & 96 & 1.002 & $[0.679, 1.299]$ & $+0.014$ & $-0.336$ \\
                \bottomrule
            \end{tabular}
        \end{adjustbox}
    \end{table}

    \subsection{Task families and scoring rules}
    \label{app:scoring_rules}

    \paragraph{Task families.}
    A task is denoted \texttt{Rung::family::variant}. Tab.~\ref{tab:task_families} lists the ten question families by rung; the families expand on \textit{CLadder}'s collection. Within a family, the \textbf{output variant} fixes what the agent must return and therefore how the task is graded: a sign, a boolean, a set, a point estimate, an interval, or an abstention-eligible value. Effect families carry several variants of the same estimand (point, $95\%$ interval, sign, agreement with the observed association), so one scene can test the same quantity at different levels of demand. Tabs.~\ref{tab:scoring_r1}--\ref{tab:scoring_r3} enumerate every variant with its metric.

    \begin{table}[ht!]
        \centering
        \caption{\textbf{Question families per rung.}}
        \label{tab:task_families}
        \normalsize
        \setlength{\tabcolsep}{6pt}
        \renewcommand{\arraystretch}{1.15}
        \begin{tabularx}{\textwidth}{@{}c >{\raggedright\arraybackslash}p{0.24\textwidth} >{\raggedright\arraybackslash}X@{}}
            \toprule
            Rung & Family & Description \\
            \midrule
            R1 & Prediction          & Fit a predictive model for outcome $Y$ from observed columns; write held-out point predictions or, for continuous $Y$, central prediction intervals. \\
            R1 & Association         & Quantify the observational relation between treatment $X$ and outcome $Y$, marginally or after conditioning on a third variable. \\
            R1 & Collider phenomenon & Test whether conditioning on a named collider opens a spurious association between two of its parents (explaining-away). \\
            \midrule
            R2 & Causal sketch       & Recover the story-implied graph as a directed edge set or as an undirected skeleton. \\
            R2 & Identification      & Decide whether the population ATE of $X$ on $Y$ is identifiable from the observed conceptual variables and by what strategy; includes adjustment-set queries. \\
            R2 & Effect estimation   & Estimate the population average treatment effect (ATE); output point, uncertainty, sign, or agreement with the observed association. \\
            R2 & Bias diagnostics    & Diagnose whether an adjustment strategy is biased (collider conditioning, forbidden controls). \\
            \midrule
            R3 & Counterfactual identification & Decide whether the target counterfactual is identifiable.\\
            R3 & Counterfactual effects        & Estimate the effect of treatment on the treated (ETT), or abstain if non-identifiable. \\
            R3 & Mediation                     & Decompose the effect into natural direct (NDE) and natural indirect (NIE) components; report point, sign, or which dominates. \\
            \bottomrule
        \end{tabularx}
    \end{table}

    \paragraph{Scoring rules.}
    We list the scoring rules used in the benchmark in this appendix. Tab.~\ref{tab:scoring_base_metrics} lists the base metrics, the normalized quantity each task contributes when pooled, and the pooled category it feeds. Tabs.~\ref{tab:scoring_r1}--\ref{tab:scoring_r3} then assign them to task variants.
    The abstain convention from Sec.~\ref{sec:scoring} applies whenever the target estimand is
    non-identifiable or the model abstains (routing details below), superseding the task-specific metrics.

    \begin{table}[ht!]
        \centering
        \caption{\textbf{Base metrics and their pooled forms.} $\mathrm{sd}(Y_{\mathrm{test}})$ is the outcome
        standard deviation on the private test set, $\theta$ and $\tau$ the GT target, $(x)_+ = \max(x, 0)$, and
        $c$ the cap of the interval pool. $N$ counts task--output variant pairs;
        $\dagger$ marks the same pair in both prediction rows. Abstention reroutes existing variants.}
        \label{tab:scoring_base_metrics}
        \footnotesize
        \setlength{\tabcolsep}{4pt}
        \renewcommand{\arraystretch}{1.15}
        \begin{adjustbox}{max width=\textwidth}
        \begin{tabular}{@{}>{\raggedright\arraybackslash}p{0.16\textwidth} >{\raggedright\arraybackslash}p{0.12\textwidth} >{\raggedright\arraybackslash}p{0.29\textwidth} >{\raggedright\arraybackslash}p{0.24\textwidth} >{\raggedright\arraybackslash}p{0.09\textwidth} c@{}}
            \toprule
            Used for & Metric & Definition & Normalized contribution & Pool & $N$ \\
            \midrule
            continuous prediction & RMSE & $\sqrt{n^{-1}\sum_j (y_j-\hat y_j)^2}$ over the test rows & $\mathrm{RMSE}/(1+\mathrm{sd}(Y_{\mathrm{test}}))$ & $T_{\mathrm{NR}}$ & $1^\dagger$ \\
            \addlinespace[2pt]
            binary prediction & ROC-AUC, Brier, log-loss & $\mathrm{Brier} = n^{-1}\sum_j (\hat p_j-y_j)^2$; ROC-AUC and log-loss reported alongside & $\sqrt{\mathrm{Brier}}/(1+\mathrm{sd}(Y_{\mathrm{test}}))$ & $T_{\mathrm{NR}}$ & $1^\dagger$ \\
            \addlinespace[2pt]
            point effects, association strengths & Absolute error & $|\hat\theta-\theta|$ & $|\hat\theta-\theta|/(1+|\theta|)$ & $T_{\mathrm{NR}}$ & $6$ \\
            \addlinespace[2pt]
            prediction intervals ($\alpha = 0.1$, mean over test rows); effect intervals ($\alpha = 0.05$) & Interval score $\mathrm{IS}_\alpha$ & $\begin{aligned}[t] & (\hat u-\hat\ell) \\ & {}+ \tfrac{2}{\alpha}(\hat\ell-\tau)_+ \\ & {}+ \tfrac{2}{\alpha}(\tau-\hat u)_+ \end{aligned}$ & $\mathrm{NRelIS} = \mathrm{IS}_\alpha/(1+s)$ with $s = \mathrm{sd}(Y_{\mathrm{test}})$ or $|\tau|$; capped mean across tasks, $c = 10$ & $T_{\mathrm{NR}}$ & $4$ \\
            \addlinespace[2pt]
            edge sets, adjustment sets, forbidden controls & $F_1$ & $2|\hat S\cap S|/(|\hat S|+|S|)$; $1$ when both sets are empty & $1-F_1$ & Med.\ $F_1$-Loss & $4$ \\
            \addlinespace[2pt]
            discrete answers & Exact match & $\mathbf{1}[\text{answer} = \text{GT}]$ & as is & Pass Rate & $18$ \\
            \addlinespace[2pt]
            non-identifiable target or null answer & Abstention binary & $\mathrm{AbstMatch}=1$ if the model's abstention decision matches target non-identifiability, and $0$ otherwise & as is & Pass Rate & --- \\
            \bottomrule
        \end{tabular}
        \end{adjustbox}
    \end{table}

    \paragraph{Interval scores and the aggregator $S_{\mathrm{NR}}$.}
    The interval score is strictly proper \citep{gneiting_strictly_2007}: its expectation is minimized by the true
    central (equal-tailed) $(1-\alpha)$ interval, so the agent picks its own estimator and the submitted
    $(\hat\ell, \hat u)$ are scored directly against the scalar target $\tau$, row-wise against $y_j$ for prediction
    intervals, with no reference interval computed or stored. Propriety is preserved by averaging, so interval scores are aggregated by means (as opposed to e.g., medians): over test rows within a prediction-interval task, and across tasks inside the aggregate numeric metric $S_{\mathrm{NR}}$ (Eq.~\ref{eq:nrel_aggregator}). For this reason, $S_{\mathrm{NR}}$  splits the numeric pool $T_{\mathrm{NR}}$ into a point pool $T_{\mathrm{NR}}^{\mathrm{pt}}$ and an interval pool $T_{\mathrm{NR}}^{\mathrm{int}}$. 
    As point-graded tasks emit normalized errors, a median over the point pool is appropriate as it tracks typical-scene performance without being  tail-driven.
    Interval-graded tasks, however, carry a miss penalty (after the above averaging) $2/\alpha$ where one overconfident miss can reach $\mathrm{NRelIS}$ (Tab.~\ref{tab:scoring_base_metrics})
    values an order of magnitude above the typical level. While a median would discard such signals (and abandon propriety), a raw mean lets one catastrophic miss dominate the metric. Hence we settle on the \textit{capped mean} $\mathrm{mean}\,\min(\mathrm{NRelIS}_i, c)$ with $c = 10$ as a robustness compromise.
    With $\alpha = 0.05$, a miss by $0.1\,|\tau|$ incurs $\approx 4|\tau|/(1+|\tau|)$ (well below the cap), while a miss by $|\tau|$ reaches the cap once $|\tau| \geq 1/3$. The cap bounds any single task's contribution to $c$, so the aggregates cannot be dominated by a single catastrophic miss.
    \begin{table}[ht!]
        \centering
        \caption{\textbf{Rung 1 scoring rules.}}
        \label{tab:scoring_r1}
        \footnotesize
        \setlength{\tabcolsep}{3pt}
        \begin{tabular}{@{}p{0.16\textwidth} p{0.22\textwidth} p{0.08\textwidth} p{0.50\textwidth}@{}}
            \toprule
            Task                                      & Output variant                                                                     & Metric      & Score definition                                                                                                  \\
            \midrule
            \texttt{prediction}                       & \texttt{point\_\allowbreak predictor}                                              & RMSE / AUC  & RMSE on the private test set for continuous $Y$; ROC-AUC for binary $Y\in\{0,1\}$ (Brier score when AUC is undefined on a one-class test split). The leaderboard NRel.\ Err contribution uses $\sqrt{\mathrm{Brier}}$ for binary $Y$ (Tab.~\ref{tab:scoring_base_metrics}). \\
            \texttt{prediction}                       & \texttt{prediction\_\allowbreak interval}                                          & Int.\ score & Proper interval score on the private test set; report coverage and mean width.                                    \\
            \texttt{association}                      & \texttt{sign\_\allowbreak only}                                                    & Exact       & $1$ iff answer \texttt{sign} matches the GT sign in $\{+,-,\texttt{unknown}\}$.                                   \\
            \texttt{association}                      & \texttt{effect\_\allowbreak size\_\allowbreak point}                               & Abs.\ err. & $|\widehat{r}-r|$ where $r$ is the GT Pearson correlation between treatment and outcome on the ground-truth data. \\
            \texttt{association}                      & \texttt{sign\_\allowbreak before\_\allowbreak after}                               & Exact       & $1$ iff both \texttt{sign\_before} and \texttt{sign\_after} match GT.                                             \\
            \texttt{association}                      & \texttt{delta\_\allowbreak point}                                                  & Abs.\ err.  & Absolute error on $\widehat{\Delta}$ where $\Delta$ is the GT change in association after conditioning.           \\
            \texttt{association}                      & \texttt{delta\_\allowbreak sign\_\allowbreak only}                                 & Exact       & $1$ iff answer \texttt{sign} matches the sign of the GT conditional-association change.                           \\
            \texttt{association}                      & \texttt{argmax\_\allowbreak change}                                                & Exact       & $1$ iff the selected conditioning variable matches the GT variable with the largest absolute association change.  \\
            \texttt{collider\_\allowbreak phenomenon} & \texttt{induced\_\allowbreak association\_\allowbreak boolean} & Exact & $1$ iff answer \texttt{association\_present} matches GT. \\
            \texttt{collider\_\allowbreak phenomenon} & \texttt{induced\_\allowbreak association\_\allowbreak sign\_\allowbreak only} & Exact & $1$ iff answer \texttt{sign} matches the GT induced-association sign after conditioning on the collider. \\
            \texttt{collider\_\allowbreak phenomenon} & \texttt{induced\_\allowbreak association\_\allowbreak strength\_\allowbreak point} & Abs.\ err. & Absolute error on the GT conditional association induced by conditioning on the collider. \\
            \bottomrule
        \end{tabular}
    \end{table}
    \begin{table}[ht!]
        \centering
        \caption{\textbf{Rung 2 scoring rules.}}
        \label{tab:scoring_r2}
        \footnotesize
        \setlength{\tabcolsep}{3pt}
        \begin{tabular}{@{}>{\raggedright\arraybackslash}p{0.21\textwidth} p{0.21\textwidth} p{0.10\textwidth} p{0.44\textwidth}@{}}
            \toprule
            Task                                  & Output variant                                                                       & Metric              & Score definition                                                                                                                                                                                              \\
            \midrule
            \texttt{causal\_\allowbreak sketch}   & \texttt{edges\_\allowbreak only}                                                     & Edge $F_1$          & Directed edge-set $F_1$ against GT edges in story-name space.                                                                                                                                                 \\
            \texttt{causal\_\allowbreak sketch}   & \texttt{skeleton\_\allowbreak edges}                                                 & Skel.\ $F_1$ & Undirected edge-set $F_1$ against the GT skeleton, ignoring direction. \\
            \texttt{identification}               & \texttt{one\_\allowbreak valid\_\allowbreak adjustment\_\allowbreak set} & Exact & $1$ iff answer \texttt{adjust} equals any GT valid backdoor set; if the ATE is identifiable but no valid backdoor set exists (e.g., front-door), $1$ iff the answer is the \texttt{no\_backdoor} sentinel. \\
            \texttt{identification}               & \texttt{method\_\allowbreak label}                                                   & Exact               & $1$ iff answer \texttt{method} matches GT, with \texttt{none} denoting not identifiable.                                                                                                                      \\
            \texttt{identification}               & \texttt{identifiable\_\allowbreak boolean}                                           & Exact               & $1$ iff answer \texttt{identifiable} matches the GT population-ATE identifiability.                                                                                                                           \\
            \texttt{identification}               & \texttt{minimal\_\allowbreak adjustment\_\allowbreak set\_\allowbreak size} & Exact & $1$ iff answer \texttt{k} equals the size of a GT minimal valid backdoor set; if the ATE is identifiable but no valid backdoor set exists, $1$ iff the answer is the \texttt{no\_backdoor} sentinel. \\
            \texttt{identification}               & \texttt{n\_\allowbreak valid\_\allowbreak adjustment\_\allowbreak sets} & Exact & $1$ iff answer \texttt{n} equals the number of GT valid backdoor sets. \\
            \texttt{identification}               & \texttt{all\_\allowbreak minimal\_\allowbreak adjustment\_\allowbreak sets} & Set $F_1$ & Set-level $F_1$ over predicted versus GT minimal valid backdoor sets. \\
            \texttt{effect\_\allowbreak estimate} & \texttt{ate\_\allowbreak point}                                                      & Abs.\ err. / Exact & If identifiable, $|\widehat{\tau}-\tau|$ where $\tau$ is the SCM Monte Carlo GT ATE; also report relative error. If non-identifiable, $1$ iff the answer abstains with \texttt{ate=null}. \\
            \texttt{effect\_\allowbreak estimate} & \texttt{ate\_\allowbreak uq\_\allowbreak 95} & Int.\ score / Exact & If identifiable, the proper interval score for $\tau$; report coverage and width. If non-identifiable, $1$ iff the answer abstains with \texttt{ate=null}. \\
            \texttt{effect\_\allowbreak estimate} & \texttt{ate\_\allowbreak sign\_\allowbreak only} & Exact & $1$ iff answer \texttt{sign} matches the GT ATE sign in $\{+,-,0,\texttt{unknown}\}$. \\
            \texttt{effect\_\allowbreak estimate} & \texttt{ate\_\allowbreak vs\_\allowbreak assoc\_\allowbreak sign\_\allowbreak match} & Exact & $1$ iff answer \texttt{matches} agrees with whether the GT ATE sign matches the GT observational-association sign. \\
            \texttt{bias\_\allowbreak diagnostic} & \texttt{collider\_\allowbreak bias\_\allowbreak boolean} & Exact & $1$ iff answer \texttt{bias\_present} matches GT. \\
            \texttt{bias\_\allowbreak diagnostic} & \texttt{forbidden\_\allowbreak controls\_\allowbreak list} & Set $F_1$ & Set-level $F_1$ over predicted versus GT forbidden observed controls. \\
            \bottomrule
        \end{tabular}
    \end{table}
    \begin{table}[ht!]
        \centering
        \caption{\textbf{Rung 3 scoring rules.}}
        \label{tab:scoring_r3}
        \footnotesize
        \setlength{\tabcolsep}{3pt}
        \begin{tabular}{@{}>{\raggedright\arraybackslash}p{0.21\textwidth} p{0.21\textwidth} p{0.10\textwidth} p{0.44\textwidth}@{}}
            \toprule
            Task                                                            & Output variant                                                               & Metric              & Score definition                                                                                                                                                                                                                                                \\
            \midrule
            \texttt{counter\allowbreak factual\_\allowbreak identification} & \texttt{identifiable\_\allowbreak boolean} & Exact & $1$ iff answer \texttt{identifiable} matches the GT identifiability of the target estimand (ETT, NDE, or NIE). \\
            \texttt{counter\allowbreak factual\_\allowbreak effect}         & \texttt{effect\_\allowbreak point}                                           & Abs.\ err. / Exact & If identifiable, $|\widehat{\mathrm{ETT}}-\mathrm{ETT}|$ using SCM GT; also report relative error. If non-identifiable, $1$ iff the answer abstains with \texttt{value=null}. \\
            \texttt{counter\allowbreak factual\_\allowbreak effect}         & \texttt{effect\_\allowbreak uq\_\allowbreak 95}                              & Int.\ score / Exact & If identifiable, the proper interval score for the ETT; report coverage and width. If non-identifiable, $1$ iff the answer abstains with \texttt{value=null}. \\
            \texttt{counter\allowbreak factual\_\allowbreak effect}         & \texttt{sign\_\allowbreak only}                                              & Exact & If identifiable, $1$ iff answer \texttt{sign} matches the GT ETT sign in $\{+,-,0,\texttt{unknown}\}$. If non-identifiable, $1$ iff \texttt{sign=unknown}. \\
            \texttt{mediation\_\allowbreak effect}                          & \texttt{effect\_\allowbreak point}                                           & Abs.\ err. / Exact & For NDE or NIE, if identifiable, absolute error on the requested SCM GT effect; also report relative error. If non-identifiable, $1$ iff the answer abstains with \texttt{value=null}. \\
            \texttt{mediation\_\allowbreak effect}                          & \texttt{effect\_\allowbreak uq\_\allowbreak 95}                              & Int.\ score / Exact & For NDE or NIE, if identifiable, the proper interval score for the requested effect; report coverage and width. If non-identifiable, $1$ iff the answer abstains with \texttt{value=null}. \\
            \texttt{mediation\_\allowbreak effect}                          & \texttt{sign\_\allowbreak only}                                              & Exact               & For NDE or NIE, if identifiable, $1$ iff answer \texttt{sign} matches the GT sign in $\{+,-,0,\texttt{unknown}\}$. If non-identifiable, $1$ iff \texttt{sign=unknown}.                                                                                          \\
            \texttt{mediation\_\allowbreak effect}                          & \texttt{direct\_\allowbreak vs\_\allowbreak indirect\_\allowbreak dominance} & Exact & If identifiable, $1$ iff answer \texttt{dominant} matches whether $|\mathrm{NDE}|$ exceeds $|\mathrm{NIE}|$, whether $|\mathrm{NIE}|$ exceeds $|\mathrm{NDE}|$, or whether they are tied within tolerance. If non-identifiable, $1$ iff \texttt{dominant=null}. \\
            \bottomrule
        \end{tabular}
    \end{table}

    \paragraph{Permissive band for sign tasks.}
    For the \texttt{sign\_only} variants we adopt a ``permissive'' criterion: the GT label is \texttt{0} when $|\tau| < 3\,\mathrm{sd}(Y)/\sqrt{n}$, i.e.\ when the effect is below three standard errors of an outcome mean estimated from $n$ directly observed outcome values. Inside this band both \texttt{0} and the true sign are accepted as correct answers.
    On views where the outcome is observed directly, $n$ is the training-row count. If the outcome is not directly measured (its values are released only through a measurement bundle), $n$ is the number of calibration rows, the only rows carrying gold outcome values.

    \paragraph{Abstention pool routing.}
    Some variants exist only in compatible scenes (interval prediction requires a continuous outcome, collider tasks a collider), and the exam builder samples among the available ones. Independently of that, the grader routes a task into the abstention pool whenever \emph{either} the GT estimand is non-identifiable \emph{or} the model returned a null-equivalent answer. In these cases, the per-variant content metric is replaced with a single uniform abstention binary $\mathrm{AbstMatch}_i$ (Sec.~\ref{sec:scoring}).
    This covers every variant whose estimand can be non-identifiable: the rung-2 \texttt{identification} family, the rung-2 \texttt{effect\_estimate}
    family, \texttt{bias\_diagnostic.\allowbreak forbidden\_controls\_list}, and the rung-3
    \texttt{counterfactual\_identification}, \texttt{counterfactual\_effect}, and \texttt{mediation\_effect} variants.
    For families that ask for identifiability this routing changes no score as the identifiability judgment is the content there. The task, however, is assigned to the abstention pool so the same binary is counted once. For the other families it replaces the content metric, which would be meaningless on a non-identifiable estimand.
    Each answered task therefore belongs to exactly one pool: a wrong abstention call receives a $0$ on the abstention binary but does not receive a substitute content score.
    The R3 abstention check uses the per-estimand R3 identifiability
    ($\texttt{counterfactual\_identification.}\!\langle\text{ETT,NDE,NIE}\rangle$); for the dominance variant we require both NDE and NIE to be identifiable. Null-equivalent labels (\texttt{none}, \texttt{null}, \texttt{unknown}, \texttt{n/a}, \texttt{na}, missing field, empty string) are accepted interchangeably.

    \subsection{Further exam results}
    \label{app:exam_results}

    Tables in this section further break down the results reported in the main paper (Sec.~\ref{sec:results}).
    Note that all slices (i.e., per-motif) are marginal views of the same exams: the axes are not varied independently, so a
    difference along one axis can also reflect the values of the other axes that happen to fall into that slice.

    \paragraph{Exam composition.}
    Tables~\ref{tab:realistic_composition}, \ref{tab:hard_composition}, and~\ref{tab:exam_composition_axes}
    compare the two exams. The realistic exam covers all ten task families, while the hard exam focuses on identification, bias diagnostics, and R2/R3 effect tasks. The hard exam has larger shares of front-door, \texttt{double\_nc}, and grafted scenes, and more
    of its tasks require data; the continuous/binary outcome mix is the same in both exams. Some realistic exam motif-by-family cells contain only one or two tasks, limiting comparisons at that level.
    The hard exam was constructed by targeting fixed identifiability and graph structure strata.
    Then, the task-family weights are specified per structure and output- and observation-variant weights are specified within each task family.
    The observation mix is tilted toward harder measurements: $143$ of the $167$ data-backed hard-exam tasks use \texttt{proxy\_hard} or the ``new'' \texttt{proxy\_severe} tier. The latter combines more complex
    measurement links, stronger noise, and a smaller calibration subset (App.~\ref{app:obsmodel_details}).

    We screened $686$ candidate scenes using one initial Claude Fable~5.1 response per scene to a fixed ATE prompt containing its story.
    Screening excluded $67$ scenes that triggered safety refusals; the hard exam was sampled from the $619$ retained scenes under the prescribed stratum quotas.
    \begin{table}[ht!]
        \centering
        \caption{\textbf{Realistic-exam composition.} Scenes by structural motif and task family.
        Pred: prediction; Assoc: association; Ident: identification; CF-ID: counterfactual identification; CFE: counterfactual effect.
        \texttt{grafted\_$k$} denotes scenes with $k$ auxiliary motifs anchor-grafted onto a host main graph.}
        \label{tab:realistic_composition}
        \footnotesize
        \begin{adjustbox}{max width=\textwidth}
            \begin{tabular}{lccccccccccc}
                \toprule
                \rowcolor{exammain!10} Motif          & Pred        & Assoc       & Collider   & Sketch      & Ident       & Effect      & Bias       & CF-ID       & CFE        & Mediation  & Total        \\
                \midrule
                arrowhead      & 0           & 0           & 0          & 1           & 0           & 0           & 0          & 1           & 0          & 0          & 2            \\
                chain          & 1           & 0           & 0          & 1           & 1           & 1           & 0          & 1           & 0          & 0          & 5            \\
                collider       & 0           & 0           & 0          & 1           & 0           & 0           & 1          & 1           & 0          & 0          & 3            \\
                confounding    & 6           & 1           & 0          & 3           & 2           & 3           & 3          & 4           & 3          & 0          & 25           \\
                diamond        & 0           & 0           & 0          & 0           & 0           & 0           & 0          & 0           & 0          & 1          & 1            \\
                diamondcut     & 0           & 0           & 0          & 0           & 2           & 0           & 0          & 0           & 0          & 0          & 2            \\
                double\_nc     & 0           & 0           & 0          & 0           & 0           & 0           & 0          & 0           & 1          & 0          & 1            \\
                fork           & 1           & 2           & 0          & 0           & 1           & 0           & 0          & 0           & 0          & 0          & 4            \\
                frontdoor      & 1           & 0           & 0          & 0           & 0           & 1           & 0          & 0           & 0          & 1          & 3            \\
                iv             & 2           & 3           & 0          & 8           & 4           & 3           & 3          & 3           & 1          & 0          & 27           \\
                mediation      & 0           & 1           & 0          & 2           & 0           & 1           & 0          & 0           & 0          & 0          & 4            \\
                triangle       & 1           & 0           & 0          & 0           & 0           & 1           & 0          & 0           & 0          & 0          & 2            \\
                grafted\_1     & 2           & 3           & 2          & 1           & 3           & 3           & 0          & 0           & 2          & 0          & 16           \\
                grafted\_2     & 1           & 1           & 0          & 0           & 1           & 0           & 0          & 2           & 0          & 0          & 5            \\
                \midrule
                \textbf{Total} & \textbf{15} & \textbf{11} & \textbf{2} & \textbf{17} & \textbf{14} & \textbf{13} & \textbf{7} & \textbf{12} & \textbf{7} & \textbf{2} & \textbf{100} \\
                \bottomrule
            \end{tabular}%
        \end{adjustbox}
    \end{table}
    \begin{table}[ht!]
        \centering
        \caption{\textbf{Hard-exam composition.} Scenes by structural motif and task family; columns as in Tab.~\ref{tab:realistic_composition}.}
        \label{tab:hard_composition}
        \footnotesize
        \begin{adjustbox}{max width=\textwidth}
            \begin{tabular}{lccccccccccc}
                \toprule
                \rowcolor{examhard!10} Motif & Pred & Assoc & Collider & Sketch & Ident & Effect & Bias & CF-ID & CFE & Mediation & Total \\
                \midrule
                arrowhead & 0 & 0 & 0 & 0 & 0 & 1 & 0 & 0 & 1 & 4 & 6 \\
                chain & 0 & 0 & 0 & 0 & 2 & 0 & 0 & 0 & 1 & 10 & 13 \\
                confounding & 0 & 0 & 0 & 0 & 2 & 8 & 1 & 0 & 5 & 0 & 16 \\
                diamond & 0 & 0 & 0 & 0 & 1 & 1 & 0 & 0 & 3 & 7 & 12 \\
                diamondcut & 0 & 0 & 0 & 0 & 0 & 1 & 0 & 0 & 1 & 0 & 2 \\
                double\_nc & 0 & 0 & 0 & 0 & 4 & 5 & 1 & 0 & 3 & 0 & 13 \\
                frontdoor & 0 & 0 & 0 & 0 & 5 & 13 & 0 & 0 & 6 & 6 & 30 \\
                iv & 0 & 0 & 0 & 0 & 3 & 10 & 1 & 0 & 8 & 0 & 22 \\
                mediation & 0 & 0 & 0 & 0 & 3 & 2 & 0 & 0 & 3 & 2 & 10 \\
                grafted\_1 & 0 & 0 & 0 & 0 & 0 & 15 & 6 & 0 & 17 & 1 & 39 \\
                grafted\_2 & 0 & 0 & 0 & 0 & 0 & 15 & 4 & 0 & 16 & 2 & 37 \\
                \midrule
                \textbf{Total} & \textbf{0} & \textbf{0} & \textbf{0} & \textbf{0} & \textbf{20} & \textbf{71} & \textbf{13} & \textbf{0} & \textbf{64} & \textbf{32} & \textbf{200} \\
                \bottomrule
            \end{tabular}%
        \end{adjustbox}
    \end{table}
%
    \begin{table}[ht!]
        \centering
        \caption{\textbf{Exam composition by observation variant, outcome type, and input mode.}
        Counts (percent). Observation-variant percentages refer to data-backed tasks alone.
        \textit{Outcome type} refers to the conceptual outcome variable.}
        \label{tab:exam_composition_axes}
        \footnotesize
        \setlength{\tabcolsep}{8pt}
        \renewcommand{\arraystretch}{1.1}
        \begin{adjustbox}{max width=\textwidth}
            \begin{tabular}{@{}llrr@{}}
                \toprule
                Axis & Level & \examMain[Realistic ($n=100$)] & \examHard[Hard ($n=200$)] \\
                \midrule
                \makecell[l]{Observation variant\\(data-backed only)} & \texttt{clean} & 22 (43.1\%) & 0 (0\%) \\
                                    & \texttt{proxy} & 10 (19.6\%) & 24 (14.4\%) \\
                                    & \texttt{proxy\_hard} & 19 (37.3\%) & 55 (32.9\%) \\
                                    & \texttt{proxy\_severe} & 0 (0\%) & 88 (52.7\%) \\
                \addlinespace[3pt]
                Outcome type & Continuous & 59 (59\%) & 118 (59\%) \\
                             & Binary & 41 (41\%) & 82 (41\%) \\
                \addlinespace[3pt]
                Input mode & Symbolic & 49 (49\%) & 33 (16.5\%) \\
                           & Data-backed & 51 (51\%) & 167 (83.5\%) \\
                \bottomrule
            \end{tabular}
        \end{adjustbox}
    \end{table}
    \paragraph{Performance by observation variant.}
    Table~\ref{tab:realistic_obs_variant} breaks Pass Rate and median normalized error down by observation variant (\texttt{clean}: every conceptual variable measured directly). Across the hard exam's three proxy tiers, pass rates shift by under twenty points and in no consistent direction, whereas median error grows five- to fifteen-fold from \texttt{proxy} to \texttt{proxy\_severe}, and under the severe tier the weakest models have medians many times those of the leaders. The hard-exam \texttt{clean} slice holds only symbolic binaries with a larger non-identifiable share, so its Pass Rate is not comparable to the proxy tiers'. On the realistic exam the small \texttt{proxy\_hard} slice (note its run-to-run SDs) is where pass rates are low for several closed and open models alike. The \texttt{proxy\_hard}-versus-\texttt{clean} gap survives adjustment for rung and output variant (App.~\ref{app:difficulty_knob}), and the matched-scene ablation of App.~\ref{app:observation_ablation} confirms the ordering within identical scenes.

\begin{table}[ht!]
        \centering
        \caption{\textbf{Pass Rate and median NRel.\ Err by observation variant.} Three-run mean $\pm$ SD per model and exam. Hard / Severe: \texttt{proxy\_hard} / \texttt{proxy\_severe}; NRel.\ Err over valid numeric answers. --- : empty pool (no severe variant on the realistic exam; hard-exam \texttt{clean} tasks are all binary). Rows in realistic-exam leaderboard order; best per column and exam in bold. \markclosed\ closed, \markopen\ open-weight.}
        \label{tab:realistic_obs_variant}
        \label{tab:realistic_obs_variant_pr}
        \footnotesize
        \begingroup
        \setlength{\tabcolsep}{3pt}
        \begin{adjustbox}{max width=\textwidth}
            \begin{tabular}{@{}ll cccc@{\hspace{10pt}}cccc@{}}
                \toprule
                & & \multicolumn{4}{c}{Pass Rate (\%) $\uparrow$} & \multicolumn{4}{c}{Med.\ NRel.\ Err $\downarrow$} \\
                \cmidrule(lr){3-6}\cmidrule(lr){7-10}
                Model & Exam & Clean & Proxy & Hard & Severe & Clean & Proxy & Hard & Severe \\
                \midrule
                \multirow{2}{*}{\markclosed\,GPT-6 Astra} & \examMain & $\boldsymbol{92.6 \pm 3.2}$ & $\boldsymbol{100.0 \pm 0.0}$ & $\boldsymbol{100.0 \pm 0.0}$ & --- & $0.053 \pm 0.006$ & $0.101 \pm 0.063$ & $0.147 \pm 0.004$ & --- \\
                 & \examHard & $\boldsymbol{100.0 \pm 0.0}$ & $\boldsymbol{100.0 \pm 0.0}$ & $\boldsymbol{91.3 \pm 4.3}$ & $82.9 \pm 2.7$ & --- & $0.060 \pm 0.017$ & $0.089 \pm 0.011$ & $0.331 \pm 0.063$ \\
                \addlinespace[2pt]
                \multirow{2}{*}{\markopen\,GLM 5.3} & \examMain & $74.1 \pm 3.2$ & $76.9 \pm 0.0$ & $50.0 \pm 16.7$ & --- & $0.041 \pm 0.006$ & $0.302 \pm 0.001$ & $\boldsymbol{0.092 \pm 0.026}$ & --- \\
                 & \examHard & $77.5 \pm 2.2$ & $62.6 \pm 14.1$ & $72.4 \pm 4.0$ & $73.2 \pm 2.7$ & --- & $\boldsymbol{0.025 \pm 0.000}$ & $\boldsymbol{0.060 \pm 0.008}$ & $0.297 \pm 0.085$ \\
                \addlinespace[2pt]
                \multirow{2}{*}{\markclosed\,Gemini 3.1 Pro} & \examMain & $83.3 \pm 5.6$ & $82.1 \pm 4.4$ & $38.9 \pm 9.6$ & --- & $0.068 \pm 0.047$ & $0.217 \pm 0.146$ & $0.168 \pm 0.067$ & --- \\
                 & \examHard & $82.6 \pm 4.3$ & $50.0 \pm 0.0$ & $55.1 \pm 5.0$ & $62.8 \pm 5.4$ & --- & $0.059 \pm 0.027$ & $0.183 \pm 0.003$ & $0.429 \pm 0.160$ \\
                \addlinespace[2pt]
                \multirow{2}{*}{\markopen\,DeepSeek V4.1 Flash} & \examMain & $75.9 \pm 3.2$ & $69.2 \pm 7.7$ & $77.8 \pm 19.2$ & --- & $0.061 \pm 0.038$ & $0.288 \pm 0.022$ & $0.175 \pm 0.027$ & --- \\
                 & \examHard & $84.1 \pm 9.1$ & $80.1 \pm 7.4$ & $83.1 \pm 0.4$ & $85.2 \pm 2.6$ & --- & $0.042 \pm 0.013$ & $0.093 \pm 0.028$ & $0.282 \pm 0.041$ \\
                \addlinespace[2pt]
                \multirow{2}{*}{\markclosed\,Claude Fable 5.1} & \examMain & $77.8 \pm 0.0$ & $59.0 \pm 4.4$ & $58.3 \pm 14.4$ & --- & $\boldsymbol{0.027 \pm 0.007}$ & $0.380 \pm 0.001$ & $0.114 \pm 0.010$ & --- \\
                 & \examHard & $82.6 \pm 4.3$ & $87.5 \pm 0.0$ & $85.9 \pm 2.3$ & $\boldsymbol{86.2 \pm 2.1}$ & --- & $0.032 \pm 0.001$ & $0.066 \pm 0.011$ & $\boldsymbol{0.212 \pm 0.055}$ \\
                \addlinespace[2pt]
                \multirow{2}{*}{\markopen\,Kimi K3} & \examMain & $64.8 \pm 6.4$ & $69.2 \pm 0.0$ & $66.7 \pm 33.3$ & --- & $0.042 \pm 0.010$ & $0.301 \pm 0.000$ & $0.146 \pm 0.017$ & --- \\
                 & \examHard & $62.8 \pm 3.3$ & $62.5 \pm 0.0$ & $63.8 \pm 5.0$ & $65.0 \pm 6.8$ & --- & $0.048 \pm 0.007$ & $0.147 \pm 0.044$ & $0.334 \pm 0.024$ \\
                \addlinespace[2pt]
                \multirow{2}{*}{\markopen\,K2-MoVA 36B-A4B} & \examMain & $74.7 \pm 7.7$ & $66.7 \pm 4.4$ & $80.6 \pm 17.3$ & --- & $0.122 \pm 0.043$ & $0.301 \pm 0.000$ & $0.180 \pm 0.070$ & --- \\
                 & \examHard & $68.6 \pm 4.7$ & $70.0 \pm 0.0$ & $68.9 \pm 4.4$ & $75.2 \pm 5.4$ & --- & $0.037 \pm 0.006$ & $0.112 \pm 0.012$ & $0.307 \pm 0.005$ \\
                \addlinespace[2pt]
                \multirow{2}{*}{\markclosed\,Claude Opus 4.8} & \examMain & $88.9 \pm 0.0$ & $79.5 \pm 4.4$ & $44.4 \pm 19.2$ & --- & $0.135 \pm 0.014$ & $\boldsymbol{0.099 \pm 0.000}$ & $0.216 \pm 0.040$ & --- \\
                 & \examHard & $82.6 \pm 4.3$ & $68.1 \pm 9.0$ & $61.1 \pm 5.0$ & $58.5 \pm 3.0$ & --- & $0.128 \pm 0.082$ & $0.196 \pm 0.036$ & $0.709 \pm 0.312$ \\
                \addlinespace[2pt]
                \multirow{2}{*}{\markopen\,Kimi K2.6} & \examMain & $78.5 \pm 13.5$ & $76.9 \pm 0.0$ & $66.7 \pm 33.3$ & --- & $0.108 \pm 0.100$ & $0.258 \pm 0.075$ & $0.180 \pm 0.018$ & --- \\
                 & \examHard & $76.8 \pm 2.5$ & $83.8 \pm 7.6$ & $72.4 \pm 8.8$ & $81.6 \pm 2.5$ & --- & $0.069 \pm 0.048$ & $0.215 \pm 0.052$ & $0.479 \pm 0.142$ \\
                \addlinespace[2pt]
                \multirow{2}{*}{\markopen\,Qwen 3.6 35B} & \examMain & $69.2 \pm 5.2$ & $79.5 \pm 4.4$ & $16.7 \pm 16.7$ & --- & $0.220 \pm 0.023$ & $0.164 \pm 0.120$ & $0.186 \pm 0.080$ & --- \\
                 & \examHard & $67.7 \pm 7.5$ & $69.2 \pm 15.9$ & $53.1 \pm 2.8$ & $52.9 \pm 4.4$ & --- & $0.233 \pm 0.025$ & $0.248 \pm 0.054$ & $1.253 \pm 0.300$ \\
                \addlinespace[2pt]
                \multirow{2}{*}{\markclosed\,GPT-5.5} & \examMain & $85.5 \pm 3.0$ & $77.3 \pm 8.2$ & $72.2 \pm 25.5$ & --- & $0.182 \pm 0.040$ & $0.215 \pm 0.100$ & $0.216 \pm 0.051$ & --- \\
                 & \examHard & $87.2 \pm 11.4$ & $80.0 \pm 0.0$ & $62.7 \pm 1.8$ & $62.0 \pm 2.4$ & --- & $0.117 \pm 0.064$ & $1.226 \pm 0.846$ & $1.606 \pm 0.598$ \\
                \addlinespace[2pt]
                \multirow{2}{*}{\markopen\,Gemma 4 26B} & \examMain & $51.9 \pm 8.5$ & $76.9 \pm 7.7$ & $27.8 \pm 25.5$ & --- & $0.213 \pm 0.014$ & $0.113 \pm 0.023$ & $0.333 \pm 0.021$ & --- \\
                 & \examHard & $50.0 \pm 2.2$ & $57.4 \pm 8.5$ & $46.4 \pm 6.2$ & $45.0 \pm 3.5$ & --- & $0.215 \pm 0.058$ & $0.947 \pm 0.475$ & $3.344 \pm 1.268$ \\
                \bottomrule
            \end{tabular}
        \end{adjustbox}
        \endgroup
    \end{table}
    \paragraph{Content correctness and abstention.}
    Table~\ref{tab:realistic_pr_split} splits Pass Rate into the content pool and the abstention pool; both can be graded because identifiability is determined by the hidden SCM (Sec.~\ref{sec:scoring}). On the realistic exam content PR is close to saturation for most models, so the spread in Pass Rate is almost entirely abstention: GPT-6 Astra abstains almost exactly when the ground truth requires it, whereas several models that answer nearly every identifiable target correctly abstain on only about half of the non-identifiable ones. On the hard exam content PR spreads as well, and Astra's lead there comes from abstention; its content PR is below that of several open models. The two abstention errors differ across exams: on the realistic exam missed abstention, committing to a number the graph does not identify, dominates for every model; on the hard exam false abstention, refusing an identifiable target, becomes substantial for weaker and for more cautious models, and for GPT-5.5 exceeds its missed abstention. App.~\ref{app:error_taxonomy} traces both errors to the identification step and to the motifs on which each occurs.

\begin{table}[ht!]
        \centering
        \caption{\textbf{Pass Rate decomposition.} Three-run mean $\pm$ SD (PR: mean only); parentheses: per-run pool size or range. The abstention pool holds non-identifiable targets and model abstentions (App.~\ref{app:scoring_rules}); its PR, missed abstention (answered, or answered invalidly, where abstention was required), and false abstention (abstained on an identifiable target) sum to $100\%$. Rows sorted by mean PR over the two exams; best per column and exam in bold. \markclosed\ closed, \markopen\ open-weight.}
        \label{tab:realistic_pr_split}
        \footnotesize
        \begingroup
        \setlength{\tabcolsep}{3pt}
        \begin{adjustbox}{max width=\textwidth}
            \begin{tabular}{@{}ll ccc@{\hspace{12pt}}cc@{}}
                \toprule
                & & \multicolumn{3}{c}{Pass rates (\%) $\uparrow$} & \multicolumn{2}{c}{Abstention errors (\%) $\downarrow$} \\
                \cmidrule(lr){3-5}\cmidrule(lr){6-7}
                Model & Exam & PR & Content PR ($N$) & Abstention PR ($N$) & Missed abstention & False abstention \\
                \midrule
                \multirow{2}{*}{\markclosed\,GPT-6 Astra} & \examMain & $\boldsymbol{96.1}$ & $96.3 \pm 3.2$ (18) & $\boldsymbol{95.8 \pm 3.6}$ (16) & $\boldsymbol{4.2 \pm 3.6}$ & $\boldsymbol{0.0 \pm 0.0}$ \\
                 & \examHard & $\boldsymbol{90.4}$ & $77.9 \pm 0.3$ (40--41) & $\boldsymbol{99.4 \pm 1.0}$ (56--57) & $\boldsymbol{0.0 \pm 0.0}$ & $\boldsymbol{0.6 \pm 1.0}$ \\
                \addlinespace[2pt]
                \multirow{2}{*}{\markopen\,DeepSeek V4.1 Flash} & \examMain & $73.5$ & $98.0 \pm 3.4$ (17--18) & $48.0 \pm 1.7$ (16--17) & $48.0 \pm 1.7$ & $3.9 \pm 3.4$ \\
                 & \examHard & $83.9$ & $85.1 \pm 2.5$ (40--41) & $83.0 \pm 2.8$ (60--62) & $8.8 \pm 4.3$ & $8.2 \pm 1.5$ \\
                \addlinespace[2pt]
                \multirow{2}{*}{\markopen\,Kimi K2.6} & \examMain & $76.8$ & $98.0 \pm 3.4$ (17--18) & $54.6 \pm 15.0$ (16--18) & $41.7 \pm 9.5$ & $3.7 \pm 6.4$ \\
                 & \examHard & $78.5$ & $85.0 \pm 4.3$ (40) & $74.2 \pm 3.4$ (60--61) & $18.1 \pm 3.3$ & $7.7 \pm 0.9$ \\
                \addlinespace[2pt]
                \multirow{2}{*}{\markclosed\,Claude Fable 5.1} & \examMain & $68.6$ & $82.5 \pm 3.0$ (19) & $52.1 \pm 3.6$ (16) & $47.9 \pm 3.6$ & $\boldsymbol{0.0 \pm 0.0}$ \\
                 & \examHard & $85.4$ & $87.1 \pm 3.5$ (41--42) & $84.1 \pm 1.8$ (56--57) & $14.7 \pm 2.1$ & $1.2 \pm 1.0$ \\
                \addlinespace[2pt]
                \multirow{2}{*}{\markclosed\,GPT-5.5} & \examMain & $81.0$ & $88.9 \pm 5.6$ (18) & $72.5 \pm 6.9$ (16--18) & $21.8 \pm 9.6$ & $5.7 \pm 5.6$ \\
                 & \examHard & $69.0$ & $63.5 \pm 4.9$ (42--47) & $72.7 \pm 2.6$ (66--70) & $10.5 \pm 4.7$ & $16.8 \pm 2.8$ \\
                \addlinespace[2pt]
                \multirow{2}{*}{\markclosed\,Claude Opus 4.8} & \examMain & $81.4$ & $\boldsymbol{100.0 \pm 0.0}$ (18) & $60.4 \pm 3.6$ (16) & $39.6 \pm 3.6$ & $\boldsymbol{0.0 \pm 0.0}$ \\
                 & \examHard & $65.2$ & $78.5 \pm 1.3$ (40--41) & $56.8 \pm 1.8$ (62--67) & $30.8 \pm 2.8$ & $12.4 \pm 3.6$ \\
                \addlinespace[2pt]
                \multirow{2}{*}{\markopen\,GLM 5.3} & \examMain & $72.8$ & $96.4 \pm 3.1$ (18--19) & $45.8 \pm 3.6$ (16) & $54.2 \pm 3.6$ & $\boldsymbol{0.0 \pm 0.0}$ \\
                 & \examHard & $73.0$ & $81.8 \pm 5.9$ (40--44) & $67.2 \pm 1.9$ (62--64) & $21.7 \pm 0.6$ & $11.1 \pm 1.4$ \\
                \addlinespace[2pt]
                \multirow{2}{*}{\markopen\,K2-MoVA 36B-A4B} & \examMain & $72.1$ & $98.0 \pm 3.4$ (16--17) & $47.9 \pm 12.9$ (17--19) & $41.1 \pm 16.0$ & $10.9 \pm 5.0$ \\
                 & \examHard & $71.7$ & $85.7 \pm 4.0$ (39--40) & $63.4 \pm 3.5$ (66--69) & $19.8 \pm 1.9$ & $16.8 \pm 1.9$ \\
                \addlinespace[2pt]
                \multirow{2}{*}{\markclosed\,Gemini 3.1 Pro} & \examMain & $78.7$ & $96.3 \pm 3.2$ (18) & $59.2 \pm 3.1$ (16--17) & $38.8 \pm 4.4$ & $2.0 \pm 3.4$ \\
                 & \examHard & $64.5$ & $\boldsymbol{87.7 \pm 0.2}$ (40--41) & $48.3 \pm 3.7$ (57--59) & $48.3 \pm 2.3$ & $3.4 \pm 1.7$ \\
                \addlinespace[2pt]
                \multirow{2}{*}{\markopen\,Kimi K3} & \examMain & $66.7$ & $96.3 \pm 6.4$ (17--18) & $34.4 \pm 11.4$ (16--17) & $63.6 \pm 14.7$ & $2.0 \pm 3.4$ \\
                 & \examHard & $64.0$ & $86.1 \pm 1.5$ (40--41) & $48.6 \pm 2.2$ (57--60) & $47.5 \pm 2.5$ & $4.0 \pm 2.5$ \\
                \addlinespace[2pt]
                \multirow{2}{*}{\markopen\,Qwen 3.6 35B} & \examMain & $66.5$ & $89.8 \pm 7.7$ (17--21) & $41.2 \pm 5.9$ (17) & $52.9 \pm 5.9$ & $5.9 \pm 0.0$ \\
                 & \examHard & $57.6$ & $76.6 \pm 3.5$ (37--40) & $46.9 \pm 5.1$ (67--71) & $34.3 \pm 4.2$ & $18.8 \pm 2.4$ \\
                \addlinespace[2pt]
                \multirow{2}{*}{\markopen\,Gemma 4 26B} & \examMain & $59.2$ & $88.5 \pm 0.4$ (17--18) & $29.1 \pm 9.0$ (16--18) & $65.2 \pm 14.0$ & $5.7 \pm 5.6$ \\
                 & \examHard & $47.4$ & $69.5 \pm 6.5$ (39--40) & $35.2 \pm 3.3$ (66--73) & $44.6 \pm 2.7$ & $20.2 \pm 4.4$ \\
                \bottomrule
            \end{tabular}
        \end{adjustbox}
        \endgroup
    \end{table}

    \begin{table}[ht!]
        \centering
        \caption{\textbf{Per-rung and per-family aggregates on the realistic exam.} Three-run means; row blocks are \texttt{rung:family} slices with the task count $n$; each rung block opens with the pooled per-rung slice. Family abbreviations as in Tab.~\ref{tab:realistic_composition}.  --- marks an empty pool, bold marks the best model per row. Model markers: \markclosed\ closed, \markopen\ open-weight.}
        \label{tab:realistic_per_task_type}\label{tab:realistic_abstention_by_family}\label{tab:realistic_perrung}
        \scriptsize
        \begingroup
        \setlength{\tabcolsep}{2.5pt}
        \begin{adjustbox}{max width=\textwidth}
            \begin{tabular}{@{}l l *{12}{c}@{}}
            \toprule
            \rowcolor{exammain!10} Family ($n$) & Metric & \markclosed\,GPT-6 & \markopen\,GLM & \markclosed\,Gemini & \markopen\,DeepSeek & \markclosed\,Fable & \markopen\,K3 & \markopen\,K2-MoVA & \markclosed\,Opus & \markopen\,K2.6 & \markopen\,Qwen & \markclosed\,GPT-5.5 & \markopen\,Gemma \\
            \midrule
            \makecell[l]{\texttt{R1:all}\\(28)} & PR (\%) & \textbf{100.0} & \textbf{100.0} & \textbf{100.0} & \textbf{100.0} & \textbf{100.0} & \textbf{100.0} & \textbf{100.0} & \textbf{100.0} & \textbf{100.0} & 91.7 & 91.7 & \textbf{100.0} \\
             & Cont.\ PR (\%) & \textbf{100.0} & \textbf{100.0} & \textbf{100.0} & \textbf{100.0} & \textbf{100.0} & \textbf{100.0} & \textbf{100.0} & \textbf{100.0} & \textbf{100.0} & 91.7 & 91.7 & \textbf{100.0} \\
             & Med.\ NRel & 0.272 & 0.281 & 0.283 & \textbf{0.271} & 0.272 & 0.291 & 0.281 & 0.318 & 0.273 & 0.283 & 0.291 & 0.348 \\
            \addlinespace[3pt]
            \makecell[l]{\texttt{R1:Pred}\\(15)} & Med.\ NRel & 0.530 & 0.530 & 0.530 & 0.530 & \textbf{0.491} & 0.530 & 0.531 & 0.531 & 0.532 & 0.546 & 0.671 & 0.538 \\
            \addlinespace[3pt]
            \makecell[l]{\texttt{R1:Assoc}\\(11)} & PR (\%) & 100.0 & 100.0 & 100.0 & 100.0 & 100.0 & 100.0 & 100.0 & 100.0 & 100.0 & 100.0 & 100.0 & 100.0 \\
             & Cont.\ PR (\%) & 100.0 & 100.0 & 100.0 & 100.0 & 100.0 & 100.0 & 100.0 & 100.0 & 100.0 & 100.0 & 100.0 & 100.0 \\
             & Med.\ NRel & 0.001 & 0.001 & 0.001 & 0.001 & 0.001 & 0.001 & 0.001 & 0.001 & 0.001 & 0.001 & 0.001 & 0.001 \\
            \addlinespace[3pt]
            \makecell[l]{\texttt{R1:Collider}\\(2)} & PR (\%) & \textbf{100.0} & \textbf{100.0} & \textbf{100.0} & \textbf{100.0} & \textbf{100.0} & \textbf{100.0} & \textbf{100.0} & \textbf{100.0} & \textbf{100.0} & 83.3 & 83.3 & \textbf{100.0} \\
             & Cont.\ PR (\%) & \textbf{100.0} & \textbf{100.0} & \textbf{100.0} & \textbf{100.0} & \textbf{100.0} & \textbf{100.0} & \textbf{100.0} & \textbf{100.0} & \textbf{100.0} & 83.3 & 83.3 & \textbf{100.0} \\
             & Med.\ NRel & 0.013 & 0.009 & 0.010 & 0.009 & \textbf{0.004} & 0.014 & 0.013 & 0.161 & 0.112 & --- & --- & 0.164 \\
            \midrule
            \makecell[l]{\texttt{R2:all}\\(51)} & PR (\%) & \textbf{100.0} & 45.2 & 62.7 & 47.6 & 54.8 & 35.7 & 56.7 & 57.1 & 54.8 & 47.8 & 88.1 & 37.1 \\
             & Cont.\ PR (\%) & \textbf{100.0} & \textbf{100.0} & \textbf{100.0} & 88.9 & \textbf{100.0} & 88.9 & \textbf{100.0} & \textbf{100.0} & \textbf{100.0} & \textbf{100.0} & \textbf{100.0} & \textbf{100.0} \\
             & Abst.\ PR (\%) & \textbf{100.0} & 30.3 & 52.8 & 36.4 & 42.4 & 21.2 & 45.5 & 45.5 & 42.4 & 34.3 & 84.8 & 20.5 \\
             & Med.\ NRel & \textbf{0.059} & 0.071 & 0.127 & 0.092 & 0.094 & 0.084 & 0.181 & 0.161 & 0.146 & 0.380 & 1.069 & 2.101 \\
            \addlinespace[3pt]
            \makecell[l]{\texttt{R2:Ident}\\(14)} & PR (\%) & \textbf{100.0} & 47.6 & 90.5 & 47.6 & 57.1 & 38.1 & 54.8 & 90.5 & 61.9 & 71.4 & 90.5 & 38.1 \\
             & Cont.\ PR (\%) & \textbf{100.0} & \textbf{100.0} & \textbf{100.0} & 83.3 & \textbf{100.0} & 83.3 & \textbf{100.0} & \textbf{100.0} & \textbf{100.0} & \textbf{100.0} & \textbf{100.0} & \textbf{100.0} \\
             & Abst.\ PR (\%) & \textbf{100.0} & 26.7 & 86.7 & 33.3 & 40.0 & 20.0 & 37.8 & 86.7 & 46.7 & 60.0 & 86.7 & 13.3 \\
            \addlinespace[3pt]
            \makecell[l]{\texttt{R2:Effect}\\(13)} & PR (\%) & \textbf{100.0} & 55.6 & 8.3 & 77.8 & 44.4 & 44.4 & 80.6 & 11.1 & 55.6 & 8.3 & 88.9 & 30.6 \\
             & Abst.\ PR (\%) & \textbf{100.0} & 55.6 & 8.3 & 77.8 & 44.4 & 44.4 & 80.6 & 11.1 & 55.6 & 8.3 & 88.9 & 30.6 \\
             & Med.\ NRel & \textbf{0.059} & 0.071 & 0.127 & 0.092 & 0.094 & 0.084 & 0.181 & 0.161 & 0.146 & 0.380 & 1.069 & 2.101 \\
            \addlinespace[3pt]
            \makecell[l]{\texttt{R2:Bias}\\(7)} & PR (\%) & \textbf{100.0} & 33.3 & 58.3 & 25.0 & 58.3 & 25.0 & 41.7 & 33.3 & 41.7 & 41.7 & 83.3 & 41.7 \\
             & Cont.\ PR (\%) & 100.0 & 100.0 & 100.0 & 100.0 & 100.0 & 100.0 & 100.0 & 100.0 & 100.0 & 100.0 & 100.0 & 100.0 \\
             & Abst.\ PR (\%) & \textbf{100.0} & 11.1 & 44.4 & 0.0 & 44.4 & 0.0 & 22.2 & 11.1 & 22.2 & 22.2 & 77.8 & 22.2 \\
            \midrule
            \makecell[l]{\texttt{R3:all}\\(21)} & PR (\%) & 92.2 & 90.4 & 88.2 & 90.2 & 74.1 & 86.3 & 80.4 & \textbf{98.0} & 90.6 & 77.5 & 73.5 & 70.6 \\
             & Cont.\ PR (\%) & 94.4 & 94.7 & 94.4 & \textbf{100.0} & 74.4 & 97.2 & 97.0 & \textbf{100.0} & 97.0 & 87.0 & 85.6 & 82.3 \\
             & Abst.\ PR (\%) & 86.7 & 80.0 & 73.3 & 71.1 & 73.3 & 63.3 & 52.4 & \textbf{93.3} & 79.0 & 55.6 & 49.0 & 46.7 \\
             & Med.\ NRel & 0.033 & \textbf{0.024} & 0.033 & 0.042 & 0.027 & 0.047 & 0.094 & 0.026 & 0.049 & 0.085 & 0.037 & 0.091 \\
            \addlinespace[3pt]
            \makecell[l]{\texttt{R3:CF-ID}\\(12)} & PR (\%) & 94.4 & 94.4 & 94.4 & 91.7 & 72.2 & 88.9 & 83.3 & \textbf{97.2} & 94.4 & 91.7 & 88.9 & 77.8 \\
             & Cont.\ PR (\%) & \textbf{100.0} & \textbf{100.0} & 96.3 & \textbf{100.0} & 74.1 & \textbf{100.0} & 96.3 & \textbf{100.0} & \textbf{100.0} & \textbf{100.0} & \textbf{100.0} & 88.4 \\
             & Abst.\ PR (\%) & 77.8 & 77.8 & \textbf{88.9} & 66.7 & 66.7 & 61.1 & 50.0 & \textbf{88.9} & 77.8 & 66.7 & 61.1 & 50.0 \\
            \addlinespace[3pt]
            \makecell[l]{\texttt{R3:CFE}\\(7)} & PR (\%) & 83.3 & 83.3 & 66.7 & \textbf{100.0} & 91.7 & 75.0 & 91.7 & \textbf{100.0} & 86.7 & 46.7 & 41.7 & 50.0 \\
             & Cont.\ PR (\%) & 66.7 & 83.3 & 83.3 & \textbf{100.0} & \textbf{100.0} & 83.3 & \textbf{100.0} & \textbf{100.0} & 83.3 & 44.4 & 44.4 & 50.0 \\
             & Abst.\ PR (\%) & \textbf{100.0} & 83.3 & 50.0 & \textbf{100.0} & 83.3 & 66.7 & 83.3 & \textbf{100.0} & 88.9 & 50.0 & 44.4 & 50.0 \\
             & Med.\ NRel & 0.046 & \textbf{0.015} & 0.025 & 0.055 & 0.027 & 0.023 & 0.097 & 0.040 & 0.058 & 0.101 & 0.069 & 0.115 \\
            \addlinespace[3pt]
            \makecell[l]{\texttt{R3:Mediation}\\(2)} & PR (\%) & \textbf{100.0} & 83.3 & \textbf{100.0} & 33.3 & 50.0 & \textbf{100.0} & 0.0 & \textbf{100.0} & 66.7 & 50.0 & 33.3 & 66.7 \\
             & Cont.\ PR (\%) & \textbf{100.0} & 83.3 & \textbf{100.0} & --- & 50.0 & \textbf{100.0} & --- & \textbf{100.0} & --- & --- & --- & --- \\
             & Abst.\ PR (\%) & --- & --- & --- & --- & --- & --- & 0.0 & --- & --- & --- & --- & --- \\
             & Med.\ NRel & \textbf{0.004} & --- & 0.040 & 0.019 & --- & 0.089 & 0.090 & 0.011 & 0.014 & --- & 0.015 & 0.012 \\
            \bottomrule
            \end{tabular}
        \end{adjustbox}
        \endgroup
    \end{table}
%
    \begin{table}[ht!]
        \centering
        \caption{\textbf{Per-rung and per-family aggregates on the hard exam.} Three-run means.  Notation same as in Tab.~\ref{tab:realistic_per_task_type}.}
        \label{tab:hard_per_task_type}
        \label{tab:hard_perrung}
        \scriptsize
        \begingroup
        \setlength{\tabcolsep}{2.5pt}
        \begin{adjustbox}{max width=\textwidth}
            \begin{tabular}{@{}l l *{12}{c}@{}}
            \toprule
            \rowcolor{examhard!10} Family ($n$) & Metric & \markopen\,GLM & \markclosed\,Fable & \markopen\,DeepSeek & \markclosed\,GPT-6 & \markopen\,K2.6 & \markclosed\,Gemini & \markopen\,K3 & \markopen\,K2-MoVA & \markclosed\,Opus & \markopen\,Qwen & \markclosed\,GPT-5.5 & \markopen\,Gemma \\
            \midrule
            \makecell[l]{\texttt{R2:all}\\(104)} & PR (\%) & 71.9 & 84.4 & 85.8 & \textbf{100.0} & 74.8 & 59.6 & 51.8 & 68.6 & 62.9 & 49.7 & 83.0 & 41.3 \\
             & Cont.\ PR (\%) & \textbf{100.0} & \textbf{100.0} & \textbf{100.0} & \textbf{100.0} & \textbf{100.0} & \textbf{100.0} & \textbf{100.0} & \textbf{100.0} & \textbf{100.0} & \textbf{100.0} & \textbf{100.0} & 80.1 \\
             & Abst.\ PR (\%) & 65.5 & 80.6 & 82.4 & \textbf{100.0} & 69.0 & 49.6 & 40.0 & 61.8 & 54.4 & 40.2 & 79.5 & 34.2 \\
             & Med.\ NRel & 0.121 & \textbf{0.105} & 0.144 & 0.183 & 0.267 & 0.271 & 0.294 & 0.157 & 0.256 & 0.414 & 0.762 & 1.322 \\
            \addlinespace[3pt]
            \makecell[l]{\texttt{R2:Ident}\\(20)} & PR (\%) & 86.7 & 88.2 & 92.2 & \textbf{100.0} & 88.2 & 90.2 & 78.8 & 84.6 & 94.1 & 76.8 & 92.5 & 57.6 \\
             & Cont.\ PR (\%) & \textbf{100.0} & \textbf{100.0} & \textbf{100.0} & \textbf{100.0} & \textbf{100.0} & \textbf{100.0} & \textbf{100.0} & \textbf{100.0} & \textbf{100.0} & \textbf{100.0} & \textbf{100.0} & 80.1 \\
             & Abst.\ PR (\%) & 74.1 & 75.0 & 83.3 & \textbf{100.0} & 75.0 & 79.2 & 55.6 & 69.4 & 87.5 & 59.1 & 86.3 & 37.4 \\
            \addlinespace[3pt]
            \makecell[l]{\texttt{R2:Effect}\\(71)} & PR (\%) & 66.1 & 86.4 & 87.4 & \textbf{100.0} & 72.9 & 35.9 & 40.3 & 68.9 & 45.2 & 33.0 & 79.3 & 34.7 \\
             & Abst.\ PR (\%) & 66.1 & 86.4 & 87.4 & \textbf{100.0} & 72.9 & 35.9 & 40.3 & 68.9 & 45.2 & 33.0 & 79.3 & 34.7 \\
             & Med.\ NRel & 0.121 & \textbf{0.105} & 0.144 & 0.183 & 0.267 & 0.271 & 0.294 & 0.157 & 0.256 & 0.414 & 0.762 & 1.322 \\
            \addlinespace[3pt]
            \makecell[l]{\texttt{R2:Bias}\\(13)} & PR (\%) & 50.0 & 66.7 & 61.1 & \textbf{100.0} & 44.4 & 61.1 & 16.7 & 22.2 & 50.0 & 38.9 & 72.2 & 27.8 \\
             & Abst.\ PR (\%) & 50.0 & 66.7 & 61.1 & \textbf{100.0} & 44.4 & 61.1 & 16.7 & 22.2 & 50.0 & 38.9 & 72.2 & 27.8 \\
            \midrule
            \makecell[l]{\texttt{R3:all}\\(96)} & PR (\%) & 73.8 & \textbf{86.2} & 82.6 & 82.1 & 81.8 & 68.8 & 74.4 & 74.3 & 67.2 & 64.8 & 58.0 & 53.2 \\
             & Cont.\ PR (\%) & 77.0 & 83.6 & 80.8 & 71.6 & 80.6 & \textbf{84.2} & 82.1 & 81.7 & 72.3 & 70.6 & 54.9 & 66.7 \\
             & Abst.\ PR (\%) & 69.9 & 90.3 & 85.1 & \textbf{98.4} & 83.3 & 46.2 & 63.2 & 66.0 & 60.7 & 58.0 & 62.9 & 36.9 \\
             & Med.\ NRel & \textbf{0.121} & 0.151 & 0.151 & 0.141 & 0.200 & 0.193 & 0.173 & 0.175 & 0.300 & 0.525 & 0.809 & 0.968 \\
            \addlinespace[3pt]
            \makecell[l]{\texttt{R3:CFE}\\(64)} & PR (\%) & 75.5 & \textbf{85.5} & 85.1 & 81.3 & 81.0 & 63.5 & 70.6 & 77.2 & 67.2 & 62.5 & 56.3 & 46.6 \\
             & Cont.\ PR (\%) & 70.6 & 78.2 & 78.7 & 64.4 & 78.3 & \textbf{80.6} & 77.4 & 79.6 & 67.2 & 64.3 & 43.5 & 56.7 \\
             & Abst.\ PR (\%) & 80.4 & 93.3 & 91.0 & \textbf{98.4} & 83.3 & 46.9 & 64.1 & 75.3 & 67.0 & 61.1 & 68.9 & 38.3 \\
             & Med.\ NRel & \textbf{0.125} & 0.139 & 0.135 & 0.151 & 0.182 & 0.170 & 0.189 & 0.206 & 0.404 & 0.378 & 0.897 & 0.999 \\
            \addlinespace[3pt]
            \makecell[l]{\texttt{R3:Mediation}\\(32)} & PR (\%) & 68.9 & \textbf{88.9} & 74.8 & 84.8 & 84.8 & 88.4 & 88.4 & 66.4 & 67.6 & 73.1 & 63.7 & 77.9 \\
             & Cont.\ PR (\%) & 88.9 & \textbf{93.9} & 84.8 & 84.8 & 84.8 & 90.9 & 90.9 & 85.6 & 81.8 & 81.8 & 76.5 & 84.8 \\
             & Abst.\ PR (\%) & 0.0 & --- & --- & --- & --- & --- & --- & 0.0 & 0.0 & 0.0 & 0.0 & --- \\
             & Med.\ NRel & \textbf{0.109} & 0.170 & 0.161 & 0.119 & 0.220 & 0.284 & 0.171 & 0.180 & 0.215 & 1.233 & 0.939 & 1.073 \\
            \bottomrule
            \end{tabular}
        \end{adjustbox}
        \endgroup
    \end{table}
%
    \paragraph{Per-rung and per-family aggregates (Tabs.~\ref{tab:realistic_perrung}, \ref{tab:hard_perrung}).}
    On the realistic exam, R1 binary performance is near ceiling: ten of twelve models score $100\%$, with Qwen 3.6 35B and GPT-5.5 at $91.7\%$. Association shows essentially no model separation, while prediction
    Med.\ NRel.\ Err spans $0.491$--$0.671$. The $17$ causal-sketch tasks are also recovered almost perfectly
    (cf. App.~\ref{app:audit_details}); set-valued losses are omitted from these tables for this reason. They are reported separately in Tab.~\ref{tab:realistic_passk_continuous}.

    On both exams, most R2 binary separation comes from abstention. Content PR is $100\%$ for ten models on
    the realistic exam and eleven on the hard exam, whereas abstention PR ranges over $20.5$--$100\%$ and $34.2$--$100\%$,
    respectively. GPT-6 Astra is the only model perfect across all R2 abstention families on both exams.
    Other models vary substantially by family: on the realistic exam, Claude Opus 4.8 reaches $86.7\%$ abstention PR
    for identification but $11.1\%$ for both effect estimation and bias diagnostics. Numeric effect-estimation
    accuracy also separates models: GPT-6 Astra has the lowest Med.\ NRel.\ Err on the realistic exam ($0.059$),
    and Claude Fable 5.1 on the hard exam ($0.105$); GPT-5.5 and Gemma 4 26B have the largest errors on both exams.
    Thus GPT-5.5's relatively strong R2 abstention performance does not extend to estimation accuracy, which is related to its near-one-shot strategy as discussed in Sec.~\ref{sec:results}.

    R3 content performance varies more than R2 on both exams: content PR starts to become a genuine challenge,  spanning $74.4$--$100\%$ on the realistic exam and $54.9$--$84.2\%$ on the hard exam. 
    Claude Opus 4.8 leads R3 abstention on the realistic exam ($93.3\%$), while
    GPT-6 Astra again leads on the hard exam ($98.4\%$). Every model has higher R3 median numeric error on the hard exam, with
    Med.\ NRel.\ Err ranging over $0.121$--$0.968$, compared with $0.024$--$0.094$ on the realistic exam.
    On the hard exam, every model has lower content PR for counterfactual effects than for mediation effects.
    The latter task family grows from two realistic-exam tasks to $32$ hard-exam tasks.
    
%
%
    \begin{table}[ht!]
        \centering
        \caption{\textbf{Per-motif aggregates.} Three-run means. Count in brackets. Grafted scenes split by graft count. Exam slices with fewer than 3 tasks not presented. Avg is the mean over the models with a value; bold marks the best model per row. Model markers: \markclosed\ closed, \markopen\ open-weight.}
        \label{tab:realistic_per_motif}
        \label{tab:hard_per_motif}
        \scriptsize
        \begingroup
        \setlength{\tabcolsep}{2.5pt}
        \begin{adjustbox}{max width=\textwidth}
            \begin{tabular}{@{}l l l *{13}{c}@{}}
            \toprule
            Motif & Exam ($n$) & Metric & \markclosed\,GPT-6 & \markopen\,GLM & \markclosed\,Gemini & \markopen\,DeepSeek & \markclosed\,Fable & \markopen\,K3 & \markopen\,K2-MoVA & \markclosed\,Opus & \markopen\,K2.6 & \markopen\,Qwen & \markclosed\,GPT-5.5 & \markopen\,Gemma & Avg \\
            \midrule
            \texttt{grafted\_1} & \examMain{} (16) & PR (\%) & 86.7 & 93.3 & 93.3 & 93.3 & 93.3 & 80.0 & 86.7 & \textbf{100.0} & 93.3 & 75.6 & 75.6 & 60.0 & 85.9 \\
             &  & Med.\ NRel & 0.169 & 0.179 & \textbf{0.139} & 0.227 & 0.333 & 0.186 & 0.177 & 0.206 & 0.364 & 0.299 & 0.217 & 0.252 & 0.229 \\
            \addlinespace[2pt]
             & \examHard{} (39) & PR (\%) & 79.2 & 56.2 & 53.2 & \textbf{81.2} & 65.3 & 36.8 & 64.6 & 43.8 & 72.2 & 52.1 & 46.9 & 31.2 & 56.9 \\
             &  & Med.\ NRel & 0.171 & \textbf{0.108} & 0.168 & 0.135 & 0.144 & 0.187 & 0.226 & 0.497 & 0.252 & 0.559 & 1.400 & 1.399 & 0.437 \\
            \midrule
            \texttt{iv} & \examMain{} (27) & PR (\%) & \textbf{95.2} & 40.5 & 54.8 & 45.2 & 47.6 & 31.0 & 52.4 & 54.8 & 50.0 & 40.5 & 76.2 & 33.3 & 51.8 \\
             &  & Med.\ NRel & 0.035 & \textbf{0.026} & 0.043 & 0.036 & 0.028 & 0.044 & 0.052 & 0.052 & 0.052 & 0.043 & 0.052 & 0.052 & 0.043 \\
            \addlinespace[2pt]
             & \examHard{} (22) & PR (\%) & \textbf{100.0} & 66.7 & 30.3 & 83.3 & 77.3 & 39.4 & 71.2 & 56.1 & 74.2 & 57.6 & 81.8 & 37.9 & 64.6 \\
            \midrule
            \texttt{grafted\_2} & \examMain{} (5) & PR (\%) & \textbf{100.0} & \textbf{100.0} & 83.3 & \textbf{100.0} & 50.0 & \textbf{100.0} & 72.2 & \textbf{100.0} & \textbf{100.0} & \textbf{100.0} & \textbf{100.0} & \textbf{100.0} & 92.1 \\
             &  & Med.\ NRel & 0.246 & 0.247 & 0.247 & 0.246 & 0.246 & 0.247 & 0.246 & 0.247 & 0.246 & \textbf{0.179} & 0.247 & 0.250 & 0.241 \\
            \addlinespace[2pt]
             & \examHard{} (37) & PR (\%) & 84.4 & 68.1 & 51.1 & \textbf{86.7} & \textbf{86.7} & 51.1 & 73.3 & 57.8 & 65.1 & 55.6 & 75.0 & 42.2 & 66.4 \\
             &  & Med.\ NRel & 0.142 & 0.109 & 0.138 & 0.083 & \textbf{0.082} & 0.155 & 0.087 & 0.180 & 0.105 & 0.224 & 0.337 & 0.433 & 0.173 \\
            \midrule
            \texttt{confounding} & \examMain{} (25) & PR (\%) & \textbf{100.0} & \textbf{100.0} & \textbf{100.0} & \textbf{100.0} & 83.3 & 94.4 & 94.4 & \textbf{100.0} & 95.2 & 95.2 & 85.7 & 88.9 & 94.8 \\
             &  & Med.\ NRel & 0.230 & \textbf{0.221} & 0.234 & 0.230 & 0.241 & 0.237 & 0.245 & 0.260 & 0.254 & 0.257 & 0.301 & 0.343 & 0.254 \\
            \addlinespace[2pt]
             & \examHard{} (16) & PR (\%) & \textbf{100.0} & \textbf{100.0} & \textbf{100.0} & \textbf{100.0} & \textbf{100.0} & \textbf{100.0} & \textbf{100.0} & \textbf{100.0} & 94.4 & 73.3 & 88.9 & 86.7 & 95.3 \\
             &  & Med.\ NRel & 0.148 & 0.154 & 0.219 & 0.125 & \textbf{0.108} & 0.223 & 0.152 & 0.171 & 0.854 & 0.409 & 0.576 & 1.647 & 0.399 \\
            \midrule
            \texttt{frontdoor} & \examMain{} (3) & PR (\%) & \textbf{100.0} & \textbf{100.0} & 83.3 & 33.3 & \textbf{100.0} & \textbf{100.0} & 0.0 & \textbf{100.0} & 66.7 & 33.3 & 33.3 & 66.7 & 68.1 \\
             &  & Med.\ NRel & \textbf{0.224} & 0.225 & 0.288 & 0.225 & 0.225 & 0.225 & 0.288 & 0.292 & 0.225 & 0.361 & 0.232 & 0.348 & 0.263 \\
            \addlinespace[2pt]
             & \examHard{} (30) & PR (\%) & \textbf{86.7} & 31.2 & 61.1 & 35.7 & 83.3 & 55.6 & 20.5 & 27.4 & 55.6 & 12.7 & 16.1 & 4.2 & 40.8 \\
             &  & Med.\ NRel & 0.163 & \textbf{0.121} & 0.544 & 0.254 & \textbf{0.121} & 0.284 & 0.327 & 0.311 & 0.346 & 1.472 & 0.548 & 2.025 & 0.543 \\
            \midrule
            \texttt{chain} & \examMain{} (5) & PR (\%) & \textbf{100.0} & \textbf{100.0} & \textbf{100.0} & \textbf{100.0} & 66.7 & \textbf{100.0} & \textbf{100.0} & \textbf{100.0} & \textbf{100.0} & \textbf{100.0} & \textbf{100.0} & \textbf{100.0} & 97.2 \\
             &  & Med.\ NRel & 0.293 & 0.322 & 0.384 & 0.312 & \textbf{0.094} & 0.337 & 0.304 & 0.332 & 0.322 & 0.390 & 0.302 & 0.296 & 0.307 \\
            \addlinespace[2pt]
             & \examHard{} (13) & PR (\%) & 91.7 & 92.6 & 95.8 & 91.7 & \textbf{100.0} & \textbf{100.0} & 92.6 & 87.5 & 91.7 & 87.5 & 92.6 & 95.8 & 93.3 \\
             &  & Med.\ NRel & \textbf{1.042} & 1.490 & 2.777 & 1.809 & 4.347 & 4.455 & 6.179 & 7.182 & 5.714 & 6.964 & 6.765 & 7.260 & 4.665 \\
            \midrule
            \texttt{mediation} & \examMain{} (4) & Med.\ NRel & 0.008 & 0.008 & 0.008 & 0.008 & 0.008 & 0.008 & 0.008 & 0.008 & 0.008 & 0.008 & 0.008 & 0.008 & 0.008 \\
            \addlinespace[2pt]
             & \examHard{} (10) & PR (\%) & \textbf{100.0} & \textbf{100.0} & 91.7 & \textbf{100.0} & \textbf{100.0} & \textbf{100.0} & \textbf{100.0} & \textbf{100.0} & \textbf{100.0} & \textbf{100.0} & 91.7 & 88.9 & 97.7 \\
             &  & Med.\ NRel & 0.217 & 0.337 & 1.171 & 0.275 & \textbf{0.149} & 0.445 & 1.128 & 0.670 & \textbf{0.149} & 3.589 & 0.983 & 1.637 & 0.896 \\
            \midrule
            \texttt{double\_nc} & \examHard{} (13) & PR (\%) & \textbf{100.0} & \textbf{100.0} & 97.4 & \textbf{100.0} & \textbf{100.0} & 94.9 & 92.3 & 97.4 & 97.4 & 61.5 & 94.9 & 64.1 & 91.7 \\
            \midrule

            \texttt{diamond} & \examHard{} (12) & PR (\%) & 77.8 & \textbf{100.0} & \textbf{100.0} & \textbf{100.0} & \textbf{100.0} & \textbf{100.0} & \textbf{100.0} & \textbf{100.0} & 90.5 & \textbf{100.0} & 90.5 & \textbf{100.0} & 96.6 \\
             &  & Med.\ NRel & 0.101 & \textbf{0.049} & 0.177 & 0.159 & 0.083 & 0.113 & 0.214 & 0.238 & 0.172 & 0.183 & 0.218 & 0.160 & 0.156 \\
            \midrule
            \texttt{arrowhead} & \examHard{} (6) & PR (\%) & 66.7 & \textbf{77.8} & \textbf{77.8} & 66.7 & \textbf{77.8} & 66.7 & 66.7 & 66.7 & 66.7 & 66.7 & 50.0 & 52.8 & 66.9 \\
             &  & Med.\ NRel & 0.696 & \textbf{0.166} & 1.057 & 0.239 & 0.529 & 1.888 & 1.324 & 0.640 & 0.183 & 0.766 & 2.802 & 0.828 & 0.927 \\
            \midrule
            \texttt{fork} & \examMain{} (4) & PR (\%) & \textbf{100.0} & \textbf{100.0} & \textbf{100.0} & 66.7 & \textbf{100.0} & 66.7 & \textbf{100.0} & \textbf{100.0} & \textbf{100.0} & \textbf{100.0} & \textbf{100.0} & \textbf{100.0} & 94.4 \\
             &  & Med.\ NRel & 0.002 & 0.002 & 0.002 & 0.002 & 0.002 & 0.002 & 0.002 & 0.002 & 0.002 & 0.002 & 0.002 & 0.002 & 0.002 \\
            \midrule
            \texttt{collider} & \examMain{} (3) & PR (\%) & \textbf{100.0} & \textbf{100.0} & \textbf{100.0} & \textbf{100.0} & \textbf{100.0} & \textbf{100.0} & \textbf{100.0} & \textbf{100.0} & \textbf{100.0} & \textbf{100.0} & \textbf{100.0} & 66.7 & 97.2 \\
            \bottomrule
            \end{tabular}
        \end{adjustbox}
        \endgroup
    \end{table}
%
    \paragraph{Per-motif aggregates (Tab.~\ref{tab:realistic_per_motif}).}
    The two motifs that are non-identifiable by construction, \texttt{iv} and \texttt{double\_nc}
    (App.~\ref{app:composition_grounding}), rank models mainly on the abstention call (solely for the hard exam).
    The two structures are not equally hard: on
    \texttt{double\_nc} ten of the twelve models abstain correctly on more than $92\%$ of the tasks (the smaller models --- Qwen 3.6 35B and Gemma 4 26B --- only achieve $62$--$64\%$), whereas the \texttt{iv} pass rates spread from $31.0\%$ to $95.2\%$ on the realistic exam and from $30.3\%$ to $100.0\%$ on the hard exam, with GPT-6 Astra alone at the top both times. Often, the instrument is what tempts a model into an identification it cannot support, the over-claim traced on the running example in  App.~\ref{app:error_examples}. \texttt{confounding} is the mirror image: its queries are identifiable, pass rates are
    saturated on both exams (averages $94.8\%$ and $95.3\%$), and the numeric error column separates the models instead, tightly on the realistic exam ($0.221$--$0.343$) and widely on the hard exam, where five (more recent) models stay at or below $0.154$ while Qwen 3.6 35B, GPT-5.5, Kimi K2.6, and Gemma 4 26B sit at $0.41$--$1.65$. \texttt{frontdoor}, whose effect is
    identifiable by construction, is the hard exam's discriminator: over its $30$ tasks the average pass rate is only $40.8\%$, from $86.7\%$ (GPT-6 Astra) and $83.3\%$ (Claude Fable 5.1) down to $4.2\%$ (Gemma 4 26B), the over-refusal on identifiable front-door targets that the error taxonomy (App.~\ref{app:error_taxonomy}) and the
    matched-scene ablation (App.~\ref{app:observation_ablation}) sections both isolate.
    The models that do answer also estimate well ($0.121$ for GLM 5.3 and Claude Fable 5.1). 
    
    Grafting is benign on the realistic exam (average pass rates $85.9\%$ and
    $92.1\%$ for one and two grafts, errors within $0.14$--$0.36$), but on the hard exam the grafted scenes ($76$ of the $200$ tasks) are where the field spreads:
    $31.2$--$81.2\%$ on \texttt{grafted\_1} and  $42.2$--$86.7\%$ on \texttt{grafted\_2}, with the \texttt{grafted\_1} errors of Qwen 3.6 35B, GPT-5.5, and Gemma 4 26B at $0.56$--$1.40$ against $0.11$--$0.50$ for the rest. The reason for this is that graft count, task family, and observation variant co-vary across these slices; for the hard exam, grafted motifs cover about half of counterfactual-effect and most of the bias-diagnostic tasks   (Tab.~\ref{tab:hard_composition}).
    The simple identifiable structures form a floor on both exams: \texttt{chain}, \texttt{fork}, and \texttt{collider} on the realistic exam and \texttt{mediation} and \texttt{diamond} on the hard exam are passed by  all but one or two models per row, and \texttt{diamond} is estimated tightly ($0.049$--$0.238$). 
    The exception is the hard-exam \texttt{chain} block, where pass rates stay above $87\%$ but the median error is the largest in the  table ($1.04$--$7.26$) as ten of its thirteen tasks are mediation effects (R3:Mediation in Tab.~\ref{tab:hard_per_task_type} spans $0.109$--$1.233$).
    Slices with fewer than three tasks are not included in the table: \texttt{arrowhead}, \texttt{diamond}, \texttt{diamondcut}, \texttt{double\_nc}, and \texttt{triangle} on the realistic exam and \texttt{diamondcut} on the hard exam.

%
%
    \begin{table}[ht!]
        \centering
        \caption{\textbf{Composite Rank by rung and observation variant.} Realistic and hard exam; $0$ is best. \emph{Overall} is the whole exam, the Comp.\ Rank of Tabs.~\ref{tab:realistic_leaderboard} and~\ref{tab:hard_leaderboard}; --- marks a slice absent from the exam. Rows follow realistic-exam Overall (ties by CausalDSScore); best per column and exam in bold. Model markers: \markclosed\ closed, \markopen\ open-weight.}
        \label{tab:realistic_subset_composite_rank}\label{tab:hard_subset_composite_rank}
        \footnotesize
        \begingroup
        \setlength{\tabcolsep}{3pt}
        \begin{adjustbox}{max width=\textwidth}
            \begin{tabular}{@{}l l ccc cccc c@{}}
            \toprule
            & & \multicolumn{3}{c}{Rung} & \multicolumn{4}{c}{Observation variant} & \\
            \cmidrule(lr){3-5}\cmidrule(lr){6-9}
            Model & Exam & R1 & R2 & R3 & \texttt{clean} & \texttt{proxy} & \texttt{proxy\_hard} & \texttt{proxy\_severe} & Overall \\
            \midrule
            \multirow{2}{*}{\markclosed\,GPT-6 Astra} & \examMain & 0.295 & \textbf{0.288} & 0.227 & \textbf{0.258} & \textbf{0.167} & 0.348 & --- & \textbf{0.288} \\
             & \examHard & --- & 0.258 & 0.182 & \textbf{0.250} & \textbf{0.091} & \textbf{0.045} & 0.273 & 0.258 \\
            \addlinespace[2pt]
            \multirow{2}{*}{\markopen\,GLM 5.3} & \examMain & 0.250 & 0.439 & 0.136 & 0.439 & 0.621 & 0.379 & --- & 0.348 \\
             & \examHard & --- & 0.318 & 0.273 & 0.523 & 0.500 & 0.182 & 0.227 & 0.288 \\
            \addlinespace[2pt]
            \multirow{2}{*}{\markopen\,DeepSeek V4.1 Flash} & \examMain & 0.386 & 0.470 & 0.455 & 0.409 & 0.576 & 0.318 & --- & 0.379 \\
             & \examHard & --- & \textbf{0.227} & 0.182 & 0.341 & 0.136 & 0.227 & 0.136 & 0.288 \\
            \addlinespace[2pt]
            \multirow{2}{*}{\markclosed\,Gemini 3.1 Pro} & \examMain & 0.341 & 0.409 & 0.364 & 0.409 & 0.379 & 0.561 & --- & 0.409 \\
             & \examHard & --- & 0.561 & 0.500 & 0.432 & 0.727 & 0.591 & 0.545 & 0.530 \\
            \addlinespace[2pt]
            \multirow{2}{*}{\markclosed\,Claude Opus 4.8} & \examMain & 0.614 & 0.500 & \textbf{0.045} & 0.439 & 0.273 & 0.621 & --- & 0.409 \\
             & \examHard & --- & 0.621 & 0.727 & 0.432 & 0.682 & 0.682 & 0.773 & 0.621 \\
            \addlinespace[2pt]
            \multirow{2}{*}{\markopen\,Kimi K3} & \examMain & 0.432 & 0.530 & 0.591 & 0.470 & 0.667 & 0.364 & --- & 0.470 \\
             & \examHard & --- & 0.561 & 0.409 & 0.705 & 0.682 & 0.545 & 0.409 & 0.591 \\
            \addlinespace[2pt]
            \multirow{2}{*}{\markclosed\,Claude Fable 5.1} & \examMain & \textbf{0.205} & 0.485 & 0.500 & 0.439 & 0.803 & 0.500 & --- & 0.500 \\
             & \examHard & --- & 0.288 & \textbf{0.045} & 0.432 & 0.227 & 0.136 & \textbf{0.045} & \textbf{0.227} \\
            \addlinespace[2pt]
            \multirow{2}{*}{\markclosed\,GPT-5.5} & \examMain & 0.932 & 0.500 & 0.682 & 0.500 & 0.379 & 0.561 & --- & 0.530 \\
             & \examHard & --- & 0.561 & 0.955 & 0.295 & 0.636 & 0.818 & 0.818 & 0.652 \\
            \addlinespace[2pt]
            \multirow{2}{*}{\markopen\,K2-MoVA 36B-A4B} & \examMain & 0.477 & 0.379 & 0.818 & 0.591 & 0.803 & \textbf{0.227} & --- & 0.561 \\
             & \examHard & --- & 0.561 & 0.545 & 0.614 & 0.545 & 0.455 & 0.500 & 0.530 \\
            \addlinespace[2pt]
            \multirow{2}{*}{\markopen\,Kimi K2.6} & \examMain & 0.705 & 0.545 & 0.455 & 0.470 & 0.561 & 0.576 & --- & 0.561 \\
             & \examHard & --- & 0.470 & 0.409 & 0.568 & 0.136 & 0.500 & 0.409 & 0.439 \\
            \addlinespace[2pt]
            \multirow{2}{*}{\markopen\,Qwen 3.6 35B} & \examMain & 0.795 & 0.652 & 0.773 & 0.773 & 0.364 & 0.742 & --- & 0.712 \\
             & \examHard & --- & 0.742 & 0.818 & 0.659 & 0.773 & 0.864 & 0.864 & 0.742 \\
            \addlinespace[2pt]
            \multirow{2}{*}{\markopen\,Gemma 4 26B} & \examMain & 0.568 & 0.803 & 0.955 & 0.803 & 0.409 & 0.803 & --- & 0.833 \\
             & \examHard & --- & 0.833 & 0.955 & 0.750 & 0.864 & 0.955 & 1.000 & 0.833 \\
            \bottomrule
            \end{tabular}
        \end{adjustbox}
        \endgroup
    \end{table}
%

    \paragraph{Per-axis Composite Rank (Tab.~\ref{tab:realistic_subset_composite_rank}).}
    Each cell of Tab.~\ref{tab:realistic_subset_composite_rank} is the leaderboards' Composite Rank recomputed on the tasks of that slice only.
    The three CausalDSScore grading pools (Pass Rate, $S_{\mathrm{NR}}$, Med.\ $F_1$-Loss) are first averaged over the three runs; then the twelve models are ranked on each pool, the ranks are scaled to $[0,1]$, and the cell is the mean over the grading pools the slice contains.
    Each column is therefore a ranking within its slice.
    On the realistic exam the columns split between models: GPT-6 Astra takes R2, \texttt{clean}, and \texttt{proxy} along with the overall rank, Claude Fable 5.1 takes R1, Claude Opus 4.8 R3 ($0.045$), and K2-MoVA 36B-A4B \texttt{proxy\_hard}, the one column an open-weight model wins there. 
    The hard exam concentrates the lead: Claude Fable 5.1 is best overall ($0.227$) and takes R3 and \texttt{proxy\_severe} (both $0.045$), GPT-6 Astra keeps the three lighter observation columns, and DeepSeek V4.1 Flash takes R2 ($0.227$), with GLM 5.3 and
    DeepSeek V4.1 Flash tied for third overall ($0.288$). The rank inversions between the exams are large:
    Claude Fable 5.1's move partly reflects its refusals on the realistic exam; Claude Opus 4.8 moves from $0.409$ to $0.621$ and from the best R3 rank
    to $0.727$, GPT-5.5 from $0.530$ to $0.652$ with an R3 rank of $0.955$, and Gemini 3.1 Pro from $0.409$ to $0.530$, while the two open-weight leaders rise from $0.348$ and $0.379$ to $0.288$. GPT-6 Astra's \texttt{clean} lead on both exams matches its perfect identification abstention (Tabs.~\ref{tab:realistic_per_task_type} and~\ref{tab:hard_per_task_type}). 
    With $F_1$-Loss  saturated for every model, the pools that move a slice rank are Pass Rate and the interval-sensitive $S_{\mathrm{NR}}$. 
    GPT-5.5's realistic R1 rank of $0.932$ combines the slice's lowest pass rate, shared with
    Qwen 3.6 35B, with the largest prediction error (Tab.~\ref{tab:realistic_per_task_type}); its hard-exam ranks of   $0.818$ on both proxy columns and $0.955$ on R3, and Gemma 4 26B's $0.955$--$1.000$ on the same columns, follow from poorly covering intervals in the field (Fig.~\ref{fig:ate_coverage}; App.~\ref{app:error_intervals}), while
    Tab.~\ref{tab:realistic_obs_variant} shows the pooled errors behind the observation columns.

    \paragraph{What the breakdowns add.}
    Detailed analyses further elucidate the models' performance along the five main axes CausalDS tests. 
    Symbolic causal reasoning is often saturated: graph recovery, association, and the R1 binaries on the realistic exam are passed by nearly every model.
    Data-science execution separates the field progressively under hard measurement: R2
    content is perfect for ten or more models on each exam, while the hard exam's errors spread by an order of magnitude on, e.g., \texttt{confounding} motifs, with the five most recent models (GPT-6 Astra, GLM 5.3, DeepSeek V4.1 Flash, Claude Fable 5.1, K2-MoVA 36B-A4B) staying accurate where GPT-5.5, Qwen 3.6 35B, Kimi K2.6, and Gemma 4 26B blow up.
    Uncertainty quantification enters these tables only through $S_{\mathrm{NR}}$, and it is what sinks the noisy observation and R3 ranks of GPT-5.5 and Gemma 4 26B (Fig.~\ref{fig:ate_coverage}; App.~\ref{app:error_intervals}). 
    Epistemic abstention  is decisive for ranking the field on both exams: abstention pass rates span $20$--$100\%$ whereas binary content can be saturated. Only GPT-6 Astra makes the
    call reliably across families and exams. 
    Tool use and coding leave no column here, but the near-one-shot closed models, GPT-5.5 and Claude Opus 4.8, sit with Qwen 3.6 35B and Gemma 4 26B at the bottom of the
    \texttt{proxy\_severe} error column (Tab.~\ref{tab:realistic_obs_variant}; App.~\ref{app:cost_efficiency}), consistent with the observation model scaling data-science difficulty independently of the other axes.
    A caveat nonetheless stands: as every slice is a marginal view of exams whose axes co-vary,  the causal reading of any single axis rests on the matched-scene
    ablation (App.~\ref{app:observation_ablation}) and the error taxonomy (App.~\ref{app:error_taxonomy}),  and cells with a handful of tasks support no reliable conclusion.

    \subsection{Detailed error taxonomy}
    \label{app:error_taxonomy}

    The leaderboard (Sec.~\ref{sec:results}) records whether an agent passed and how far its numbers were off.
    It does not say which part of an integrated item failed.
    Because every scene is generated from a known SCM, each task has an exact answer --- a graph, an identifiability verdict, an effect value --- and abstention is graded as a first-class answer (Sec.~\ref{sec:scoring}); a miss can therefore be attributed to the decision that produced it rather than to a disagreement about the reference.
    This section does that per failed task, for all twelve models and all three replications of both exams.

    \paragraph{Input mode.}
    The coarsest cut is whether a task needs the data at all.
    \texttt{causal\_sketch}, \texttt{identification}, \texttt{counterfactual\_identification}, and the forbidden-controls \texttt{bias\_diagnostic} variant are solvable from the story-implied graph (\emph{symbolic}); every other family needs the released files (\emph{data-backed}).
    The two Pass Rate columns of Tab.~\ref{tab:realistic_input_mode} do not move together on either exam. GPT-6 Astra leads both. On the realistic exam DeepSeek~V4.1 Flash and K2-MoVA rank in the bottom half on symbolic tasks and in the top half on data-backed ones, while GPT-5.5, Gemini~3.1 Pro, and Qwen~3.6 35B are the reverse; on the hard exam GPT-5.5, Gemini, and Claude Opus~4.8 keep symbolic Pass Rates near the top with data-backed Pass Rates among the lowest.
    Structure recovery is saturated ($F_1$-Loss $0$ for every model; the per-family view is in App.~\ref{app:exam_results}), so the symbolic column measures identification alone.

    \begin{table}[ht!]
        \centering
        \caption{\textbf{Performance by task input mode.} \emph{Symbolic} tasks (\texttt{causal\_sketch}, \texttt{identification}, \texttt{counterfactual\_identification}, forbidden-controls \texttt{bias\_diagnostic}) are solvable from the story-implied graph; \emph{data-backed} tasks need the released files. Pass Rate pools the content and abstention binaries of the slice; NRel.\ Err is over valid numeric answers. Three-run mean $\pm$ SD, with the per-run pool size in parentheses. Median $F_1$-loss on the symbolic set/graph tasks is $0.000$ for every model on both exams and is omitted. Rows follow the realistic-exam leaderboard; bold marks each exam's best per column. Model markers: \markclosed\ closed, \markopen\ open-weight.}
        \label{tab:realistic_input_mode}
        \footnotesize
        \begingroup
        \setlength{\tabcolsep}{4pt}
        \begin{adjustbox}{max width=\textwidth}
            \begin{tabular}{@{}ll c@{\hspace{10pt}}cc@{}}
                \toprule
                & & Symbolic & \multicolumn{2}{c}{Data-backed} \\
                \cmidrule(lr){3-3}\cmidrule(lr){4-5}
                Model & Exam & Pass Rate (\%) $\uparrow$ & Pass Rate (\%) $\uparrow$ & Med.\ NRel.\ Err $\downarrow$ \\
                \midrule
                \multirow{2}{*}{\markclosed\,GPT-6 Astra} & \examMain & $\boldsymbol{97.0 \pm 2.6}$\,{\scriptsize(22)} & $\boldsymbol{94.4 \pm 4.8}$\,{\scriptsize(12)} & $0.084 \pm 0.041$\,{\scriptsize(39)} \\
                 & \examHard & $\boldsymbol{100.0 \pm 0.0}$\,{\scriptsize(23)} & $\boldsymbol{87.4 \pm 0.8}$\,{\scriptsize(74)} & $0.155 \pm 0.014$\,{\scriptsize(93)} \\
                \addlinespace[2pt]
                \multirow{2}{*}{\markopen\,GLM 5.3} & \examMain & $68.2 \pm 0.0$\,{\scriptsize(22)} & $81.0 \pm 5.2$\,{\scriptsize(12--13)} & $\boldsymbol{0.080 \pm 0.011}$\,{\scriptsize(38--39)} \\
                 & \examHard & $77.5 \pm 2.2$\,{\scriptsize(23--24)} & $71.6 \pm 3.2$\,{\scriptsize(80--83)} & $\boldsymbol{0.121 \pm 0.021}$\,{\scriptsize(82--84)} \\
                \addlinespace[2pt]
                \multirow{2}{*}{\markclosed\,Gemini 3.1 Pro} & \examMain & $86.4 \pm 4.5$\,{\scriptsize(22)} & $65.0 \pm 3.0$\,{\scriptsize(12--13)} & $0.140 \pm 0.081$\,{\scriptsize(38--39)} \\
                 & \examHard & $82.6 \pm 4.3$\,{\scriptsize(23)} & $59.0 \pm 1.6$\,{\scriptsize(75--77)} & $0.232 \pm 0.047$\,{\scriptsize(90--92)} \\
                \addlinespace[2pt]
                \multirow{2}{*}{\markopen\,DeepSeek V4.1 Flash} & \examMain & $65.2 \pm 2.6$\,{\scriptsize(22)} & $88.9 \pm 4.8$\,{\scriptsize(12)} & $0.149 \pm 0.042$\,{\scriptsize(39)} \\
                 & \examHard & $84.1 \pm 9.1$\,{\scriptsize(23)} & $83.8 \pm 1.8$\,{\scriptsize(77--80)} & $0.140 \pm 0.012$\,{\scriptsize(87--90)} \\
                \addlinespace[2pt]
                \multirow{2}{*}{\markclosed\,Claude Fable 5.1} & \examMain & $63.6 \pm 4.5$\,{\scriptsize(22)} & $76.9 \pm 0.0$\,{\scriptsize(13)} & $0.103 \pm 0.012$\,{\scriptsize(35)} \\
                 & \examHard & $82.6 \pm 4.3$\,{\scriptsize(23)} & $86.2 \pm 1.9$\,{\scriptsize(74--76)} & $0.125 \pm 0.009$\,{\scriptsize(88--92)} \\
                \addlinespace[2pt]
                \multirow{2}{*}{\markopen\,Kimi K3} & \examMain & $60.6 \pm 2.6$\,{\scriptsize(22)} & $77.8 \pm 12.7$\,{\scriptsize(12)} & $0.099 \pm 0.013$\,{\scriptsize(39)} \\
                 & \examHard & $62.8 \pm 3.3$\,{\scriptsize(23--24)} & $64.3 \pm 3.2$\,{\scriptsize(75--76)} & $0.221 \pm 0.061$\,{\scriptsize(90--92)} \\
                \addlinespace[2pt]
                \multirow{2}{*}{\markopen\,K2-MoVA 36B-A4B} & \examMain & $65.7 \pm 6.2$\,{\scriptsize(22--23)} & $83.8 \pm 8.4$\,{\scriptsize(12--13)} & $0.166 \pm 0.059$\,{\scriptsize(38--39)} \\
                 & \examHard & $68.6 \pm 4.7$\,{\scriptsize(23--24)} & $72.5 \pm 4.4$\,{\scriptsize(83--85)} & $0.172 \pm 0.022$\,{\scriptsize(81--83)} \\
                \addlinespace[2pt]
                \multirow{2}{*}{\markclosed\,Claude Opus 4.8} & \examMain & $83.3 \pm 2.6$\,{\scriptsize(22)} & $77.8 \pm 4.8$\,{\scriptsize(12)} & $0.186 \pm 0.033$\,{\scriptsize(38--39)} \\
                 & \examHard & $82.6 \pm 4.3$\,{\scriptsize(23)} & $60.3 \pm 2.3$\,{\scriptsize(79--85)} & $0.275 \pm 0.034$\,{\scriptsize(82--88)} \\
                \addlinespace[2pt]
                \multirow{2}{*}{\markopen\,Kimi K2.6} & \examMain & $74.2 \pm 2.6$\,{\scriptsize(22)} & $81.8 \pm 24.6$\,{\scriptsize(12--13)} & $0.186 \pm 0.038$\,{\scriptsize(38--39)} \\
                 & \examHard & $76.8 \pm 2.5$\,{\scriptsize(23)} & $79.0 \pm 3.1$\,{\scriptsize(77--78)} & $0.213 \pm 0.023$\,{\scriptsize(88--90)} \\
                \addlinespace[2pt]
                \multirow{2}{*}{\markopen\,Qwen 3.6 35B} & \examMain & $75.8 \pm 2.6$\,{\scriptsize(22)} & $51.5 \pm 6.2$\,{\scriptsize(12--16)} & $0.203 \pm 0.041$\,{\scriptsize(28--39)} \\
                 & \examHard & $67.7 \pm 7.5$\,{\scriptsize(24--25)} & $54.7 \pm 4.6$\,{\scriptsize(81--84)} & $0.375 \pm 0.055$\,{\scriptsize(83--86)} \\
                \addlinespace[2pt]
                \multirow{2}{*}{\markclosed\,GPT-5.5} & \examMain & $87.9 \pm 5.2$\,{\scriptsize(22)} & $69.3 \pm 6.7$\,{\scriptsize(12--14)} & $0.223 \pm 0.042$\,{\scriptsize(37--38)} \\
                 & \examHard & $87.2 \pm 11.4$\,{\scriptsize(23--24)} & $64.2 \pm 0.8$\,{\scriptsize(85--89)} & $0.437 \pm 0.057$\,{\scriptsize(76--80)} \\
                \addlinespace[2pt]
                \multirow{2}{*}{\markopen\,Gemma 4 26B} & \examMain & $57.6 \pm 2.6$\,{\scriptsize(22)} & $62.2 \pm 4.2$\,{\scriptsize(12--13)} & $0.267 \pm 0.041$\,{\scriptsize(38--39)} \\
                 & \examHard & $50.0 \pm 2.2$\,{\scriptsize(23--24)} & $46.7 \pm 1.4$\,{\scriptsize(83--88)} & $1.022 \pm 0.122$\,{\scriptsize(79--83)} \\
                \bottomrule
            \end{tabular}
        \end{adjustbox}
        \endgroup
    \end{table}

    \paragraph{Mechanisms.}
    Every graded task ($10{,}800$ task-grades) gets one label.
    A task is a \emph{combined failure} if it is an official discrete miss (\texttt{correct=False}, abstention included), a non-submitted or unparseable exit, or a material continuous miss: point relative error above $0.25$, an interval that misses the SCM value or has $\mathrm{NRelIS}>2$ (Sec.~\ref{sec:scoring}), or prediction $n$-relative RMSE above $0.50$.
    The mechanism is \emph{tool/runtime} if the agent loop produced no usable submission, \emph{reasoning} if a valid discrete answer was wrong, and \emph{data science} if an identifiable numeric answer was materially wrong.
    Missing files count as tool/runtime and schema errors after a \texttt{Submitted} exit are ${\approx}0$; the one persistent tool case is Claude Fable~5.1 exiting \texttt{RepeatedFormatError} on every run of realistic \texttt{scene\_000018} without a bash call (App.~\ref{app:harness_ops}).
    The official Pass Rate sees only the first two mechanisms; a bad interval is never a binary fail.
    Trajectories of the eight leading models (GPT-6 Astra, Claude Fable 5.1, DeepSeek V4.1 Flash, GLM 5.3, Claude Opus 4.8, Gemini 3.1 Pro, GPT-5.5, Kimi K3; $7{,}200$ traces) annotate these grade-level labels --- a recovered bash error does not relabel a submitted discrete miss --- and supply the first-commitment audit of App.~\ref{app:error_tracks}.

    Combined failures are $28.9{\pm}0.3\%$ of task-grades ($15.8\%$ realistic, $35.4\%$ hard), and their mix is stable across replications (Tab.~\ref{tab:error_mechanism}): half data science, somewhat less reasoning, one in twenty tool/runtime.
    Restricted to the official discrete pool, misses are $88\%$ reasoning and $12\%$ tool/runtime.
    The realistic exam is the more tool-exposed ($15.9\%$ of its combined fails in run~1, then $8.2\%$ and $7.8\%$); on the hard exam tool/runtime is $3.5\%$ and data science $55.3\%$.

    \begin{table}[ht!]
        \centering
        \caption{\textbf{Combined-failure mix by mechanism.} Share of combined failures (\%), mean $\pm$ sample SD, over three replications; $N$ is the per-run combined-fail count. Combined failure as defined in the text. \textcolor{paperViolet!80!black}{Violet} marks the realistic exam, \textcolor{paperOrange!80!black}{orange} the hard exam.}
        \label{tab:error_mechanism}
        \footnotesize
        \begin{adjustbox}{max width=\textwidth}
            \begin{tabular}{@{}l c c c c@{}}
                \toprule
                Slice & $N$ / run & Reasoning (\%) & Data science (\%) & Tool / runtime (\%) \\
                \midrule
                Both exams & $1039 \pm 12$ & $43.0 \pm 0.8$ & $52.1 \pm 0.9$ & $4.8 \pm 0.9$ \\
                \examMain[Realistic] & $190 \pm 16$ & $51.4 \pm 3.3$ & $37.8 \pm 2.6$ & $10.6 \pm 4.5$ \\
                \examHard[Hard] & $849 \pm 15$ & $41.2 \pm 0.6$ & $55.3 \pm 0.3$ & $3.5 \pm 0.3$ \\
                \bottomrule
            \end{tabular}
        \end{adjustbox}
    \end{table}

\begin{figure}[ht!]
        \centering
        \includegraphics[width=.55\textwidth]{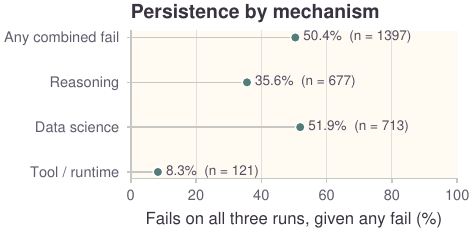}
        \caption{\textbf{Persistence by mechanism.} Share of complete (exam, model, scene, task) tuples that fail on all three replications, among those that fail on at least one ($n$ in parentheses; $3{,}600$ triples, twelve models, both exams). Teal dots are the measured shares.}
        \label{fig:error_persistence}
    \end{figure}

    \paragraph{Persistence.}
    Over the $3{,}600$ complete (exam, model, scene, task) tuples, a data-science fail that occurs once recurs on all three runs $51.9\%$ of the time, a reasoning fail $35.6\%$, a tool/runtime fail $8.3\%$ ($10$ of $121$; Fig.~\ref{fig:error_persistence}).
    The sandbox is not the exam.
    The same separation appears in the metrics Sec.~\ref{sec:results} summarizes: the gap between any-of-3 and all-of-3 in Tabs.~\ref{tab:realistic_leaderboard} and~\ref{tab:hard_leaderboard} is the share of the binary pool that passes on some runs and not others; point estimates reproduce almost exactly across restarts (median per-task NRelErr SD below $0.001$ for every model on the realistic exam and below $0.03$ on the hard exam, over tasks answered in all three runs), whereas the per-task SD of NRelIS is one to two orders of magnitude larger, driven by a few \texttt{ate\_uq\_95} items that swing between ${\approx}0$ and the cap (the point/interval split by $k$ is Tab.~\ref{tab:realistic_passk_continuous}).
    The worst/best CausalDSScore in the leaderboards scores each restart as a complete exam, which keeps the covariance between score components.

    \texttt{scene\_000700} (realistic exam; ATE with $95\%$ CI; grafted confounding host; \texttt{proxy\_hard}) is the extreme case.
    Treatment and confounder are latent behind noisy bundles, with gold values on the $1{,}600$ calibration rows only, and the treatment is a sharp threshold of the confounder (treated share by confounder decile $1, 1, 1, 1, 1, 0.95, 0.68, 0.42, 0.09, 0$), so the ATE must be extrapolated where there is no overlap and the answer is decided by the adjustment form: linear or quadratic adjustment on the gold rows gives $0.34$--$0.41$ (SCM value $0.42$), an additive spline $0.11$, a reconstructed-confounder regression $0.20$ or $0.82$.
    Agents choose anew on each restart.
    Claude Opus~4.8 scores $\mathrm{NRelIS}$ $0.18$, $5.0$, $9.9$ on its three runs, GPT-6 Astra $0.22$, $10$, $0.23$, and six of twelve models flip between below $0.6$ and above $3.5$.
    These are shifted-point misses in the sense of App.~\ref{app:error_intervals} (Opus's point moves $0.41 \to 0.20 \to 0.82$), the mirror image of hard \texttt{scene\_000070} in App.~\ref{app:error_examples}, where every model lands the same attenuated point on every run.
    Each flip moves CausalDSScore by $0.04$--$0.10$; with \texttt{scene\_000126}, this item accounts for the whole per-run spread of Opus ($0.276$, $0.370$, $0.415$) and of Gemini~3.1~Pro ($0.368$, $0.232$, $0.225$).

\begin{figure}[ht!]
        \centering
        \includegraphics[width=\textwidth]{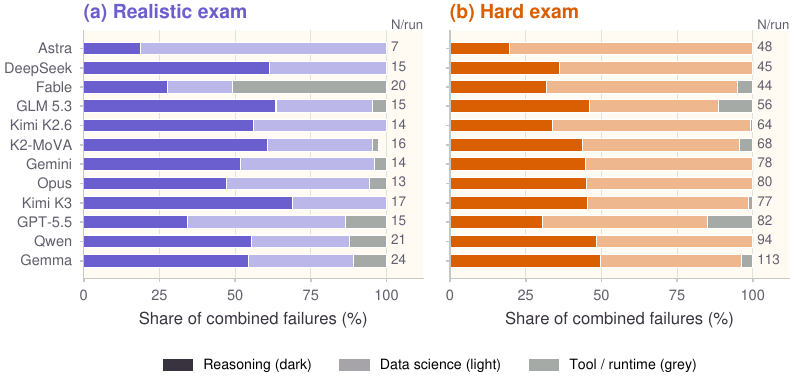}
        \caption{\textbf{Combined-failure mix by model and exam.} One bar per model: reasoning (full exam color), data science (light), tool / runtime (gray); shares of that model's combined failures, mean over three replications, with the per-run count $N$ at the bar end. (a) realistic exam in violet, (b) hard exam in orange. Models are ordered by pooled combined fails per run.}
        \label{fig:error_mechanism_by_model}
    \end{figure}

    \paragraph{By model.}
    The field mix is not a model property (Fig.~\ref{fig:error_mechanism_by_model}, Tab.~\ref{tab:error_by_model}).
    GPT-6 Astra's fails are $80\%$ data science with no tool/runtime, and its rare discrete misses are wrong effect signs ($26$ of $32$); only $2$ are false identifiability.
    Gemma~4 26B, GLM~5.3, Kimi~K3, and Qwen~3.6 35B fail at least as often on reasoning as on data science.
    GPT-5.5 inverts the field's abstention error: $41\%$ of its reasoning fails are false abstentions against $19\%$ overall --- the same signature as its column in Tab.~\ref{tab:realistic_pr_split} --- and it is the only leading model with a double-digit tool share ($15\%$).
    Claude Opus~4.8 sits on the field mean, which here means two large piles rather than balance.
    Within data science, UQ misses with an acceptable point are $54$--$66\%$ for every model: models differ in which class they enter, while the breakdown within the class is the same.

    \begin{table}[ht!]
        \centering
        \caption{\textbf{Subtype shares by model.} Both exams, mean over three replications; $N$ is combined fails per run, mean $\pm$ SD. False ID / false abstention are shares of that model's \emph{reasoning} fails; UQ-ok / uncovered$+$bad are shares of its \emph{data-science} fails. The mechanism split itself is Fig.~\ref{fig:error_mechanism_by_model}. Model markers: \markclosed\ closed, \markopen\ open-weight.}
        \label{tab:error_by_model}
        \footnotesize
        \begin{adjustbox}{max width=\textwidth}
            \begin{tabular}{@{}l c c c c c@{}}
                \toprule
                 &  & \multicolumn{2}{c}{Of reasoning fails (\%)} & \multicolumn{2}{c}{Of data-science fails (\%)} \\
                \cmidrule(lr{3pt}){3-4}\cmidrule(l{3pt}){5-6}
                \makecell{Model\\{\scriptsize by $N$}} & \makecell{$N$\\{\scriptsize / run}} & False ID & False abs. & UQ-ok & Uncov.$+$bad \\
                \midrule
                \markclosed\,GPT-6 Astra & $55 \pm 5$ & 6 & 3 & 62 & 20 \\
                \markopen\,DeepSeek V4.1 Flash & $60 \pm 2$ & 53 & 22 & 62 & 19 \\
                \markclosed\,Claude Fable 5.1 & $64 \pm 4$ & 72 & 3 & 66 & 21 \\
                \markopen\,GLM 5.3 & $71 \pm 3$ & 64 & 20 & 66 & 16 \\
                \markopen\,Kimi K2.6 & $78 \pm 4$ & 61 & 18 & 61 & 24 \\
                \markopen\,K2-MoVA 36B-A4B & $84 \pm 3$ & 51 & 34 & 59 & 25 \\
                \markclosed\,Gemini 3.1 Pro & $92 \pm 4$ & 82 & 5 & 64 & 22 \\
                \markclosed\,Claude Opus 4.8 & $93 \pm 5$ & 61 & 19 & 56 & 24 \\
                \markopen\,Kimi K3 & $94 \pm 3$ & 80 & 6 & 66 & 20 \\
                \markclosed\,GPT-5.5 & $96 \pm 4$ & 22 & 41 & 54 & 30 \\
                \markopen\,Qwen 3.6 35B & $115 \pm 7$ & 58 & 25 & 60 & 22 \\
                \markopen\,Gemma 4 26B & $137 \pm 6$ & 57 & 22 & 56 & 27 \\
                \bottomrule
            \end{tabular}
        \end{adjustbox}
    \end{table}

    \subsubsection{Reasoning and data-science failure tracks}
    \label{app:error_tracks}

\begin{figure}[ht!]
        \centering
        \includegraphics[width=\textwidth]{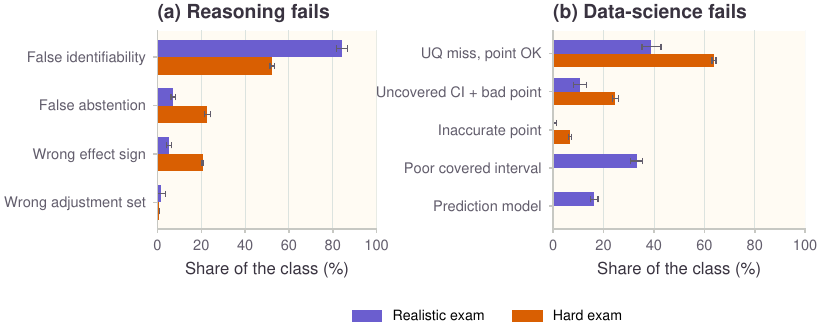}
        \caption{\textbf{Failure subtypes by exam.} (a) Reasoning-fail subtypes as a share of reasoning fails; (b) data-science-fail subtypes as a share of data-science fails. Bars are means over three replications, whiskers the sample SD; violet realistic exam, orange hard exam. ``UQ miss, point OK'': the $95\%$ interval misses the SCM value while point relative error is ${\le}0.25$. Pooled counts are given in the text.}
        \label{fig:error_subtypes}
    \end{figure}

    \paragraph{Reasoning.}
    A reasoning fail is a valid discrete answer that is wrong: the grader could score the JSON and the binary, set, or sign target was missed.
    The pool is the discrete pool, which is wider than ``binary tasks'': an effect item that was estimated when it should have been abstained on, or abstained on when identifiable, is graded here; a wrong identification strategy behind a submitted number is not, and the census below exists to catch it.
    Both abstention errors are gradable only because identifiability is fixed by the hidden graph: \texttt{y0.id} and \texttt{dowhy} decide it from the SCM rather than a textbook default, so ``estimated when it should have abstained'' is a fact about the scene (App.~\ref{app:scoring_rules}).
    Three subtypes account for $96\%$ of the pool (Fig.~\ref{fig:error_subtypes}a): false identifiability ($59\%$, $796$ grades), false abstention ($19\%$), and a wrong effect sign ($17\%$).
    The first two are the abstention errors of Tab.~\ref{tab:realistic_pr_split} under other names --- a missed abstention there is a false-identifiability grade here --- with a different denominator (that table divides by the abstention pool per exam, this section by reasoning fails over both exams); the model ordering agrees, with GPT-6 Astra lowest on both and GPT-5.5 the false-abstention outlier on the hard exam.
    Wrong adjustment sets are $14$ grades in all, and graph reconstruction fails $3$ times in $10{,}800$.

    The two abstention errors are not one skill on different motifs.
    Of the $796$ false-identifiability grades, $502$ sit on an IV host and $243$ on a grafted graph that still carries an instrument-like unused cause; front-door hosts contribute none.
    False abstention concentrates on front-door mediation and counterfactual-effect items (Tab.~\ref{tab:error_reason_family}): the agent sees an unobserved common cause, calls it non-identification, and never applies the front-door functional.
    The first is the over-claim diagnosed trace by trace on the running example (App.~\ref{app:error_examples}); the per-motif view already ranks models on the \texttt{iv} row purely by that call (Tab.~\ref{tab:realistic_per_motif}).
    The second is the exit the matched-scene ablation isolates under \texttt{proxy\_hard} (App.~\ref{app:observation_ablation}).
    Wrong signs are a hard-exam phenomenon ($217$ \texttt{sign\_only} grades on the hard exam against $15$ on the realistic exam); the realistic reasoning pile is $84\%$ false identifiability.

    \begin{table}[ht!]
        \centering
        \caption{\textbf{Reasoning fails by task family.} Pooled over three replications, twelve models, both exams. Absolute counts; families have different exam weights.}
        \label{tab:error_reason_family}
        \footnotesize
        \begin{adjustbox}{max width=\textwidth}
            \begin{tabular}{@{}l c c c c@{}}
                \toprule
                Family & False ID & False abs. & Wrong sign & Wrong adj.\ set \\
                \midrule
                Effect estimate & 304 & 89 & 0 & 0 \\
                Identification & 146 & 21 & 0 & 14 \\
                Bias diagnostic & 172 & 0 & 0 & 0 \\
                CF effect & 146 & 88 & 216 & 0 \\
                CF identification & 28 & 4 & 0 & 0 \\
                Mediation effect & 0 & 59 & 16 & 0 \\
                \bottomrule
            \end{tabular}
        \end{adjustbox}
    \end{table}

    \paragraph{First-commitment census.}
    The grade cannot see a wrong meta-approach behind a submitted number: backdoor-only on a front-door host is data science if a number came out.
    We therefore coded the first causal commitment --- the first assistant turn that names identifiability or a method, including the reasoning summary and the first bash, or the first planning turn after an inspect-only opening --- on every leading-model fail with a trajectory plus six effect-item passes per exam and model: $1{,}971$ traces (Tab.~\ref{tab:error_pretool}).
    Most traces inspect the files before committing; GPT-5.5 and Astra more often commit in the first message, before any tool result.

    \begin{table}[ht!]
        \centering
        \caption{\textbf{First-commitment census: strata and first-turn regimes.} Leading-model traces with files (eight models, both exams, three replications; corrected grades). Fail strata are a census; effect-item passes are $6$ per exam$\times$model. Inspect-first $=$ the first assistant turn has no identifiability or method claim; pre-tool commit $=$ the causal plan is stated before any tool result. Hidden-reasoning rates per stratum and per model are in Fig.~\ref{fig:error_first_commitment}.}
        \label{tab:error_pretool}
        \footnotesize
        \begin{adjustbox}{max width=\textwidth}
            \begin{tabular}{@{}l c c c@{}}
                \toprule
                Stratum & \makecell{$n$\\{\scriptsize traces}} & Inspect-first & Pre-tool commit \\
                \midrule
                Discrete reasoning fail & 756 & 531 & 225 \\
                DS fail, effect / CF / mediation & 941 & 781 & 160 \\
                DS fail, prediction & 69 & 69 & 0 \\
                Effect-item pass (sample) & 96 & 75 & 21 \\
                Tool / runtime & 109 & 60 & 20 \\
                \bottomrule
            \end{tabular}
        \end{adjustbox}
    \end{table}

\begin{figure}[ht!]
        \centering
        \includegraphics[width=\textwidth]{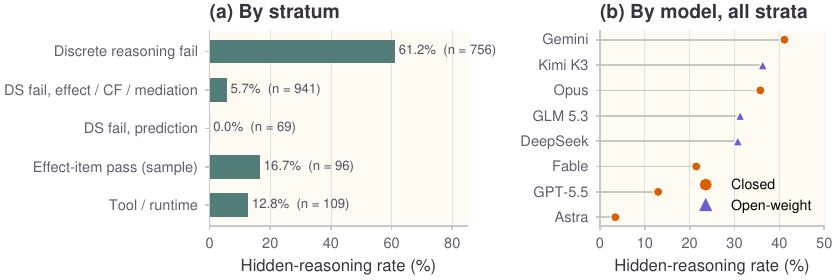}
        \caption{\textbf{Hidden reasoning in the first commitment.} Share of traces whose first causal commitment is a false-identifiability plan, false-abstention plan, wrong identification strategy, linearity rescue, or wrong estimand. (a) By census stratum, with $n$; (b) by model over all strata: orange circles closed models, violet triangles open-weight.}
        \label{fig:error_first_commitment}
    \end{figure}

    On discrete reasoning fails the first commitment already contains the error $61\%$ of the time ($56\%$ on the hard exam, $80\%$ on the realistic), typically IV on a non-identifiable ATE.
    On data-science effect fails it is motif-consistent in $94\%$ of traces; the $54$ exceptions are almost all one thing, a front-door host whose first sentence treats unobserved $U$ as fatal --- and that reading stays in the first sentence; the submitted script does not act on it, as hard \texttt{scene\_000070} in App.~\ref{app:error_examples} shows.
    Prediction fails have no causal plan at all.
    So the continuous pile is not a relabeled reasoning pile; its one leak is front-door graphs read as non-identified backdoors.
    The flags are not uniform (Fig.~\ref{fig:error_first_commitment}b): GPT-6 Astra almost never states a false first plan, GPT-5.5 rarely, while Gemini~3.1 Pro, Kimi~K3, and Claude Opus~4.8 do so in more than a third of coded traces.
    That is a process difference; at the grade level (Tab.~\ref{tab:error_by_model}) Astra's fails are still mostly data science.

    \paragraph{Data science.}
    A data-science fail is an identifiable numeric, interval, or prediction answer that is materially wrong, and the dominant subtype is not a blown point estimate.
    Across both exams $60\%$ of these fails are \emph{UQ misses with an acceptable point}: the $95\%$ interval misses the SCM value while point relative error is ${\le}0.25$ (Fig.~\ref{fig:error_subtypes}b; $982$ of $1{,}623$).
    Another $23\%$ miss the interval and the point; the rest are isolated inaccurate points, poor covered intervals, and prediction-model misses, the last two almost entirely on the realistic exam, which still has a prediction family and few severe-proxy interval items.
    The observation layer moves the pile from covered-but-poor intervals on \texttt{clean} to uncovered intervals on \texttt{proxy\_hard} and \texttt{proxy\_severe}; App.~\ref{app:error_intervals} gives the geometry by view.
    Among the eight leading models with traces, data-science fails run the longest tool loops (mean $14.5$ bash calls versus $7.8$ on passes and $10.1$ on reasoning fails; per-task budgets are in App.~\ref{app:cost_efficiency}), and $56\%$ show a recovered bash or format error.
    The typical data-science miss is not ``never opened the data''.
    It is a submitted, calibration-aware estimate whose interval is too tight, too shifted, or both; App.~\ref{app:error_estimators} shows the script and App.~\ref{app:error_intervals} the interval.

    \subsubsection{Failure-mode examples}
    \label{app:error_examples}

    One scene per large subtype makes the tracks concrete.
    Table~\ref{tab:error_examples} gives every example in one format --- ground truth, the outcome of all twelve models on each run, the stable split, a one-line reading --- and Tab.~\ref{tab:realistic_scene154} adds the per-agent diagnoses for the running example.
    They are illustrations and are not a representative sample.
    What the tables do not carry is the traces.

    \begin{table}[ht!]
        \centering
        \caption{\textbf{Scene-level forks.} One row per example scene, twelve models, three replications. Outcome strips show runs 1--3 per model: \textcolor{paperTeal}{$\bullet$} pass, {$\bullet$} reasoning miss (false identifiability or false abstention), {$\circ$} data-science miss (uncovered interval), {\color{gray}$\times$} tool / runtime exit. Scene IDs are colored by exam (\textcolor{paperViolet!80!black}{Realistic}, \textcolor{paperOrange!80!black}{Hard}); K3, K2.6, MoVA, GLM abbreviate Kimi K3, Kimi K2.6, K2-MoVA 36B-A4B, GLM 5.3. \markclosed\ Closed; \markopen\ open-weight. Quotes are in the text.}
        \label{tab:error_examples}
        \begingroup

\newcommand{\ETSlicePlain}[1]{{\LBPrimary #1}}
\newcommand{\ETSliceMain}[1]{{\LBPrimary\color{paperViolet!80!black}#1}}
\newcommand{\ETSliceHard}[1]{{\LBPrimary\color{paperOrange!80!black}#1}}
\newcommand{\ETGroupHead}[2]{{\color{#1}\sffamily\bfseries\fontsize{7}{8}\selectfont #2}}
\newcommand{\ETDash}{{\color{LBMuted}---}}
\definecolor{ETPointer}{HTML}{A6AAA6}
\makeatletter
\@namedef{ETDot@P}{\tikz[baseline=-0.6ex]\fill[paperTeal] (0,0) circle[radius=2pt];}
\@namedef{ETDot@R}{\tikz[baseline=-0.6ex]\fill[LBInk] (0,0) circle[radius=2pt];}
\@namedef{ETDot@D}{\tikz[baseline=-0.6ex]\draw[LBInk,line width=.7pt] (0,0) circle[radius=1.7pt];}
\@namedef{ETDot@T}{\tikz[baseline=-0.6ex]\draw[ETPointer,line width=.8pt] (-1.5pt,-1.5pt)--(1.5pt,1.5pt) (-1.5pt,1.5pt)--(1.5pt,-1.5pt);}
\newcommand{\ETDots}[1]{\@tfor\ETc:=#1\do{\csname ETDot@\ETc\endcsname\hspace{1pt}}}
\@namedef{ETExamColor@main}{paperViolet}\@namedef{ETExamColor@hard}{paperOrange}
\newcommand{\ETExamColor}[1]{\csname ETExamColor@#1\endcsname}
\makeatother
\newcommand{\ETStrip}[3]{%
  \begin{minipage}[t]{.235\linewidth}\raggedright\vspace{0pt}%
    {\LBDetail\color{LBMuted}#1\,#2}\\[-0.5pt]\ETDots{#3}%
  \end{minipage}}
\newcommand{\ETStrips}[1]{\begin{minipage}[t]{\linewidth}\raggedright\vspace{0pt}#1\end{minipage}}
\newcommand{\ETScene}[4]{%
  \begin{minipage}[t]{\linewidth}\raggedright\vspace{0pt}
    {\sffamily\bfseries\fontsize{7.2}{9}\selectfont\color{\ETExamColor{#1}!80!black}\texttt{#2}}\\[-0.5pt]
    {\LBDetail\color{LBMuted}#3}\\[-0.5pt]
    {\LBDetail\color{LBInk}#4}
  \end{minipage}}
\newcommand{\ETRight}[1]{{\color{LBValidTop}\texttt{#1}}}
\newcommand{\ETWrong}[1]{{\color{LBValidBottom}\texttt{#1}}}

        \LBTableSetup
        \begin{tabular}{@{}>{\raggedright\arraybackslash}m{.15\linewidth}@{\hspace{5pt}} >{\LBDetail\color{LBInk}\raggedright\arraybackslash}m{.14\linewidth}@{\hspace{5pt}} >{\raggedright\arraybackslash}m{.33\linewidth}@{\hspace{5pt}} >{\LBDetail\color{LBInk}\raggedright\arraybackslash}m{.17\linewidth}@{\hspace{5pt}} >{\LBDetail\color{LBInk}\raggedright\arraybackslash}m{.17\linewidth}@{}}
            \LBHeader{Scene}{task; host, view; class} & \LBHeader{Ground truth}{} & \LBHeader{Outcomes, runs 1--3}{} & \LBHeader{Stable split}{} & \LBHeader{Reading}{}\\[3pt]
            \midrule
                \ETScene{main}{scene\_000154}{Forbidden-controls bias diagnostic; IV, \texttt{proxy\_hard}}{False identifiability} & Not point-identified; correct: \texttt{non\_id} & \ETStrips{\ETStrip{\markclosed}{Astra}{PPP} \ETStrip{\markclosed}{GPT-5.5}{PPR} \ETStrip{\markclosed}{Opus}{RRR} \ETStrip{\markclosed}{Fable}{RRP} \ETStrip{\markclosed}{Gemini}{PPP} \ETStrip{\markopen}{DeepSeek}{RRR} \ETStrip{\markopen}{GLM}{RRR} \ETStrip{\markopen}{K3}{RRR} \ETStrip{\markopen}{K2.6}{RRR} \ETStrip{\markopen}{MoVA}{RRR} \ETStrip{\markopen}{Qwen}{RRR} \ETStrip{\markopen}{Gemma}{RRP}} & Astra, Gemini $3/3$ \texttt{non\_id}; Opus, DeepSeek, GLM, K2-MoVA, both Kimis, Qwen $3/3$ \texttt{no\_backdoor}; GPT-5.5 $2/3$, Fable $1/3$ correct & IV over-claim: an instrument is read as identifying the population ATE \\
                \addlinespace[4pt]
                \ETScene{hard}{scene\_000470}{ATE with 95\% CI; IV, \texttt{proxy\_severe}}{False identifiability} & Not point-identified; correct: null & \ETStrips{\ETStrip{\markclosed}{Astra}{PPP} \ETStrip{\markclosed}{GPT-5.5}{PPP} \ETStrip{\markclosed}{Opus}{RRR} \ETStrip{\markclosed}{Fable}{PPP} \ETStrip{\markclosed}{Gemini}{RRR} \ETStrip{\markopen}{DeepSeek}{PPP} \ETStrip{\markopen}{GLM}{RPR} \ETStrip{\markopen}{K3}{RTR} \ETStrip{\markopen}{K2.6}{PPP} \ETStrip{\markopen}{MoVA}{PRR} \ETStrip{\markopen}{Qwen}{PPP} \ETStrip{\markopen}{Gemma}{RPP}} & Astra, Fable, DeepSeek, GPT-5.5, Kimi K2.6, Qwen $3/3$ null; Opus, Gemini $3/3$ number & The same fork when the requested object is an interval \\
                \addlinespace[4pt]
                \ETScene{hard}{scene\_000236}{NIE point; front-door, \texttt{proxy}}{False abstention} & NIE identifiable (front-door); correct: a number & \ETStrips{\ETStrip{\markclosed}{Astra}{PPP} \ETStrip{\markclosed}{GPT-5.5}{RRR} \ETStrip{\markclosed}{Opus}{PRP} \ETStrip{\markclosed}{Fable}{PPP} \ETStrip{\markclosed}{Gemini}{PPP} \ETStrip{\markopen}{DeepSeek}{RRP} \ETStrip{\markopen}{GLM}{RRR} \ETStrip{\markopen}{K3}{PPP} \ETStrip{\markopen}{K2.6}{PPP} \ETStrip{\markopen}{MoVA}{RRR} \ETStrip{\markopen}{Qwen}{RPP} \ETStrip{\markopen}{Gemma}{PPP}} & GLM, GPT-5.5, K2-MoVA $3/3$ null; Astra, Fable, Gemini, both Kimis, Gemma $3/3$ number & Front-door false abstention: latent $U$ read as fatal \\
                \addlinespace[4pt]
                \ETScene{hard}{scene\_000070}{ATE with 95\% CI; front-door, \texttt{proxy\_hard}}{UQ miss, point OK} & SCM ATE $-0.121$; correct: a covering front-door CI & \ETStrips{\ETStrip{\markclosed}{Astra}{DDD} \ETStrip{\markclosed}{GPT-5.5}{DDD} \ETStrip{\markclosed}{Opus}{DDD} \ETStrip{\markclosed}{Fable}{DDD} \ETStrip{\markclosed}{Gemini}{DDD} \ETStrip{\markopen}{DeepSeek}{DDD} \ETStrip{\markopen}{GLM}{DDD} \ETStrip{\markopen}{K3}{DDD} \ETStrip{\markopen}{K2.6}{DDD} \ETStrip{\markopen}{MoVA}{RRD} \ETStrip{\markopen}{Qwen}{DDD} \ETStrip{\markopen}{Gemma}{DDR}} & All eight leading models, all runs: uncovered CI, point relative error ${\approx}0.09$ & Identification held; the interval did not \\
                \addlinespace[4pt]
                \ETScene{main}{scene\_000700}{ATE with 95\% CI; grafted confounding, \texttt{proxy\_hard}}{Shifted-point interval miss} & SCM ATE $0.42$; no overlap in the confounder tails & \ETStrips{\ETStrip{\markclosed}{Astra}{PDP} \ETStrip{\markclosed}{GPT-5.5}{PDP} \ETStrip{\markclosed}{Opus}{PDD} \ETStrip{\markclosed}{Fable}{DDD} \ETStrip{\markclosed}{Gemini}{DDD} \ETStrip{\markopen}{DeepSeek}{DDD} \ETStrip{\markopen}{GLM}{DDP} \ETStrip{\markopen}{K3}{PDD} \ETStrip{\markopen}{K2.6}{PDD} \ETStrip{\markopen}{MoVA}{PPT} \ETStrip{\markopen}{Qwen}{TPP} \ETStrip{\markopen}{Gemma}{DDD}} & Six of twelve models flip between covered and far off across runs & Shifted-point misses; the adjustment form decides \\
                \addlinespace[4pt]
                \ETScene{main}{scene\_000018}{Prediction interval; ---, \texttt{clean}}{Tool / runtime} & Any valid $90\%$ prediction interval & \ETStrips{\ETStrip{\markclosed}{Astra}{PPP} \ETStrip{\markclosed}{GPT-5.5}{PPP} \ETStrip{\markclosed}{Opus}{PPP} \ETStrip{\markclosed}{Fable}{TTT} \ETStrip{\markclosed}{Gemini}{PPP} \ETStrip{\markopen}{DeepSeek}{PPP} \ETStrip{\markopen}{GLM}{PPP} \ETStrip{\markopen}{K3}{PPP} \ETStrip{\markopen}{K2.6}{PPP} \ETStrip{\markopen}{MoVA}{PPP} \ETStrip{\markopen}{Qwen}{PPP} \ETStrip{\markopen}{Gemma}{PPP}} & Fable $3/3$ \texttt{RepeatedFormatError}; eleven models pass every run & The tool pile in miniature \\
            \bottomrule[.55pt]
        \end{tabular}\par
        \LBFinish
        \endgroup
    \end{table}

    On the running example (\texttt{scene\_000154}, Fig.~\ref{fig:running_example}) the population ATE is not point-identified --- the subsidy instrument identifies at most a compliance-local effect (\texttt{y0.id} and \texttt{dowhy} agree) --- and most models answer \texttt{no\_backdoor}: no adjustment set exists, therefore the instrument identifies the ATE.
    It is the modal discrete miss of the field ($502$ of $796$ false-identifiability grades sit on an IV host), and Tab.~\ref{tab:realistic_scene154} shows the routes to it: Kimi~K2.6 derives the LATE-versus-ATE gap and then bets against it on guessed grader convention, Claude Opus~4.8 bridges it with a linearity assumption the story never states, Qwen and Gemma never surface the estimand question, Gemini~3.1 Pro starts from the same over-claim and reasons its way back to \texttt{non\_id}, and GPT-5.5 abstains in its first and only step.

    \begin{table}[ht!]
        \centering
        \caption{\textbf{Per-agent responses on the running example's bias diagnostic.} Forbidden-controls bias diagnostic for \texttt{scene\_000154} (Fig.~\ref{fig:running_example});
        quotes are verbatim from the agents' own reasoning traces. All six agents reconstruct the same graph and correctly conclude that no observed
        backdoor adjustment set exists; the answers diverge only on the final step---whether the subsidy instrument point-identifies the \emph{population} ATE.
        Failing agents ordered by how much of the correct argument appears in their trace.
         \textcolor{hbFullColor}{\texttt{non\_id}} is the correct answer; \textcolor{hbNoneColor}{\texttt{no\_backdoor}} over-claims. \markclosed\ Closed; \markopen\ open-weight.}
        \label{tab:realistic_scene154}
        \begingroup
        
        \LBTableSetup
        \begin{tabular}{@{}>{\raggedright\arraybackslash}m{.18\linewidth} >{\raggedright\arraybackslash}m{.13\linewidth} >{\raggedright\arraybackslash}m{.69\linewidth}@{}}
            \LBHeader{Model}{} & \LBHeader{Answer}{} & \LBHeader{Diagnosis}{verbatim quotes from the reasoning trace}\\[3pt]
            \midrule
                \LBModel{}{\markclosed\,Gemini 3.1 Pro} & \ETRight{non\_id} & {\LBDetail\color{LBInk}Starts where the failures end (``S is a valid instrument''), then prosecutes its own claim at length: recalls that nonparametric IV with a binary treatment yields only Balke--Pearl bounds, notices that the story's threshold mechanism (subsidy sufficient \emph{and} permeable soil) ``definitively breaks the linear SEM assumptions'' that would rescue point identification, and reverses to \texttt{non\_id}.} \\
                \addlinespace[3pt]
                \LBModel{}{\markclosed\,GPT-5.5} & \ETRight{non\_id} & {\LBDetail\color{LBInk}Answers in its first and only step, without opening anything in the workspace: the subsidy is ``an instrument-like cause of treatment, not a variable that blocks the backdoor path'', so the ATE ``is not nonparametrically identifiable from this graph by adjustment or another graph-based strategy''. The nonparametric standard is applied by default, with no deliberation about what the grader might accept.} \\
                \addlinespace[3pt]
                \LBModel{}{\markopen\,Kimi K2.6} & \ETWrong{no\_backdoor} & {\LBDetail\color{LBInk}Its thinking trace contains the correct answer in full: it derives that IV identifies ``the local average treatment effect (LATE)\ldots but the population ATE may not be'', and even works out the bias-amplification algebra for conditioning on an instrument---then bets against its own analysis on guessed grader convention: ``the question likely comes from a course or dataset where they expect \texttt{no\_backdoor}'' and ``the problem might not be that nuanced''.\footnotemark} \\
                \addlinespace[3pt]
                \LBModel{}{\markclosed\,Claude Opus 4.8} & \ETWrong{no\_backdoor} & {\LBDetail\color{LBInk}Reaches the correct fork, then bridges it with parametric assumptions it reads into the narration: ``the population ATE may be identifiable via IV under linearity/additivity assumptions. The story describes [the outcome] as linear-additive combination\ldots So ATE is identifiable by IV''. The story states no functional form (the outcome ``reflects the combined action'' of treatment and confounder), and the task is symbolic---there is no released data against which linearity could be checked, and its sole exploration is reading the story. The ``rescue'' rests on an assumption that is neither stated nor checkable.} \\
                \addlinespace[3pt]
                \LBModel{}{\markopen\,Qwen 3.6 35B} & \ETWrong{no\_backdoor} & {\LBDetail\color{LBInk}Reverses its verdict four times (``Let me go with \texttt{non\_id} as the safer answer\ldots OK let me just commit to \texttt{no\_backdoor}''), in between spending most of a 60-message trajectory re-printing the 1.6\,kB story it already had verbatim in the prompt. The final commitment verifies the three instrument conditions---relevance, exclusion, independence, all of which \emph{do} hold---and treats that checklist as the finish line: ``So IV identification is valid! The answer should be \texttt{no\_backdoor}.'' But those conditions (plus the monotonicity it \textit{never} invokes) deliver only the complier-local effect, not the population ATE the task asks for. Unlike the two agents above, it never registers that a bridging step is missing: where Kimi K2.6 derived the LATE-vs-ATE gap and gambled against it, and Claude Opus 4.8 filled it with an invented linearity assumption, Qwen never surfaces the estimand question at all---LATE, monotonicity, and effect homogeneity appear nowhere in its trace.} \\
                \addlinespace[3pt]
                \LBModel{}{\markopen\,Gemma 4 26B} & \ETWrong{no\_backdoor} & {\LBDetail\color{LBInk}Never questions the step at all: ``Is the ATE identifiable? Yes, via IV'' is asserted flatly, and the entire deliberation---written into a 22\,kB draft of the answer file itself, whose first version answered \texttt{[]}---concerns which answer label fits the definition of a forbidden control, not whether the instrument identifies anything.} \\
            \bottomrule[.55pt]
        \end{tabular}\par
        \LBFinish
        \endgroup
    \end{table}
\footnotetext{Kimi K2.6's meta-reasoning is  \textit{precisely} the failure mode \textit{CausalDS}'s combination of fully synthetic generation and first-class abstention grading is designed to expose: fitting the answer to an imagined exam sourced from a textbook rather than to the problem at hand.}

    On hard \texttt{scene\_000470} the same fork appears on an interval item.
    Astra's first turn is the whole argument (``Fossil reserves alone do not identify the population ATE without additional assumptions''); Opus writes the same nonparametric sentence and then runs a Wald ratio off a first stage with $R^2{=}0.006$, rescued by a homogeneity assumption read out of an OLS fit --- the model that states the principle is not the one that acts on it.
    On \texttt{scene\_000236} the front-door NIE is identifiable and GLM abstains with the mediation-formula condition applied one graph too early (``with $U$ latent, $E[Y_{x0,m}]$ cannot be identified via the mediation formula''); GPT-5.5 makes the same exit on \texttt{scene\_000188} from a single bundled script that decides identifiability and writes the null together.
    On \texttt{scene\_000070} identification held and the interval did not.
    Fable names the front-door criterion and calls the effect ``essentially null'' (${\approx}-0.017$, CI ${\approx}[-0.048, 0.013]$, SCM $-0.121$); Opus reaches the same number from the other side --- first sentence an open backdoor, submitted script a front-door functional, the near-zero located in the reconstruction (``the coefficient of $E\mid G$ is nearly zero \ldots measurement noise attenuation'').
    The field agrees on a tight interval around a mediator regression that the noisy bundles have flattened.

    \subsubsection{What the agent actually ran}
    \label{app:error_estimators}

    The census above codes the named plan; this section codes the bash that produced the number, on every leading-model data-science fail and interval pass with a trajectory ($1{,}598$ traces: $941$ fails, $657$ passes).
    The label is the identification functional the last substantial Python script implements --- front-door, $g$-formula, AIPW / double ML, IPW, 2SLS, OLS with covariates, an unadjusted mean difference, or a mediation product; a front-door mention in that script counts as a front-door attempt, since parametric front-door is often \texttt{sm.OLS} of $Y$ on $M$ and $X$, and \texttt{LinearRegression} imports are calibration helpers, not the estimator.

     $99.5\%$ of data-science fails read \texttt{calibration.parquet}, $40\%$ mention attenuation, reliability, or measurement error, and scikit-learn and statsmodels are used as calibration and OLS helpers rather than as a causal API.
    The functional does not separate a pass from a fail (Tab.~\ref{tab:error_estimator}).
    Pooled, fails look more front-door-heavy only because front-door hosts are over-represented among them ($226$ fails vs.\ $109$ interval passes); within a host the mixes line up, and about one in six last scripts match no pattern (Astra $40\%$) --- custom one-shots, with no hidden second functional.
    GPT-5.5 is the only model that leans on AIPW as a named object ($21\%$ of its data-science fails).
    The interval recipe is even more uniform: $95\%$ of data-science fails compute a bootstrap percentile interval, an analytic Wald interval, or both, and passes use the same recipes in similar proportions.
    So the continuous pile is not ``OLS on a front-door graph''.
    The agents reconstructed the concepts and ran a motif-typical script about as often as the passers; what differs is the interval they then submitted.

    \begin{table}[ht!]
        \centering
        \caption{\textbf{Last-script functional on leading-model traces, within structural motif.} Front-door $=$ the last Python script names or implements a front-door formula; OLS $=$ \texttt{sm.OLS}/\texttt{smf.ols} in that script and no front-door mention.}
        \label{tab:error_estimator}
        \footnotesize
        \begin{adjustbox}{max width=\textwidth}
            \begin{tabular}{@{}l c >{\raggedright\arraybackslash}p{.34\linewidth} >{\raggedright\arraybackslash}p{.34\linewidth}@{}}
                \toprule
                Host & \makecell{$n$\\{\scriptsize fail / pass}} & Leading functional, fails & Leading functional, passes \\
                \midrule
                Front-door & $226$ / $109$ & front-door $67.3\%$; OLS $20.4\%$ & front-door $55.0\%$; OLS $23.9\%$ \\
                Confounding & $115$ / $67$ & OLS $44.3\%$; $g$-formula $20.0\%$; AIPW $15.7\%$ & OLS $50.7\%$; $g$-formula $19.4\%$; AIPW $14.9\%$ \\
                \bottomrule
            \end{tabular}
        \end{adjustbox}
    \end{table}

    \subsubsection{Interval coverage and error geometry}
    \label{app:error_intervals}

\begin{figure}[ht!]
        \centering
        \includegraphics[width=.8\textwidth]{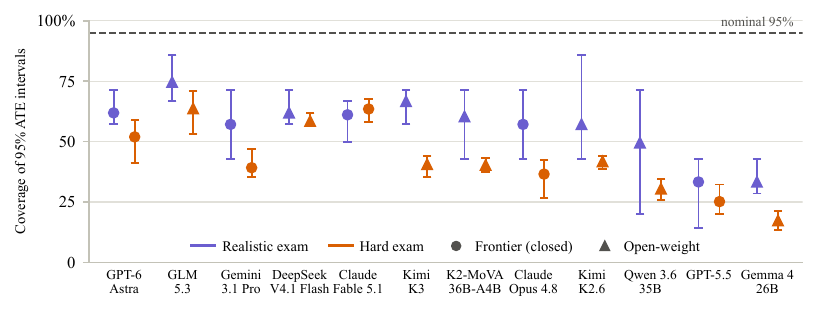}
        \caption{\textbf{Coverage of nominal $95\%$ ATE intervals.} Per model on the realistic (violet) and hard (orange) exam, over identifiable \texttt{ate\_uq\_95} tasks with a valid interval ($7$ and $34$ per run; fewer where a run returned no interval). Marker: mean over three runs; whisker: worst to best run. Circles: closed models; triangles: open-weight. Models in realistic-exam leaderboard order.}
        \label{fig:ate_coverage}
    \end{figure}

    Interval calibration is the weakest axis for every model. Two pools are used. Fig.~\ref{fig:ate_coverage} covers the identifiable \texttt{ate\_uq\_95} tasks with a valid interval ($7$ per run on the realistic exam, $34$ on the hard): coverage of the nominal $95\%$ interval is below nominal for every model on both exams, well below for most, and only loosely related to the leaderboard (the best-covered model on the realistic exam is not the score leader). The geometry below uses the wider pool of every graded effect, counterfactual, and mediation interval on both exams, all twelve models and three replications ($2{,}231$ usable triples $(\hat\tau, [L,U], \tau)$ recovered from the grade report, $15$ unparseable, observation variants joined from \texttt{exam.json} where the report omits them); its empirical coverage is $39\%$ ($37\%$ hard, $56\%$ realistic). The two coverage numbers are therefore not the same statistic; the matched-scene ablation (App.~\ref{app:observation_ablation}) sees the same under-coverage on a fixed set of scenes. The pattern is consistent with recent uncertainty-elicitation benchmarks, in which nominal intervals from frontier models cover far below their nominal level \citep{hobor_bayesian_2026, epstein_llms_2025, renda_openestimate_2025}.

    \paragraph{Error geometry.}
    An uncovered interval almost always contains the agent's own point and is almost never off-center relative to it.
    The miss therefore splits on whether the point is acceptable, the same cut as Fig.~\ref{fig:error_subtypes}b: $72\%$ of uncovered intervals are \emph{centered too narrow} --- point and midpoint within the point-OK band of $\tau$, interval too short --- and $27\%$ are a \emph{shifted point} (Tab.~\ref{tab:error_interval_geo}).
    Among the centered misses the median half-width would have to grow by $2.7\times$, at the same center, to cover.

    \begin{table}[ht!]
        \centering
        \caption{\textbf{Geometry of uncovered effect-family $95\%$ intervals.} Twelve models, both exams, three replications. Centered too narrow $=$ point and midpoint both within relative error $0.25$ of the SCM value; shifted point $=$ the point itself is outside that band. Width is the median normalized width; expansion is the factor on half-width, at the submitted center, that would have covered.}
        \label{tab:error_interval_geo}
        \footnotesize
        \begin{adjustbox}{max width=\textwidth}
            \begin{tabular}{@{}l c c c c@{}}
                \toprule
                Class & $n$ & \makecell{Share of\\{\scriptsize uncovered \%}} & \makecell{Med.\ norm.\\{\scriptsize width}} & \makecell{Med.\\{\scriptsize expansion}} \\
                \midrule
                Centered too narrow & 984 & 72.5 & 0.054 & 2.70 \\
                Shifted point & 371 & 27.3 & 0.100 & 7.10 \\
                Off-center interval & 3 & 0.2 & 0.260 & 1.95 \\
                \midrule
                All uncovered & 1358 & 100 & 0.065 & 3.59 \\
                All covered & 873 & --- & 0.074 & 0.44 \\
                \bottomrule
            \end{tabular}
        \end{adjustbox}
    \end{table}

\begin{figure}[ht!]
        \centering
        \includegraphics[width=\textwidth]{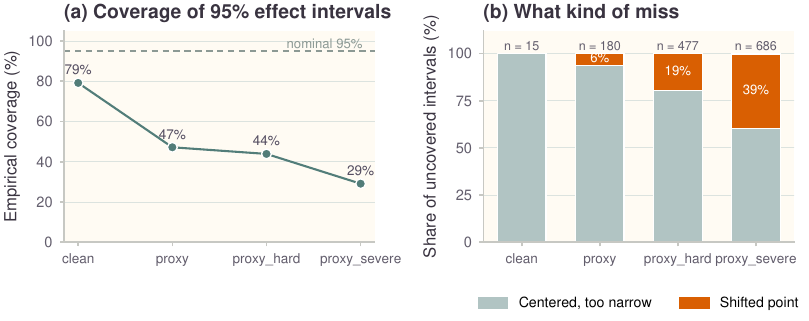}
        \caption{\textbf{Interval coverage by observation variant.} (a) Empirical coverage of the nominal $95\%$ effect-family intervals per observation variant (teal, measured; dashed gray line is the nominal level). (b) Share of uncovered intervals that are centered too narrow (light teal) versus a shifted point (orange), with the number of uncovered intervals above each bar.}
        \label{fig:error_interval_variant}
    \end{figure}

    The observation layer changes how often the point breaks; the shape of the typical miss stays the same (Fig.~\ref{fig:error_interval_variant}).
    Coverage falls from $79\%$ on \texttt{clean} to $29\%$ on \texttt{proxy\_severe}, and the shifted-point share of the misses rises from zero to $39\%$; even under the worst measurement the majority of misses stay centered too narrow, and every one of the $15$ \texttt{clean} misses does.
    A pooled covered-versus-uncovered width comparison is the wrong statistic, for the compositional reason of App.~\ref{app:error_estimators}: uncovered rows sit on the noisier views, which have wider intervals either way.
    Within a view, the models that cover on \texttt{proxy\_hard} and \texttt{proxy\_severe} did widen (median normalized width $0.062$ vs.\ $0.048$ and $0.165$ vs.\ $0.098$).

\begin{figure}[ht!]
        \centering
        \includegraphics[width=.62\textwidth]{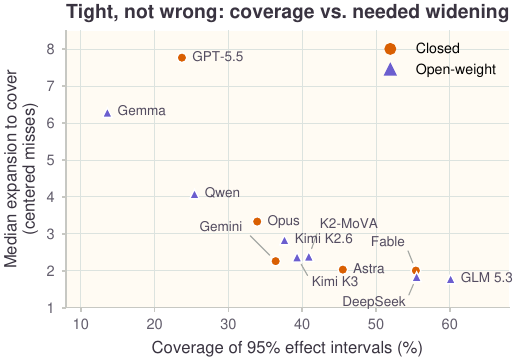}
        \caption{\textbf{Coverage versus the widening that would have covered.} Per model, empirical coverage of effect-family $95\%$ intervals against the median expansion factor on the centered-too-narrow misses (both exams, three replications). Orange circles closed models, violet triangles open-weight.}
        \label{fig:error_interval_expansion}
    \end{figure}

    Models differ in how tight they are willing to be; the shape of the error is shared (Fig.~\ref{fig:error_interval_expansion}).
    GLM~5.3 covers most often and would have needed a $1.8\times$ expansion on its centered misses; GPT-5.5 and Gemma~4 26B cover least and would have needed $6$--$8\times$.
    Claude Opus~4.8 is not the calibration leader once the target is a $95\%$ interval rather than a point, and GPT-6 Astra, the most data-science-heavy model, covers $46\%$ with intervals that are merely somewhat too tight.
    The data-science class is therefore a UQ class in a precise sense: overconfidence after a correct identification decision, with no second helping of false identifiability.

\definecolor{exammain}{HTML}{6B5ECF}
\definecolor{examhard}{HTML}{D95F02}

    \subsection{Stability analysis of exam generation}
    \label{app:leaderboard_ci}

    Because CausalDS is a generator rather than a fixed dataset, an exam is one draw of tasks from the scene collection, and a leaderboard is only as meaningful as it is stable under redrawing that exam. Sec.~\ref{sec:results} reports run-to-run variation as an SD over three replications, and App.~\ref{app:error_taxonomy} shows it concentrates in a few interval items and abstention flips; this section holds the models fixed and varies the exam instead. Tables~\ref{tab:realistic_leaderboard_ci} and~\ref{tab:hard_leaderboard_ci} repeat the per-run means of Tabs.~\ref{tab:realistic_leaderboard} and~\ref{tab:hard_leaderboard} with $95\%$ finite-exam intervals. Pass Rate uses a Wilson interval on the run-averaged rate, with the mean per-run denominator as sample size ($34$--$35.7$ on the realistic exam, $97$--$111$ on the hard). Every other column uses percentile intervals of a task bootstrap stratified by grading category: $B{=}10{,}000$ pseudo-exams redraw tasks with replacement within each of the five ground-truth scoring pools, so that a pseudo-exam has the same number of exact-match, interval, abstention, point, and graph/set tasks as the real one (realistic: exact-match content 18, interval 14, abstention 16, point 25, graph/set 27; hard: exact-match content 41, interval 59, abstention 56, point 34, graph/set 10), each replication is rescored as a full exam under the leaderboard routing, and the three values are averaged. The Rank column is the CausalDSScore rank over the same pseudo-exams; all-of-3 and any-of-3 are computed on the ground-truth binary pool of App.~\ref{app:pass@k} with the same intervals; Med.\ $F_1$-Loss is omitted because point estimate and interval are $0.000$ for all but two hard-exam models. Table~\ref{tab:rank_stability} uses the same pseudo-exams, in which every model is scored on the same drawn tasks, to give the probability that one model beats another on CausalDSScore (ties count one half). Table~\ref{tab:leaderboard_sensitivity} recomputes the ranks under four scoring-rule variants: interval cap $c{=}5$ or $20$ instead of $10$ in $S_{\mathrm{NR}}$ (Eq.~\ref{eq:nrel_aggregator}), the median instead of the capped mean for the interval term, and equal pool weights instead of count-proportional weights in Eq.~\ref{eq:causalds_score}.
    \begin{table}[ht!]
        \centering
        \caption{\textbf{Realistic-exam leaderboard with finite-exam intervals (100 tasks).} Per-run means over three replications with $95\%$ finite-exam intervals in brackets. Rank is the CausalDSScore rank with its interval; all-of-3 and any-of-3 are the fractions of the ground-truth binary pool ($n{=}34$) passed in all three and in at least one replication. Rows ordered by CausalDSScore; model names abbreviated as in Tab.~\ref{tab:realistic_per_task_type}. Model markers: \markclosed\ closed, \markopen\ open-weight.}
        \label{tab:realistic_leaderboard_ci}
        \footnotesize
        \begin{adjustbox}{width=\textwidth}
        \begin{tabular}{r l c c c c c c c}
            \toprule
            \rowcolor{exammain!10} \# & Model & CausalDSScore $\downarrow$ & Rank & Pass Rate $\uparrow$ & Med.\ NRel.\ Err $\downarrow$ & $S_{\mathrm{NR}}$ $\downarrow$ & all-of-3 $\uparrow$ & any-of-3 $\uparrow$ \\
            \midrule
            1 & \markclosed\,GPT-6 & \makecell{$0.211$\\{\scriptsize$[0.137,\,0.293]$}} & \makecell{1\\{\scriptsize$[1,\,2]$}} & \makecell{$96.1\%$\\{\scriptsize$[83.6,\,99.2]$}} & \makecell{$0.084$\\{\scriptsize$[0.038,\,0.281]$}} & \makecell{$0.507$\\{\scriptsize$[0.324,\,0.713]$}} & \makecell{$94.1\%$\\{\scriptsize$[85.3,\,100.0]$}} & \makecell{$100.0\%$\\{\scriptsize$[100.0,\,100.0]$}} \\
            2 & \markopen\,GLM & \makecell{$0.268$\\{\scriptsize$[0.193,\,0.353]$}} & \makecell{2\\{\scriptsize$[2,\,6]$}} & \makecell{$72.8\%$\\{\scriptsize$[56.2,\,84.8]$}} & \makecell{$0.080$\\{\scriptsize$[0.030,\,0.294]$}} & \makecell{$0.454$\\{\scriptsize$[0.279,\,0.657]$}} & \makecell{$64.7\%$\\{\scriptsize$[52.9,\,76.5]$}} & \makecell{$82.4\%$\\{\scriptsize$[70.6,\,94.1]$}} \\
            3 & \markclosed\,Gemini & \makecell{$0.275$\\{\scriptsize$[0.197,\,0.368]$}} & \makecell{3\\{\scriptsize$[2,\,7]$}} & \makecell{$78.7\%$\\{\scriptsize$[62.5,\,89.1]$}} & \makecell{$0.140$\\{\scriptsize$[0.040,\,0.301]$}} & \makecell{$0.525$\\{\scriptsize$[0.341,\,0.753]$}} & \makecell{$67.6\%$\\{\scriptsize$[52.9,\,79.4]$}} & \makecell{$88.2\%$\\{\scriptsize$[76.5,\,97.1]$}} \\
            4 & \markopen\,DeepSeek & \makecell{$0.281$\\{\scriptsize$[0.202,\,0.371]$}} & \makecell{4\\{\scriptsize$[2,\,8]$}} & \makecell{$73.5\%$\\{\scriptsize$[56.9,\,85.4]$}} & \makecell{$0.149$\\{\scriptsize$[0.041,\,0.291]$}} & \makecell{$0.489$\\{\scriptsize$[0.306,\,0.704]$}} & \makecell{$64.7\%$\\{\scriptsize$[50.0,\,79.4]$}} & \makecell{$85.3\%$\\{\scriptsize$[73.5,\,94.1]$}} \\
            5 & \markclosed\,Fable & \makecell{$0.294$\\{\scriptsize$[0.204,\,0.396]$}} & \makecell{5\\{\scriptsize$[2,\,9]$}} & \makecell{$68.6\%$\\{\scriptsize$[52.0,\,81.4]$}} & \makecell{$0.103$\\{\scriptsize$[0.028,\,0.318]$}} & \makecell{$0.501$\\{\scriptsize$[0.293,\,0.745]$}} & \makecell{$61.8\%$\\{\scriptsize$[47.1,\,76.5]$}} & \makecell{$79.4\%$\\{\scriptsize$[64.7,\,91.2]$}} \\
            6 & \markopen\,K3 & \makecell{$0.294$\\{\scriptsize$[0.225,\,0.374]$}} & \makecell{6\\{\scriptsize$[4,\,9]$}} & \makecell{$66.7\%$\\{\scriptsize$[49.9,\,80.1]$}} & \makecell{$0.099$\\{\scriptsize$[0.050,\,0.307]$}} & \makecell{$0.464$\\{\scriptsize$[0.302,\,0.653]$}} & \makecell{$52.9\%$\\{\scriptsize$[41.2,\,64.7]$}} & \makecell{$82.4\%$\\{\scriptsize$[70.6,\,94.1]$}} \\
            7 & \markopen\,K2-MoVA & \makecell{$0.300$\\{\scriptsize$[0.231,\,0.369]$}} & \makecell{7\\{\scriptsize$[1,\,9]$}} & \makecell{$72.1\%$\\{\scriptsize$[55.5,\,84.2]$}} & \makecell{$0.166$\\{\scriptsize$[0.065,\,0.307]$}} & \makecell{$0.527$\\{\scriptsize$[0.376,\,0.678]$}} & \makecell{$58.8\%$\\{\scriptsize$[44.1,\,73.5]$}} & \makecell{$85.3\%$\\{\scriptsize$[73.5,\,94.1]$}} \\
            8 & \markclosed\,Opus & \makecell{$0.354$\\{\scriptsize$[0.238,\,0.492]$}} & \makecell{8\\{\scriptsize$[2,\,10]$}} & \makecell{$81.4\%$\\{\scriptsize$[65.4,\,91.0]$}} & \makecell{$0.186$\\{\scriptsize$[0.052,\,0.343]$}} & \makecell{$0.747$\\{\scriptsize$[0.465,\,1.087]$}} & \makecell{$76.5\%$\\{\scriptsize$[64.7,\,88.2]$}} & \makecell{$88.2\%$\\{\scriptsize$[76.5,\,97.1]$}} \\
            9 & \markopen\,K2.6 & \makecell{$0.385$\\{\scriptsize$[0.264,\,0.536]$}} & \makecell{9\\{\scriptsize$[6,\,11]$}} & \makecell{$76.8\%$\\{\scriptsize$[60.5,\,87.8]$}} & \makecell{$0.186$\\{\scriptsize$[0.067,\,0.304]$}} & \makecell{$0.790$\\{\scriptsize$[0.489,\,1.177]$}} & \makecell{$61.8\%$\\{\scriptsize$[50.0,\,73.5]$}} & \makecell{$91.2\%$\\{\scriptsize$[82.4,\,100.0]$}} \\
            10 & \markopen\,Qwen & \makecell{$0.392$\\{\scriptsize$[0.288,\,0.518]$}} & \makecell{10\\{\scriptsize$[3,\,11]$}} & \makecell{$66.5\%$\\{\scriptsize$[50.1,\,79.7]$}} & \makecell{$0.203$\\{\scriptsize$[0.068,\,0.324]$}} & \makecell{$0.764$\\{\scriptsize$[0.489,\,1.109]$}} & \makecell{$55.9\%$\\{\scriptsize$[44.1,\,67.6]$}} & \makecell{$82.4\%$\\{\scriptsize$[70.6,\,94.1]$}} \\
            11 & \markclosed\,GPT-5.5 & \makecell{$0.560$\\{\scriptsize$[0.341,\,0.804]$}} & \makecell{11\\{\scriptsize$[9,\,12]$}} & \makecell{$81.0\%$\\{\scriptsize$[65.2,\,90.6]$}} & \makecell{$0.223$\\{\scriptsize$[0.069,\,0.353]$}} & \makecell{$1.308$\\{\scriptsize$[0.739,\,1.946]$}} & \makecell{$70.6\%$\\{\scriptsize$[55.9,\,85.3]$}} & \makecell{$94.1\%$\\{\scriptsize$[85.3,\,100.0]$}} \\
            12 & \markopen\,Gemma & \makecell{$0.662$\\{\scriptsize$[0.434,\,0.929]$}} & \makecell{12\\{\scriptsize$[11,\,12]$}} & \makecell{$59.2\%$\\{\scriptsize$[42.7,\,73.9]$}} & \makecell{$0.267$\\{\scriptsize$[0.081,\,0.404]$}} & \makecell{$1.352$\\{\scriptsize$[0.768,\,2.038]$}} & \makecell{$41.2\%$\\{\scriptsize$[29.4,\,52.9]$}} & \makecell{$79.4\%$\\{\scriptsize$[67.6,\,91.2]$}} \\
            \bottomrule
        \end{tabular}
        \end{adjustbox}
    \end{table}
    \begin{table}[ht!]
        \centering
        \caption{\textbf{Hard-exam leaderboard with finite-exam intervals (200 tasks).} Per-run means over three replications with $95\%$ finite-exam intervals in brackets. Rank is the CausalDSScore rank with its interval; all-of-3 and any-of-3 are the fractions of the ground-truth binary pool ($n{=}97$) passed in all three and in at least one replication. Rows ordered by CausalDSScore; model names abbreviated as in Tab.~\ref{tab:realistic_per_task_type}. Model markers: \markclosed\ closed, \markopen\ open-weight.}
        \label{tab:hard_leaderboard_ci}
        \footnotesize
        \begin{adjustbox}{width=\textwidth}
        \begin{tabular}{r l c c c c c c c}
            \toprule
            \rowcolor{examhard!10} \# & Model & CausalDSScore $\downarrow$ & Rank & Pass Rate $\uparrow$ & Med.\ NRel.\ Err $\downarrow$ & $S_{\mathrm{NR}}$ $\downarrow$ & all-of-3 $\uparrow$ & any-of-3 $\uparrow$ \\
            \midrule
            1 & \markopen\,GLM & \makecell{$0.559$\\{\scriptsize$[0.422,\,0.708]$}} & \makecell{1\\{\scriptsize$[1,\,3]$}} & \makecell{$73.0\%$\\{\scriptsize$[63.8,\,80.5]$}} & \makecell{$0.121$\\{\scriptsize$[0.082,\,0.196]$}} & \makecell{$0.988$\\{\scriptsize$[0.677,\,1.336]$}} & \makecell{$68.0\%$\\{\scriptsize$[58.8,\,77.3]$}} & \makecell{$89.7\%$\\{\scriptsize$[83.5,\,95.9]$}} \\
            2 & \markclosed\,Fable & \makecell{$0.627$\\{\scriptsize$[0.426,\,0.852]$}} & \makecell{2\\{\scriptsize$[1,\,4]$}} & \makecell{$85.4\%$\\{\scriptsize$[77.1,\,91.0]$}} & \makecell{$0.125$\\{\scriptsize$[0.086,\,0.161]$}} & \makecell{$1.221$\\{\scriptsize$[0.786,\,1.714]$}} & \makecell{$81.4\%$\\{\scriptsize$[73.2,\,88.7]$}} & \makecell{$90.7\%$\\{\scriptsize$[84.5,\,95.9]$}} \\
            3 & \markopen\,DeepSeek & \makecell{$0.639$\\{\scriptsize$[0.447,\,0.855]$}} & \makecell{3\\{\scriptsize$[1,\,4]$}} & \makecell{$83.9\%$\\{\scriptsize$[75.5,\,89.8]$}} & \makecell{$0.140$\\{\scriptsize$[0.111,\,0.204]$}} & \makecell{$1.258$\\{\scriptsize$[0.832,\,1.737]$}} & \makecell{$81.4\%$\\{\scriptsize$[73.2,\,88.7]$}} & \makecell{$92.8\%$\\{\scriptsize$[87.6,\,96.9]$}} \\
            4 & \markclosed\,GPT-6 & \makecell{$0.733$\\{\scriptsize$[0.524,\,0.965]$}} & \makecell{4\\{\scriptsize$[2,\,4]$}} & \makecell{$90.4\%$\\{\scriptsize$[82.9,\,94.8]$}} & \makecell{$0.155$\\{\scriptsize$[0.112,\,0.249]$}} & \makecell{$1.475$\\{\scriptsize$[1.029,\,1.974]$}} & \makecell{$86.6\%$\\{\scriptsize$[80.4,\,92.8]$}} & \makecell{$94.8\%$\\{\scriptsize$[90.7,\,99.0]$}} \\
            5 & \markopen\,K2.6 & \makecell{$0.950$\\{\scriptsize$[0.715,\,1.205]$}} & \makecell{5\\{\scriptsize$[5,\,8]$}} & \makecell{$78.5\%$\\{\scriptsize$[69.5,\,85.4]$}} & \makecell{$0.213$\\{\scriptsize$[0.147,\,0.364]$}} & \makecell{$1.886$\\{\scriptsize$[1.361,\,2.452]$}} & \makecell{$73.2\%$\\{\scriptsize$[63.9,\,82.5]$}} & \makecell{$88.7\%$\\{\scriptsize$[81.4,\,94.8]$}} \\
            6 & \markclosed\,Gemini & \makecell{$0.963$\\{\scriptsize$[0.744,\,1.207]$}} & \makecell{6\\{\scriptsize$[5,\,9]$}} & \makecell{$64.5\%$\\{\scriptsize$[54.7,\,73.3]$}} & \makecell{$0.232$\\{\scriptsize$[0.136,\,0.372]$}} & \makecell{$1.727$\\{\scriptsize$[1.253,\,2.253]$}} & \makecell{$57.7\%$\\{\scriptsize$[49.5,\,67.0]$}} & \makecell{$73.2\%$\\{\scriptsize$[64.9,\,81.4]$}} \\
            7 & \markopen\,K3 & \makecell{$0.966$\\{\scriptsize$[0.752,\,1.195]$}} & \makecell{7\\{\scriptsize$[5,\,8]$}} & \makecell{$64.0\%$\\{\scriptsize$[54.2,\,72.8]$}} & \makecell{$0.221$\\{\scriptsize$[0.148,\,0.333]$}} & \makecell{$1.728$\\{\scriptsize$[1.264,\,2.228]$}} & \makecell{$54.6\%$\\{\scriptsize$[45.4,\,63.9]$}} & \makecell{$75.3\%$\\{\scriptsize$[67.0,\,83.5]$}} \\
            8 & \markopen\,K2-MoVA & \makecell{$0.967$\\{\scriptsize$[0.744,\,1.207]$}} & \makecell{8\\{\scriptsize$[5,\,8]$}} & \makecell{$71.7\%$\\{\scriptsize$[62.5,\,79.3]$}} & \makecell{$0.172$\\{\scriptsize$[0.120,\,0.288]$}} & \makecell{$1.972$\\{\scriptsize$[1.436,\,2.548]$}} & \makecell{$69.1\%$\\{\scriptsize$[59.8,\,78.4]$}} & \makecell{$88.7\%$\\{\scriptsize$[82.5,\,94.8]$}} \\
            9 & \markclosed\,Opus & \makecell{$1.136$\\{\scriptsize$[0.888,\,1.398]$}} & \makecell{9\\{\scriptsize$[8,\,10]$}} & \makecell{$65.2\%$\\{\scriptsize$[55.7,\,73.7]$}} & \makecell{$0.275$\\{\scriptsize$[0.190,\,0.444]$}} & \makecell{$2.237$\\{\scriptsize$[1.663,\,2.841]$}} & \makecell{$61.9\%$\\{\scriptsize$[52.6,\,71.1]$}} & \makecell{$78.4\%$\\{\scriptsize$[70.1,\,86.6]$}} \\
            10 & \markopen\,Qwen & \makecell{$1.337$\\{\scriptsize$[1.101,\,1.579]$}} & \makecell{10\\{\scriptsize$[10,\,11]$}} & \makecell{$57.6\%$\\{\scriptsize$[48.2,\,66.5]$}} & \makecell{$0.375$\\{\scriptsize$[0.244,\,1.084]$}} & \makecell{$2.640$\\{\scriptsize$[2.083,\,3.214]$}} & \makecell{$47.4\%$\\{\scriptsize$[38.1,\,56.7]$}} & \makecell{$77.3\%$\\{\scriptsize$[69.1,\,85.6]$}} \\
            11 & \markclosed\,GPT-5.5 & \makecell{$1.416$\\{\scriptsize$[1.175,\,1.658]$}} & \makecell{11\\{\scriptsize$[10,\,11]$}} & \makecell{$69.0\%$\\{\scriptsize$[59.9,\,76.9]$}} & \makecell{$0.437$\\{\scriptsize$[0.287,\,1.641]$}} & \makecell{$3.171$\\{\scriptsize$[2.563,\,3.787]$}} & \makecell{$63.9\%$\\{\scriptsize$[54.6,\,73.2]$}} & \makecell{$90.7\%$\\{\scriptsize$[85.6,\,95.9]$}} \\
            12 & \markopen\,Gemma & \makecell{$1.594$\\{\scriptsize$[1.334,\,1.867]$}} & \makecell{12\\{\scriptsize$[11,\,12]$}} & \makecell{$47.4\%$\\{\scriptsize$[38.3,\,56.7]$}} & \makecell{$1.022$\\{\scriptsize$[0.502,\,2.177]$}} & \makecell{$3.249$\\{\scriptsize$[2.610,\,3.905]$}} & \makecell{$32.0\%$\\{\scriptsize$[23.7,\,40.2]$}} & \makecell{$75.3\%$\\{\scriptsize$[67.0,\,83.5]$}} \\
            \bottomrule
        \end{tabular}
        \end{adjustbox}
    \end{table}

    Resampling within grading categories describes a redrawn exam of exactly the same grading composition. A freshly generated exam would not have that: the generator fixes scene strata, not scoring pools, so the closer analogue resamples within the groups the exam was assembled from --- on the hard exam its seven construction strata (identifiable front-door scenes (30), identifiable grafted scenes (46), identifiable mediator scenes (54), identifiable simple scenes (14), non-identifiable double-negative-control scenes (13), non-identifiable IV scenes (22), other non-identifiable scenes (21)), on the realistic exam a plain resample of its 100 tasks --- letting the pool mix and the CausalDSScore weights vary as they would from one generated exam to the next. Under that scheme the CausalDSScore intervals widen by 1.3--1.6$\times$ on the realistic exam and 1.1--1.4$\times$ on the hard, and no rank interval moves by more than one place at either end. The ranking is therefore stable to how the exam was composed, not only to which tasks happened to fill a fixed composition.
    \begin{table}[ht!]
        \centering
        \caption{\textbf{Paired rank stability.} Probability that the row model has the lower (better) CausalDSScore than the column model over the pseudo-exams of Tabs.~\ref{tab:realistic_leaderboard_ci} and~\ref{tab:hard_leaderboard_ci}. \textcolor{exammain}{realistic exam above the diagonal}, \textcolor{examhard}{hard exam below the diagonal}. Rows and columns follow the realistic-exam rank; the hard-exam rank is in parentheses. Model markers: \markclosed\ closed, \markopen\ open-weight.}
        \label{tab:rank_stability}
        \footnotesize
        \begin{adjustbox}{width=\textwidth}
        \begin{tabular}{l cccccccccccc}
            \toprule
             & 1 & 2 & 3 & 4 & 5 & 6 & 7 & 8 & 9 & 10 & 11 & 12 \\
            \midrule
            1~\markclosed\,GPT-6 Astra (4) & -- & \textcolor{exammain}{0.98} & \textcolor{exammain}{0.98} & \textcolor{exammain}{0.99} & \textcolor{exammain}{0.99} & \textcolor{exammain}{1.00} & \textcolor{exammain}{0.96} & \textcolor{exammain}{1.00} & \textcolor{exammain}{1.00} & \textcolor{exammain}{1.00} & \textcolor{exammain}{1.00} & \textcolor{exammain}{1.00} \\
            2~\markopen\,GLM 5.3 (1) & \textcolor{examhard}{0.96} & -- & \textcolor{exammain}{0.59} & \textcolor{exammain}{0.75} & \textcolor{exammain}{0.78} & \textcolor{exammain}{0.97} & \textcolor{exammain}{0.76} & \textcolor{exammain}{0.92} & \textcolor{exammain}{0.99} & \textcolor{exammain}{0.95} & \textcolor{exammain}{0.99} & \textcolor{exammain}{1.00} \\
            3~\markclosed\,Gemini 3.1 Pro (6) & \textcolor{examhard}{0.00} & \textcolor{examhard}{0.00} & -- & \textcolor{exammain}{0.56} & \textcolor{exammain}{0.69} & \textcolor{exammain}{0.78} & \textcolor{exammain}{0.70} & \textcolor{exammain}{0.94} & \textcolor{exammain}{0.98} & \textcolor{exammain}{0.94} & \textcolor{exammain}{0.99} & \textcolor{exammain}{1.00} \\
            4~\markopen\,DeepSeek V4.1 Flash (3) & \textcolor{examhard}{0.89} & \textcolor{examhard}{0.19} & \textcolor{examhard}{1.00} & -- & \textcolor{exammain}{0.66} & \textcolor{exammain}{0.82} & \textcolor{exammain}{0.68} & \textcolor{exammain}{0.89} & \textcolor{exammain}{0.99} & \textcolor{exammain}{0.92} & \textcolor{exammain}{0.99} & \textcolor{exammain}{1.00} \\
            5~\markclosed\,Claude Fable 5.1 (2) & \textcolor{examhard}{0.88} & \textcolor{examhard}{0.23} & \textcolor{examhard}{1.00} & \textcolor{examhard}{0.56} & -- & \textcolor{exammain}{0.52} & \textcolor{exammain}{0.56} & \textcolor{exammain}{0.81} & \textcolor{exammain}{0.96} & \textcolor{exammain}{0.89} & \textcolor{exammain}{0.98} & \textcolor{exammain}{1.00} \\
            6~\markopen\,Kimi K3 (7) & \textcolor{examhard}{0.00} & \textcolor{examhard}{0.00} & \textcolor{examhard}{0.48} & \textcolor{examhard}{0.00} & \textcolor{examhard}{0.00} & -- & \textcolor{exammain}{0.54} & \textcolor{exammain}{0.81} & \textcolor{exammain}{0.95} & \textcolor{exammain}{0.90} & \textcolor{exammain}{0.98} & \textcolor{exammain}{1.00} \\
            7~\markopen\,K2-MoVA 36B-A4B (8) & \textcolor{examhard}{0.00} & \textcolor{examhard}{0.00} & \textcolor{examhard}{0.48} & \textcolor{examhard}{0.00} & \textcolor{examhard}{0.00} & \textcolor{examhard}{0.51} & -- & \textcolor{exammain}{0.78} & \textcolor{exammain}{0.92} & \textcolor{exammain}{0.97} & \textcolor{exammain}{0.99} & \textcolor{exammain}{1.00} \\
            8~\markclosed\,Claude Opus 4.8 (9) & \textcolor{examhard}{0.00} & \textcolor{examhard}{0.00} & \textcolor{examhard}{0.03} & \textcolor{examhard}{0.00} & \textcolor{examhard}{0.00} & \textcolor{examhard}{0.02} & \textcolor{examhard}{0.01} & -- & \textcolor{exammain}{0.64} & \textcolor{exammain}{0.74} & \textcolor{exammain}{0.97} & \textcolor{exammain}{1.00} \\
            9~\markopen\,Kimi K2.6 (5) & \textcolor{examhard}{0.01} & \textcolor{examhard}{0.00} & \textcolor{examhard}{0.56} & \textcolor{examhard}{0.00} & \textcolor{examhard}{0.00} & \textcolor{examhard}{0.58} & \textcolor{examhard}{0.57} & \textcolor{examhard}{0.98} & -- & \textcolor{exammain}{0.54} & \textcolor{exammain}{0.94} & \textcolor{exammain}{1.00} \\
            10~\markopen\,Qwen 3.6 35B (10) & \textcolor{examhard}{0.00} & \textcolor{examhard}{0.00} & \textcolor{examhard}{0.00} & \textcolor{examhard}{0.00} & \textcolor{examhard}{0.00} & \textcolor{examhard}{0.00} & \textcolor{examhard}{0.00} & \textcolor{examhard}{0.02} & \textcolor{examhard}{0.00} & -- & \textcolor{exammain}{0.96} & \textcolor{exammain}{1.00} \\
            11~\markclosed\,GPT-5.5 (11) & \textcolor{examhard}{0.00} & \textcolor{examhard}{0.00} & \textcolor{examhard}{0.00} & \textcolor{examhard}{0.00} & \textcolor{examhard}{0.00} & \textcolor{examhard}{0.00} & \textcolor{examhard}{0.00} & \textcolor{examhard}{0.01} & \textcolor{examhard}{0.00} & \textcolor{examhard}{0.25} & -- & \textcolor{exammain}{0.92} \\
            12~\markopen\,Gemma 4 26B (12) & \textcolor{examhard}{0.00} & \textcolor{examhard}{0.00} & \textcolor{examhard}{0.00} & \textcolor{examhard}{0.00} & \textcolor{examhard}{0.00} & \textcolor{examhard}{0.00} & \textcolor{examhard}{0.00} & \textcolor{examhard}{0.00} & \textcolor{examhard}{0.00} & \textcolor{examhard}{0.01} & \textcolor{examhard}{0.02} & -- \\
            \bottomrule
        \end{tabular}
        \end{adjustbox}
    \end{table}

    \begin{table}[ht!]
        \centering
        \caption{\textbf{Rank under scoring-rule variants.} CausalDSScore rank (1 = best) under the scoring-rule variants defined in the text, both exams; base is the published rule. Rows ordered by the realistic-exam CausalDSScore. Model markers: \markclosed\ closed, \markopen\ open-weight.}
        \label{tab:leaderboard_sensitivity}
        \footnotesize
        \begin{adjustbox}{max width=\textwidth}
        \begin{tabular}{l ccccc c ccccc}
            \toprule
             & \multicolumn{5}{c}{\examMain[Realistic exam]} & & \multicolumn{5}{c}{\examHard[Hard exam]} \\
            \cmidrule(lr){2-6}\cmidrule(lr){8-12}
            Model & base & $c{=}5$ & $c{=}20$ & median & equal $w$ & & base & $c{=}5$ & $c{=}20$ & median & equal $w$ \\
            \midrule
            \markclosed\,GPT-6 Astra & 1 & 1 & 1 & 1 & 1 &  & 4 & 4 & 4 & 3 & 4 \\
            \markopen\,GLM 5.3 & 2 & 3 & 2 & 3 & 2 &  & 1 & 1 & 1 & 4 & 1 \\
            \markclosed\,Gemini 3.1 Pro & 3 & 2 & 4 & 2 & 3 &  & 6 & 7 & 8 & 8 & 5 \\
            \markopen\,DeepSeek V4.1 Flash & 4 & 4 & 3 & 7 & 4 &  & 3 & 3 & 3 & 2 & 3 \\
            \markclosed\,Claude Fable 5.1 & 5 & 5 & 5 & 6 & 7 &  & 2 & 2 & 2 & 1 & 2 \\
            \markopen\,Kimi K3 & 6 & 6 & 6 & 8 & 5 &  & 7 & 8 & 5 & 7 & 6 \\
            \markopen\,K2-MoVA 36B-A4B & 7 & 7 & 7 & 5 & 6 &  & 8 & 6 & 6 & 6 & 8 \\
            \markclosed\,Claude Opus 4.8 & 8 & 9 & 8 & 4 & 8 &  & 9 & 9 & 9 & 9 & 9 \\
            \markopen\,Kimi K2.6 & 9 & 8 & 9 & 9 & 9 &  & 5 & 5 & 7 & 5 & 7 \\
            \markopen\,Qwen 3.6 35B & 10 & 10 & 10 & 11 & 10 &  & 10 & 11 & 10 & 10 & 10 \\
            \markclosed\,GPT-5.5 & 11 & 11 & 11 & 10 & 11 &  & 11 & 10 & 11 & 11 & 11 \\
            \markopen\,Gemma 4 26B & 12 & 12 & 12 & 12 & 12 &  & 12 & 12 & 12 & 12 & 12 \\
            \bottomrule
        \end{tabular}
        \end{adjustbox}
    \end{table}


    On the realistic exam the intervals support three orderings: GPT-6 Astra first (rank interval $[1,2]$, beating every other model in at least $96\%$ of pseudo-exams), Gemma~4 26B last ($[11,12]$), GPT-5.5 eleventh ($[9,12]$).
    Ranks $2$--$7$ (GLM~5.3, Gemini~3.1 Pro, DeepSeek~V4.1 Flash, Claude Fable~5.1, Kimi~K3, K2-MoVA) are one band --- every rank interval contains $2$ and reaches $6$--$9$, and neighbors beat each other with probability $0.52$--$0.78$ --- and Claude Opus~4.8, Kimi~K2.6, and Qwen~3.6 35B form a second band at $8$--$10$.
    A $100$-task exam cannot order either band, and its Pass Rate intervals are $\pm 15$ points wide on a binary pool of $34$.
    The hard exam is a different instrument with larger pools, and its intervals are correspondingly narrower: GLM~5.3 is first with probability $0.77$--$0.96$ against the next three, Fable and DeepSeek are tied for second ($0.56$), Astra is fourth ($[2,4]$), Kimi~K2.6, Gemini, Kimi~K3, and K2-MoVA are a band at $5$--$8$ with pairwise probabilities near $0.5$, and the bottom four are ordered except Qwen and GPT-5.5 ($[10,11]$ each).
    The scoring rule adds little uncertainty of its own on either exam.
    Under the interval-cap and equal-weight variants no rank moves more than two places and the extremes are fixed; only the median variant reshuffles the middle (Opus $8 \to 4$ on the realistic exam, GLM $1 \to 4$ on the hard), because it discards the capped interval tail that the mean deliberately keeps (Sec.~\ref{sec:scoring}).
    What survives both the task draw and the rule is, on the realistic exam, Astra first and Gemma last with GPT-5.5 just above it; on the hard exam, GLM, Fable, DeepSeek, and Astra as the top four and Gemma, GPT-5.5, and Qwen as the bottom three.

    \subsection{Per-axis reliability across three runs}
    \label{app:pass@k}

    Three replications already separate the models on reliability. In Tabs.~\ref{tab:realistic_leaderboard} and~\ref{tab:hard_leaderboard} all-of-3 spreads much more than any-of-3, widest for the weaker open models, and the best--worst CausalDSScore range runs from a few percent of the mean to a factor of two on the realistic exam. The variance is not in the point estimates, which reproduce almost exactly across restarts (App.~\ref{app:pass@k}, Tab.~\ref{tab:realistic_passk_continuous}); it comes from a few interval tasks that swing between near-correct and the cap, and from abstention flips re-scored in a small pass/fail pool. Persistence follows the same pattern: a data-science miss recurs on all three runs about half the time, a reasoning miss about a third, a tool exit rarely (Fig.~\ref{fig:error_persistence}).

    \paragraph{Setup.}
    The finite-exam intervals of those leaderboard columns are in App.~\ref{app:leaderboard_ci}; here we give their definitions, the continuous analogue, and the per-axis breakdowns.
    Each task has $N{=}3$ complete independent restarts, each a fresh single-shot trajectory.
    Because the ground truth is generated, a task can be assigned to a scoring pool by what the SCM says rather than by what the model answered: non-identifiable targets form the binary abstention pool, where a committed answer is scored as wrong; identifiable targets keep their native metric, with failed attempts handled according to that metric.
    The pools are therefore fixed and common to all twelve models --- binary $34$ / $97$, point $25$ / $34$, interval $14$ / $59$, graph/set $27$ / $10$ on the realistic / hard exam.
    Point and interval losses are pooled separately here while the leaderboard's Med.~NRel.~Err mixes them, so the continuous numbers below are not comparable to Tab.~\ref{tab:realistic_leaderboard}.
    Any-of-$N$ is the fraction of tasks passed in at least one of the $N$ observed runs; all-of-$N$ is the fraction passed in every run. The Mean rows average each task's success fraction over the fixed ground-truth binary pool. These summaries use all $N{=}3$ runs. The main-text mini-exam instead uses $N{=}10$ attempts to estimate pass@$k$ and pass\^{}$k$ over size-$k$ subsets for $k<N$ (Sec.~\ref{sec:results}); at $k=N$, those estimators reduce to any-of-$N$ and all-of-$N$.
    For losses, Mean averages each task's valid losses, best-of-3 takes its smallest valid loss, and worst-of-3 its largest when all three attempts are valid; these are the $k=1$ and $k=N=3$ cases of Eq.~\ref{eq:subset_extrema}. Best-of-3 is an oracle selection using the graded loss \citep{brown_monkeys_2024}. The reported values are equal-task means over eligible tasks, whose pools can differ between summaries. Set/graph failures receive loss $1$; point and interval failures have no valid loss.

    \paragraph{Subset estimators.}
    The mini-exam figure of Sec.~\ref{sec:results} (Fig.~\ref{fig:targeted_reliability}) uses the following estimators with $N{=}10$ attempts; the Mean, best-of-3, and worst-of-3 columns above are their $k=1$ and $k=N=3$ cases. On the numeric pools an invalid attempt has no loss and is handled by the estimators' conditioning.
    For the binary pool we use the unbiased hypergeometric pass@$k$ and pass\^{}$k$ estimators over the $N{=}10$ attempts \citep{chen_codex_2021, yao_tau_bench_2024}. For the loss-scored tasks we use the continuous analogue, as in eff@$k$ \citep{qiu_how_2025} and worst-at-$k$ \citep{arora_healthbench_2025}: the expected best (smallest) and worst (largest) loss over a uniformly drawn $k$-subset of the attempts. With a task's $v$ valid losses sorted as $\ell_{(1)}\le\dots\le\ell_{(v)}$,
    {\small
    \begin{equation}
        b(k)=\frac{\sum_{r=1}^{v}\ell_{(r)}\binom{N-r}{k-1}}{\binom{N}{k}-\binom{N-v}{k}},
        \qquad
        w(k)=\frac{\sum_{r=1}^{v}\ell_{(r)}\binom{r-1}{k-1}}{\binom{v}{k}},
        \label{eq:subset_extrema}
    \end{equation}
    }
    where $b(k)$ (\textit{best-of-$k$}) conditions on the subset containing at least one valid loss and $w(k)$ (\textit{worst-of-$k$}) on all $k$ losses being valid; both reduce to the mean loss at $k=1$, so no attempt is privileged. Curves are equal-task means over the tasks for which the statistic is defined, so that a single failing task stays visible. The strips under the loss panels show the expected number of such tasks, the sum over tasks of the conditioning probability, which is the denominator in Eq.~\ref{eq:subset_extrema} divided by $\binom{N}{k}$. Failures on bounded losses (set-valued $1-F_1$) score $1$ and count as valid; failures on the unbounded point and interval losses stay excluded.

\begingroup
\footnotesize
\setlength{\tabcolsep}{2pt}
\begin{longtable}{@{}l ccc @{\hspace{12pt}} ccc@{}}
\caption{\textbf{Continuous-task reliability.} Equal-task means of the per-task mean, best-of-3, and worst-of-3 losses over three full-exam runs. Parentheses give mean valid tasks/run for Mean, tasks with at least one valid loss for best-of-3, and tasks valid in all three runs for worst-of-3, each over the pool total. Set/graph failures have loss $1$; point and interval losses are uncapped. Models follow realistic-exam CausalDSScore; lowest per exam and pool in bold. Model markers: \markclosed\ closed, \markopen\ open-weight.}
\label{tab:realistic_passk_continuous}\\
\toprule
& \multicolumn{3}{c}{\examMain} & \multicolumn{3}{c}{\examHard} \\
\cmidrule(lr){2-4}\cmidrule(lr){5-7}
Model & Mean & best-of-3 & worst-of-3 & Mean & best-of-3 & worst-of-3 \\
\midrule
\endfirsthead
\multicolumn{7}{l}{\tablename~\thetable\ (continued)} \\
\toprule
& \multicolumn{3}{c}{\examMain} & \multicolumn{3}{c}{\examHard} \\
\cmidrule(lr){2-4}\cmidrule(lr){5-7}
Model & Mean & best-of-3 & worst-of-3 & Mean & best-of-3 & worst-of-3 \\
\midrule
\endhead
\midrule
\multicolumn{7}{r}{Continued on next page} \\
\endfoot
\bottomrule
\endlastfoot
\addlinespace[3pt]
\multicolumn{7}{l}{\textit{Point (NRelErr)}} \\*
\markclosed\,GPT-6 Astra & $0.125$\,{\scriptsize(25.0/25)} & $0.120$\,{\scriptsize(25/25)} & $0.132$\,{\scriptsize(25/25)} & $0.131$\,{\scriptsize(34.0/34)} & $0.069$\,{\scriptsize(34/34)} & $0.221$\,{\scriptsize(34/34)} \\*
\markopen\,GLM 5.3 & $0.126$\,{\scriptsize(24.7/25)} & $0.121$\,{\scriptsize(25/25)} & $0.136$\,{\scriptsize(24/25)} & $0.088$\,{\scriptsize(30.0/34)} & $0.063$\,{\scriptsize(33/34)} & $\mathbf{0.104}$\,{\scriptsize(26/34)} \\*
\markclosed\,Gemini 3.1 Pro & $0.127$\,{\scriptsize(24.3/25)} & $0.122$\,{\scriptsize(25/25)} & $0.139$\,{\scriptsize(23/25)} & $0.111$\,{\scriptsize(32.3/34)} & $0.083$\,{\scriptsize(33/34)} & $0.143$\,{\scriptsize(31/34)} \\*
\markopen\,DeepSeek V4.1 Flash & $0.129$\,{\scriptsize(25.0/25)} & $0.121$\,{\scriptsize(25/25)} & $0.144$\,{\scriptsize(25/25)} & $0.105$\,{\scriptsize(32.0/34)} & $0.073$\,{\scriptsize(34/34)} & $0.150$\,{\scriptsize(30/34)} \\*
\markclosed\,Claude Fable 5.1 & $\mathbf{0.120}$\,{\scriptsize(23.0/25)} & $\mathbf{0.117}$\,{\scriptsize(23/25)} & $\mathbf{0.125}$\,{\scriptsize(23/25)} & $\mathbf{0.081}$\,{\scriptsize(34.0/34)} & $\mathbf{0.057}$\,{\scriptsize(34/34)} & $0.108$\,{\scriptsize(34/34)} \\*
\markopen\,Kimi K3 & $0.134$\,{\scriptsize(25.0/25)} & $0.125$\,{\scriptsize(25/25)} & $0.144$\,{\scriptsize(25/25)} & $0.110$\,{\scriptsize(32.7/34)} & $0.082$\,{\scriptsize(33/34)} & $0.146$\,{\scriptsize(32/34)} \\*
\markopen\,K2-MoVA 36B-A4B & $0.137$\,{\scriptsize(24.7/25)} & $0.128$\,{\scriptsize(25/25)} & $0.152$\,{\scriptsize(24/25)} & $0.109$\,{\scriptsize(30.0/34)} & $0.077$\,{\scriptsize(32/34)} & $0.135$\,{\scriptsize(26/34)} \\*
\markclosed\,Claude Opus 4.8 & $0.146$\,{\scriptsize(24.7/25)} & $0.141$\,{\scriptsize(25/25)} & $0.154$\,{\scriptsize(24/25)} & $0.148$\,{\scriptsize(31.7/34)} & $0.127$\,{\scriptsize(34/34)} & $0.180$\,{\scriptsize(29/34)} \\*
\markopen\,Kimi K2.6 & $0.137$\,{\scriptsize(24.7/25)} & $0.125$\,{\scriptsize(25/25)} & $0.153$\,{\scriptsize(24/25)} & $0.123$\,{\scriptsize(31.3/34)} & $0.089$\,{\scriptsize(34/34)} & $0.156$\,{\scriptsize(29/34)} \\*
\markopen\,Qwen 3.6 35B & $0.148$\,{\scriptsize(22.3/25)} & $0.133$\,{\scriptsize(25/25)} & $0.141$\,{\scriptsize(18/25)} & $0.181$\,{\scriptsize(30.0/34)} & $0.119$\,{\scriptsize(32/34)} & $0.237$\,{\scriptsize(28/34)} \\*
\markclosed\,GPT-5.5 & $0.141$\,{\scriptsize(23.7/25)} & $0.134$\,{\scriptsize(25/25)} & $0.154$\,{\scriptsize(21/25)} & $0.170$\,{\scriptsize(27.0/34)} & $0.137$\,{\scriptsize(31/34)} & $0.183$\,{\scriptsize(21/34)} \\*
\markopen\,Gemma 4 26B & $0.164$\,{\scriptsize(24.7/25)} & $0.148$\,{\scriptsize(25/25)} & $0.174$\,{\scriptsize(24/25)} & $0.157$\,{\scriptsize(28.3/34)} & $0.123$\,{\scriptsize(30/34)} & $0.207$\,{\scriptsize(27/34)} \\
\addlinespace[3pt]
\multicolumn{7}{l}{\textit{Interval (NRelIS)}} \\*
\markclosed\,GPT-6 Astra & $1.526$\,{\scriptsize(14.0/14)} & $1.038$\,{\scriptsize(14/14)} & $2.475$\,{\scriptsize(14/14)} & $41.787$\,{\scriptsize(59.0/59)} & $1.525$\,{\scriptsize(59/59)} & $121.394$\,{\scriptsize(59/59)} \\*
\markopen\,GLM 5.3 & $\mathbf{1.206}$\,{\scriptsize(13.7/14)} & $1.038$\,{\scriptsize(14/14)} & $1.406$\,{\scriptsize(13/14)} & $\mathbf{1.552}$\,{\scriptsize(52.7/59)} & $\mathbf{0.671}$\,{\scriptsize(58/59)} & $\mathbf{2.440}$\,{\scriptsize(45/59)} \\*
\markclosed\,Gemini 3.1 Pro & $1.459$\,{\scriptsize(14.0/14)} & $\mathbf{0.958}$\,{\scriptsize(14/14)} & $2.250$\,{\scriptsize(14/14)} & $3.177$\,{\scriptsize(59.0/59)} & $1.945$\,{\scriptsize(59/59)} & $4.348$\,{\scriptsize(59/59)} \\*
\markopen\,DeepSeek V4.1 Flash & $1.290$\,{\scriptsize(14.0/14)} & $1.087$\,{\scriptsize(14/14)} & $1.409$\,{\scriptsize(14/14)} & $2.245$\,{\scriptsize(56.7/59)} & $1.511$\,{\scriptsize(59/59)} & $2.647$\,{\scriptsize(54/59)} \\*
\markclosed\,Claude Fable 5.1 & $1.407$\,{\scriptsize(12.0/14)} & $1.268$\,{\scriptsize(12/14)} & $1.647$\,{\scriptsize(12/14)} & $2.100$\,{\scriptsize(56.0/59)} & $1.465$\,{\scriptsize(59/59)} & $2.793$\,{\scriptsize(52/59)} \\*
\markopen\,Kimi K3 & $1.220$\,{\scriptsize(14.0/14)} & $1.027$\,{\scriptsize(14/14)} & $\mathbf{1.363}$\,{\scriptsize(14/14)} & $3.035$\,{\scriptsize(58.3/59)} & $1.900$\,{\scriptsize(59/59)} & $4.561$\,{\scriptsize(57/59)} \\*
\markopen\,K2-MoVA 36B-A4B & $1.349$\,{\scriptsize(13.7/14)} & $1.088$\,{\scriptsize(14/14)} & $1.959$\,{\scriptsize(13/14)} & $3.828$\,{\scriptsize(52.0/59)} & $2.923$\,{\scriptsize(57/59)} & $5.038$\,{\scriptsize(45/59)} \\*
\markclosed\,Claude Opus 4.8 & $1.955$\,{\scriptsize(14.0/14)} & $1.344$\,{\scriptsize(14/14)} & $2.503$\,{\scriptsize(14/14)} & $4.150$\,{\scriptsize(54.0/59)} & $3.143$\,{\scriptsize(58/59)} & $5.293$\,{\scriptsize(50/59)} \\*
\markopen\,Kimi K2.6 & $2.098$\,{\scriptsize(14.0/14)} & $0.963$\,{\scriptsize(14/14)} & $3.223$\,{\scriptsize(14/14)} & $3.493$\,{\scriptsize(57.7/59)} & $2.538$\,{\scriptsize(59/59)} & $4.287$\,{\scriptsize(56/59)} \\*
\markopen\,Qwen 3.6 35B & $2.097$\,{\scriptsize(12.7/14)} & $1.334$\,{\scriptsize(14/14)} & $2.484$\,{\scriptsize(10/14)} & $5.145$\,{\scriptsize(54.0/59)} & $3.585$\,{\scriptsize(57/59)} & $6.999$\,{\scriptsize(51/59)} \\*
\markclosed\,GPT-5.5 & $4.410$\,{\scriptsize(14.0/14)} & $3.098$\,{\scriptsize(14/14)} & $6.159$\,{\scriptsize(14/14)} & $5.887$\,{\scriptsize(50.7/59)} & $3.791$\,{\scriptsize(59/59)} & $7.335$\,{\scriptsize(38/59)} \\*
\markopen\,Gemma 4 26B & $6.495$\,{\scriptsize(14.0/14)} & $5.932$\,{\scriptsize(14/14)} & $7.205$\,{\scriptsize(14/14)} & $6.183$\,{\scriptsize(52.0/59)} & $4.854$\,{\scriptsize(56/59)} & $7.038$\,{\scriptsize(47/59)} \\
\addlinespace[3pt]
\multicolumn{7}{l}{\textit{Set/graph ($1-F_1$)}} \\*
\markclosed\,GPT-6 Astra & $\mathbf{0.000}$ & $\mathbf{0.000}$ & $\mathbf{0.000}$ & $\mathbf{0.000}$ & $\mathbf{0.000}$ & $\mathbf{0.000}$ \\*
\markopen\,GLM 5.3 & $0.006$ & $\mathbf{0.000}$ & $0.019$ & $0.067$ & $\mathbf{0.000}$ & $0.100$ \\*
\markclosed\,Gemini 3.1 Pro & $\mathbf{0.000}$ & $\mathbf{0.000}$ & $\mathbf{0.000}$ & $\mathbf{0.000}$ & $\mathbf{0.000}$ & $\mathbf{0.000}$ \\*
\markopen\,DeepSeek V4.1 Flash & $\mathbf{0.000}$ & $\mathbf{0.000}$ & $\mathbf{0.000}$ & $\mathbf{0.000}$ & $\mathbf{0.000}$ & $\mathbf{0.000}$ \\*
\markclosed\,Claude Fable 5.1 & $0.074$ & $0.074$ & $0.074$ & $\mathbf{0.000}$ & $\mathbf{0.000}$ & $\mathbf{0.000}$ \\*
\markopen\,Kimi K3 & $0.010$ & $\mathbf{0.000}$ & $0.031$ & $0.033$ & $\mathbf{0.000}$ & $0.100$ \\*
\markopen\,K2-MoVA 36B-A4B & $0.027$ & $\mathbf{0.000}$ & $0.080$ & $0.067$ & $\mathbf{0.000}$ & $0.200$ \\*
\markclosed\,Claude Opus 4.8 & $\mathbf{0.000}$ & $\mathbf{0.000}$ & $\mathbf{0.000}$ & $\mathbf{0.000}$ & $\mathbf{0.000}$ & $\mathbf{0.000}$ \\*
\markopen\,Kimi K2.6 & $\mathbf{0.000}$ & $\mathbf{0.000}$ & $\mathbf{0.000}$ & $\mathbf{0.000}$ & $\mathbf{0.000}$ & $\mathbf{0.000}$ \\*
\markopen\,Qwen 3.6 35B & $\mathbf{0.000}$ & $\mathbf{0.000}$ & $\mathbf{0.000}$ & $0.167$ & $0.100$ & $0.200$ \\*
\markclosed\,GPT-5.5 & $\mathbf{0.000}$ & $\mathbf{0.000}$ & $\mathbf{0.000}$ & $0.133$ & $\mathbf{0.000}$ & $0.300$ \\*
\markopen\,Gemma 4 26B & $0.062$ & $\mathbf{0.000}$ & $0.185$ & $0.267$ & $0.100$ & $0.600$ \\
\end{longtable}
\endgroup

    \paragraph{Continuous reliability.}
    Restarts barely move point estimates.
    On the realistic exam the twelve Mean values span $0.120$--$0.164$ and best-of-3 $0.117$--$0.148$, with Claude Fable~5.1 lowest on both exams; the agents refit the same data to the same number, consistent with the run-to-run SDs in App.~\ref{app:error_taxonomy}.
    What restarts do change is eligibility and intervals.
    Qwen~3.6 35B's missing submissions are transient --- $22.3$ of $25$ realistic point tasks valid per run on average, all $25$ in at least one of the three runs --- so the tool-output exits behind its low valid-answer rate (App.~\ref{app:harness_ops}) are recoverable by restarting; GLM~5.3 and GPT-5.5 show the same pattern on the hard exam.
    Intervals are the unstable pool: best-of-3 and worst-of-3 differ by up to $3\times$ (Kimi~K2.6 $0.96$ vs.\ $3.22$, GPT-6 Astra $1.04$ vs.\ $2.48$ on the realistic exam), and because these losses are uncapped a few blown intervals dominate the means --- Astra's hard-exam interval loss is $41.8$ for Mean and $121$ for worst-of-3 against a best-of-3 of $1.5$.
    That is the interval swing of App.~\ref{app:error_taxonomy} seen from the retry side, and the reason the leaderboard caps the interval term (Sec.~\ref{sec:scoring}).
    Set/graph losses are zero or near it for every model but Gemma~4 26B (worst-of-3 $0.19$ realistic, $0.60$ hard).
\begin{table}[ht!]
    \centering
    \caption{\textbf{Binary reliability by rung (\%).} Mean is the three-run mean success rate on the fixed ground-truth pool; any-of-3/all-of-3 are the shares passed in at least one/all three runs. $n$ is the ground-truth pool size; model columns follow realistic-exam CausalDSScore, with best per row in bold. Model markers: \markclosed\ closed, \markopen\ open-weight.}
    \label{tab:realistic_passk_passrate}
    \scriptsize
    \begingroup
    \setlength{\tabcolsep}{2.5pt}
    \begin{adjustbox}{max width=\textwidth}
    \begin{tabular}{@{}l l *{12}{c}@{}}
\toprule
Rung ($n$) & Metric & \markclosed\,GPT-6 & \markopen\,GLM & \markclosed\,Gemini & \markopen\,DeepSeek & \markclosed\,Fable & \markopen\,K3 & \markopen\,K2-MoVA & \markclosed\,Opus & \markopen\,K2.6 & \markopen\,Qwen & \markclosed\,GPT-5.5 & \markopen\,Gemma \\
\midrule
\multicolumn{14}{@{}l}{\textbf{\examMain}} \\
\multirow{3}{*}{R1 (3)} & Mean & $\mathbf{100.0}$ & $\mathbf{100.0}$ & $\mathbf{100.0}$ & $\mathbf{100.0}$ & $\mathbf{100.0}$ & $\mathbf{100.0}$ & $\mathbf{100.0}$ & $\mathbf{100.0}$ & $\mathbf{100.0}$ & $\mathbf{100.0}$ & $\mathbf{100.0}$ & $\mathbf{100.0}$ \\
 & any-of-3 & $\mathbf{100.0}$ & $\mathbf{100.0}$ & $\mathbf{100.0}$ & $\mathbf{100.0}$ & $\mathbf{100.0}$ & $\mathbf{100.0}$ & $\mathbf{100.0}$ & $\mathbf{100.0}$ & $\mathbf{100.0}$ & $\mathbf{100.0}$ & $\mathbf{100.0}$ & $\mathbf{100.0}$ \\
 & all-of-3 & $\mathbf{100.0}$ & $\mathbf{100.0}$ & $\mathbf{100.0}$ & $\mathbf{100.0}$ & $\mathbf{100.0}$ & $\mathbf{100.0}$ & $\mathbf{100.0}$ & $\mathbf{100.0}$ & $\mathbf{100.0}$ & $\mathbf{100.0}$ & $\mathbf{100.0}$ & $\mathbf{100.0}$ \\
\addlinespace[2pt]
\multirow{3}{*}{R2 (14)} & Mean & $\mathbf{100.0}$ & $45.2$ & $64.3$ & $47.6$ & $54.8$ & $35.7$ & $59.5$ & $57.1$ & $54.8$ & $50.0$ & $88.1$ & $38.1$ \\
 & any-of-3 & $\mathbf{100.0}$ & $57.1$ & $78.6$ & $71.4$ & $71.4$ & $64.3$ & $78.6$ & $71.4$ & $78.6$ & $71.4$ & $\mathbf{100.0}$ & $64.3$ \\
 & all-of-3 & $\mathbf{100.0}$ & $35.7$ & $50.0$ & $28.6$ & $42.9$ & $14.3$ & $42.9$ & $50.0$ & $28.6$ & $28.6$ & $71.4$ & $21.4$ \\
\addlinespace[2pt]
\multirow{3}{*}{R3 (17)} & Mean & $92.2$ & $92.2$ & $88.2$ & $90.2$ & $78.4$ & $86.3$ & $80.4$ & $\mathbf{98.0}$ & $92.2$ & $80.4$ & $76.5$ & $70.6$ \\
 & any-of-3 & $\mathbf{100.0}$ & $\mathbf{100.0}$ & $94.1$ & $94.1$ & $82.4$ & $94.1$ & $88.2$ & $\mathbf{100.0}$ & $\mathbf{100.0}$ & $88.2$ & $88.2$ & $88.2$ \\
 & all-of-3 & $88.2$ & $82.4$ & $76.5$ & $88.2$ & $70.6$ & $76.5$ & $64.7$ & $\mathbf{94.1}$ & $82.4$ & $70.6$ & $64.7$ & $47.1$ \\
\midrule
\multicolumn{14}{@{}l}{\textbf{\examHard}} \\
\multirow{3}{*}{R2 (45)} & Mean & $\mathbf{100.0}$ & $74.1$ & $60.0$ & $88.1$ & $84.4$ & $52.6$ & $74.1$ & $67.4$ & $79.3$ & $56.3$ & $90.4$ & $48.9$ \\
 & any-of-3 & $\mathbf{100.0}$ & $86.7$ & $66.7$ & $95.6$ & $88.9$ & $62.2$ & $86.7$ & $77.8$ & $86.7$ & $75.6$ & $\mathbf{100.0}$ & $75.6$ \\
 & all-of-3 & $\mathbf{100.0}$ & $64.4$ & $53.3$ & $77.8$ & $80.0$ & $44.4$ & $62.2$ & $57.8$ & $71.1$ & $33.3$ & $77.8$ & $24.4$ \\
\addlinespace[2pt]
\multirow{3}{*}{R3 (52)} & Mean & $82.1$ & $82.7$ & $70.5$ & $87.2$ & $\mathbf{87.8}$ & $76.3$ & $83.3$ & $72.4$ & $83.3$ & $70.5$ & $69.2$ & $57.7$ \\
 & any-of-3 & $90.4$ & $\mathbf{92.3}$ & $78.8$ & $90.4$ & $\mathbf{92.3}$ & $86.5$ & $90.4$ & $78.8$ & $90.4$ & $78.8$ & $82.7$ & $75.0$ \\
 & all-of-3 & $75.0$ & $71.2$ & $61.5$ & $\mathbf{84.6}$ & $82.7$ & $63.5$ & $75.0$ & $65.4$ & $75.0$ & $59.6$ & $51.9$ & $38.5$ \\
\bottomrule
    \end{tabular}
    \end{adjustbox}
    \endgroup
\end{table}
\begin{table}[ht!]
    \centering
    \caption{\textbf{Binary reliability by observation condition (\%).} Definitions and model order follow Tab.~\ref{tab:realistic_passk_passrate}. Best per row in bold. Model markers: \markclosed\ closed, \markopen\ open-weight.}
    \label{tab:passk_by_observation}
    \scriptsize
    \begingroup
    \setlength{\tabcolsep}{2.5pt}
    \begin{adjustbox}{max width=\textwidth}
    \begin{tabular}{@{}l l *{12}{c}@{}}
\toprule
Observation ($n$) & Metric & \markclosed\,GPT-6 & \markopen\,GLM & \markclosed\,Gemini & \markopen\,DeepSeek & \markclosed\,Fable & \markopen\,K3 & \markopen\,K2-MoVA & \markclosed\,Opus & \markopen\,K2.6 & \markopen\,Qwen & \markclosed\,GPT-5.5 & \markopen\,Gemma \\
\midrule
\multicolumn{14}{@{}l}{\textbf{\examMain}} \\
\multirow{3}{*}{\texttt{clean} (18)} & Mean & $\mathbf{92.6}$ & $74.1$ & $83.3$ & $75.9$ & $77.8$ & $64.8$ & $75.9$ & $88.9$ & $79.6$ & $70.4$ & $87.0$ & $51.9$ \\
 & any-of-3 & $\mathbf{100.0}$ & $88.9$ & $94.4$ & $88.9$ & $83.3$ & $83.3$ & $88.9$ & $94.4$ & $94.4$ & $77.8$ & $88.9$ & $72.2$ \\
 & all-of-3 & $\mathbf{88.9}$ & $61.1$ & $66.7$ & $66.7$ & $72.2$ & $50.0$ & $61.1$ & $83.3$ & $61.1$ & $61.1$ & $83.3$ & $33.3$ \\
\addlinespace[2pt]
\multirow{3}{*}{\texttt{proxy} (13)} & Mean & $\mathbf{100.0}$ & $76.9$ & $82.1$ & $69.2$ & $59.0$ & $69.2$ & $66.7$ & $79.5$ & $76.9$ & $79.5$ & $79.5$ & $76.9$ \\
 & any-of-3 & $\mathbf{100.0}$ & $76.9$ & $84.6$ & $76.9$ & $69.2$ & $76.9$ & $76.9$ & $84.6$ & $84.6$ & $92.3$ & $\mathbf{100.0}$ & $92.3$ \\
 & all-of-3 & $\mathbf{100.0}$ & $76.9$ & $76.9$ & $61.5$ & $46.2$ & $61.5$ & $53.8$ & $76.9$ & $69.2$ & $61.5$ & $61.5$ & $61.5$ \\
\addlinespace[2pt]
\multirow{3}{*}{\texttt{proxy\_hard} (3)} & Mean & $\mathbf{100.0}$ & $55.6$ & $44.4$ & $77.8$ & $77.8$ & $66.7$ & $88.9$ & $44.4$ & $66.7$ & $22.2$ & $77.8$ & $33.3$ \\
 & any-of-3 & $\mathbf{100.0}$ & $66.7$ & $66.7$ & $\mathbf{100.0}$ & $\mathbf{100.0}$ & $\mathbf{100.0}$ & $\mathbf{100.0}$ & $66.7$ & $\mathbf{100.0}$ & $66.7$ & $\mathbf{100.0}$ & $66.7$ \\
 & all-of-3 & $\mathbf{100.0}$ & $33.3$ & $33.3$ & $66.7$ & $66.7$ & $33.3$ & $66.7$ & $33.3$ & $33.3$ & $0.0$ & $33.3$ & $0.0$ \\
\midrule
\multicolumn{14}{@{}l}{\textbf{\examHard}} \\
\multirow{3}{*}{\texttt{clean} (23)} & Mean & $\mathbf{100.0}$ & $79.7$ & $82.6$ & $84.1$ & $82.6$ & $63.8$ & $69.6$ & $82.6$ & $76.8$ & $72.5$ & $88.4$ & $50.7$ \\
 & any-of-3 & $\mathbf{100.0}$ & $91.3$ & $91.3$ & $95.7$ & $87.0$ & $69.6$ & $78.3$ & $91.3$ & $82.6$ & $91.3$ & $\mathbf{100.0}$ & $78.3$ \\
 & all-of-3 & $\mathbf{100.0}$ & $69.6$ & $73.9$ & $69.6$ & $78.3$ & $56.5$ & $60.9$ & $73.9$ & $69.6$ & $47.8$ & $69.6$ & $30.4$ \\
\addlinespace[2pt]
\multirow{3}{*}{\texttt{proxy} (8)} & Mean & $\mathbf{100.0}$ & $75.0$ & $50.0$ & $95.8$ & $87.5$ & $62.5$ & $87.5$ & $79.2$ & $87.5$ & $79.2$ & $\mathbf{100.0}$ & $62.5$ \\
 & any-of-3 & $\mathbf{100.0}$ & $87.5$ & $50.0$ & $\mathbf{100.0}$ & $87.5$ & $75.0$ & $87.5$ & $87.5$ & $\mathbf{100.0}$ & $87.5$ & $\mathbf{100.0}$ & $75.0$ \\
 & all-of-3 & $\mathbf{100.0}$ & $62.5$ & $50.0$ & $87.5$ & $87.5$ & $50.0$ & $87.5$ & $62.5$ & $75.0$ & $75.0$ & $\mathbf{100.0}$ & $50.0$ \\
\addlinespace[2pt]
\multirow{3}{*}{\texttt{proxy\_hard} (23)} & Mean & $\mathbf{91.3}$ & $79.7$ & $55.1$ & $85.5$ & $88.4$ & $63.8$ & $79.7$ & $65.2$ & $75.4$ & $60.9$ & $75.4$ & $55.1$ \\
 & any-of-3 & $\mathbf{95.7}$ & $87.0$ & $69.6$ & $91.3$ & $91.3$ & $73.9$ & $91.3$ & $69.6$ & $87.0$ & $73.9$ & $87.0$ & $69.6$ \\
 & all-of-3 & $\mathbf{87.0}$ & $69.6$ & $39.1$ & $82.6$ & $82.6$ & $52.2$ & $69.6$ & $60.9$ & $65.2$ & $47.8$ & $56.5$ & $39.1$ \\
\addlinespace[2pt]
\multirow{3}{*}{\texttt{proxy\_severe} (43)} & Mean & $82.9$ & $78.3$ & $65.1$ & $\mathbf{89.1}$ & $86.8$ & $67.4$ & $82.2$ & $64.3$ & $86.0$ & $58.1$ & $72.1$ & $52.7$ \\
 & any-of-3 & $90.7$ & $90.7$ & $69.8$ & $90.7$ & $\mathbf{93.0}$ & $79.1$ & $\mathbf{93.0}$ & $74.4$ & $90.7$ & $69.8$ & $86.0$ & $76.7$ \\
 & all-of-3 & $76.7$ & $67.4$ & $60.5$ & $\mathbf{86.0}$ & $81.4$ & $55.8$ & $69.8$ & $55.8$ & $79.1$ & $41.9$ & $58.1$ & $25.6$ \\
\bottomrule
    \end{tabular}
    \end{adjustbox}
    \endgroup
\end{table}
\begin{table}[ht!]
    \centering
    \caption{\textbf{Abstention reliability by task family (\%).} Restricted to non-identifiable targets; definitions and model order follow Tab.~\ref{tab:realistic_passk_passrate}. Family abbreviations follow Tab.~\ref{tab:realistic_composition}; best per row in bold. Model markers: \markclosed\ closed, \markopen\ open-weight.}
    \label{tab:realistic_passk_abstention}
    \scriptsize
    \begingroup
    \setlength{\tabcolsep}{2.5pt}
    \begin{adjustbox}{max width=\textwidth}
    \begin{tabular}{@{}l l *{12}{c}@{}}
\toprule
Family ($n$) & Metric & \markclosed\,GPT-6 & \markopen\,GLM & \markclosed\,Gemini & \markopen\,DeepSeek & \markclosed\,Fable & \markopen\,K3 & \markopen\,K2-MoVA & \markclosed\,Opus & \markopen\,K2.6 & \markopen\,Qwen & \markclosed\,GPT-5.5 & \markopen\,Gemma \\
\midrule
\multicolumn{14}{@{}l}{\textbf{\examMain}} \\
\multirow{3}{*}{Ident (5)} & Mean & $\mathbf{100.0}$ & $26.7$ & $86.7$ & $33.3$ & $40.0$ & $20.0$ & $40.0$ & $86.7$ & $46.7$ & $60.0$ & $86.7$ & $13.3$ \\
 & any-of-3 & $\mathbf{100.0}$ & $40.0$ & $\mathbf{100.0}$ & $80.0$ & $60.0$ & $60.0$ & $60.0$ & $\mathbf{100.0}$ & $80.0$ & $\mathbf{100.0}$ & $\mathbf{100.0}$ & $40.0$ \\
 & all-of-3 & $\mathbf{100.0}$ & $20.0$ & $60.0$ & $0.0$ & $20.0$ & $0.0$ & $20.0$ & $80.0$ & $20.0$ & $20.0$ & $80.0$ & $0.0$ \\
\addlinespace[2pt]
\multirow{3}{*}{Effect (3)} & Mean & $\mathbf{100.0}$ & $55.6$ & $11.1$ & $77.8$ & $44.4$ & $44.4$ & $88.9$ & $11.1$ & $55.6$ & $11.1$ & $88.9$ & $33.3$ \\
 & any-of-3 & $\mathbf{100.0}$ & $66.7$ & $33.3$ & $\mathbf{100.0}$ & $66.7$ & $\mathbf{100.0}$ & $\mathbf{100.0}$ & $33.3$ & $\mathbf{100.0}$ & $33.3$ & $\mathbf{100.0}$ & $66.7$ \\
 & all-of-3 & $\mathbf{100.0}$ & $33.3$ & $0.0$ & $66.7$ & $33.3$ & $0.0$ & $66.7$ & $0.0$ & $0.0$ & $0.0$ & $66.7$ & $0.0$ \\
\addlinespace[2pt]
\multirow{3}{*}{Bias (3)} & Mean & $\mathbf{100.0}$ & $11.1$ & $44.4$ & $0.0$ & $44.4$ & $0.0$ & $22.2$ & $11.1$ & $22.2$ & $22.2$ & $77.8$ & $22.2$ \\
 & any-of-3 & $\mathbf{100.0}$ & $33.3$ & $66.7$ & $0.0$ & $66.7$ & $0.0$ & $66.7$ & $33.3$ & $33.3$ & $33.3$ & $\mathbf{100.0}$ & $66.7$ \\
 & all-of-3 & $\mathbf{100.0}$ & $0.0$ & $33.3$ & $0.0$ & $33.3$ & $0.0$ & $0.0$ & $0.0$ & $0.0$ & $0.0$ & $33.3$ & $0.0$ \\
\addlinespace[2pt]
\multirow{3}{*}{CF-ID (3)} & Mean & $77.8$ & $77.8$ & $\mathbf{88.9}$ & $66.7$ & $66.7$ & $66.7$ & $55.6$ & $\mathbf{88.9}$ & $77.8$ & $66.7$ & $66.7$ & $55.6$ \\
 & any-of-3 & $\mathbf{100.0}$ & $\mathbf{100.0}$ & $\mathbf{100.0}$ & $66.7$ & $66.7$ & $66.7$ & $66.7$ & $\mathbf{100.0}$ & $\mathbf{100.0}$ & $66.7$ & $66.7$ & $66.7$ \\
 & all-of-3 & $\mathbf{66.7}$ & $\mathbf{66.7}$ & $\mathbf{66.7}$ & $\mathbf{66.7}$ & $\mathbf{66.7}$ & $\mathbf{66.7}$ & $33.3$ & $\mathbf{66.7}$ & $\mathbf{66.7}$ & $\mathbf{66.7}$ & $\mathbf{66.7}$ & $33.3$ \\
\addlinespace[2pt]
\multirow{3}{*}{CFE (2)} & Mean & $\mathbf{100.0}$ & $83.3$ & $50.0$ & $\mathbf{100.0}$ & $83.3$ & $66.7$ & $83.3$ & $\mathbf{100.0}$ & $\mathbf{100.0}$ & $50.0$ & $50.0$ & $50.0$ \\
 & any-of-3 & $\mathbf{100.0}$ & $\mathbf{100.0}$ & $50.0$ & $\mathbf{100.0}$ & $\mathbf{100.0}$ & $\mathbf{100.0}$ & $\mathbf{100.0}$ & $\mathbf{100.0}$ & $\mathbf{100.0}$ & $\mathbf{100.0}$ & $\mathbf{100.0}$ & $\mathbf{100.0}$ \\
 & all-of-3 & $\mathbf{100.0}$ & $50.0$ & $50.0$ & $\mathbf{100.0}$ & $50.0$ & $50.0$ & $50.0$ & $\mathbf{100.0}$ & $\mathbf{100.0}$ & $0.0$ & $0.0$ & $0.0$ \\
\midrule
\multicolumn{14}{@{}l}{\textbf{\examHard}} \\
\multirow{3}{*}{Ident (8)} & Mean & $\mathbf{100.0}$ & $83.3$ & $79.2$ & $83.3$ & $75.0$ & $58.3$ & $75.0$ & $87.5$ & $75.0$ & $79.2$ & $95.8$ & $41.7$ \\
 & any-of-3 & $\mathbf{100.0}$ & $87.5$ & $87.5$ & $\mathbf{100.0}$ & $87.5$ & $75.0$ & $87.5$ & $\mathbf{100.0}$ & $87.5$ & $87.5$ & $\mathbf{100.0}$ & $62.5$ \\
 & all-of-3 & $\mathbf{100.0}$ & $75.0$ & $62.5$ & $75.0$ & $62.5$ & $37.5$ & $62.5$ & $75.0$ & $62.5$ & $62.5$ & $87.5$ & $12.5$ \\
\addlinespace[2pt]
\multirow{3}{*}{Effect (22)} & Mean & $\mathbf{100.0}$ & $68.2$ & $36.4$ & $92.4$ & $86.4$ & $40.9$ & $78.8$ & $51.5$ & $81.8$ & $39.4$ & $92.4$ & $47.0$ \\
 & any-of-3 & $\mathbf{100.0}$ & $81.8$ & $40.9$ & $95.5$ & $90.9$ & $54.5$ & $95.5$ & $63.6$ & $90.9$ & $59.1$ & $\mathbf{100.0}$ & $72.7$ \\
 & all-of-3 & $\mathbf{100.0}$ & $59.1$ & $31.8$ & $86.4$ & $81.8$ & $31.8$ & $63.6$ & $40.9$ & $72.7$ & $18.2$ & $86.4$ & $18.2$ \\
\addlinespace[2pt]
\multirow{3}{*}{Bias (6)} & Mean & $\mathbf{100.0}$ & $50.0$ & $61.1$ & $61.1$ & $66.7$ & $16.7$ & $22.2$ & $50.0$ & $44.4$ & $38.9$ & $72.2$ & $27.8$ \\
 & any-of-3 & $\mathbf{100.0}$ & $83.3$ & $83.3$ & $83.3$ & $66.7$ & $16.7$ & $33.3$ & $66.7$ & $50.0$ & $83.3$ & $\mathbf{100.0}$ & $83.3$ \\
 & all-of-3 & $\mathbf{100.0}$ & $33.3$ & $50.0$ & $16.7$ & $66.7$ & $16.7$ & $16.7$ & $33.3$ & $33.3$ & $0.0$ & $33.3$ & $0.0$ \\
\addlinespace[2pt]
\multirow{3}{*}{CFE (20)} & Mean & $\mathbf{100.0}$ & $88.3$ & $50.0$ & $\mathbf{100.0}$ & $93.3$ & $68.3$ & $90.0$ & $75.0$ & $91.7$ & $75.0$ & $83.3$ & $46.7$ \\
 & any-of-3 & $\mathbf{100.0}$ & $\mathbf{100.0}$ & $65.0$ & $\mathbf{100.0}$ & $95.0$ & $90.0$ & $\mathbf{100.0}$ & $85.0$ & $\mathbf{100.0}$ & $90.0$ & $\mathbf{100.0}$ & $75.0$ \\
 & all-of-3 & $\mathbf{100.0}$ & $75.0$ & $35.0$ & $\mathbf{100.0}$ & $90.0$ & $45.0$ & $80.0$ & $60.0$ & $85.0$ & $55.0$ & $60.0$ & $20.0$ \\
\bottomrule
    \end{tabular}
    \end{adjustbox}
    \endgroup
\end{table}

    \paragraph{Binary reliability by slice.}
    Rung~1 is saturated: its three realistic binary tasks are solved by every model on every run.
    Rung~2 is where restarts add coverage --- realistic Mean $\to$ any-of-3 takes DeepSeek from $47.6$ to $71.4\%$, Kimi~K2.6 from $54.8$ to $78.6\%$, Kimi~K3 from $35.7$ to $64.3\%$ --- while all-of-3 stays at $14$--$43\%$ for the open models against $100\%$ for GPT-6 Astra; on the hard exam ($n{=}45$) all-of-3 runs from $24.4\%$ (Gemma) to $100\%$ (Astra).
    Rung~3 has high coverage everywhere (any-of-3 $\ge 75\%$) over a wide reliability floor (all-of-3 $38.5$--$94.1\%$).
    Across observation conditions the any-of-3 minus all-of-3 gap is a model property more than a view property: on the hard exam's \texttt{proxy\_severe} slice ($n{=}43$) it is under $12$ points for DeepSeek, Fable, Gemini, and Kimi~K2.6 and $51$ points for Gemma, whose gap is about the same on \texttt{clean}.

    Abstention is the clearest pattern (Tab.~\ref{tab:realistic_passk_abstention}), and it is only measurable because non-identifiability is a property of the generated scene rather than a matter of opinion.
    Astra abstains correctly on every run of every non-identifiable task but one.
    The open models recover much of what they miss on a typical run --- realistic \texttt{identification} abstention has any-of-3 of $100\%$ for Qwen and $80\%$ for DeepSeek and Kimi~K2.6 --- yet their all-of-3 on the same pool is $0$--$20\%$: they recognize a non-identifiable scene on some runs only.
    The forbidden-controls \texttt{bias\_diagnostic} family is the hardest abstention call under all three summaries on both exams.
    The realistic per-family pools are small ($n{=}2$--$5$), so those rows are indicative; the hard-exam pools ($n{=}6$--$22$) carry the claim.

    \subsection{Cost and efficiency}
    \label{app:cost_efficiency}

\begin{figure}[ht!]
        \centering
        \includegraphics[width=.75\textwidth]{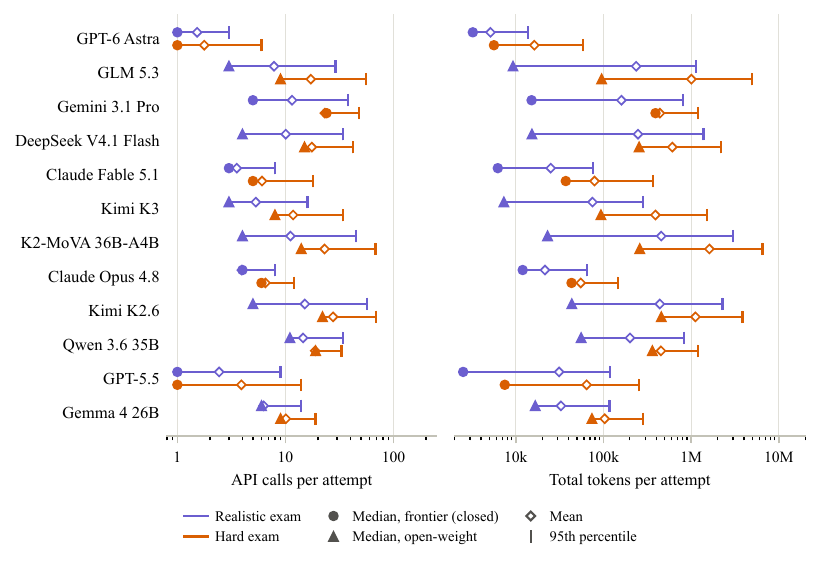}
        \caption{\textbf{Per-attempt cost by model.} Realistic exam (violet) and hard exam (orange): API calls (left) and total prompt$+$completion tokens (right), log scale. Each line runs from the median attempt (circle --- closed model, triangle --- open-weight model) to the 95th percentile (tick); the hollow diamond is the mean; the token mean is the Tokens column of the leaderboards. Rows follow the realistic-exam leaderboard (mean CausalDSScore), as in Tab.~\ref{tab:effort_quantiles}.}
        \label{fig:effort_quantiles}
    \end{figure}

    Fig.~\ref{fig:effort_quantiles} and Tab.~\ref{tab:effort_quantiles} unpack the mean costs in the leaderboards (Tabs.~\ref{tab:realistic_leaderboard} and~\ref{tab:hard_leaderboard}), showing per-attempt calls and total tokens through termination, regardless of outcome. Each model contributes $300$ realistic-exam and $600$ hard-exam attempts across three replications. Token accounting follows Sec.~\ref{sec:scoring}.
    Cumulative token usage can exceed a model's context-window size even when every individual request fits within it.
    Quantiles use the empirical inverse CDF.
    We wish to shed light on how the benchmark's fifth axis, tool use and coding, is reflected in these metrics: how each model spends, where the spending concentrates, and whether it buys anything.

    \begin{table}[ht!]
        \centering
        \caption{\textbf{Per-attempt cost quantiles.} Mean, median (med.), and p95 of API calls and total prompt$+$completion tokens (thousands) across three replications; rows follow the realistic-exam leaderboard (mean CausalDSScore), with all attempts pooled in the last row. N/I denote near one-shot/iterative ($<5$ mean calls on the realistic exam and $<10$ on the hard exam for N). Column minima in bold. Model markers: \markclosed\ closed, \markopen\ open-weight.}
        \label{tab:effort_quantiles}
        \footnotesize
        \setlength{\tabcolsep}{3pt}
        \begin{adjustbox}{max width=\textwidth}
        \begin{tabular}{l c rrr rrr c rrr rrr}
            \toprule
            & & \multicolumn{6}{c}{\examMain[Realistic exam] (100 tasks, 300 attempts per model)} & & \multicolumn{6}{c}{\examHard[Hard exam] (200 tasks, 600 attempts per model)} \\
            \cmidrule(lr){3-8}\cmidrule(lr){10-15}
            & & \multicolumn{3}{c}{Calls} & \multicolumn{3}{c}{Tokens (k)} & & \multicolumn{3}{c}{Calls} & \multicolumn{3}{c}{Tokens (k)} \\
            \cmidrule(lr){3-5}\cmidrule(lr){6-8}\cmidrule(lr){10-12}\cmidrule(lr){13-15}
            Model & Strategy & mean & med. & p95 & mean & med. & p95 & & mean & med. & p95 & mean & med. & p95 \\
            \midrule
            \markclosed\,GPT-6 Astra & N & \textbf{1.5} & \textbf{1} & \textbf{3} & \textbf{5.1} & 3.2 & \textbf{14} &  & \textbf{1.8} & \textbf{1} & \textbf{6} & \textbf{16} & \textbf{5.6} & \textbf{59} \\
            \markopen\,GLM 5.3 & I & 7.8 & 3 & 29 & 236 & 9.3 & 1,146 &  & 17.2 & 9 & 56 & 1,007 & 95 & 4,919 \\
            \markclosed\,Gemini 3.1 Pro & I & 11.5 & 5 & 38 & 160 & 15 & 813 &  & 23.2 & 24 & 48 & 440 & 392 & 1,208 \\
            \markopen\,DeepSeek V4.1 Flash & I & 10.1 & 4 & 34 & 247 & 15 & 1,382 &  & 17.5 & 15 & 42 & 610 & 255 & 2,180 \\
            \markclosed\,Claude Fable 5.1 & N & 3.5 & 3 & 8 & 25 & 6.2 & 76 &  & 6.1 & 5 & 18 & 79 & 37 & 369 \\
            \markopen\,Kimi K3 & I & 5.3 & 3 & 16 & 75 & 7.3 & 284 &  & 11.8 & 8 & 34 & 392 & 93 & 1,519 \\
            \markopen\,K2-MoVA 36B-A4B & I & 11.1 & 4 & 45 & 456 & 23 & 3,021 &  & 23.0 & 14 & 68 & 1,616 & 259 & 6,510 \\
            \markclosed\,Claude Opus 4.8 & N & 4.0 & 4 & 8 & 22 & 12 & 64 &  & 6.5 & 6 & 12 & 55 & 43 & 146 \\
            \markopen\,Kimi K2.6 & I & 15.1 & 5 & 57 & 439 & 44 & 2,272 &  & 27.6 & 22 & 69 & 1,119 & 457 & 3,862 \\
            \markopen\,Qwen 3.6 35B & I & 14.5 & 11 & 34 & 201 & 56 & 836 &  & 18.9 & 19 & 33 & 452 & 361 & 1,187 \\
            \markclosed\,GPT-5.5 & N & 2.4 & \textbf{1} & 9 & 31 & \textbf{2.5} & 118 &  & 3.9 & \textbf{1} & 14 & 64 & 7.5 & 252 \\
            \markopen\,Gemma 4 26B & I & 6.3 & 6 & 14 & 33 & 17 & 117 &  & 10.1 & 9 & 19 & 103 & 74 & 285 \\
            \midrule
            All models & --- & 7.8 & 4 & 31 & 161 & 12 & 798 &  & 14.0 & 8 & 45 & 496 & 71 & 2,304 \\
            \bottomrule
        \end{tabular}
        \end{adjustbox}
    \end{table}
    \paragraph{Interaction style.}
    The call number distributions sort the field into three styles. GPT-6 Astra and GPT-5.5 answer with a single call at the median on both exams: one script, one submission. Claude Fable 5.1 and Claude Opus 4.8 run short loops of bounded length, with 95th-percentile call counts of $8$ on the realistic exam and $12$--$18$ on the hard exam. The remaining eight models iterate open-endedly: their 95th percentiles run from $14$ to $57$ calls on the realistic exam and from $19$ to $69$ on the hard exam, and Gemini 3.1 Pro, Kimi K2.6, and Qwen 3.6 35B need $19$--$24$ calls for a typical hard-exam attempt. The mean-call thresholds of Tab.~\ref{tab:effort_quantiles} draw the same line, near one-shot (N) for the first two styles and iterative (I) for the third. Because the harness re-sends the growing context on every call, cumulative tokens follow calls: the median across models of the within-model Spearman correlation between an attempt's calls and its total tokens is $0.95$ on the realistic exam and $0.98$ on the hard exam, so the call count is the primary quantity and the tokens are a secondary, tail-dominated one. For the near one-shot models the mean token cost lies well above the median attempt, twelvefold for GPT-5.5 on the realistic exam, so their Tok./task entries on the leaderboards are set by a minority of long attempts (Fig.~\ref{fig:effort_quantiles}).

    \paragraph{Where the cost goes.}
    The hard exam raises typical cost more than extreme cost: pooled over models, the median attempt grows sixfold, from $12$k to $71$k tokens, and the 95th percentile threefold (Tab.~\ref{tab:effort_quantiles}). The styles absorb this differently. The iterative models lengthen their \textit{typical} attempt, Gemini 3.1 Pro's median moving from $5$ to $24$ calls and Kimi K2.6's from $5$ to $22$, whereas GPT-6 Astra and GPT-5.5 keep the one-call median and grow only in the tail, their 95th-percentile calls moving from $3$ to $6$ and from $9$ to $14$. Tab.~\ref{tab:effort_composition} locates the costly attempts, those above their model's own 95th percentile. On both exams they are interval and numerical-point targets under the harder observation variants. The absence of costly \texttt{clean} attempts on the hard exam coincides with that slice holding only identification and bias-diagnostic tasks, i.e., symbolic tasks. Cost therefore concentrates where the data-science and uncertainty-quantification axes meet, on numeric estimation under measurement error. 

    \paragraph{What spending buys.}
    Between models, tokens buy score in the Pareto sense of Fig.~\ref{fig:token_efficiency}: GLM 5.3 attains the best hard-exam CausalDSScore at $1{,}007$k tokens per attempt, over sixty times GPT-6 Astra's. Within a model they do not. The trajectory-level taxonomy shows that \textit{the failing attempts are the long ones}: among the leading models, data-science fails run the longest tool loops and reasoning fails the next longest (App.~\ref{app:error_tracks}), and the first-commitment census finds GPT-6 Astra and GPT-5.5 committing to a plan before any tool result more often than the other models, i.e., in a near one-shot fashion. The abstention decision shows the same asymmetry in cost. On the hard exam an over-claim, a committed answer on a non-identifiable target, costs a median $116$k tokens against $26$k for a correct abstention and is costly sixteen times as often ($6.5\%$ against $0.4\%$ of attempts; Tab.~\ref{tab:effort_composition}). 
    The hard exam's non-identifiable structures, \texttt{iv} and \texttt{double\_nc}, carry mostly effect and counterfactual-effect tasks (Tab.~\ref{tab:hard_composition}), so the over-claim is an estimation pipeline run to completion on a quantity the graph does not identify; on the realistic exam, where most non-identifiable items are symbolic (Tab.~\ref{tab:realistic_composition}), the two outcomes have nearly the same median cost, though over-claims are still costly five times as often. The routes into the over-claim are traced in App.~\ref{app:error_examples}, where GPT-5.5 reaches the abstention in its first and only step. Read together with the abstention axis, the iterative open models pay for a missed abstention twice, in the answer and in the tokens, whereas the near one-shot models miss cheaply, and their misses fall mostly on the uncertainty-quantification axis (Sec.~\ref{sec:results}; App.~\ref{app:error_taxonomy}).

    \paragraph{Comparability across benchmarks.}
    Call counts are not commensurate across harnesses: a single bash call here can run a complete analysis script, so the quantity that transfers is the interaction style, single-pass script submission versus multi-round iteration. The split is one instance of the model- and benchmark-specific tool-use patterns reported elsewhere: RoadmapBench and AutoLab report opposite exploration failures for GPT-5.4 \citep{xu_roadmapbench_2026, xu_autolab_2026}, GPT-5 uses tools heavily on MCP-Bench \citep{wang_mcp-bench_2025}, and more tool use improves long-horizon planning on DeepPlanning \citep{zhang_deepplanning_2026}. AutoLab also finds that the GPT family's one-pass style persists under a prompt that discourages it (App.~\ref{app:harness_ops}).
%
    \begin{table}[ht!]
        \centering
        \caption{\textbf{Task composition of costly attempts.} Pooled over models and replications. Share is the category's percentage of the exam's attempts, Median its median total tokens (thousands), and Costly the percentage of the category's attempts whose total tokens exceed their model's own within-exam 95th percentile. --- marks a category absent from the exam.}
        \label{tab:effort_composition}
        \footnotesize
        \setlength{\tabcolsep}{4pt}
        \begin{adjustbox}{max width=\textwidth}
        \begin{tabular}{@{}l rr rr rr@{}}
            \toprule
            & \multicolumn{2}{c}{Share (\%)} & \multicolumn{2}{c}{Median (k)} & \multicolumn{2}{c}{Costly (\%)} \\
            \cmidrule(lr){2-3}\cmidrule(lr){4-5}\cmidrule(lr){6-7}
            Category & \examMain[Real.] & \examHard & \examMain[Real.] & \examHard & \examMain[Real.] & \examHard \\
            \midrule
            Interval & 14.0 & 29.5 & 70 & 243 & 13.5 & 9.8 \\
            Numerical point & 25.0 & 17.0 & 47 & 256 & 9.0 & 7.7 \\
            Non-identifiable & 16.0 & 28.0 & 14 & 34 & 3.5 & 2.1 \\
            \quad abstained & 8.8 & 20.1 & 13 & 26 & 1.3 & 0.4 \\
            \quad over-claimed & 6.8 & 7.6 & 16 & 116 & 6.5 & 6.5 \\
            Graph/set & 27.0 & 5.0 & 3.7 & 8.7 & 0.0 & 0.0 \\
            Binary content & 18.0 & 20.5 & 8.8 & 42 & 1.7 & 1.0 \\
            \midrule
            \texttt{clean} & 48.0 & 16.5 & 9.5 & 8.3 & 1.7 & 0.0 \\
            \texttt{proxy} & 32.0 & 12.0 & 7.5 & 100 & 2.3 & 4.7 \\
            \texttt{proxy\_hard} & 20.0 & 27.5 & 93 & 100 & 17.2 & 5.8 \\
            \texttt{proxy\_severe} & --- & 44.0 & --- & 124 & --- & 6.5 \\
            \bottomrule
        \end{tabular}
        \end{adjustbox}
    \end{table}

    \subsection{Robustness to the CauseNet-seeded verbalization}
    \label{app:causenet_ablation}

    CauseNet seeding (Sec.~\ref{sec:verbalization}) injects real-world causal pairs into the story
    and variable names. One concern is whether this semantic layer leaks the answer, letting a model solve a  scene from a familiar causal template in the prose rather than from the story-implied graph and data --- the ``causal parrot'' failure mode. The more general concern is stability: scores should be a function of the formal problem, not of the particular story drawn for it. Because every story is audited to be faithful to
    its graph (cf. App.~\ref{app:audit_details}), a model whose answers move under a story swap is failing on its own terms --- exactly the behavior the benchmark is meant to expose --- but a persistent per-scene movement across models could indicate dependence on the story draw. We probe both with a \emph{matched verbalization-swap} ablation that holds the formal problem fixed and varies only the seeded verbal grounding.

    \paragraph{Construction.}
    From the main scene pool we form ten \emph{bundles}. Within a bundle the motif, the raw node order, and the
    per-node variable-type signature are identical, and one representative member's clean data and private ground
    truth are mapped back onto every member's variable names, so all members share the same numbers, the same
    scoring target, and the same private truth. The members thus differ only in their CauseNet-seeded story and variable names, giving four verbalizations of a single fixed formal problem. Each bundle contributes one task, for $40$ exam items spanning nine distinct motifs and both symbolic and data-backed families. We evaluate two open-weight agents, Kimi~K2.6 and Qwen~3.6 35B, with one attempt per item each.

    \paragraph{Diagnostic loss.}
    To place heterogeneous task outputs on one scale we use a capped ``diagnostic loss'' (higher is worse): $1-\text{score}$
    for boolean, $F_1$, and abstention tasks; normalized relative error for scalar estimates, normalized RMSE for prediction, and normalized interval score for intervals, each capped at $1$; and $1$ for a missing or unparseable answer or a runtime failure, including context overflow. Binary prediction uses $\sqrt{\mathrm{Brier}}/(1+\mathrm{sd}(Y_{\mathrm{test}}))$ (Eq.~\ref{eq:nrel_error}). The quantity of interest is the \emph{within-bundle range} of this loss: how much the outcome moves when only the verbalization changes while the formal problem is held fixed.
    \begin{table}[ht!]
        \centering
        \caption{\textbf{Matched verbalization-swap ablation.} Capped diagnostic loss of each model on the four verbalizations $s_1$--$s_4$ of a bundle's fixed formal problem (ordered by source scene ID) and the within-bundle range (\mbox{max$-$min}). Slice is the bundle task's \texttt{rung:family}, abbreviated as in Tab.~\ref{tab:realistic_composition}. \emph{abst} marks an abstention on an identifiable target, \emph{miss} a missing or unparseable answer or a runtime failure, and \emph{id} an identifiable call on a non-identifiable target; each scores loss $1$. The lower block counts bundles per model by range. }
        \label{tab:causenet_ablation}
        \footnotesize
        \begin{adjustbox}{max width=\textwidth}
        \begin{tabular}{@{}c l l c c c c c c c c c c c@{}}
            \toprule
            & & & \multicolumn{5}{c}{\markopen\,Kimi K2.6} & & \multicolumn{5}{c}{\markopen\,Qwen 3.6 35B} \\
            \cmidrule(lr){4-8}\cmidrule(lr){10-14}
            Bundle & Slice & Motif & $s_1$ & $s_2$ & $s_3$ & $s_4$ & Range & & $s_1$ & $s_2$ & $s_3$ & $s_4$ & Range \\
            \midrule
            1 & R2:Effect & arrowhead & $0.009$ & $0.009$ & $0.009$ & $0.009$ & $0.000$ &  & $0.009$ & $0.009$ & $0.006$ & $0.006$ & $0.003$ \\
            2 & R2:Sketch & chain & $0.000$ & $0.000$ & $0.000$ & $0.000$ & $0.000$ &  & $0.000$ & $0.000$ & $0.000$ & $0.000$ & $0.000$ \\
            3 & R1:Assoc & collider & $0.000$ & $0.000$ & $0.000$ & $0.000$ & $0.000$ &  & $0.000$ & $0.000$ & $0.000$ & $0.000$ & $0.000$ \\
            4 & R2:Ident & confounding & $0.000$ & $0.000$ & $0.000$ & $0.000$ & $0.000$ &  & $0.000$ & $0.000$ & $0.000$ & $0.000$ & $0.000$ \\
            5 & R1:Pred & diamond & $0.167$ & $0.168$ & $0.167$ & $0.173$ & $0.006$ &  & $0.171$ & $0.279$ & $0.172$ & $0.174$ & $0.107$ \\
            6 & R2:Bias & diamondcut & $0.000$ & $0.000$ & $0.000$ & $0.000$ & $0.000$ &  & $0.000$ & $0.000$ & $0.000$ & $0.000$ & $0.000$ \\
            7 & R3:CFE & frontdoor & $0.109$ & \emph{abst} & \emph{abst} & $0.109$ & $0.891$ &  & $0.386$ & \emph{abst} & \emph{abst} & \emph{miss} & $0.614$ \\
            8 & R3:CF-ID & double\_nc & $0.000$ & $0.000$ & $0.000$ & $0.000$ & $0.000$ &  & $0.000$ & \emph{id} & $0.000$ & $0.000$ & $1.000$ \\
            9 & R1:Collider & collider & $0.000$ & $0.000$ & $0.000$ & $0.000$ & $0.000$ &  & $0.000$ & $0.000$ & $0.000$ & $0.000$ & $0.000$ \\
            10 & R2:Effect & mediation & $0.002$ & $0.002$ & $0.002$ & $0.002$ & $0.000$ &  & $0.001$ & $0.002$ & $0.002$ & $0.002$ & $0.002$ \\
            \midrule
            \multicolumn{7}{l}{Bundles with nontrivial spread ($>0.001$)} & 2/10 &  &  &  &  &  & 5/10 \\
            \multicolumn{7}{l}{Bundles with large spread ($>0.05$)} & 1/10 &  &  &  &  &  & 3/10 \\
            \bottomrule
        \end{tabular}
        \end{adjustbox}
    \end{table}

    \paragraph{The stronger agent is more stable.}
    Table~\ref{tab:causenet_ablation} shows that most bundles are flat: every discrete-output bundle outside
    Rung~3 is invariant for both models, and the ordinary Rung-1/2 numeric bundles move only marginally, determined by the particular numeric approach the model picks on the scene.
    Kimi (the stronger agent, cf. Tab.~\ref{tab:realistic_leaderboard}) is largely invariant, with nontrivial spread on $2/10$ bundles and a single large
    ($>0.05$) swing.
    Qwen is materially more sensitive: $5/10$ bundles move and three swing widely.
    The instability concentrates in the two Rung-3 bundles (counterfactual effect and identification), where the ranges reflect
    discrete flips (commit vs. abstain, identifiable vs. not) rather than numeric dispersion, and, for Qwen, one binary prediction bundle where the ranking of test cases is preserved across stories (AUC $0.73$--$0.75$, with the     discrepant story in fact the highest) while the emitted probabilities become poorly calibrated (Brier $0.128$
    against $0.048$--$0.050$ on the other three verbalizations): the story changed how confident the predictions are, not which cases the model scores higher.

    \paragraph{Shared false abstentions.}
    The movement that does exist is attributable to the story layer, not to sampling variation. On the stable
    numeric bundles, Kimi's estimates are identical across all four verbalizations to ten or more significant
    digits and Qwen's vary by at most $3\times10^{-3}$ in relative error: given the same numbers, the agents run
    essentially deterministic pipelines. In bundle~7, by contrast, both models abstain on the same two stories
    and attempt the estimate on the other two (Kimi with numerically identical estimates; Qwen with one estimate
    and one context-overflow failure) --- two agents agreeing on \emph{which} stories trigger an unwarranted abstention is not single-run stochasticity. This is precisely the persistent cross-model movement flagged above as the possible signature of story-draw dependence, and the only place the ablation finds it. We nevertheless argue that the coordinated behavior is mostly model-driven: it is a residue of the ``causal parrot'' failure mode (also cf. the next paragraph), and a competent agent would score identically on all four members. That this happens less often overall for the model otherwise benchmarked as the stronger one speaks in favor of this interpretation and against the ``null hypothesis'' that the verbalization seeding makes the benchmark unstable. What remains on the benchmark side is limited: for susceptible models, the story a scene is told with can make it harder or easier.
    This is thus a latent factor in the \textit{difficulty} of the benchmark; its \textit{stability} is unaffected.

    \paragraph{A concrete semantic-lure failure.}
    The clearest case is bundle~8 (\texttt{double\_nc} motif, symbolic counterfactual identification of the ETT), where the estimand is \emph{non-identifiable}. Kimi answers non-identifiable
    throughout. Qwen also answers non-identifiable on the three stories seeded with the
    hypothermia--unconsciousness, illness--disability, and events--arrest pairs, but flips to \emph{identifiable}
    on the remaining one ($s_2$ in Tab.~\ref{tab:causenet_ablation}, source scene \texttt{scene\_000409}), whose prose describes
    rainfall-gauge and satellite-precipitation measurements of a latent precipitation confounder:
    \begin{quote}
        \small ``Because researchers cannot directly observe the true annual precipitation totals across the
        entire heterogeneous mountainscape, they rely on two complementary assessments: the Rainfall Gauge
        Measurement collected at fixed field stations, and the Satellite Precipitation Estimate derived from
        orbital remote sensing algorithms. [\dots] The true precipitation levels driving these fire dynamics
        remain latent variables inferred through the gauge and satellite observations rather than measured
        continuously across all terrain.''
    \end{quote}
    From those measurement descendants Qwen constructs a proxy/backdoor identification argument. The
    lure is a sophisticated one: the two-proxy prose mirrors the double-negative-control setting of proximal
    causal inference \citep{miao_identifying_2018, tchetgen_introduction_2024}, and the conditions the model
    asserts have exactly that shape. Even in that framework, however, point identification rests on completeness
    assumptions that are not encoded in the graph and that the model neither states nor checks.
    Under the benchmark's graph-level  identifiability policy, observed descendants of a latent confounder do not license identification, so this is a template imported from the story without the assumptions it needs, rather than a property of the graph --- exactly the behavior the fully synthetic  construction is meant to expose.

    \paragraph{Pair commonness does not explain the variation.}
    We recovered the CauseNet support count and salience percentile of each scene's seeded pair and regressed the
    diagnostic loss on commonness with bundle fixed effects. The within-bundle slope is negative for both models
    (Kimi $-0.05$, bootstrap $95\%$ CI $[-0.18, 0.00]$; Qwen $-0.45$, CI $[-1.08, 0.00]$), and the within-bundle
    Spearman correlation is $+0.05$ (Kimi) and $-0.27$ (Qwen). The sign is consistent with more common pairs being
    marginally easier, but the intervals include zero, and two bundles carry no within-bundle commonness variation. We therefore report commonness as an inconclusive moderator: the seeded verbal grounding affects the weaker agent, but that effect is not sufficiently explained by CauseNet pair commonness in this diagnostic.

    \paragraph{Takeaway.}
    The two results read as the two halves of the motivating concern. The stronger agent's near-invariance --- exact agreement of its answers across seeded pairs with different real-world scenarios --- indicates that the score measures the formal problem rather than the seeded verbalization: the benchmark is robust to CauseNet seeding for an agent that works from the released structure and data. The weaker agent's movement is likely causal-parrot susceptibility itself: the example above indicates that the model inferred semantics that the graph and data did not support.

    \subsection{Observation-layer hardness under matched scenes}
    \label{app:observation_ablation}

    The observation layer (Sec.~\ref{sec:observation_layer}) is designed to change how hard an identified
    functional is to estimate from the released files while preserving the conceptual SCM, the target estimand,
    and the identifiability label. Here, we present a matched ablation in which the same conceptual scene/task is evaluated under all three released observation views.

    \paragraph{Construction.}
    From the main scene pool we select ten matched scene/task pairs whose source scene releases all three  observation variants, spanning the data-backed continuous families. Only identifiable targets enter, so abstention is never the correct response.
    The $10 \times 3 = 30$ instances form a single randomized exam. The observation model label used is hidden from the agent, as are previously submitted answers. We
    evaluate the two open-weight agents, Kimi~K2.6 and Qwen~3.6 35B (the pair also used in
    App.~\ref{app:causenet_ablation}) with one run each. The paired deltas are descriptive contrasts on these ten selected pairs, not estimates of a population-level effect of the observation layer.

    \paragraph{Diagnostic loss.}
    For evaluation purposes, we define a (capped) \textit{diagnostic loss} (higher is worse): the normalized relative error
    (point/scalar outputs) or normalized interval score (intervals), capped at $1$; a missing, unparseable, or runtime-failed answer scores $1$; and --- because every selected target is identifiable --- an abstention also scores $1$, as it is always incorrect.
    This convention is specific to this ablation and differs from the main leaderboard grading, where
    abstention is scored as its own metric and the continuous pools contain only valid numeric answers. The paired contrasts are:
    $\Delta_{\mathrm{proxy}} = L_{\mathrm{proxy}} - L_{\mathrm{clean}}$ and
    $\Delta_{\mathrm{hard}} = L_{\mathrm{proxy\_hard}} - L_{\mathrm{clean}}$ per pair, with paired-bootstrap $95\%$ intervals over the ten pairs.
    The survivor-only view --- deltas over pairs with \textit{valid} answers in both arms --- is reported alongside.
    \begin{table}[ht!]
        \centering
        \caption{\textbf{Matched observation-layer ablation.} 
        (a) Mean paired change in capped diagnostic loss relative to \texttt{clean}, with paired-bootstrap $95\%$ CI; Worse counts pairs with $\Delta>0$. Survivor-only means condition on valid numeric answers in both arms, with their contributing count $n$.
        (b) Valid numeric answers, false abstentions on the six abstention-eligible tasks, context-overflow exits, mean API calls per task, and median capped loss among valid numeric answers.}
        \label{tab:observation_ablation_summary}
        \small
        \setlength{\tabcolsep}{4pt}
        \begin{subtable}[t]{\textwidth}
            \centering
            \caption{Paired contrasts against \texttt{clean}.}
            \label{tab:observation_ablation}
            \begin{adjustbox}{max width=\linewidth}
            \begin{tabular}{@{}llcccc@{}}
                \toprule
                Model & View & Mean $\Delta$ & $95\%$ CI & Worse & Survivor-only ($n$) \\
                \midrule
                \multirow{2}{*}{\markopen\,Kimi K2.6} & \texttt{proxy} & $+0.003$ & $[-0.275, +0.255]$ & 7/10 & $+0.003$ (10) \\
                 & \texttt{proxy\_hard} & $+0.200$ & $[+0.054, +0.380]$ & 8/10 & $+0.147$ (8) \\
                \midrule
                \multirow{2}{*}{\markopen\,Qwen 3.6 35B} & \texttt{proxy} & $+0.304$ & $[+0.014, +0.602]$ & 6/10 & $+0.019$ (3) \\
                 & \texttt{proxy\_hard} & $+0.332$ & $[+0.070, +0.629]$ & 6/10 & $+0.111$ (3) \\
                \bottomrule
            \end{tabular}
            \end{adjustbox}
        \end{subtable}
        \par\medskip
        \begin{subtable}[t]{\textwidth}
            \centering
            \caption{Per-view behavior.}
            \label{tab:observation_ablation_validity}
            \begin{adjustbox}{max width=\linewidth}
            \begin{tabular}{@{}llrrrrr@{}}
                \toprule
                Model & View & Valid/10 & Abst./6 & Ctx./10 & Calls/task & Med.\ valid loss \\
                \midrule
                \multirow{3}{*}{\markopen\,Kimi K2.6} & \texttt{clean} & 10 & 0 & 0 & 16.5 & 0.005 \\
                 & \texttt{proxy} & 10 & 0 & 0 & 26.7 & 0.039 \\
                 & \texttt{proxy\_hard} & 8 & 2 & 0 & 26.0 & 0.149 \\
                \midrule
                \multirow{3}{*}{\markopen\,Qwen 3.6 35B} & \texttt{clean} & 8 & 2 & 0 & 13.3 & 0.003 \\
                 & \texttt{proxy} & 4 & 1 & 5 & 19.6 & 0.028 \\
                 & \texttt{proxy\_hard} & 3 & 2 & 5 & 19.7 & 0.016 \\
                \bottomrule
            \end{tabular}
            \end{adjustbox}
        \end{subtable}
    \end{table}
    \begin{figure}[ht!]
        \centering
        \begin{subfigure}[t]{.443\textwidth}
            \centering
            \includegraphics[width=\linewidth]{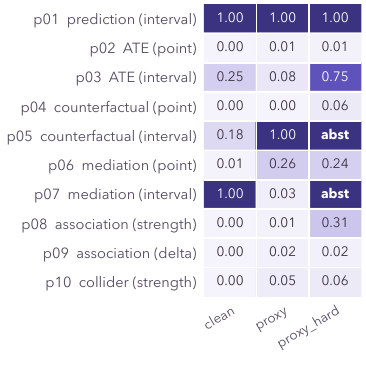}
            \caption{Kimi K2.6.}
            \label{fig:observation_outcome_matrix}
        \end{subfigure}\hfill
        \begin{subfigure}[t]{.204\textwidth}
            \centering
            \includegraphics[width=\linewidth]{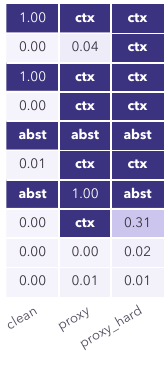}
            \caption{Qwen 3.6 35B.}
            \label{fig:observation_outcome_matrix_qwen}
        \end{subfigure}\hfill
        \begin{subfigure}[t]{.348\textwidth}
            \centering
            \includegraphics[width=\linewidth]{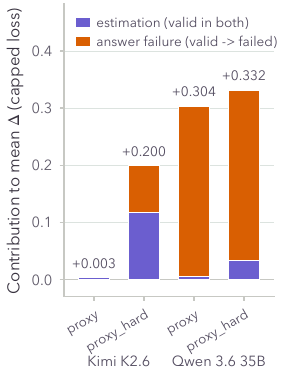}
            \caption{Contributions to the mean paired contrast.}
            \label{fig:observation_delta_decomposition}
        \end{subfigure}
        \caption{\textbf{Matched observation-layer ablation.}
        (a,~b) Capped diagnostic loss for all $60$ instances. Shading and printed values give the loss; \emph{abst} marks false abstention and \emph{ctx} context-window failure.
        (c) Exact decomposition of the mean paired loss change relative to \texttt{clean}.}
        \label{fig:observation_ablation}
    \end{figure}
    \paragraph{The matched contrast confirms the penalty under harder observation.}
    Table~\ref{tab:observation_ablation} shows both models scoring worse under \texttt{proxy\_hard} (per-instance
    results in Fig.~\ref{fig:observation_ablation}), with
    paired-bootstrap intervals excluding zero (Kimi $+0.20$, $8/10$ pairs worse; Qwen $+0.33$). The middle tier
    separates the models: Qwen already degrades at \texttt{proxy} ($+0.304$, CI $[+0.014, +0.602]$), while Kimi's
    \texttt{proxy} contrast is flat on average ($+0.003$ against a $+0.01$ median, the mean pulled down by one    interval pair whose \texttt{clean} answer was already at the loss cap while its \texttt{proxy} answer happened  to score well). Median deltas are moderate throughout ($\leq 0.06$).

    \paragraph{Hardness acts through two separate effects.}
    The first effect is statistical: among instances that produce a valid answer, estimation degrades. The median survivor loss rises $0.005 \to 0.149$ (Kimi) and $0.003 \to 0.016$ (Qwen) from \texttt{clean} to \texttt{proxy\_hard} (Tab.~\ref{tab:observation_ablation_validity}), and Kimi's survivor-only  $\Delta_{\mathrm{hard}}$ is $+0.15$ with $7/8$ pairs worse.

    The second effect is agentic: the harder views increasingly prevent a valid answer at all. Over \texttt{clean}/\texttt{proxy}/\texttt{proxy\_hard}, valid answers fall $10/10/8$ (Kimi) and $8/4/3$ (Qwen).
    The underlying driver is workload: both models work harder as the view hardens (Kimi $16.5 \to {\sim}26$
    calls per task, Qwen $13.3 \to {\sim}20$), and the extra work ends differently for the two agents. For Qwen,
    whose deployment serves a 32k-token window, the longer exploration is precisely what exhausts the context:
    ten of its fifteen failures are context-window overflows (five in each proxy tier), and the remaining five are false abstentions. Kimi never exhausts its context; instead it starts
    abstaining under \texttt{proxy\_hard} --- plausibly the added measurement uncertainty tipping it into judging an identifiable target unanswerable (its abstention cases are examined below).
    Failing to finish is a second, separate signal that the observation layer stresses the whole agentic
    workflow as well as estimator accuracy. The decomposition in Fig.~\ref{fig:observation_delta_decomposition}
    separates the two contributions to the capped mean --- for Qwen the failing pairs account for $+0.30$ of the
    $+0.33$, while Kimi's splits into $+0.12$ from surviving estimates and $+0.08$ from abstention onset ---
    though these shares reflect the failure-scores-$1$ convention as much as the data and should not be read as a ranking of the two effects.

    \paragraph{Abstention as a hardness-triggered exit.}
    All seven abstentions in the ablation fall on the same two pairs (twelve instances: two models under three variants each) --- an ETT interval and an NIE interval, both on frontdoor-motif scenes whose targets the \textit{ID*/IDC*}-based algorithm certifies as identifiable.
    The submitted explanations argue from textbook sequential-ignorability and backdoor conditions: the models     locate the latent confounder named in the story and declare the estimand non-identifiable, missing the frontdoor-type identification.
    Kimi's abstentions are hardness-triggered exits rather than up-front identification calls:
    it commits to estimates on both pairs under \texttt{clean} and \texttt{proxy}, and its \texttt{proxy\_hard} trajectories
    show it fitting proxy-reconstruction models and implementing the estimators before submitting a null answer whose graph-level rationale contradicts its own attempts, mirroring the broader trend discussed in App.~\ref{app:error_tracks}.

    \paragraph{Interval calibration and caveats.}
    Kimi's per-row prediction intervals stay near their nominal $90\%$ coverage across all three variants
    ($88.2\%$--$89.95\%$). Qwen achieves $89.6\%$ on \texttt{clean} but produces no valid prediction intervals
    under either proxy view. Among submitted causal-effect intervals with nominal $95\%$ coverage, neither
    model covers any target on \texttt{clean}; Kimi nevertheless covers two of three targets under \texttt{proxy}. 
    Causal-interval undercoverage is therefore already present without the observation layer (cf. App.~\ref{app:error_intervals}). The ablation is deliberately small, supporting mostly a descriptive observation.

    \subsection{Limitations}
    \label{app:limitations}

    \paragraph{Synthetic realism.}
    Every scene is generated. Story fidelity to the hidden graph is audited by LLMs and confirmed post hoc by near-perfect structure recovery (App.~\ref{app:audit_details}), and exam composition is grounded in real-study corpora (App.~\ref{app:composition_grounding}), but no human study has rated the stories or data for domain plausibility, and the mechanism registries (App.~\ref{app:scm_profiles}) span only the functional families we chose to implement. A model that performs well here may still fail on a real study, whose data, documentation, and confounding are not drawn from a known SCM; the benchmark measures causal data-science competence on problems with a certified answer, which is the price of having one.

    \paragraph{Identifiability policy.}
    Identifiability is decided at the graph level by \texttt{dowhy} and the \texttt{y0} \textit{ID*/IDC*} algorithms on the conceptual DAG. Partial identification, sensitivity analysis, and proximal or negative-control strategies are not credited: an agent that returns bounds or a proxy-based estimate on a non-identifiable target is graded as a missed abstention. This is a deliberate policy (App.~\ref{app:scoring_rules}, App.~\ref{app:obsmodel_details}), and it makes the abstention axis a test of whether the agent adopts the benchmark's identification standard rather than of every defensible analysis.

    \paragraph{Exam size and slices.}
    Each exam is one draw of $100$ or $200$ tasks. The finite-exam intervals of App.~\ref{app:leaderboard_ci} are wide on the realistic exam --- ranks two through seven are not separable --- and many per-slice cells in Apps.~\ref{app:exam_results} and~\ref{app:pass@k} hold fewer than five tasks. The observation-variant slices are marginal views of exams whose other axes co-vary, so the per-variant comparisons in Tab.~\ref{tab:realistic_obs_variant} and the additive adjustment of App.~\ref{app:difficulty_knob} are descriptive; the matched-scene evidence (App.~\ref{app:observation_ablation}) rests on ten pairs, two open-weight models, and one run each.

    \paragraph{Scoring choices.}
    CausalDSScore weights its three pools by task count and caps each interval term at $c{=}10$; both are design decisions. App.~\ref{app:leaderboard_ci} shows that the leaderboard extremes are stable under the cap and the pool weights but that the middle reshuffles if the interval term is summarized by a median instead of the capped mean, and Composite Rank orders the hard-exam top four differently from CausalDSScore (Sec.~\ref{sec:results}). The permissive band for sign tasks and the null-equivalent labels accepted as abstention (App.~\ref{app:scoring_rules}) are likewise conventions.

    \paragraph{Agents and harness.}
    All models run in one scaffold: a bash sandbox with a fixed task prompt, one reasoning setting per model, and call and cost limits (App.~\ref{app:harness_ops}). Tool-call and token counts are therefore properties of this harness and are not comparable to other benchmarks' (App.~\ref{app:cost_efficiency}). Provider-side refusals are graded as failures, one model's tool-output handling cost it valid submissions, and closed models were reached through a routing provider at a fixed point in time; the twelve models are a snapshot, and the three replications and the three-model, $N{=}10$ mini-exam bound how much we can say about reliability.

    \subsection{Technical details}
    \label{app:harness_ops}

    \paragraph{Containerization and package pinning.}
    Evaluation runs use a Singularity image built from Docker whose Python 3.11 stack is pinned to exact versions to avoid mistakes
    due to API changes: \texttt{pandas==3.0.2}, \texttt{numpy==2.4.4}, \texttt{scipy==1.17.1},
    \texttt{scikit-learn==1.8.0}, \texttt{statsmodels==0.14.6}, \texttt{pyarrow==24.0.0}, \texttt{matplotlib==3.10.9},
    \texttt{seaborn==0.13.2}, \texttt{networkx==3.6.1}, \texttt{xgboost==3.2.0}.

    The benchmark task prompts list the versions so the agent does not waste budget probing the environment for the right API.

    \paragraph{Execution limits.}
    Per-task limits are 100 model queries for the realistic exam and targeted runs, and 150 for the hard exam,
    with tracked cost caps of \$10--15. Each single shell command has a 120-second timeout.
    An attempt that reaches a limit is graded on whatever answer files it wrote and stays in the cost summaries of App.~\ref{app:cost_efficiency}; twelve did so, one on the realistic exam (K2-MoVA 36B-A4B, call limit) and eleven on the hard exam (K2-MoVA 36B-A4B six times at the call limit, Claude Fable 5.1 four times at the cost cap, Gemma 4 26B once at the call limit).

    \paragraph{Workspace layout and answer validation.}
    Each task starts a fresh model conversation; the container is reused within a replication, with scenes mounted read-only.
    Each command runs as a separate shell process and state persists through files.
    Singularity is launched with \texttt{--no-home} and \texttt{HOME=/workspace} so that submitted answer files land in
    a path the grader will actually read. After each task the runner re-checks that the expected answer file exists
    under \texttt{/workspace/answers} and is minimally parseable as JSON or CSV; failures are recorded as
    \texttt{MissingAnswerAfterSubmit} rather than counting silently as submitted. When present, top-level scratch under
    \texttt{/workspace} is archived after each task to \path{<output_dir>/workspace_scratch/<scene_id>_<task_id>.tar.gz}, then removed before the next task starts.
    The given scenes (\texttt{scenes/}), the model's answer directory (\texttt{answers/}), and \texttt{INSTRUCTIONS.md}
    persist as the benchmark state.

    \paragraph{Provider routing and inference settings.}
    Five models are routed through OpenRouter with \texttt{provider.only} pinned: Claude Opus 4.8 and Claude Fable 5.1
    to \texttt{anthropic}, Gemini 3.1 Pro to \texttt{google-ai-studio}, and GPT-5.5 and GPT-6 Astra to \texttt{openai}.
    All five use \texttt{allow\_fallbacks=false} and request high reasoning effort; Opus additionally uses adaptive thinking in accordance with the benchmark's policy that runs models on high reasoning (or equivalent) and uses the defaults for the rest (high reasoning is the default for most models anyway).
    Seven models use self-hosted OpenAI-compatible endpoints: Qwen 3.6 35B, Kimi K2.6, Gemma 4 26B, GLM 5.3,
    DeepSeek V4.1 Flash, Kimi K3, and K2-MoVA 36B-A4B. GLM, DeepSeek, and Kimi K3 receive
    \texttt{reasoning\_effort=high} through the extra request body. K2-MoVA receives high effort through its chat-template
    settings, with temperature $1.0$ and top-$p=0.95$. Qwen and Kimi K2.6 use thinking, while Gemma uses its non-thinking
    configuration, as that is its default. Kimi K3 was served with a 256k-token context window.
    Token usage and tool-call counts are reported for all models to make inference-compute differences visible.

    \paragraph{Retries and failure accounting.}
    Model-query retries use exponential waits bounded between 4 and 60 seconds. Terminal endpoint failures, including
    service outages, are excluded from grading.
    Model-side failures, including refusals, malformed responses,
    and context overflow, remain graded.

    \paragraph{Tool-output handling.}
    One behavior specific to Qwen 3.6 35B partly explains its incomplete continuous submissions
    (Tab.~\ref{tab:realistic_leaderboard}): it manages tool output poorly, frequently emitting very long outputs
    --- for instance, printing the entire provided dataset into the conversation instead of using \texttt{head}/\texttt{tail}
    or computing aggregates over it. The harness does not progressively trim accumulated context; when a \emph{single}
    command's output exceeds a length threshold it instead shows only that command's head and tail, together with an
    explicit warning to use \texttt{head}/\texttt{tail}/\texttt{grep} or redirect to a file. Qwen repeatedly ignored this
    and lost the numeric values it needed on the hardest continuous estimands. We re-ran the exam with it at several output-budget
    thresholds ($1$k, $3.2$k, and the default $10$k characters used for the other models) and the behavior persisted,
    so we count these lost submissions as a genuine tool-use failure rather than a harness artifact.
    On the near one-shot submission style, \citet{xu_autolab_2026} find that even after explicitly discouraging first-pass submissions via the prompt
    when using the GPT-family with \texttt{mini-swe-agent}, the behavior nonetheless persists, indicating that it is not purely
    harness-induced.

    \clearpage

    \subsection{Prompt templates}
    \label{app:prompt_templates}

    \subsubsection{Generation pipeline prompt templates}
    \label{app:pipeline_prompts}

    The generation pipeline (Sec.~\ref{sec:verbalization}) uses separate LLM sessions for variable mapping, feasibility
    checking, consistency auditing, story generation, and story verification. Below we reproduce the core system and
    user prompt templates as used at the current benchmark version. The variable-mapping templates
    live in \texttt{causalds/\allowbreak{}schemas.py}; feasibility-check, consistency-audit, story-generation, and
    story-verification templates live in \texttt{causalds/\allowbreak{}pre\_auditor.py},
    \texttt{causalds/\allowbreak{}mapping\_audit.py}, \texttt{causalds/\allowbreak{}verbalization\_story.py}, and
    \texttt{causalds/\allowbreak{}verbalization\_verify.py}, respectively. Runtime template variables (e.g.,
    \texttt{\{format\_description\}}, \texttt{\{serialized\_graph\}}) are filled with the graph serialization and
    scene-specific details at generation time.

    \paragraph{Variable mapping system prompt.}
    Instructs the mapper LLM to choose a coherent domain and assign human-friendly variable names to abstract node
    identifiers.
    The prompt enforces that latent/unobserved nodes are correctly marked based on the graph's \texttt{observed\_nodes}
    list. Mapping runs in unstructured (prompt) mode: \texttt{\{output\_format\_block\}} inlines the requested
    output format --- JSON for the dataset presented --- as a skeleton (\texttt{proposed\_domain} plus one
    \texttt{variable\_mapping} entry of \texttt{id} / \texttt{story\_name} / \texttt{observed} / \texttt{type} /
    \texttt{unit} per node) followed by per-field instructions. The runtime template variables
    \texttt{\{format\_description\}} and \texttt{\{independence\_note\}} expand to a graph-format reminder and
    (optionally) a list of conditional independencies.

        {\footnotesize
    \begin{promptblock}
        You are an expert in causal inference, domain modeling, and
        statistical dependencies.
        You will be given a set of abstract variables (possibly as a directed
        acyclic graph). Your goal is to map these variables to a realistic,
        coherent, and scientifically plausible domain (e.g., epidemiology,
        economics, physics) that is consistent with the provided dependencies.

        Format Specification:
            {format_description}
            {independence_note}

        Guidelines:
        1) Analyze the fixed nodes (if any) to infer the domain context.
        2) Rename the target nodes to specific, measurable, and realistic
        variables that fit this domain. Avoid generic names like "Factor A"
        or "Variable X".
        3) Ensure the 'unit' field contains realistic measurement units
        (e.g., "mmHg", "years", "kg", "counts") and 'type' is appropriate
        (e.g., "continuous", "binary"). This is not as crucial for the
        UNOBSERVED variables.
        4) The chosen variables must make sense together in a single scenario.
        5) CRITICAL for "observed" field: Check the "observed_nodes" list in the
        graph. Set observed=true ONLY for nodes in that list. Any node NOT in
        observed_nodes is LATENT/unobserved - you MUST set observed=false
        for these.

            {output_format_block}
    \end{promptblock}
    }

    \paragraph{Variable mapping user prompt template.}
    Optional placeholders are filled by the grafting and anchor logic. The auxiliary stage substitutes
    \texttt{\{additional\_requirements\_block\}} with a directive to keep the shared anchor immutable, reuse its
    existing story name, and avoid meanings that would imply unintended direct links to existing variables; the
    final merged-graph audit substitutes a softer directive that lets minor wording refinements pass while
    preserving anchor identity, scope, type, and unit. The \texttt{\{output\_format\_name\}} expands to either
    \texttt{JSON} or \texttt{XML}.

        {\footnotesize
    \begin{promptblock}
        You will be given a graph of variables represented as {format_label}.
        Your task is to choose a coherent domain and assign human-friendly
        names to node ids.{tool_note}
        The graph representation follows below:

            {serialized_graph}

            {independence_section}
        Context:
        - Fixed nodes (already named): {fixed_nodes}
        -> {fixed_nodes_instruction}
        - Nodes needing names: {needs_names}
        -> Rename these to fit the context of the fixed nodes.
            {fixed_name_assignments_block}{forbidden_story_names_block}
            {domain_hint_block}{additional_requirements_block}
            {existing_graph_mapping_block}{anchor_context_block}

        Return ONLY the {output_format_name} object{extra_instruction}.
    \end{promptblock}
    }

    \paragraph{Variable mapping feedback prompt.}
    Used in the audit/repair loop (Alg.~\ref{alg:mapstage}) when the previous mapping fails consistency
    checks. \texttt{\{output\_format\_name\}} again expands to \texttt{JSON} or \texttt{XML}.

        {\footnotesize
    \begin{promptblock}
        CRITICAL: Your previous mapping FAILED a causal-consistency audit.
        You MUST revise variable meanings so that the following violations
        no longer apply:
            {violation_block}

        Especially: for every NON-EDGE pair, ensure there is NO plausible
        direct causal link in either direction.
        Return ONLY the {output_format_name} mapping object in the same
        schema as before.
    \end{promptblock}
    }

    \paragraph{Pre-audit system prompt.}
    The pre-auditor (Alg.~\ref{alg:preaudit}) checks whether CauseNet-fixed node concepts can plausibly
    satisfy the graph's structural constraints before the mapper runs.
    The prompt defines a mediator-aware working definition of ``direct causal link'' and a HIGH/MEDIUM/LOW
    classification for non-edge inevitability that explicitly considers whether a mediator could absorb the
    effect.
    The block below is condensed: we omit the in-prompt salvageable / hopeless examples, the surrounding
    context notes, and the output-format instructions, and the inline \texttt{feasible=true/false}
    annotations summarize the prompt's decision-summary lines.
        {\footnotesize
    \begin{promptblock}
        You are a causal structure feasibility checker.

        You will be given:
        1. A causal graph structure with FIXED NODE CONCEPTS (already named
        from a knowledge base)
        2. The structural constraints imposed by the graph:
        - DIRECT EDGES: Causal relationships that MUST be plausible
        - NON-EDGES: Pairs where NO direct causal link must exist

        YOUR TASK: Decide whether there is at least one MAINSTREAM,
        non-contrived interpretation under which the FIXED NODE concepts
        satisfy the structural constraints. Ask: "Can a skilled mapper
        typically find a coherent interpretation that makes the constraints
        work?" This is a PRE-FILTER; a separate, thorough audit runs AFTER
        mapping. Catch only OBVIOUS failures; default to letting borderline
        cases through.

        WORKING DEFINITION of DIRECT CAUSAL LINK:
        'There exists an intervention on U that changes V while holding fixed
        the other variables in the graph (especially the other nodes that
        could mediate the effect).'

        NON-EDGE CHECK:
        For each NON-EDGE pair of fixed nodes, classify the inevitability of
        a direct causal link in mainstream usage:
        - HIGH: A direct effect would likely persist even after introducing
        a mediator (residual direct pathway remains plausible).
        -> feasible=false
        - MEDIUM: A direct effect is plausible but a mainstream
        operationalization (e.g., administrative proxy, eligibility rule,
        time-indexed exposure) can reasonably remove it.
        -> feasible=true, confidence="low"
        - LOW: A direct effect is not a typical interpretation.
        -> feasible=true

        EDGE PLAUSIBILITY CHECK (apply very generously):
        For each direct edge, accept ANY mechanism (even indirect, weak, or
        context-dependent). Only flag as HOPELESS if the edge is physically
        impossible or logically contradictory.

        DECISION RULE: Default to FEASIBLE with confidence="low" when
        uncertain; let the full audit catch actual problems.
    \end{promptblock}
    }

    \paragraph{Mapping audit system prompt.}
    The mapping auditor enforces causal consistency of the proposed variable mapping against the graph
    structure.
    Non-edge violations are treated as hard constraints; direct-edge plausibility is judged leniently
    (only clearly implausible edges are flagged as violations); CI and type consistency are soft
    constraints.
    For grafted graphs, the prompt is rendered with a stage-specific instruction block (auxiliary-stage anchor
    immutability or final merged-graph leniency) that the same template substitutes in via
    \texttt{\{stage\_specific\_instruction\_block\}}.
    The block below is lightly condensed: we omit the conditional interpretation-note inserts, the
    output-format instructions, and the JSON/XML schemas.
        {\footnotesize
    \begin{promptblock}
        You are a strict causal consistency auditor.

            {stage_specific_instruction_block}You will be given:
        1) DIRECT edges (u -> v) that must be plausible as direct causal effects.
        2) NON-EDGE pairs (u, v) where NO plausible direct causal link may exist.
        3) Optionally, conditional independence (CI) relations from the graph.
        4) A proposed variable mapping from node ids to real-world meanings.
        5) Optionally, expected node type hints from generation.

        Your job:
        1. For each NON-EDGE pair (U, V), decide if ANY DIRECT CAUSAL LINK
        exists between the mapped variables. Working definition:
        'There exists an intervention on U that changes V while holding
        fixed the other variables in the graph (especially the other
        nodes that could mediate the effect).'
        You MUST include an entry in "non_edge_attestations" for EVERY
        non-edge pair (no_direct_link true/false plus a brief
        justification); if a direct link DOES exist, also record a
        VIOLATION.

        2. For each DIRECT edge (U -> V), decide if a plausible direct causal
        link exists. Be GENEROUS: any reasonable mechanism (weak, partial,
        or context-dependent) is sufficient. Only flag a VIOLATION if
        clearly implausible.

        3. For each CI statement, judge if plausible given variable meanings.
        (Soft constraint.) When flagging a CI violation, name the variables,
        explain why the chosen meanings make the independence implausible
        (e.g., uncovered alternative pathways), and HINT at a reframing
        that would restore it.

        4. Type consistency: be lenient (treat close families as compatible,
        skip when missing); only flag clear contradictions. (Soft.)

        INTERPRETATION NOTES:
        We mostly care about the NON-EDGE violations. Use the definition of
        the DIRECT CAUSAL LINK above very strictly. For example, a verbal
        aggregation of a detailed effect X -> M -> Y is NOT a separate
        direct effect X -> Y! For that, we require a SEPARATE direct
        pathway!

        FEEDBACK QUALITY REQUIREMENT:
        For every violation, the "explanation" must be actionable. End each
        with a short 'HINT:' suggesting how it might be resolved. Do not
        suggest changing the graph structure or fixed nodes; if the violation
        is between two fixed nodes, say so explicitly so the caller can reject
        the sample.
    \end{promptblock}
    }

    \paragraph{Story generation system prompt.}
    Instructs the story generation LLM to produce a short narrative from the renamed graph and proposed domain.
    The runtime template variable \texttt{\{format\_system\}} expands to a reminder of the graph serialization
    format used in the user prompt.

        {\footnotesize
    \begin{promptblock}
        You are an expert in narrating a cohesive story based on a causal graph
        with a domain and domain-relevant variables.
        The graph will be provided in the following structured format:
            {format_system}

        Your task is to write a SHORT, CONCRETE STORY (2-4 paragraphs) that:
        - is situated in the provided PROPOSED DOMAIN
        - uses ONLY the provided STORY NAMES from the VARIABLE MAPPING
        (NEVER use raw node IDs like V0, V1, etc.)
        - mentions EVERY mapped variable explicitly in the story,
        including latent/background factors when present
        - makes EVERY direct causal relationship in the graph explicit in
        the story itself
        - does NOT introduce unsupported extra direct causal relationships
        - avoids graph jargon (no "edges", "nodes", "colliders", "DAG", etc.)
        - is scientifically plausible and engaging
        - may be slightly denser than usual if needed to cover the full
        graph exactly
    \end{promptblock}
    }

    \paragraph{Story generation user prompt template.}

    {\footnotesize
        \begin{promptblock}
        You will be given a PROPOSED domain, a VARIABLE MAPPING (ids -> story
        names) in JSON, and a CAUSAL GRAPH in the format prescribed by the
        system prompt.

        Your task is to write a short, concrete STORY (2-4 paragraphs) using
        ONLY the provided story names from the VARIABLE MAPPING (no raw ids).
        The story needs to reflect the causal relationships in the graph and
        mention every mapped variable explicitly within the narrative, BUT
        WITHOUT using any graph jargon revealing its structure.
        Every direct edge in the graph must be stated or clearly implied in
        the STORY itself. Do not rely on the appendix/justifications to cover
        missing edges. Do not introduce unsupported extra direct causes.
        Retain the units and variable types (e.g., categorical, continuous)
        given by the mapping. Additionally return CAUSAL JUSTIFICATIONS
        explaining how the story reflects the causal structure.

        Return ONLY a JSON object: {"story": "...",
        "causal_justifications": "..."}

        Proposed domain:        {proposed_domain}
        Variable mapping (JSON):
            {variable_mapping_json}
        Graph structure:        {graph_structure}
            {extra_instruction}
    \end{promptblock}
    }

    \paragraph{Story audit system prompt.}
    The story auditor verifies that the generated narrative faithfully represents the causal graph. Hard
    failures populate \texttt{violations}; lower-priority issues populate \texttt{warnings}; the auditor must
    additionally fill a \texttt{node\_attestations} entry for every required variable and an
    \texttt{edge\_attestations} entry for every direct edge. The block below is condensed: we omit the
    input list, the interpretation notes, the per-item sub-bullets, and the JSON/XML schemas.
        {\footnotesize
    \begin{promptblock}
        You are a strict story-to-DAG auditor.

        HARD REQUIREMENTS:
        1. Every required variable must be explicitly mentioned in the STORY
        using its provided story name. Latent/unobserved variables may be
        narrated as hidden/background factors, but they still must be
        mentioned. Emit one entry in "node_attestations" for EVERY
        required variable.
        2. Every DIRECT edge (U -> V) must be clearly supported by the STORY
        itself. A path-level claim does NOT automatically cover every edge
        on the path. Emit one entry in "edge_attestations" for EVERY
        direct edge (supported / contradicted, with a justification).
        3. If the STORY contradicts the stated direction of an edge, mark
        that as a hard violation.

        SOFT CHECKS:
        4. Warn if the STORY introduces extra direct causal claims between
        NON-EDGE pairs.
        5. Warn if the STORY drifts from the proposed domain.
        6. Warn if the STORY uses graph jargon.
        7. Warn if the STORY is implausible, globally incoherent, or mixes
        incompatible domains in a way that makes the scenario hard to
        believe as a real setting.

        DECISION RULE:
        - Set "pass" to true only if all hard requirements are satisfied.
        - Put ONLY hard failures in "violations"; put soft issues in
        "warnings".
        - Each issue must be actionable, ending in a "HINT:" clause that
        suggests how to rewrite the story to remove it.
    \end{promptblock}
    }

    \subsubsection{Benchmark task prompt templates}
    \label{app:task_prompts}
    Every benchmark task is delivered to the agent as a fixed \emph{system message} followed by a
    \emph{user message} assembled by \texttt{build\_single\_task\_prompt}. The user message wraps a
    family-specific body in shared scaffolding that is identical across tasks: a data-location header, the
    scenario \texttt{\#\#\#~Background} (the story), a \texttt{\#\#\#~Data} section, a shared
    \texttt{\#\#\#~Measurement Note} on data-backed tasks, the family-specific \texttt{\#\#\#~Question} and
    \texttt{\#\#\#~Output Format}, and a footer giving the answer-file path and submit instruction. To avoid
    repetition, we reproduce the shared blocks once below and then list only the per-family
    \texttt{\#\#\#~Question}/\texttt{\#\#\#~Output Format} bodies. Template variables (\texttt{\{story\}},
    \texttt{\{treatment\_name\}}, \texttt{\{outcome\_name\}}, etc.) are filled from the scene at generation
    time; the formal JSON response schema is not sent to the model (the agent writes an answer file via
    \texttt{bash}), so each body shows only the inline JSON example. The full \texttt{(family, output variant)}
    catalog is enumerated in App.~\ref{app:scoring_rules}.

    \paragraph{Shared system message.}
    Identical for every task; lists the available Python stack and the one-bash-call interaction contract.

        {\footnotesize
    \begin{promptblock}
        You are an expert data scientist solving causal reasoning tasks.
        You have Python 3.11 with: pandas 3.0.2, numpy 2.4.4, scipy 1.17.1,
        scikit-learn 1.8.0, statsmodels 0.14.6, pyarrow 24.0.0,
        matplotlib 3.10.9, seaborn 0.13.2, networkx 3.6.1, xgboost 3.2.0.
        Do not assume unlisted packages are installed.

        <rules>
        1. EVERY response must include EXACTLY ONE bash tool call. No exceptions.
        2. Before each tool call, briefly state what you are about to check and why.
        3. For multi-line Python, use a heredoc:
        python3 << 'PYEOF'
        import pandas as pd
        df = pd.read_parquet("scenes/<scene_id>/data.parquet")
        print(df.describe())
        PYEOF
        4. To write files, use cat with heredoc:
        cat > /workspace/answers/my_answer.json << 'EOF'
        {"key": "value"}
        EOF
        5. The grader reads only /workspace/answers. Do not write answers under /home.
        6. When done, verify your answer file exists at the requested path, then submit: echo DONE
        </rules>
    \end{promptblock}
    }

    \paragraph{Shared user-message skeleton.}
    The data-location bullets list only the files released for the task: the
    \texttt{data.parquet}/\texttt{calibration.parquet}/\texttt{schema.json} lines appear only for data-backed
    tasks, and a \texttt{test\_features.parquet} line is added for prediction. The story-only Rung-2/3
    identification and graph-recovery (causal-sketch) tasks omit the \texttt{\#\#\#~Data} section and the Measurement Note.

        {\footnotesize
    \begin{promptblock}
        # Task

        **Data location:** `scenes/{scene_id}/`
        - Story/context: `scenes/{scene_id}/story.md`
        - Observational data: `scenes/{scene_id}/data.parquet`               (data-backed only)
        - Calibration subset (if present): `scenes/{scene_id}/calibration.parquet`   (data-backed only)
        - Column metadata: `scenes/{scene_id}/schema.json`                   (data-backed only)
        - Test features (no outcome column): `scenes/{scene_id}/test_features.parquet`   (prediction only)

        ### Background
            {story}

        ### Data
        You are provided with observational data in `data.parquet`.        (data-backed only)

            {### Measurement Note - shown below; data-backed tasks only}

            {family-specific ### Question and ### Output Format - see per-family blocks}

        **Answer file:** `/workspace/answers/{answer_file}`

        Write a JSON file matching the schema described above.   (CSV with columns `prediction`[, `lower`, `upper`] for prediction)

        The grader reads only files under `/workspace/answers/`; do not write answers under `/home`.
        When done, verify your answer file exists at the exact path above, then submit:
        ```bash
        echo DONE
        ```
    \end{promptblock}
    }

    \paragraph{Shared Measurement Note (data-backed tasks).}
    Inserted before the \texttt{\#\#\#~Question} on every task that ships a \texttt{.parquet} file. The first
    paragraph (released data scale) is always present; the remaining paragraphs appear only when the released
    observation variant contains noisy measurement columns.

        {\footnotesize
    \begin{promptblock}
        ### Measurement Note
        The story names conceptual variables in the causal graph. Numeric values in the released `.parquet` files, including `data.parquet`, and any question-specified values are on the benchmark's released data scale for those conceptual variables. This scale may differ from the units, ranges, or examples implied by the story; values may be centered, standardized, or otherwise transformed. For answering data-backed questions, use the released data scale.

        In this released dataset, some conceptual variables are not directly measured in `data.parquet`. Each such conceptual variable is replaced by a bundle of noisy measurement columns. Each bundle has exactly one conceptual parent and should not be treated as a set of separate causal variables. Examples in this scene: Covariate1: Covariate1_meas_a, Covariate1_meas_b; Mediator1: Mediator1_meas_a.

        A smaller `calibration.parquet` contains the measurement columns together with gold-standard measurements of the corresponding conceptual variables for a subset of rows. Use the shared name stem, `schema.json`, and the calibration rows to determine which measurement columns belong to each conceptual variable.

        For determining causal identifiability (when necessary), reason only over the conceptual variables described in the story. Same goes for any causal quantity. The observation layer affects solely the statistical estimation difficulty of the population quantity from provided data.

        When estimating a quantity from data, if a named conceptual variable is measured by a bundle, use that bundle and the calibration data to estimate or reconstruct the conceptual variable before estimating the requested statistical or causal quantity.
    \end{promptblock}
    }

    \medskip\noindent\textbf{Rung 1 (Associational).} The bodies below give the \texttt{\#\#\#~Question} and
    \texttt{\#\#\#~Output Format} that replace the \texttt{\{family-specific\ldots\}} placeholder above.\par

    \paragraph{R1: Prediction (\texttt{point\_predictor}; \texttt{prediction\_interval} adds a central 90\% interval).}
    Prediction is the one Kaggle-style family: it writes a CSV of held-out predictions rather than a JSON
    answer, so its body replaces \texttt{\#\#\#~Question}/\texttt{\#\#\#~Output Format} with an
    \texttt{\#\#\#~Objective}. The \texttt{prediction\_interval} variant appends the
    \texttt{\#\#\#~Uncertainty Quantification} block.

        {\footnotesize
    \begin{promptblock}
        ### Data
        You are provided with training data in `data.parquet` containing the following columns:
        "Treatment", "Outcome", "Covariate1", "Covariate2"

        ### Objective
        Build a predictive model for the conceptual outcome **Outcome** using the variables available in the released data.
        If the conceptual outcome is measured by a bundle in the released data, use the calibration rows to learn how those measurements relate to the conceptual outcome.

        ### Uncertainty Quantification (Prediction Intervals)        [prediction_interval variant only]
        Also produce a **central 90% prediction interval** for each prediction.

        ### Notes
        - Evaluation uses held-out test data.
    \end{promptblock}
    }

    \paragraph{R1: Association --- sign.}
    The \texttt{effect\_size\_point} variant instead asks for the numeric Pearson strength
    (\texttt{\{"value": 0.45, "method": "pearson"\}}); it is explicitly framed as ``not a causal-effect
    estimate.''

        {\footnotesize
    \begin{promptblock}
        ### Question
        Estimate the sign of the observational association between the conceptual variables **Treatment** and **Outcome**.

        Report `"+"` if higher values of **Treatment** are associated with higher values of **Outcome**, `"-"` if higher values of **Treatment** are associated with lower values of **Outcome**, and `"unknown"` only if the empirical association is too unclear to determine.

        ### Output Format
        Provide your answer as a JSON object:
        ```json
            {
            "sign": "+",  // or "-" or "unknown"
            "stat": {"method": "correlation", "value": 0.45}  // optional, encouraged
        }
        ```
    \end{promptblock}
    }

    \paragraph{R1: Conditional association --- sign before/after conditioning.}
    Same \texttt{association} family; the \texttt{delta\_point}/\texttt{delta\_sign\_only} variants ask for the
    magnitude/sign of the \emph{change} in association after conditioning, and \texttt{argmax\_change} asks
    which of several offered conditioning variables changes it most.

        {\footnotesize
    \begin{promptblock}
        ### Question
        Consider the association between **Treatment** and **Outcome**.

        1. What is the sign of the marginal (unconditional) association?
        2. What is the sign of the association after conditioning on the conceptual variable **Covariate1**?

        Conditioning here means statistical conditioning, not intervention.

        ### Output Format
        Provide your answer as a JSON object:
        ```json
            {
            "sign_before": "+",  // marginal association sign: "+", "-", or "unknown"
            "sign_after": "+",   // association after conditioning: "+", "-", or "unknown"
            "conditioning_var": "Covariate1",
            "explanation": "Brief explanation of why the association changes or stays the same"
        }
        ```
    \end{promptblock}
    }

    \paragraph{R1: Collider phenomenon (explaining away).}
    The \texttt{induced\_\allowbreak association\_\allowbreak sign\_\allowbreak only}/\allowbreak\texttt{induced\_\allowbreak association\_\allowbreak strength\_\allowbreak point} variants
    instead ask for the sign / numeric strength of the association induced by conditioning on the collider.

        {\footnotesize
    \begin{promptblock}
        ### Question
        Consider the conceptual variables **Treatment**, **Outcome**, and **Collider1**.

        After conditioning on the conceptual variable **Collider1**, is there a nonzero observational association between **Treatment** and **Outcome**?

        ### Output Format
        Provide your answer as a JSON object:
        ```json
            {
            "association_present": true,  // whether a nonzero association is present after conditioning on Collider1
            "explanation": "..."
        }
        ```
    \end{promptblock}
    }

    \needspace{8\baselineskip}
    \medskip\noindent\textbf{Rung 2 (Interventional).}\par

    \paragraph{R2: Causal sketch --- directed edges (\texttt{skeleton\_edges} drops direction).}
    A story-only task (no \texttt{\#\#\#~Data}/Measurement Note); the prompt states the exact variable count.

        {\footnotesize
    \begin{promptblock}
        ### Question
        Based only on the scenario description, identify the conceptual causal variables it describes and the direct causal relationships among them.

        The conceptual causal graph contains exactly 5 causal variables.

        Important:
        - Include only direct causal effects, not indirect effects through intermediate variables.
        - A causes B means that an intervention on A would directly change B while holding fixed other variables that could mediate the effect.
        - Do not include associations caused only by common causes, selection, or conditioning.
        - Include every causal variable the story describes, including any that is described only as a hidden, background, or unmeasured factor.

        ### Output Format
        Provide your answer as a JSON object (use the exact variable names as they appear in the story):
        ```json
            {"edges": [{"from": "Variable1", "to": "Variable2"}, {"from": "Variable1", "to": "Variable3"}]}
        ```
        [skeleton_edges variant: {"skeleton_edges": [{"a": "Variable1", "b": "Variable2"}]} instead]
    \end{promptblock}
    }

    \paragraph{R2: Identification --- method label.}
    A story-only task: the graph drives a population-ATE identifiability judgment and no estimator is run.
    Sibling variants are \texttt{identifiable\_boolean} (\texttt{\{"identifiable": true\}}) and the
    adjustment-set queries \texttt{one\_valid\_adjustment\_set}
    (\texttt{\{"adjust": [\ldots] |\allowbreak{} "no\_backdoor" |\allowbreak{} "non\_id"\}}), \texttt{minimal\_adjustment\_set\_size}
    (\texttt{\{"k": \ldots |\allowbreak{} "no\_backdoor" |\allowbreak{} "non\_id"\}}), \texttt{n\_valid\_adjustment\_sets}
    (\texttt{\{"n": \ldots |\allowbreak{} 0 |\allowbreak{} "no\_backdoor" |\allowbreak{} "non\_id"\}}), and
    \texttt{all\_minimal\_adjustment\_sets}
    (\texttt{\{"adjustment\_sets": [\ldots] |\allowbreak{} "no\_backdoor" |\allowbreak{} "non\_id"\}}). For the all-minimal variant,
    \texttt{[[]]} denotes the empty adjustment set as the only minimal valid set; for the one-valid variant,
    \texttt{[]} denotes that no adjustment is needed. The size, count, and all-minimal variants list the offered
    covariates under an \texttt{\#\#\#~Available Conceptual Covariates} header, while the one-valid variant
    names them in the question prose.

        {\footnotesize
    \begin{promptblock}
        ### Conceptual Observed Variables
        The following conceptual variables are observed for causal-identification purposes, in no particular order: "Treatment", "Outcome", "Covariate1", "Covariate2", "Mediator1"

        ### Question
        Is the population Average Treatment Effect (ATE) of **Treatment** on **Outcome** identifiable from the observational distribution over the conceptual observed variables? Return the first applicable identification label.

        Target estimand:
        ATE = E[Y | do(Treatment=x1)] - E[Y | do(Treatment=x0)]

        Treat the story as specifying the qualitative conceptual causal graph. Identifiability is a property of that graph and the observational distribution over the observed conceptual variables; do not assume parametric forms, effect homogeneity, or access to the structural model.

        Return the first applicable label:
        1. "trivial_zero": Treatment has no directed causal path to Outcome, so the population ATE is identifiable as zero.
        2. "backdoor": Otherwise, a valid backdoor adjustment set among the observed conceptual variables identifies the population ATE.
        3. "frontdoor": Otherwise, a valid front-door formula using observed conceptual variables identifies the population ATE.
        4. "other_id": Otherwise, the population ATE is identifiable by another valid do-calculus / ID argument.
        5. "none": The population ATE is not identifiable.

        ### Output Format
        Provide your answer as a JSON object:
        ```json
            {"method": "backdoor"}
        ```
    \end{promptblock}
    }

    \paragraph{R2: Effect estimate --- population ATE (point).}
    Data-backed (Measurement Note shown). Variants share this stem and swap the output contract:
    \texttt{ate\_sign\_only} (\texttt{\{"sign": "+"/\allowbreak{}"-"/\allowbreak{}"0"/\allowbreak{}"unknown"\}}), \texttt{ate\_vs\_assoc\_sign\_match}
    (\texttt{\{"matches": true|\allowbreak{}false|\allowbreak{}null\}}), and \texttt{ate\_uq\_95}, which appends ``also provide a central 95\%
    confidence interval'' and returns \texttt{\{"ate": .., "ci\_lower": .., "ci\_upper": ..\}}. Non-identifiable
    scenes (e.g.\ bare-IV) require the \texttt{null}/\texttt{unknown} abstention value.
        {\footnotesize
    \begin{promptblock}
        ### Data
        You are provided with observational data in `data.parquet` containing the following columns:
        "Treatment", "Outcome", "Covariate1", "Covariate2"

        ### Question
        Estimate the population Average Treatment Effect (ATE) of the conceptual treatment **Treatment** on the conceptual outcome **Outcome**.

        Use treatment levels x0=0.0 and x1=1.0.

        ATE = E[Y | do(Treatment=1.0)] - E[Y | do(Treatment=0.0)]

        ### Output Format
        Provide your answer as JSON:
        ```json
            {"ate": 1.23}
        ```
        If the population ATE is not identifiable from the observational distribution over the conceptual variables, return:
        ```json
            {"ate": null}
        ```
    \end{promptblock}
    }

    \paragraph{R2: Bias diagnostic --- collider bias.}
    The \texttt{forbidden\_controls\_list} variant instead lists the offered conceptual covariates and asks
    which must \emph{not} be conditioned on, returning
    \texttt{\{"forbidden": [\ldots] | [] | "no\_backdoor" | "non\_id"\}} (set-$F_1$ graded, with the
    non-identifiable branch routed to abstention). The empty list means that adjustment is available and no
    listed variable must be excluded.

        {\footnotesize
    \begin{promptblock}
        ### Question
        A researcher wants to estimate the population causal effect of **Treatment** on **Outcome**.
        They propose to condition on the conceptual variable **Collider1** in their analysis.

        Would conditioning on **Collider1** introduce collider bias or otherwise open a noncausal path between treatment and outcome?

        Answer using the story-implied conceptual causal graph. The released data may help inspect associations, but the bias judgment is causal and graph-based. Measurement columns are measurements of conceptual variables, not separate controls.

        ### Output Format
        Provide your answer as a JSON object:
        ```json
            {"bias_present": true, "explanation": "..."}
        ```
    \end{promptblock}
    }

    \medskip\noindent\textbf{Rung 3 (Counterfactual).}\par

    \paragraph{R3: Counterfactual identification --- ETT (\texttt{identifiable\_boolean}).}
    A story-only judgment of whether the target counterfactual estimand is identifiable; instantiated for ETT,
    NDE, and NIE, swapping in the corresponding estimand name, formula, and mediator list.

        {\footnotesize
    \begin{promptblock}
        ### Question
        Treat the story as specifying the qualitative conceptual causal graph.

        Target estimand: **Effect of the Treatment on the Treated (ETT)**

        Formal notation:
        ETT(x1, x0) = E[Y_{x1} - Y_{x0} | X=x1], with x0=0, x1=1

        Is the effect of treatment on the treated for **Treatment** on **Outcome** identifiable from the observational distribution under the story-implied conceptual graph?

        Identifiability of this counterfactual estimand is a property of the graph and the population observational distribution alone; do not assume parametric forms, effect homogeneity, monotonicity, or access to the structural model.

        ### Output Format
        Provide your answer as a JSON object:
        ```json
            {"identifiable": true, "explanation": "..."}
        ```
    \end{promptblock}
    }

    \paragraph{R3: Counterfactual effect --- ETT (point).}
    Data-backed. The \texttt{sign\_only} variant returns \texttt{\{"sign": "+"/\allowbreak{}"-"/\allowbreak{}"0"/\allowbreak{}"unknown"\}}, and
    \texttt{effect\_uq\_95} appends ``estimate the population ETT and provide a central 95\% confidence interval'',
    returning \texttt{\{"value": .., "ci\_lower": .., "ci\_upper": ..\}}. A non-identifiable ETT requires the
    \texttt{value=null} (or \texttt{sign="unknown"}) abstention value.
        {\footnotesize
    \begin{promptblock}
        ### Question
        Estimate the population Effect of the Treatment on the Treated (ETT) of the conceptual treatment **Treatment** on the conceptual outcome **Outcome**.

        Use treatment levels **x0=0** and **x1=1**.

        ETT(x1, x0) = E[Y_{x1} - Y_{x0} | X=x1]

        If the ETT is not identifiable from the observational distribution over the conceptual variables, return `{"value": null}`.

        ### Output Format
        Provide your answer as a JSON object:
        ```json
            {"value": 1.23, "explanation": "Briefly state the identifiability judgment and estimation approach."}
        ```
    \end{promptblock}
    }

    \needspace{10\baselineskip}
    \paragraph{R3: Mediation effect --- NDE (point).}
    Shown for the NDE point estimate; the NIE prompt is analogous with $\mathrm{NIE}(x_1, x_0) =
    E[Y_{x_0,M_{x_1}} - Y_{x_0,M_{x_0}}]$. For each estimand, \texttt{sign\_only} returns
    \texttt{\{"sign": "+"/\allowbreak{}"-"/\allowbreak{}"0"/\allowbreak{}"unknown"\}} and \texttt{effect\_uq\_95} appends a central 95\% confidence interval
    (\texttt{\{"value": .., "ci\_lower": .., "ci\_upper": ..\}}); a non-identifiable effect requires the
    \texttt{value=null} / \texttt{sign="unknown"} abstention.
        {\footnotesize
    \begin{promptblock}
        ### Question
        Estimate the population Natural Direct Effect (NDE) of the conceptual treatment **Treatment** on the conceptual outcome **Outcome**, with mediator(s): **Mediator1**.

        Use treatment levels **x0=0** and **x1=1**.

        NDE(x1, x0) = E[Y_{x1,M_{x0}} - Y_{x0,M_{x0}}]

        If the NDE is not identifiable from the observational distribution over the conceptual variables, return `{"value": null}`.

        ### Output Format
        Provide your answer as a JSON object:
        ```json
            {"value": 0.85, "explanation": "Briefly state the identifiability judgment and estimation approach."}
        ```
    \end{promptblock}
    }

    The \texttt{direct\_vs\_indirect\_dominance} variant instead asks which component (NDE or NIE) has larger
    absolute magnitude, returning \texttt{\{"dominant": "direct"/"indirect"/"tie"\}} (or \texttt{null} when
    either the NDE or NIE is non-identifiable).

\end{document}